\documentclass[preprint,12pt]{elsarticle}
\usepackage[utf8]{inputenc}
\usepackage[nottoc]{tocbibind}
\DeclareUnicodeCharacter{02BC}{'}

\usepackage{amssymb}
\usepackage[utf8]{inputenc}
\usepackage[T1]{fontenc}
\usepackage{textcomp}
\usepackage{algorithm}
\usepackage{algpseudocode}

\usepackage{amsmath,amsthm,amsfonts,amscd} 
\usepackage{upgreek}
\usepackage{mdframed}
\usepackage{multicol}
\usepackage{graphicx}
\usepackage{nomencl}
\usepackage{booktabs}
\usepackage{setspace}
\usepackage{subfig}
\usepackage{verbatim}   	
\usepackage{makeidx}    	
\usepackage{color}
\usepackage{float}
\usepackage{enumerate}
\usepackage{url}	
\usepackage{tcolorbox}
\usepackage{xcolor}
\usepackage{multirow}
\usepackage{ulem}
\usepackage[pdftex,bookmarks]{hyperref}
\usepackage{bbm, dsfont}

\usepackage[percent]{overpic}
\usepackage[table]{xcolor}
\usepackage{tikz} 
\usetikzlibrary{shapes.geometric, arrows, positioning, fit, calc}
\usetikzlibrary{shapes,arrows,positioning,calc}
\usetikzlibrary{arrows.meta, positioning, calc}

\definecolor{myred}{RGB}{190, 40, 40}
\definecolor{myblue}{RGB}{40, 40, 190}
\definecolor{mygreen}{RGB}{40, 190, 40}
\definecolor{inputcolor}{RGB}{200, 220, 255}
\definecolor{outputcolor}{RGB}{255, 220, 200}
\definecolor{modelcolor}{RGB}{200, 255, 200}
\definecolor{projectioncolor}{RGB}{255, 230, 150}

\newcommand{\bs}{\boldsymbol}
\newtheorem{ribnotheorem}{Theorem}

\usepackage[margin=1.2in]{geometry} 

\usepackage{etoolbox} 
\makeatletter
\patchcmd{\@startsection}
  {\@afterindenttrue}
  {\@afterindentfalse}
  {}{}
\makeatother

\journal{}

\begin{document}

\begin{frontmatter}



\title{
\textbf{Learning inadequacy of Bayesian neural operators for PDE-constrained inference and design}
}


\author[inst1]{Pratyush Kumar Singh}
\author[inst2]{Lianghao Cao}
\author[inst3]{E. Bruce Pitman}
\author[inst1]{Danial Faghihi\corref{cor1}}

\affiliation[inst1]{organization={Department of Mechanical and Aerospace Engineering, \\University at Buffalo},
            city={Buffalo},
            state={NY},
            country={USA}}
            
\affiliation[inst2]{organization={Department of Computing and Mathematical Sciences, \\California Institute of Technology},
            city={Pasadena},
            state={CA},
            country={USA}}

\affiliation[inst3]{organization={Department of Materials Design and Innovation, \\University at Buffalo},
            city={Buffalo},
            state={NY},
            country={USA}}

\cortext[cor1]{Corresponding Author, \texttt{danialfa@buffalo.edu} (D. Faghihi)}            

\begin{abstract}
Partial differential equations (PDEs) underpin predictive modeling across science, engineering and medicine. Their use in digital twins, parameter inference and optimal design requires repeated solutions across parameters and scenarios. For nonlinear, transient and multiphysics systems, these calculations can make numerical simulation prohibitively expensive.
Here we introduce the residual-informed Bayesian neural operator (RI-BNO), which learns operator inadequacy using a posteriori error estimates from the governing PDEs and incorporates its correction and uncertainty into Bayesian training and prediction. We establish local error bounds for the residual-informed estimates under stability and smoothness assumptions.
We demonstrate RI-BNO on a nonlinear reaction-diffusion benchmark with DeepONet and latent-subspace architectures, Markov chain Monte Carlo inference of glioma parameter fields from magnetic resonance imaging, and Monte Carlo risk-averse optimization of three-dimensional thermal breaks in building envelopes under manufacturing uncertainty. 
Across the evaluated examples, RI-BNO improves agreement with finite element (FE) reference predictions, parameter posteriors and optimized designs while using half as many full nonlinear or transient training simulations as the vanilla Bayesian neural operator (vBNO). For thermal break design, online optimization time decreases from 178 hours with the FE model to 18 minutes with RI-BNO.
These results demonstrate the value of explicit inadequacy modeling in neural operators for reliable PDE-based inference and design.

\end{abstract}

\begin{keyword}
Bayesian neural operators \sep
model inadequacy \sep
operator learning \sep
a posteriori error estimation \sep
uncertainty quantification
\end{keyword}

\end{frontmatter}



\section{Introduction}
\label{sec:introduction}
Partial differential equations (PDEs) provide mechanistic descriptions
of physical and biological systems and underpin scientific inference,
engineering design and digital twins, supporting consequential decisions. Applying these models to Bayesian inversion, uncertainty quantification, model-based control, and optimization requires repeated forward solutions across uncertain parameters, evolving states, and candidate interventions. Finite element (FE) methods provide accurate reference solutions, but repeatedly solving nonlinear, transient, or multiphysics PDEs can make these many-query calculations computationally prohibitive for realistic deployment \cite{brunton2024promising,ouyang2026noem}. Neural operators reduce this cost by learning maps from input parameters or functions to PDE solution fields using a finite collection of numerical simulations, amortizing the offline cost over rapid online evaluations. Prominent architectures include DeepONet \cite{lu2021learning}, Fourier neural operators \cite{li2021fourier}, latent-subspace constructions for high-dimensional parameter fields \cite{bhattacharya2021model,OLeary2022derivative,OLeary2024derivative}, and geometry-aware formulations for complex computational domains \cite{yin2024scalable}. 
Their usefulness in downstream analysis, however, depends on more than average forward accuracy. Errors that appear modest in the predicted state can be amplified by inversion, control or design optimization, potentially biasing inferred parameters or altering the selected intervention or design \cite{pickering2022discovering,cao2025derivative,luo2025efficient}. 
For these applications, a central question is whether accelerated
computations preserve the parameter estimates, risk measures and
designs obtained with the reference PDE model.
Bayesian neural operators represent uncertainty in learned weights arising from finite training data and propagate this uncertainty to predicted states \cite{yang2022scalable,garg2023vb,moya2023deeponet}.
The resulting predictive distributions remain conditional
on the adopted representation, architecture, training procedure and inference approximation. Weight uncertainty alone does not explicitly represent systematic discrepancy between the learned operator and the PDE solution operator.
We refer to structured error arising from finite-dimensional representations, architectural restrictions and imperfect training as neural operator inadequacy. Such inadequacy can persist as the training set grows, producing accuracy plateaus even as uncertainty associated with finite data decreases.
Bayesian calibration studies show that flexible discrepancy models can improve predictions within the observed data range without eliminating bias in inferred physical parameters \cite{kennedy2001bayesian,brynjarsdottir2014learning}. Parameter changes and model discrepancy can explain the same observations, making their effects difficult to distinguish. Resolving this ambiguity requires informative assumptions about the discrepancy, such as its expected dependence on the inputs or constraints arising from known physical mechanisms.
This motivates learning neural operator inadequacy from information
supplied by the governing equations, complementary to the solution
data used for training.
%

In this work, we introduce the residual-informed Bayesian neural operator (RI-BNO), which uses a posteriori information from the governing PDE to account for operator inadequacy in Bayesian training and prediction. Building on residual-based estimation of modeling error \cite{CAO_jcp2023,JHA_cmame2024,oden2002estimation,prudhomme1999goal}, RI-BNO obtains approximate observations of neural operator error through linearized variational solves, providing additional information about inadequacy without corresponding full nonlinear FE simulations. These observations inform a probabilistic inadequacy operator that learns an input dependent correction and its uncertainty. 
The learned inadequacy operator enter the likelihood used
to infer the neural operator weights, so residual information shapes
both Bayesian training and the final predictive distribution of the corrected state.
We establish a local error bound for the residual-informed observations under stability and smoothness assumptions, supporting their use within this Bayesian formulation. RI-BNO accommodates different neural operator architectures whenever the governing residual and its state linearization are accessible. 
Its modular implementation constructs the residual observations and
learns the inadequacy model offline, allowing subsequent inference
and optimization to use the learned neural operator and inadequacy
model without further PDE residual solves.

We evaluate RI-BNO on a nonlinear reaction–diffusion benchmark and two applications requiring repeated PDE evaluations within Bayesian inference and risk-averse optimization. The benchmark tests the approach with DeepONet and latent-subspace architectures. The first application uses Markov chain Monte Carlo sampling to infer spatially varying glioma parameters from magnetic resonance imaging. The second optimizes three-dimensional thermal breaks in building envelopes using Monte Carlo estimates of tail risks under manufacturing uncertainty. Both applications require many forward model evaluations and examine how operator inadequacy affects inferred parameter fields, posterior predictions, risk estimates and selected designs.
Across these settings, RI-BNO achieves closer agreement with FE reference predictions, parameter posteriors and optimized designs than vBNO, while using half as many full nonlinear or transient FE training simulations. In the thermal break application, online optimization time decreases from approximately 178 hours with the FE model to 18 minutes with RI-BNO.

\section{Method overview}\label{sec:method}
This section introduces residual-informed Bayesian neural operators (RI-BNOs) for learning reliable solution operators of parametric PDEs. 

\subsection{Problem formulation}
We consider a parametric PDE written in residual form as,
\begin{equation}
    \text{Given } m \in \mathcal{M}, \text{ find } u \in \mathcal{U} \text{ such that } \mathcal{R}(m, u) = 0,
\end{equation}
where \( \mathcal{R}(m, u) \) is the residual functional, $u$ is the PDE state, $m$ is a spatially varying parameter, and $\mathcal{U}$ and $\mathcal{M}$ are separable Hilbert spaces.
When the problem is well posed, this equation defines a solution operator $\mathcal{F}:\mathcal{M}\rightarrow\mathcal{U}$.
Let \(\mathcal{F}_h:\mathcal{M}\to\mathcal{U}\) denote a numerical
approximation of this operator, such as a finite element approximation.
Given training pairs
\(\mathcal{D}=\{(m^{(i)},u_h^{(i)})\}_{i=1}^{N_{\mathrm{train}}}\),
where \(u_h^{(i)}=\mathcal{F}_h(m^{(i)})\), one can train a neural operator
\(\mathcal{F}_{\bs w}:\mathcal{M}\to\mathcal{U}\), parameterized by weights
\(\bs w\in\mathbb{R}^{N_w}\), to approximate \(\mathcal{F}_h\), providing state prediction as \(u_{\bs w}=\mathcal{F}_{\bs w}(m)\).
While the formulation is stated in function spaces, in computations, input fields and predicted states are represented on computational meshes, with FE outputs identified with their corresponding functions in \(\mathcal U\).

%

\subsection{Scalable Bayesian neural operators}
Deterministic neural operators provide a point estimate of the operator parameters and do not directly quantify uncertainty in the learned map. In a Bayesian neural operator, the weights \(\bs w\in\mathbb{R}^{N_w}\) are treated as random variables and inferred from the training data:
\begin{equation}\label{eq:bayes}
 \pi_{\rm post}(\bs w|\mathcal{D}) \propto
 \pi_{\rm like}(\mathcal{D}|\bs w) \,  
 \pi_{\rm pr}(\bs w).
\end{equation}
Here, the posterior probability distribution over weights $\pi_{\rm post}(\bs w|\mathcal D)$ is calculated from the prior  $\pi_{\rm pr}(\bs w)$ and likelihood $\pi_{\rm like}(\mathcal{D}|\bs w)$, inducing a posterior predictive distribution over PDE states. Direct Bayesian inference, however, is not tractable for modern neural operators because \(N_w\) is large and the posterior covariance cannot be formed or inverted explicitly. Exploring the high-dimensional weight posterior by Markov chain Monte Carlo (MCMC) can require many likelihood evaluations \cite{psaros2023uncertainty}, motivating a computationally tractable approximation.
%
We use a Laplace approximation, which represents the posterior locally by a Gaussian distribution centered at a MAP estimate,
$
\pi_{\mathrm{post}}(\bs w | \mathcal D)
\approx
\mathcal{N}
\left(
\bs w_{\mathrm{MAP}},
\bs \Gamma_{\mathrm{post}}
\right),
$
where the MAP estimate is obtained from
\begin{equation}\label{eq:map_estimate}
\bs w_{\mathrm{MAP}}
=
\underset{\bs{w}}{\rm{argmin}}
\left\{
-\log \pi_{\mathrm{like}}(\mathcal{D}|\bs w)
-\log \pi_{\mathrm{pr}}(\bs w)
\right\}.
\end{equation}
The central computational challenge is then to approximate the posterior covariance
\(\bs\Gamma_{\mathrm{post}} \in \mathbb{R}^{N_w \times N_w}\), which requires local curvature information in the neural operator weight space.
We obtain the MAP estimate using a matrix-free Low-Rank Saddle-Free Newton (LRSFN) optimizer \cite{oleary2019inexact}. For covariance construction, we approximate the likelihood curvature at the MAP point in a low-rank subspace to construct the Laplace covariance.
Let $\mathbf{H}_{\mathrm{like}}(\bs w_{\rm MAP})$ be the Hessian of the negative log-likelihood at the MAP point. We approximate its dominant rank-\(r\) subspace as
$ \mathbf{H}_{\mathrm{like}} \approx \mathbf{V}_r \bs\Lambda_r \mathbf{V}_r^{T}$
with
$\mathbf{V}_r\in\mathbb{R}^{N_w\times r}$,
$\bs\Lambda_r\in\mathbb{R}^{r\times r}$,
and
$r\ll N_w$.
For a Gaussian prior with covariance
\(\bs\Gamma_{\mathrm{pr}}\), the posterior covariance is then approximated by
\begin{equation}\label{eq:low_rank_laplace_covariance}
\bs \Gamma_{\mathrm{post}}
=
\left(
\bs \Gamma_{\mathrm{pr}}^{-1}
+
\mathbf{H}_{\mathrm{like}}
\right)^{-1}
\approx
\bs \Gamma_{\mathrm{pr}}
-
\bs \Gamma_{\mathrm{pr}}
\mathbf{V}_r
\left(
\bs \Lambda_r^{-1}
+
\mathbf{V}_r^{T}
\bs \Gamma_{\mathrm{pr}}
\mathbf{V}_r
\right)^{-1}
\mathbf{V}_r^{T}
\bs \Gamma_{\mathrm{pr}} .
\end{equation}
The low-rank approximation avoids forming or factorizing a dense \(N_w\times N_w\) covariance matrix.
This provides a scalable Bayesian neural operator baseline, referred to here as the vanilla Bayesian neural operator (vBNO) with the posterior predictive state as $u_{\rm vBNO}=\mathcal{F}_{\bs w}(m)$ with $\bs w \sim \pi_{\rm post}(\bs w|\mathcal D)$. 
This predictive distribution captures uncertainty arising from finite training
data and weight estimation. Full implementation details for LRSFN and the
low-rank covariance construction are provided in the
\textit{Supplementary Information}.

\subsection{Residual-informed Bayesian neural operators (RI-BNO)}
The vBNO described above quantifies uncertainty in its learned weights, but its prediction can retain systematic discrepancy between the learned
operator $\mathcal F_{\bs w}$ and the solution operator $\mathcal F$.
Such discrepancy can arise from restricted representations, architectural constraints and imperfect training. 
RI-BNO represents this inadequacy explicitly and uses the governing PDE residual to inform its correction and uncertainty. We describe the numerical solution data through the additive statistical model
\begin{equation}\label{eq:additive_inadequacy}
    \mathcal F_h(m)
    =
    \mathcal F_{\bs w}(m)
    +
    \eta(m)
    +
    \varepsilon_{\rm data},
\end{equation}
where $\eta:\mathcal M\to\mathcal U$ represents structured neural operator
inadequacy relative to the solution operator $\mathcal F$, and
$\varepsilon_{\rm data}$ accounts for numerical approximation and other
data generation errors in the training states.
Solution data alone cannot generally distinguish structured inadequacy
from changes in the neural operator weights. RI-BNO therefore uses the
governing PDE residual to construct complementary observations of
operator error.

\begin{ribnotheorem}[Residual-informed operator error estimation]
\label{thm:residual_inadequacy_estimate}
Let $\eta(m)=\mathcal F(m)-\mathcal F_{\bs w}(m)$ and let $\nu_{\mathcal M}$ be the
input probability measure. The residual-based estimate of this error
satisfies
\begin{equation}\label{eq:error_estimate}
    \delta_u\mathcal R\left(m,\mathcal F_{\bs w}(m)\right)
    [e_{\rm est}(m)]
    =-\mathcal R\left(m,\mathcal F_{\bs w}(m)\right).
\end{equation}
Suppose that, for $\nu_{\mathcal M}$-almost every $m$, the state derivative of
$\mathcal R(m,\cdot)$ is $L$-Lipschitz between
$\mathcal F_{\bs w}(m)$ and $\mathcal F(m)$, and its linearization at
$\mathcal F_{\bs w}(m)$ has an inverse with norm at most $\gamma$,
where $L$ and $\gamma$ are uniform in $m$. If
$\mathbb E_{m\sim\nu_{\mathcal M}}\|\eta(m)\|_{\mathcal U}^{2}<\infty$, then
\begin{equation}\label{eq:main_residual_observation_bound}
    \mathbb E_{m\sim\nu_{\mathcal M}}
    \left[\|e_{\rm est}(m)-\eta(m)\|_{\mathcal U}\right]
    \leq
    \omega_{\mathcal R}
    \mathbb E_{m\sim\nu_{\mathcal M}}
    \left[\|\eta(m)\|_{\mathcal U}^{2}\right],
    \qquad
    \omega_{\mathcal R}:=\frac{\gamma L}{2}.
\end{equation}
Equation~\eqref{eq:error_estimate} is one Newton correction initialized
at the neural operator prediction
\cite{Ortega1968,CAO_jcp2023,JHA_cmame2024}.
The bound shows that the error remaining after this ideal correction
is second order in the original prediction error. The proof is given
in the \textit{Supplementary Information}.
\end{ribnotheorem}
%


The residual estimates from \eqref{eq:error_estimate} provide
information about neural operator inadequacy with respect to PDE solution, subject to
linearization and numerical approximation errors. 
We therefore interpret $e_{\rm est}$ as a noisy observation of
$\eta$ through the statistical observation model:
\begin{equation}\label{eq:error_observation}
    e_{\rm est}(m)
    =
    \eta(m)
    +
    \varepsilon_{\rm err},
\end{equation}
where $\varepsilon_{\rm err}$ represents unresolved error in
relating the residual estimates to the inadequacy field.
Collecting these observations gives
$\mathcal E=\{(m^{(i)},e_{\rm est}^{(i)})\}_{i=1}^{N{\rm err}}$,
which supplies information from the governing PDE complementary
to the solution data $\mathcal D$.
From $\mathcal E$, we learn a probabilistic inadequacy operator
$\mathcal G_{\bs g}:\mathcal M\to\mathcal U$, parameterized by
$\bs g$. This operator predicts an input-dependent correction to
the neural operator state and quantifies uncertainty in that correction.

A fully Bayesian formulation would infer the neural operator weights,
the inadequacy model, and the hyper-parameters jointly:
\begin{multline}\label{eq:full_bayesian}
    \pi_{\rm post}(\bs w,\bs g,\bs\Sigma_{\rm data},
    \bs\Sigma_{\rm err}\mid\mathcal D,\mathcal E)
    \propto
    \pi_{\rm like}(\mathcal D,\mathcal E\mid
    \bs w,\bs g,\bs\Sigma_{\rm data},\bs\Sigma_{\rm err})\\
    {}\times
    \pi_{\rm pr}(\bs w)\pi_{\rm pr}(\bs g)
    \pi_{\rm pr}(\bs\Sigma_{\rm data})
    \pi_{\rm pr}(\bs\Sigma_{\rm err}).
\end{multline}
Here $\pi_{\rm like}$ is a joint likelihood for the solution data and
residual-informed observations; $\bs\Sigma_{\rm data}$ and
$\bs\Sigma_{\rm err}$ parameterize their respective noise models.
Under the specified likelihood and priors, this hierarchy propagates
the modeled uncertainties in weights, inadequacy, and observations.
The RI-BNO posterior predictive state is
$u_{\rm RI}(m)=\mathcal F_{\bs w}(m)+\mathcal G_{\bs g}(m)$,
with $(\bs w,\bs g)$ drawn from their joint posterior distributions.
The formulation accommodates neural operator architectures whose
predictions admit evaluation of the governing PDE residual and its
state linearization.

\subsection{Algorithms and practical considerations}
\label{sec:algorithms}
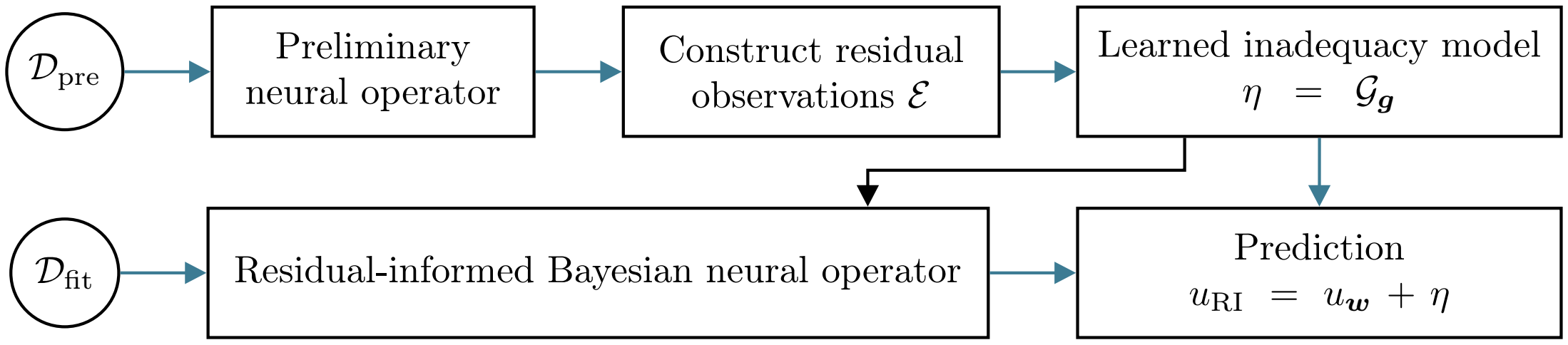
\begin{figure}
        \centering
    \footnotesize
    \begin{tikzpicture}[
        >=latex,
        node distance=0.8cm,
        block/.style={rectangle, draw, thick, text width=2.75cm, align=center, minimum height=1.2cm},
        blockb/.style={rectangle, draw, thick, text width=4.25cm, align=center, minimum height=1.2cm},
        wideblockb/.style={rectangle, draw, thick, text width=4.25cm, align=center, minimum height=1.2cm},
        wideblock/.style={rectangle, draw, thick, text width=3.25cm, align=center, minimum height=1.2cm},
        longblock/.style={rectangle, draw, thick, text width=7.0cm, align=center, minimum height=1.2cm},
        circle_node/.style={circle, draw, thick, minimum size=1cm, align=center},
        arrow/.style={-{Triangle[length=2mm, width=2mm]}, thick, color=cyan!50!black}
    ]

        \node[circle_node] (D_pre) {$\mathcal{D}_{\text{pre}}$};
        \node[block, right=0.8cm of D_pre] (pre_op) {Preliminary\\neural operator};
        \node[wideblock, right=0.8cm of pre_op] (res_obs) {Construct residual\\observations $\mathcal{E}$};
        \node[wideblockb, right=0.7cm of res_obs] (inc_model) {Learned inadequacy model\\ $\eta=\mathcal{G}_{\bs g}$};

        \node[circle_node, below=0.8cm of D_pre] (D_fit) {$\mathcal{D}_{\text{fit}}$};
        \node[longblock, right=0.8cm of D_fit] (ribno) {Residual-informed Bayesian neural operator};
        \node[blockb, right=0.8cm of ribno] (pred) {Prediction\\$u_{\text{RI}} = u_{\bs{w}} + \eta$};

        \draw[arrow] (D_pre) -- (pre_op);
        \draw[arrow] (pre_op) -- (res_obs);
        \draw[arrow] (res_obs) -- (inc_model);

        \draw[arrow] (D_fit) -- (ribno);
        \draw[arrow] (ribno) -- (pred);

        \draw[arrow,color=black] ([xshift=-1.25cm]inc_model.south) -- ++(0,-0.3) -| ([xshift=2.5cm]ribno.north);
        \draw[arrow] (inc_model.south) -- (pred.north);

    \end{tikzpicture}
        \caption{
        \textbf{Residual-informed Bayesian neural operator (RI-BNO) workflow.} A preliminary neural operator trained on $\mathcal D_{\rm pre}$ supplies predictions at additional, independently sampled inputs. Solving \eqref{eq:error_estimate} generate residual-informed observations $\mathcal E$, from which the inadequacy model $\eta$ is learned. This model is held fixed during final Bayesian training on the disjoint solution dataset $\mathcal D_{\rm fit}$. The learned inadequacy informs both the Bayesian likelihood and the predictive state.
        }
        \label{fig:workflow}
\end{figure}

Direct inference in the joint posterior \eqref{eq:full_bayesian} is
computationally demanding for neural operators. We therefore implement
RI-BNO in three stages (Figure \ref{fig:workflow}): construct residual-informed observations, learn
an inadequacy model, and train the Bayesian neural operator conditional
on that model. Bayesian calibration studies emphasize that discrepancy models need information about the expected structure and behavior of model error \cite{brynjarsdottir2014learning}. In RI-BNO, the governing PDE residual supplies this information through error estimates that complement the solution training data.
We partition the finite element solution data into two disjoint sets,
$\mathcal D=\mathcal D_{\rm pre}\cup\mathcal D_{\rm fit}$,
reserved for preliminary operator training and final Bayesian inference,
respectively. A preliminary neural operator is trained on
$\mathcal D_{\rm pre}$ to provide reference predictions for residual
evaluation and linearization. Its representative fitted weights---a MAP
estimate or variational mean, as appropriate---are held fixed while
predictions are evaluated at additional inputs sampled independently
from $\nu_{\mathcal M}$, distinct from those in $\mathcal D$.
Applying \eqref{eq:error_estimate} at these predictions generates
residual-informed observations $\mathcal E$, without generating full
reference solutions (or trajectories in time-dependent PDEs) at those inputs.
The inadequacy model is learned from $\mathcal E$ and held fixed
during final RI-BNO training using only $\mathcal D_{\rm fit}$.
Under independent sampling, this separation keeps $\mathcal D_{\rm fit}$
statistically independent of the data used to construct the inadequacy
model. It prevents the final training states from informing both the
learned correction and the Bayesian likelihood when the learned inadequacy
operator is held fixed.
It also allows the residual observations and inadequacy model to be
constructed once, before final Bayesian training.
Direct FE prediction errors at the additional inputs would require
full PDE simulations and measure discrepancy relative to the numerical
operator $\mathcal F_h$. 
The function space residual formulation targets error relative to the exact solution operator $\mathcal F$. Linearization and discretization errors in the computed estimates are represented by $\varepsilon_{\rm err}$ in the noise model \eqref{eq:error_observation}.

For the numerical experiments, we adopt a common allocation,
$|\mathcal D_{\rm pre}|=|\mathcal D_{\rm fit}|=N_{\rm train}/2$
and $N_{\rm err}=N_{\rm train}$, to assess accuracy and computational
cost under a consistent construction protocol across the examples.
More generally, these allocations can be adapted to the data
requirements of the preliminary operator, inadequacy model and final
Bayesian inference, together with the relative costs of generating solution
data and residual-informed observations.
In this work, we represent the inadequacy model $\mathcal G_{\bs g}$ using a sparse
variational Gaussian process (GP) \cite{titsias09a}. 
The GP is fitted to the residual-informed observation set $\mathcal E$, using the fixed-basis representation of the parameter fields together with spatial coordinates as inputs and the corresponding residual estimates as observations. The basis representation provides a common discretization-independent parameterization of the functional inputs, while the GP learns spatially varying inadequacy corrections through the operator product kernel.
Let $\bs e_{\rm est}$ stack the estimated error fields across the parameter samples. The GP observation
model has marginal likelihood
\begin{equation}\label{eq:gp_like}
    \pi_{\mathrm{like}}(\mathcal E\mid\bs g,\bs\Sigma_{\rm err})
    =
    \mathcal N\!\left(
        \bs e_{\rm est}-\bs\mu_\eta;
        0,\bs K_\eta+\bs\Sigma_{\rm err}
    \right).
\end{equation}
Here $\bs\mu_\eta$ and $\bs K_\eta$ are the GP mean and covariance
at the residual-observation inputs before conditioning on
$\mathcal E$. We estimate the GP parameters $\bs g$ and
residual-observation error parameters by maximizing a sparse
variational lower bound on the log marginal likelihood in
\eqref{eq:gp_like}.
After learning the GP from $\mathcal E$, let
$\bs\mu_{\eta\mid\mathcal E}$ and $\bs C_{\eta\mid\mathcal E}$
denote its predictive mean and latent covariance at the inputs in
$\mathcal D_{\rm fit}$, reconstructed in the state representation. 
Holding the fitted GP fixed and treating the latent correction and training data error as independent Gaussian terms gives the inadequacy-aware likelihood
\begin{equation}\label{eq:bno_like}
\pi_{\mathrm{like}}
(\mathcal D_{\rm fit}\mid\bs w,\bs\Sigma_{\rm data},\mathcal E)
    =
    \mathcal N\!\left(
        \bs u_h-\bs u_{\bs w}-\bs\mu_{\eta\mid\mathcal E};
        0,\bs C_{\eta\mid\mathcal E}+\bs\Sigma_{\rm data}
    \right),
\end{equation}
where $\bs u_h$ and $\bs u_{\bs w}$ stack the numerical training
states and neural operator predictions, respectively, at the inputs
in $\mathcal D_{\rm fit}$. 
All terms in this likelihood are expressed in the same state representation. We estimate \(\bs\Sigma_{\rm data}\) by evidence maximization \cite{singh2024framework}, while the residual-observation covariance \(\bs\Sigma_{\rm err}\) is estimated during GP fitting. With the fitted GP held fixed, we infer the RI-BNO weight posterior using the original weight prior and \eqref{eq:bno_like}. This modular approximation incorporates the learned inadequacy into both the mean and covariance of the likelihood. The GP mean accounts for systematic discrepancy, while its covariance reduces the weight assigned to discrepancies in directions where the correction is uncertain.

\section{Example 1: Nonlinear reaction-diffusion equation}\label{sec:example1}
We first consider a benchmark PDE to isolate neural operator inadequacy and compare vBNO and RI-BNO predictions.
The nonlinear reaction-diffusion problem is given by
\begin{eqnarray}
    -\nabla \cdot \left(\exp(m(\bs x)) \, \nabla u(\bs x) \right) + u(\bs x)^3 & = f(\bs x), \qquad & \bs x \in \Omega \nonumber\\
    u(\bs x) &= 0  \qquad & \bs x \in \Gamma_b \nonumber\\
    \exp(m(\bs x)) \, \nabla u(\bs x) \cdot \bs n &= 0  \qquad & \bs x \in \Gamma_c
\end{eqnarray}
where $\Omega=[0,1]^2$, $\Gamma_b$ denotes the bottom edge,
$\Gamma_c=\partial\Omega\setminus\Gamma_b$, and $\bs n$ is the outward unit normal.
The prescribed source term is
$f(\bs x) = \exp\left(-4(0.75 - x_1)^2\right) \cos^2(8\pi x_1)$, where $\bs x=(x_1, x_2)$.
The parameter field $m \sim \nu_{\mathcal M} =\mathcal N(0,\mathcal C)$ is modeled as a zero-mean Gaussian random field with a Matérn-type covariance, parameterized to yield a marginal variance of $0.2$ and correlation length of $0.3$. Further details of the Gaussian random field are provided in the \textit{Supplementary Information}.
%
The residual used in the a posteriori neural operator inadequacy estimate in \eqref{eq:error_estimate} is
\begin{equation}
\begin{aligned}
\left\langle \mathcal{R}(m,u),v\right\rangle_{\mathcal V^\ast,\mathcal V}
:={}&
\left\langle \exp(m)\nabla u,\nabla v\right\rangle_{\Omega}
+
\left\langle u^3,v\right\rangle_{\Omega}
-
\left\langle f,v\right\rangle_{\Omega},
\qquad \forall v\in\mathcal V,
\end{aligned}
\label{eq:rd_residual}
\end{equation}
where
$\mathcal V=\{v\in H^1(\Omega):v=0\text{ on }\Gamma_b\}$,
$\langle\cdot,\cdot\rangle_{\mathcal V^\ast,\mathcal V}$ denotes the
duality pairing, and
$\langle a,b\rangle_{\Omega}$
denotes the $L^2(\Omega)$ inner product.

\subsection{Training data and neural operator architectures}
Independent realizations of the parameter field $m$ were sampled from
$\nu_{\mathcal M}$. For the comparisons in
Figure~\ref{fig:rd_pde}, vBNO is trained on
$N_{\rm train}=1024$ parameter--state pairs in $\mathcal D$, with
reference states computed using the finite element solution operator
$\mathcal F_h$.
For RI-BNO, $\mathcal D$ contains $N_{\rm train}=512$ FE solution
pairs, divided equally into $\mathcal D_{\rm pre}$ and
$\mathcal D_{\rm fit}$, with 256 pairs in each subset.
The preliminary operator is trained on $\mathcal D_{\rm pre}$.
Its predictions at 512 additional, independently sampled inputs,
distinct from those in $\mathcal D$, are used in
\eqref{eq:error_estimate} to generate $N_{\rm err}=512$
residual-informed observations in $\mathcal E$.
The inadequacy model is learned from $\mathcal E$ and held fixed
during final Bayesian training using only $\mathcal D_{\rm fit}$,
following Section~\ref{sec:algorithms}.
Thus, RI-BNO construction uses 512 full nonlinear FE solutions and
512 linearized variational solves, replacing half of
the nonlinear solution evaluations used by vBNO with lower-cost
linearized solves.

We consider two distinct neural operator architectures to assess whether residual-informed inadequacy modeling depends on the neural operator classes.
The first is a latent-subspace neural operator
\cite{CAO_jcp2023,ohippyflow,OLeary2022derivative,OLeary2024derivative,
cao2025derivative}. The input field is encoded into a reduced space using active
subspace \cite{constantine2014active}, the state is decoded by proper orthogonal decomposition (POD)
\cite{manzoni2016dimensionality}, and a stack of ResNet layers \cite{he2016deep} learns the map between the latent coordinates. This
simple and scalable construction replaces the full operator map by truncated
linear input and output representations. 
The discarded modes and resulting reduced-dimensional learning between input and output constitute a potential source of neural operator inadequacy.
The second architecture is DeepONet \cite{lu2021learning}, a well-established
neural operator framework motivated by the universal approximation theorem
for nonlinear operators \cite{chen1995universal}. Its branch network encodes
the input field $m$, while its trunk network learns spatial basis functions
evaluated at $\bs x$. Unlike the latent-subspace construction, DeepONet does
not rely on a prescribed reduced basis for the output field and is therefore less
directly constrained by a fixed reduced representation. It provides a more
expressive baseline for testing whether residual-informed learning remains
beneficial when the underlying neural operator already attains comparatively
high predictive accuracy.
Both architecture classes are illustrated in Figure~\ref{fig:rd_pde}(b), with additional implementation details provided in the \textit{Supplementary Information}. For this problem, the latent-subspace neural operator uses 20 active-subspace and POD modes with a 10-layer ResNet (width 50), while the DeepONet employs a branch network of depth 3 and width 40 together with a trunk network of depth 4 and width 50.

\subsection{Prediction, inadequacy representation, and data efficiency}
\begin{figure}[H]
    \centering
\begin{minipage}[t]{0.58\textwidth}
        \vspace{0pt} 
\noindent\textbf{a}\vspace{-0.3in}\\[0.0em]
\begingroup 
\setlength{\tabcolsep}{0.4pt} 
\fontsize{9.25}{11.1}\selectfont
\def\fw{0.175\textwidth}    
\def\cw{0.065\textwidth}    
\def\cwvar{0.065\textwidth}  

\begin{tabular}{c c c c c c c}
& & \multicolumn{2}{c}{Latent-subspace} & \multicolumn{2}{c}{DeepONet} & \\
\cmidrule(lr){3-4} \cmidrule(lr){5-6}
& & vBNO & RI-BNO & vBNO & RI-BNO & \\

\multirow{3}{*}[0.475in]{
\begin{tabular}{@{}c@{\hspace{0pt}}c@{}}
$m_1$ & \\
\includegraphics[width=\fw]{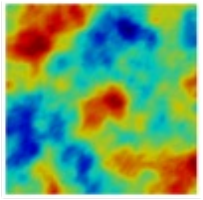} &
\hspace{0.02in}\includegraphics[width=0.075\textwidth]{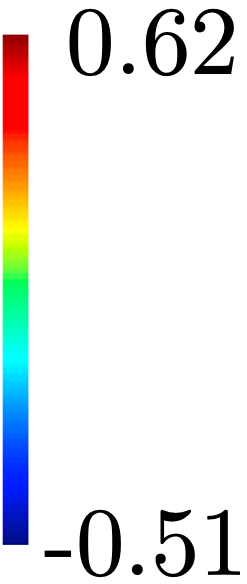} \\[1ex]
\shortstack{FE $u_h$} & \\
\includegraphics[width=\fw]{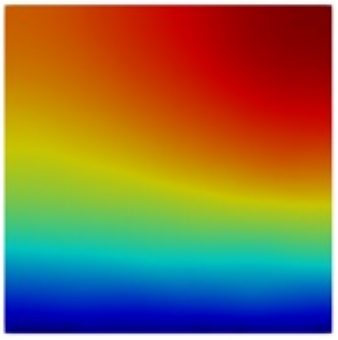} &
\hspace{-0.0in}\includegraphics[width=\cw]{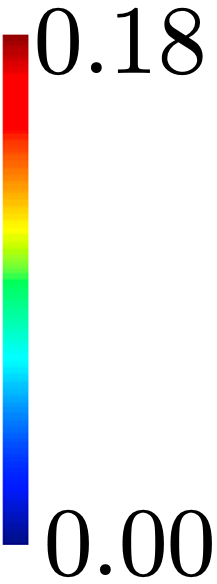}
\end{tabular}%
}
& \raisebox{0.13in}{\hspace{0.05in}\shortstack{State\\prediction\\$\overline{u}$}}
& \includegraphics[width=\fw]{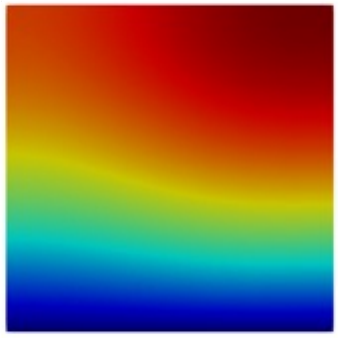}
& \includegraphics[width=\fw]{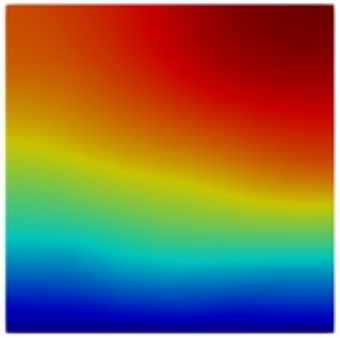}
& \includegraphics[width=\fw]{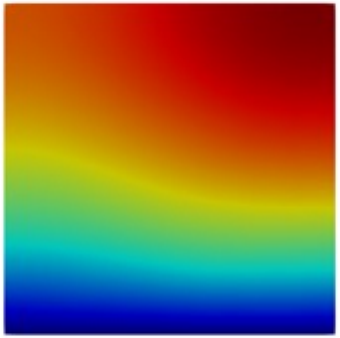}
& \includegraphics[width=\fw]{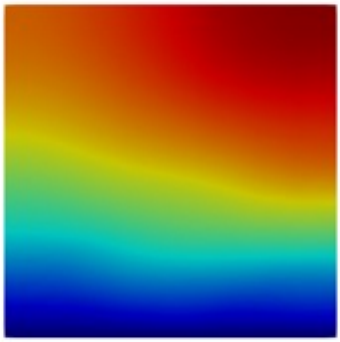}
& \hspace{0.025in}\includegraphics[width=\cw]{Figure_1/u_cont.png} \\

& \raisebox{0.16in}{\hspace{0.05in}\shortstack{Relative\\error\\$\varepsilon$}}
& \includegraphics[width=\fw]{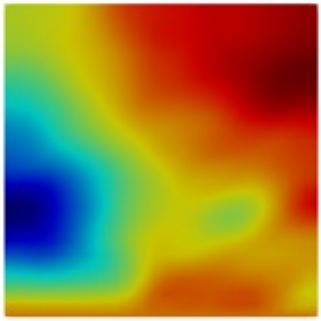}
& \includegraphics[width=\fw]{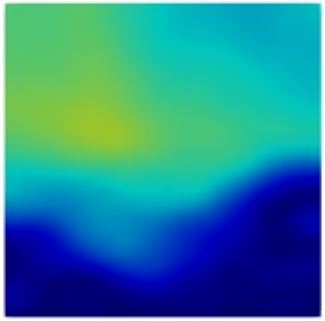}
& \includegraphics[width=\fw]{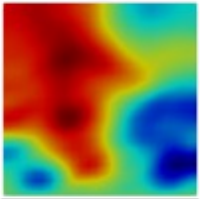}
& \includegraphics[width=\fw]{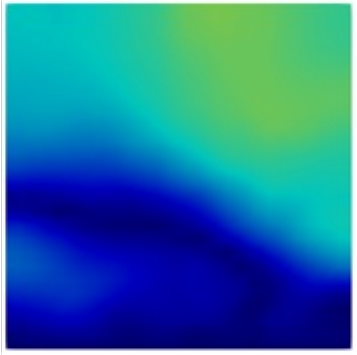}
& \hspace{0.025in}\includegraphics[width=\cw]{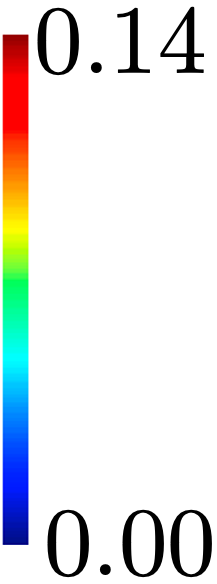} \\

& \raisebox{0.13in}{\hspace{0.05in}\shortstack{State\\variance\\$\sigma^2$}}
& \includegraphics[width=\fw]{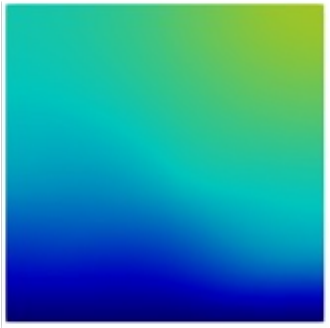}
& \includegraphics[width=\fw]{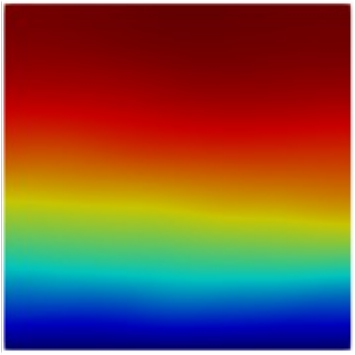}
& \includegraphics[width=\fw]{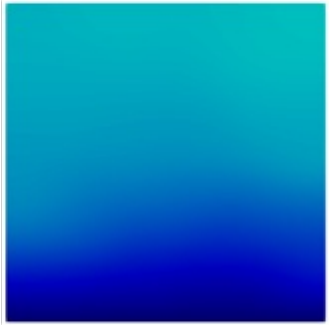}
& \includegraphics[width=\fw]{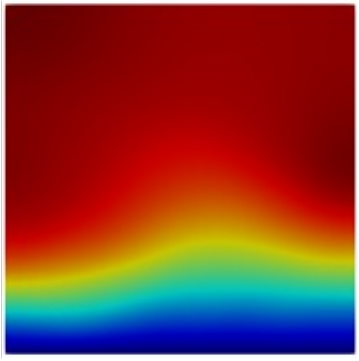}
& \hspace{0.025in}\includegraphics[width=\cwvar]{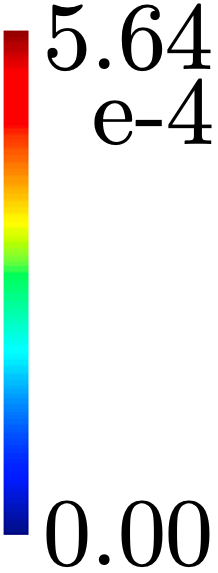} \\

\multirow{3}{*}[0.475in]{%
\raisebox{1in}{
\begin{tabular}{@{}c@{\hspace{2pt}}c@{}}
$m_2$ & \\
\includegraphics[width=\fw]{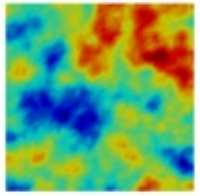} &
\hspace{-0.01in}\includegraphics[width=0.075\textwidth]{Figure_1/m_cont.png} \\[1ex]
\shortstack{FE $u_h$} & \\
\includegraphics[width=\fw]{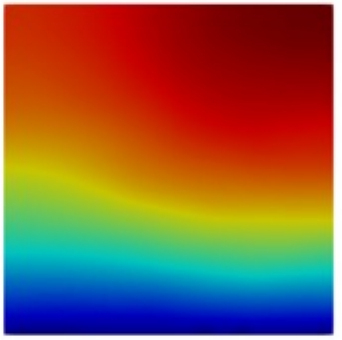} &
\hspace{-0.04in}\includegraphics[width=\cw]{Figure_1/u_cont.png}
\end{tabular}%
}%
}
& \raisebox{0.13in}{\hspace{0.05in}\shortstack{State\\prediction\\$\overline{u}$}}
& \includegraphics[width=\fw]{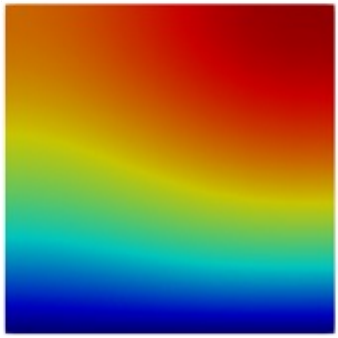}
& \includegraphics[width=\fw]{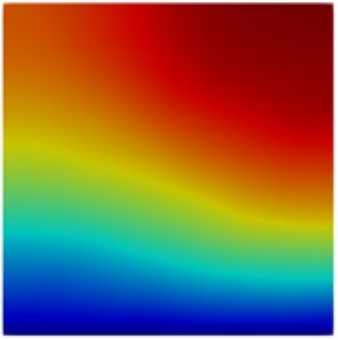}
& \includegraphics[width=\fw]{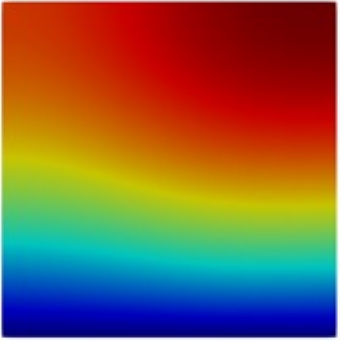}
& \includegraphics[width=\fw]{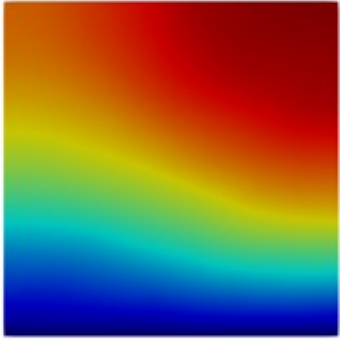}
& \hspace{0.025in}\includegraphics[width=\cw]{Figure_1/u_cont.png} \\

& \raisebox{0.16in}{\hspace{0.05in}\shortstack{Relative\\error\\$\varepsilon$}}
& \includegraphics[width=\fw]{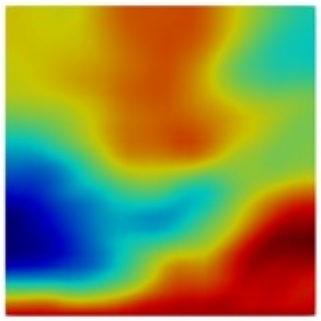}
& \includegraphics[width=\fw]{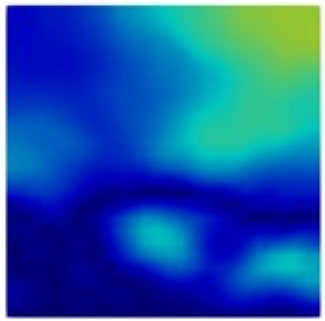}
& \includegraphics[width=\fw]{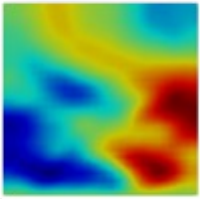}
& \includegraphics[width=\fw]{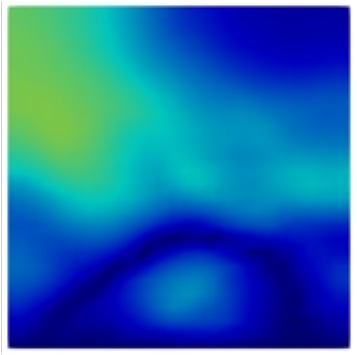}
& \hspace{0.025in}\includegraphics[width=\cw]{Figure_1/relative_error_cont.png} \\

& \raisebox{0.13in}{\hspace{0.05in}\shortstack{State\\variance\\$\sigma^2$}}
& \includegraphics[width=\fw]{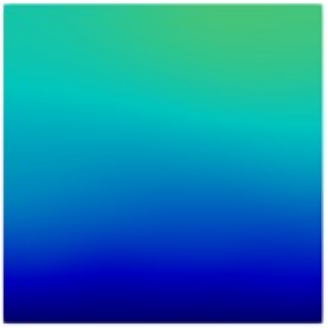}
& \includegraphics[width=\fw]{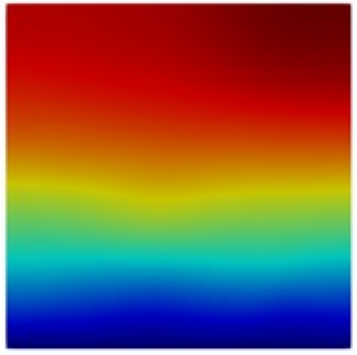}
& \includegraphics[width=\fw]{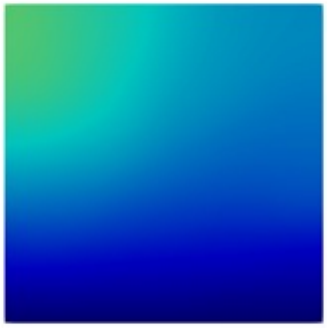}
& \includegraphics[width=\fw]{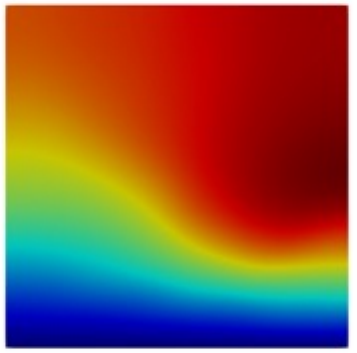}
& \hspace{0.025in}\includegraphics[width=\cwvar]{Figure_1/var_cont.png} \\
\end{tabular}
\endgroup
\vspace{0.0em} 
        
\noindent\large\hspace{-0.05in}\textbf{c}\vspace{-0.35in}\\[0.5em]
\begingroup 
\setlength{\tabcolsep}{4pt} 

\def\fw{0.18\linewidth} 
\def\cw{0.07\linewidth}

\fontsize{9.25}{11.1}\selectfont
\begin{tabular}{c c c c c}
\multicolumn{2}{c}{Latent-subspace} & & \multicolumn{2}{c}{DeepONet} \\
\cmidrule(lr){1-2} \cmidrule(lr){4-5}

\shortstack{\hspace*{-0.125in}Inadequacy\\\hspace*{-0.125in}mean} & \shortstack{\hspace*{-0.125in}Inadequacy\\\hspace*{-0.125in}variance} & & \shortstack{\hspace*{-0.125in}Inadequacy\\\hspace*{-0.125in}mean} & \shortstack{\hspace*{-0.125in}Inadequacy\\\hspace*{-0.125in}variance} \\[1ex]

\hspace*{0.05in}\includegraphics[width=\fw]{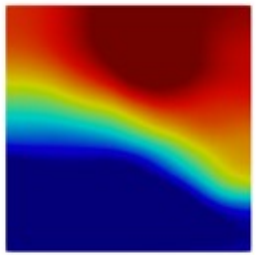}\hspace*{0.05in}\includegraphics[width=\cw]{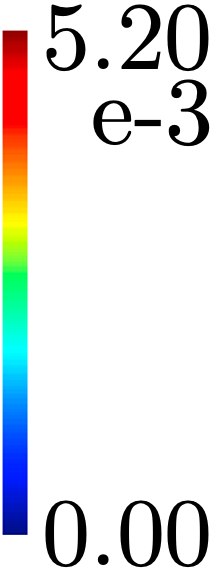} &
\hspace*{0.05in}\includegraphics[width=\fw]{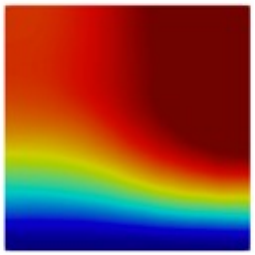}\hspace*{0.05in}\includegraphics[width=\cw]{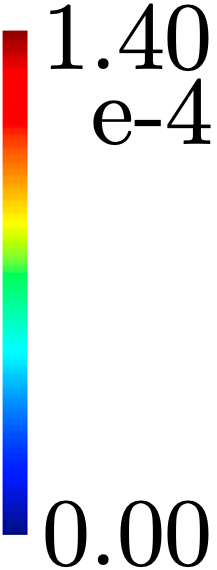} &
&
\hspace*{0.05in}\includegraphics[width=\fw]{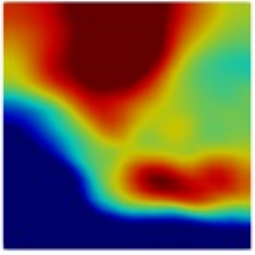}\hspace*{0.05in}\includegraphics[width=\cw]{Figure_1/gp_mean_cont.png} &
\hspace*{0.05in}\includegraphics[width=\fw]{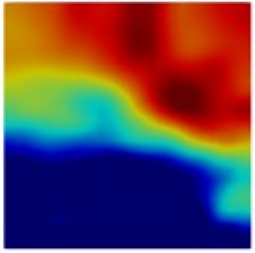}\hspace*{0.05in}\includegraphics[width=\cw]{Figure_1/gp_var_cont.png} \\

\end{tabular}
\endgroup
\end{minipage}%
    \hfill
    \begin{minipage}[t]{0.35\textwidth}
        \vspace{0pt} 
        
        \centering
\large
\hspace{-1.05in}\textbf{b}\\
\hspace*{0.575in}\includegraphics[trim={0.025in 0in 0.05in 0in}, clip, width=0.9\linewidth]{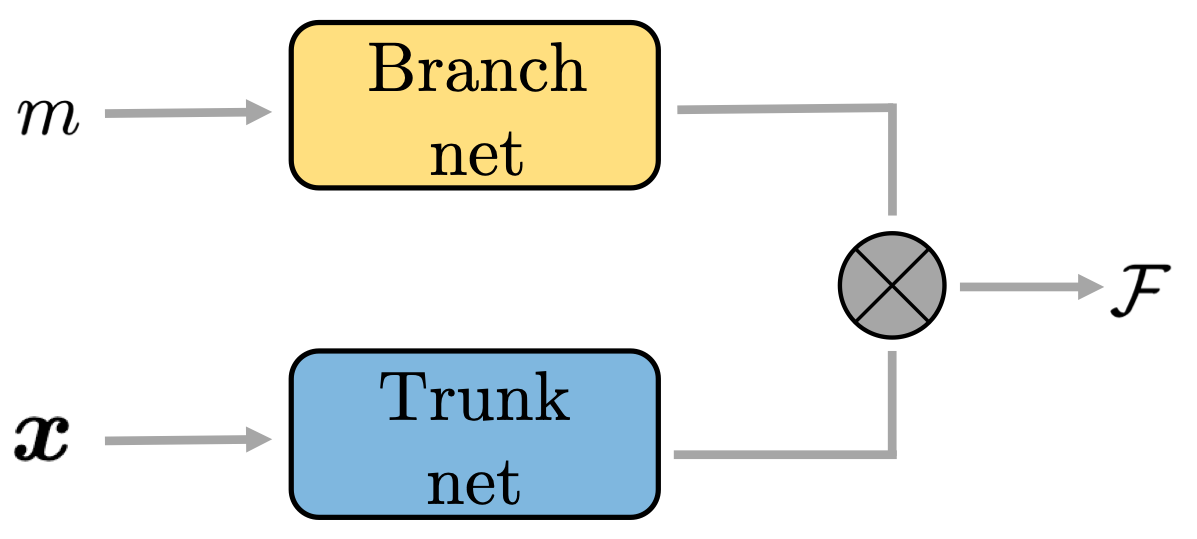}\vspace{0.25in}\\
\hspace*{0.575in}\includegraphics[trim={0in 0in 0in 0in}, clip, width=0.9\linewidth]{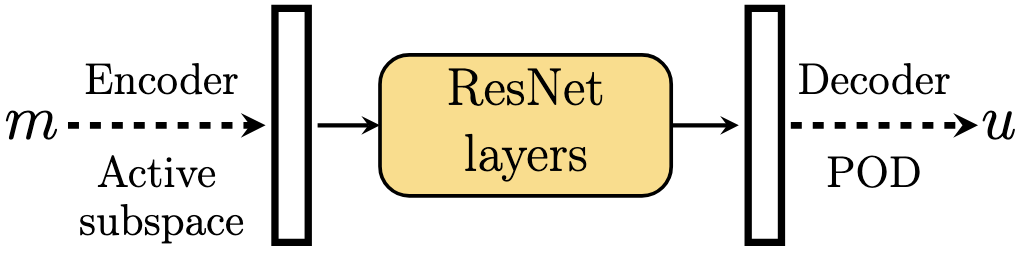}\vspace{0.05in}

\vspace{0.5em} 

\centering
\large
\hspace{-1.05in}\textbf{d}\vspace{-0.2in}\\
\vspace{0.15in}
\hspace*{0.7in}\includegraphics[width=0.75\linewidth]{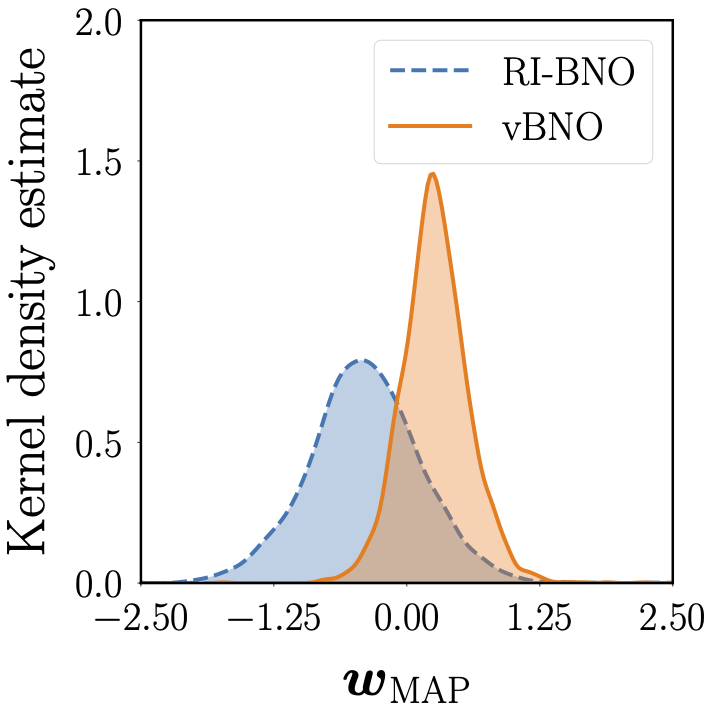}\vspace{0.15in}\\
\hspace*{0.575in}\includegraphics[width=0.8\linewidth]{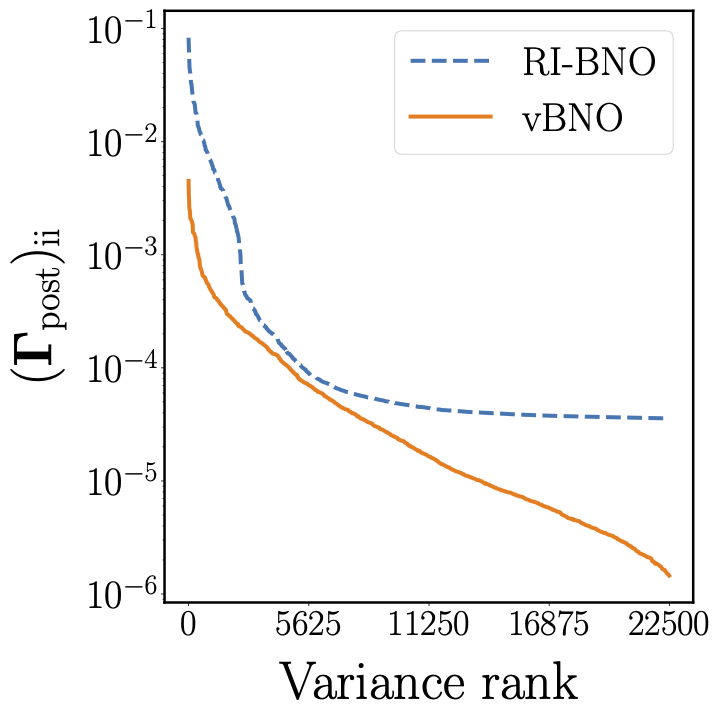}
    \end{minipage}
\caption{
\textbf{Residual-informed Bayesian neural operators for learning solution operator of a nonlinear reaction--diffusion PDE.}
Panel (a) shows predictions for two unseen in-distribution parameter fields,
$m_1$ and $m_2$. 
For each input, the finite element (FE) state \(u_h\) is compared with the state predictions of vBNO,
\(\overline{u}_{\rm vBNO}=\mathcal F_{\bs w_{\rm MAP}}(m)\),
and RI-BNO,
\(\overline{u}_{\rm RI}=\mathcal F_{\bs w_{\rm MAP}}(m)+\mu_\eta(m)\),
using the
latent-subspace and DeepONet architectures.
The pointwise relative error
$\varepsilon_{\rm vBNO}=\left|u_h-\overline{u}_{\rm vBNO}\right| / \left|u_h\right|$
and
$\varepsilon_{\rm RI}=\left|u_h-\overline{u}_{\rm RI}\right| / \left|u_h\right|$,
and posterior predictive state variance 
$\sigma_{\rm vBNO}^2 =\mathbb{V}_{\pi_{post}(\bs w\mid\mathcal D)}
\!\left[\mathcal F_{\bs w}(m)\right]$
and
$\sigma_{\rm RI}^2=
\mathbb{V}_{\pi_{\rm post}(\bs w |\mathcal D, \mathcal E)}\!\left[\mathcal F_{\bs w}(m)\right]+ \bs C_{\eta}$ are also presented.
Panel (b) illustrates the DeepONet architecture, comprising branch and trunk networks, and the latent-subspace architecture, comprising an active-subspace input encoder, a ResNet map in latent coordinates and a POD state decoder.
Panel (c) shows the inadequacy mean \(\mu_\eta\) and pointwise marginal variance \(\bs{C}_\eta\) of the sparse Gaussian process for each architecture. The mean provides the residual-informed correction, whereas the marginal variance quantifies uncertainty in that correction.
Panel (d) compares, for the latent-subspace architecture, kernel density estimates of the components of the MAP weight vectors \(\bs w_{\rm MAP}\) between vBNO and RI-BNO, together with the posterior marginal variances \((\bs\Gamma_{\rm post})_{\rm ii}\) ordered by their variance rank. Their differences illustrate that the inadequacy-aware likelihood modifies the posterior of the weights in RI-BNO compared to vBNO.
}
    \label{fig:rd_pde}
\end{figure}

Figure~\ref{fig:rd_pde}(a) compares the two architecture classes for two unseen in-distribution test samples. For each input field $m$, the finite element reference state $u_h$ is shown alongside the point prediction of the vBNO and RI-BNO computed using $w_{\rm MAP}$ and mean of the learned inadequacy $\mu_\eta$, their relative errors, and their posterior predictive state variances. 
For vBNO, DeepONet is more accurate than the latent-subspace operator, although both exhibit spatially structured prediction errors. RI-BNO substantially reduces these errors for both architectures, demonstrating that residual-informed inadequacy modeling is not tied to a particular neural operator class.
Figure~\ref{fig:rd_pde}(c) shows the learned inadequacy mean $\mu_\eta$ and marginal variance associated with $\bs C_\eta$. 
The mean captures the spatial structure of the systematic correction, whereas the marginal variance quantifies uncertainty in that correction. Consistent with Bayesian discrepancy modeling \cite{kennedy2001bayesian}, omitting structured inadequacy can produce overconfident predictions because weight uncertainty alone does not represent systematic operator error. RI-BNO instead propagates uncertainty from both the re-inferred weight posterior and the learned inadequacy, providing a more complete representation of predictive uncertainty, particularly where the residual-informed correction is itself uncertain.
Figure~\ref{fig:rd_pde}(d) illustrates differences in fitted weights and posterior variances between the two inference workflows. Because the inadequacy mean and covariance enter the likelihood, they modify both the data misfit and its uncertainty, leading to distinct MAP weights and posterior covariance.

For the latent-subspace architecture, the shifted distributions of the MAP values and the altered spectra of the posterior marginal variances for the weight parameters illustrate this re-inference. Thus, RI-BNO modifies both the learned neural operator and its posterior uncertainty, in addition to supplying the residual-informed state correction.

\begin{figure}[h]
    \centering
    \includegraphics[width=0.4\linewidth]{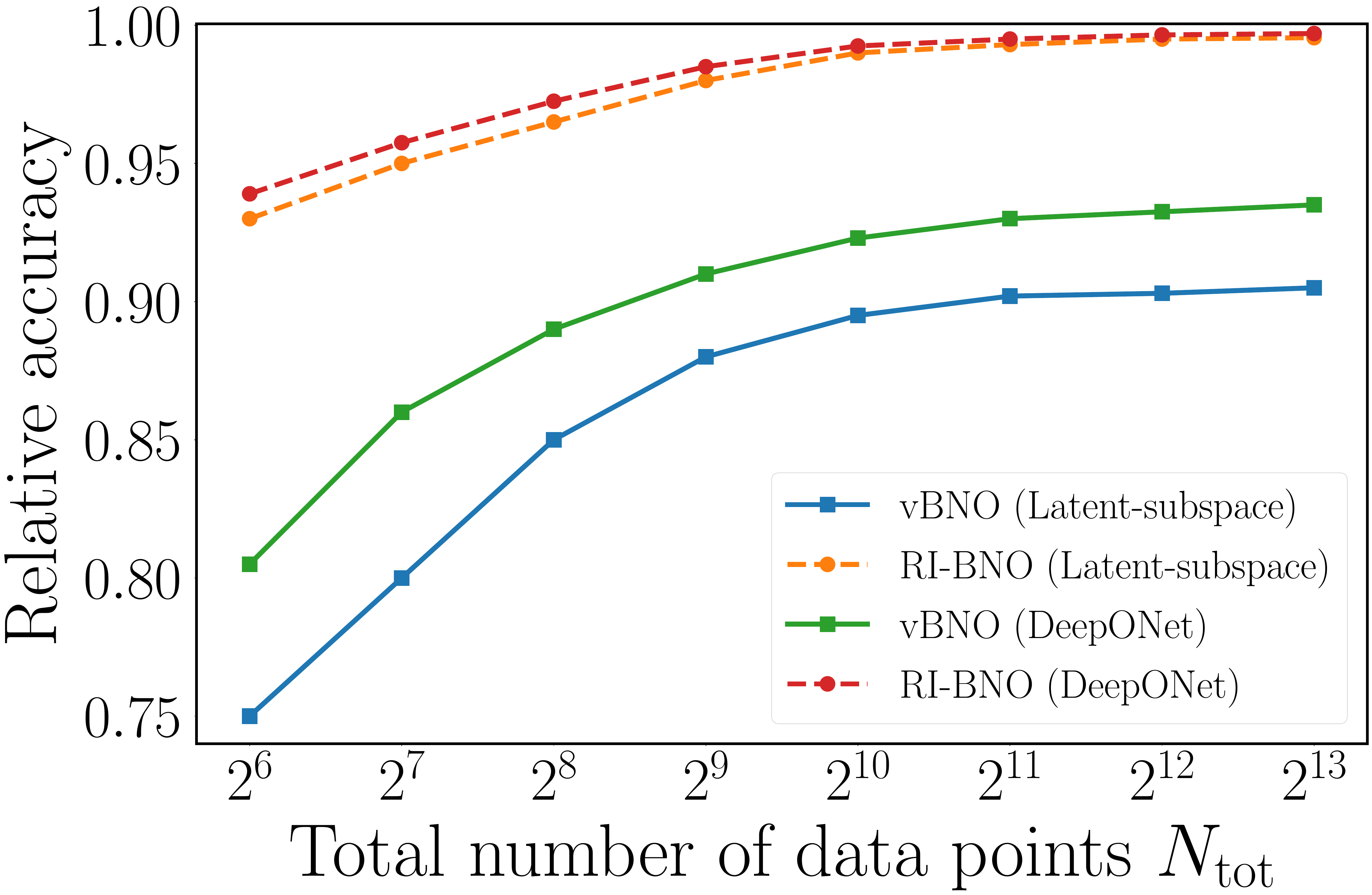} \hspace{0.5in}
    \includegraphics[width=0.4\linewidth]{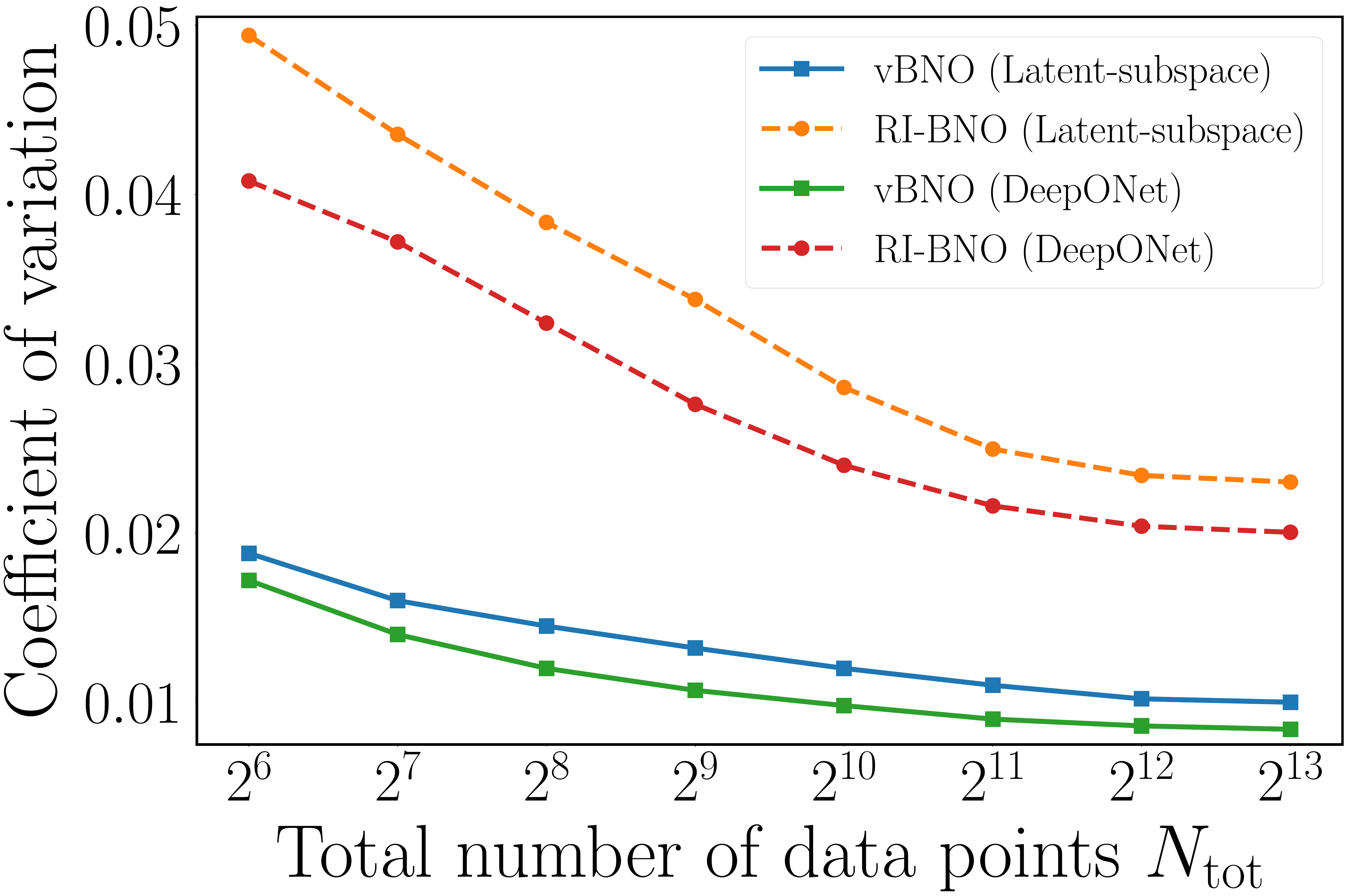}\vspace{-0.025in}\\
    \hspace{0.2in}(a) \hspace{2.75in} (b) \vspace{-0.125in}\\
    \caption{\textbf{Predictive accuracy and uncertainty of vBNO and RI-BNO using latent-subspace and DeepONet architectures}: (a) Relative accuracy versus the total number of data points $N_{\rm{tot}}$ (b) Coefficient of Variation (CV), measuring relative predictive spread, versus $N_{\rm{tot}}$.}
    \label{fig:rd_metrics}
\end{figure}

Figure~\ref{fig:rd_metrics} examines how predictive accuracy and uncertainty vary with the total observation budget. We define the relative accuracy and coefficient of variation (CV) as
\begin{equation}
\begin{aligned}
\text{Relative accuracy}
&=1-\mathbb E_{m\sim\nu_{\mathcal M}}
\left[
\frac{\|u_h(m)-\bar u(m)\|_{L^2(\Omega)}}
     {\|u_h(m)\|_{L^2(\Omega)}}
\right],\\
\text{Coefficient of variation}
&=\mathbb E_{m\sim\nu_{\mathcal M}}
\left[
\frac{\|\sigma(m)\|_{L^2(\Omega)}}
     {\|\bar u(m)\|_{L^2(\Omega)}}
\right].
\end{aligned}
\label{eq:accuracy}
\end{equation}
where $\overline{u} \in \{\overline{u}_{vBNO}, \overline{u}_{RI}\}$ and 
$\sigma \in \{\sigma_{vBNO}, \sigma_{RI}\}$ denote the pointwise posterior predictive standard-deviation field.
Unit relative accuracy corresponds to zero mean relative error with respect to the finite element reference.
%
%
%
We denote the total number of data points by
$N_{\rm tot}=N_{\rm train}+N_{\rm err}$,
where $N_{\rm train}$ is the number of parameter-state
pairs in training data $\mathcal D$ and $N_{\rm err}$ is the number of
residual-informed observations in $\mathcal E$. 
For a fixed $N_{\rm tot}$, the vBNO uses
$N_{\rm train}=N_{\rm tot}$,
whereas the RI-BNO uses equal numbers of parameter-state and
residual-informed observations,
$N_{\rm train}=N_{\rm err}=N_{\rm tot}/2$.
Across all total number of data points, the DeepONet vBNO consistently outperforms the latent-subspace architecture, demonstrating the strong standalone performance of DeepONet and remaining consistent with its more flexible output representation. 
As the training set grows, however, the latent-subspace and DeepONet vBNO models approach architecture-dependent accuracy ceilings of approximately \(0.905\) and \(0.935\), respectively, even as \(N_{\rm train}\) increases to \(2^{13}\).
In contrast, residual-informed learning changes the scaling behavior. With only $2^6$ total observations, the RI-BNO already achieves accuracies of approximately $0.93$ and $0.94$ for the latent-subspace and DeepONet architectures, respectively. 
For both architecture classes, RI-BNO reaches approximately \(0.99\) accuracy near \(N_{\mathrm{tot}}=2^{10}\), corresponding to a mean relative error of about \(1\%\) using 512 nonlinear PDE solution pairs and 512 residual-informed observations. At the largest budget, \(N_{\mathrm{tot}}=2^{13}\), accuracy reaches approximately \(0.997\), corresponding to a mean relative error of about \(0.3\%\) with 4,096 observations of each type.
Figure~\ref{fig:rd_metrics}(b) shows that the coefficient of variation (CV) decreases with increasing observation budget for all models, reflecting contraction of posterior uncertainty as additional observations are introduced. RI-BNO retains a higher CV because it explicitly propagates uncertainty in the learned inadequacy in addition to uncertainty in the neural operator weights.

%
To assess residual-informed observation generation, we also learned
the inadequacy model from direct FE prediction errors at the same
additional inputs, using a common preliminary operator and the same
GP and final Bayesian fitting procedures (see \textit{Supplementary Information}).
For the latent-subspace architecture, residual-informed observations
reduced data generation time by approximately $45\%$ and
total construction time by approximately $25\%$. Across both
architectures, both constructions achieved mean relative test errors
below $2\%$, with lower errors observed for RI-BNO. The final Bayesian
inference further reduced test error relative to correcting the preliminary
prediction with the GP mean alone.

\section{Example 2: Glioma growth prediction and MRI-informed inversion}\label{sec:tumor}
We next consider a time dependent phase field model of glioma growth to determine whether neural operator inadequacy in forward tumor prediction propagates into inference of patient-specific parameter fields from Magnetic Resonance Imaging (MRI).

\subsection{Phase field model of glioma growth}
The multiphase model describes the spatiotemporal evolution of the tumor and nutrient volume fractions, \(\phi_T(\bs x,t)\) and \(\phi_N(\bs x,t)\), in the brain domain \(\Omega\) and time interval \((0,t_F]\) \cite{lima2015analysis,lima2016selection,faghihi2020coupled,oden2010general}:
\begin{align}
\frac{\partial \phi_T(\bs x,t)}{\partial t}
&=
G\,\phi_T(1-\phi_T)\phi_N
+
\nabla\cdot
\left[
D\,\phi_T^2(1-\phi_T)^2\nabla\mu_T
\right],
&& (\bs x,t)\in\Omega\times(0,t_F],
\nonumber\\
\mu_T(\bs x,t)
&=
2K \phi_T(2\phi_T^2-3\phi_T+1)
-
H^2\nabla \cdot (\nabla\phi_T),
&& (\bs x,t)\in\Omega\times(0,t_F],
\nonumber\\
\frac{\partial \phi_N(\bs x,t)}{\partial t}
&=
-R\phi_T\phi_N
+ \nabla\cdot
\left(
M\nabla\phi_N
\right),
&& (\bs x,t)\in\Omega\times(0,t_F],
\nonumber\\
\nabla\phi_T\cdot\bs n
&=0,
\quad
\nabla\mu_T\cdot\bs n=0, 
\quad \phi_N = 0.5
&& (\bs x,t)\in\Gamma\times(0,t_F],
\nonumber\\
\phi_T (\bs x, 0) &=\phi_T^0,
\quad
\phi_N (\bs x, 0) =\phi_N^0,
&& \bs x \in\Omega
\label{eq:governing_tumor}
\end{align}
where $\Gamma=\partial\Omega$ and $\bs n$ is the outward unit normal.
Here, the spatially varying mobility \(D(\bs x)\) and proliferation rate \(G(\bs x)\) control tumor invasion and growth, respectively, and constitute the two physical parameters considered in this study. The remaining coefficients govern nutrient transport and consumption (\(M\) and \(R\)) and phase separation and diffuse-interface thickness (\(K\) and \(H\)); their fixed values are provided in the \textit{Supplementary Information}.
For FE solution, we used a first-order, semi-implicit, unconditionally energy-stable time integration scheme \cite{faghihi2020coupled}. Generating a full tumor growth trajectory requires nonlinear FE solves at successive time steps, making repeated forward evaluations for training and Bayesian inversion computationally demanding.

The neural operator maps $m=(\log D,\log G)$ to the terminal tumor $\phi_T(\bs x,t_F)$. Let $u$ denote this field, and let $u^-$ and $n$ denote approximations to $\phi_T(\bs x,t_F-\Delta t)$ and $\phi_N(\bs x,t_F)$, respectively. For residual evaluation of time dependent PDEs, these auxiliary fields are supplied by preliminary neural operator predictions, using models trained entirely on $\mathcal D_{\rm pre}$. The auxiliary fields are held fixed when linearizing the terminal residual. Implementation details are provided in the \textit{Supplementary Information}.
Suppressing the auxiliary-field arguments ,the terminal
residual used in \eqref{eq:error_estimate} is
\begin{equation}
\begin{aligned}
\left\langle
\mathcal R_T(m,u),v
\right\rangle_{\mathcal V^\ast,\mathcal V}
:={}&
\left\langle
\frac{u-u^-}{\Delta t},v
\right\rangle_{\Omega}
-
\left\langle
G\,u(1-u)n,v
\right\rangle_{\Omega}
\\
&+
\left\langle
D\,u^2(1-u)^2\nabla\mu_T(u),\nabla v
\right\rangle_{\Omega},
\qquad \forall v\in\mathcal V = H^1(\Omega) .
\end{aligned}
\label{eq:tumor_residual}
\end{equation}
Linearizing $\mathcal R_T$ with respect to the terminal tumor field
about the neural operator prediction $u_{\bs w}$, gives the residual-informed error estimate:
\begin{equation}
\left\langle
\delta_u\mathcal R_T(m,u_{\bs w})[e_{\mathrm{est}}],v
\right\rangle_{\mathcal V^\ast,\mathcal V}
=
-
\left\langle
\mathcal R_T(m,u_{\bs w}),v
\right\rangle_{\mathcal V^\ast,\mathcal V},
\qquad \forall v\in\mathcal V.
\label{eq:tumor_error_estimate}
\end{equation}
\subsection{Forward prediction of synthetic glioma growth}
\begin{figure}[H] 
    \centering
    \hspace{-6in}\large{\textbf{a}}\vspace{-0.13in}\\
    \normalsize{tumor evolution $\phi_T(\bs x,t)$}\\[0.025in]
    \begin{tabular}{@{}c c c c c c c@{}}
        
        \includegraphics[width=0.11\linewidth]{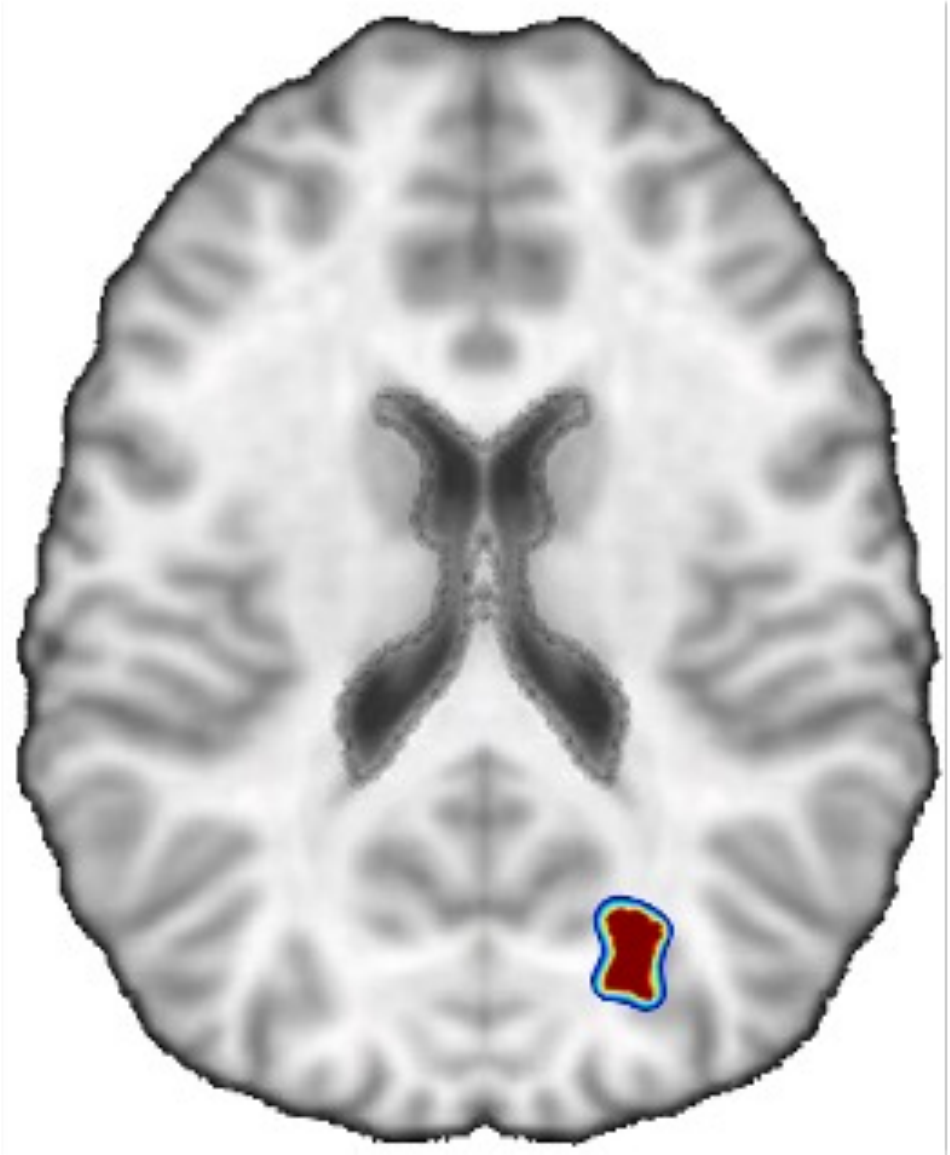} &
        \includegraphics[width=0.11\linewidth]{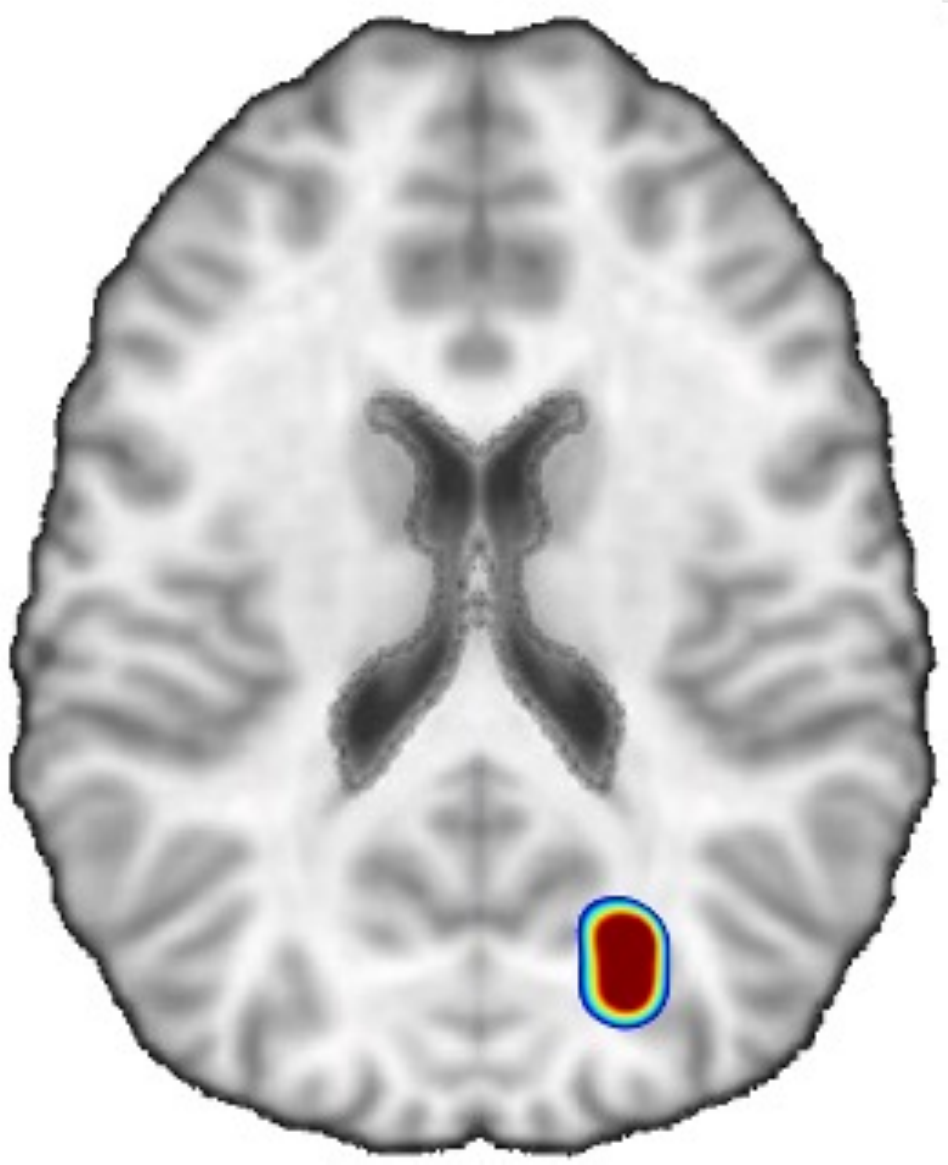} &
        \includegraphics[width=0.11\linewidth]{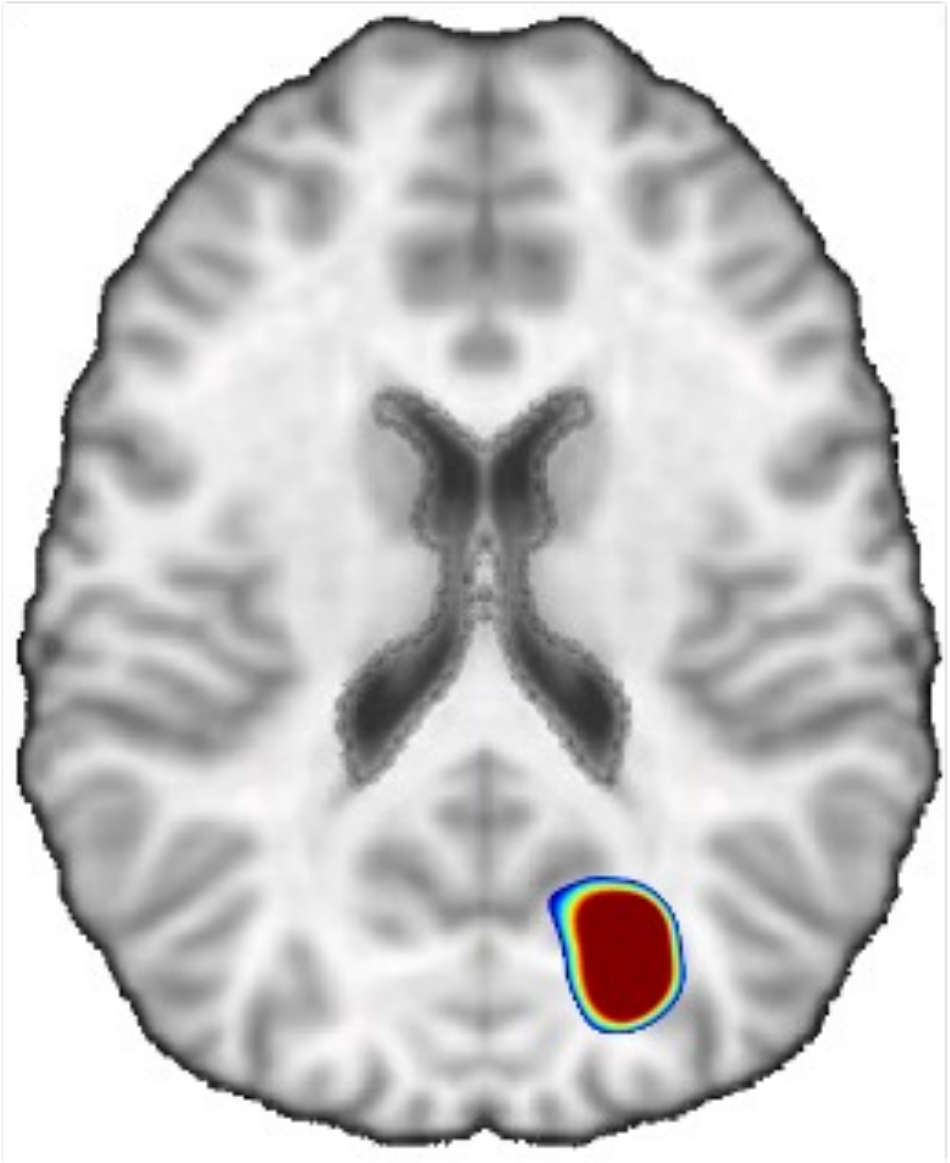} &
        \includegraphics[width=0.11\linewidth]{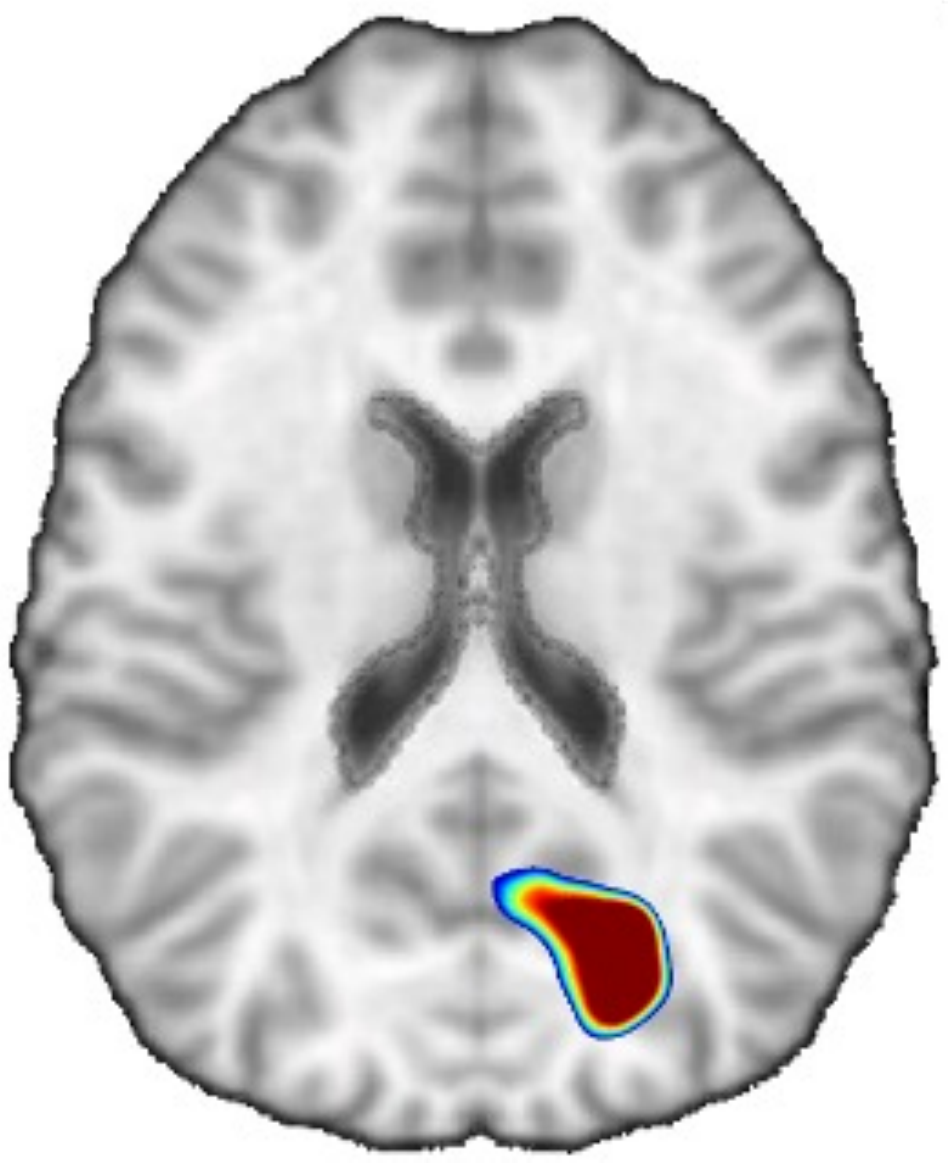} &
        \includegraphics[width=0.11\linewidth]{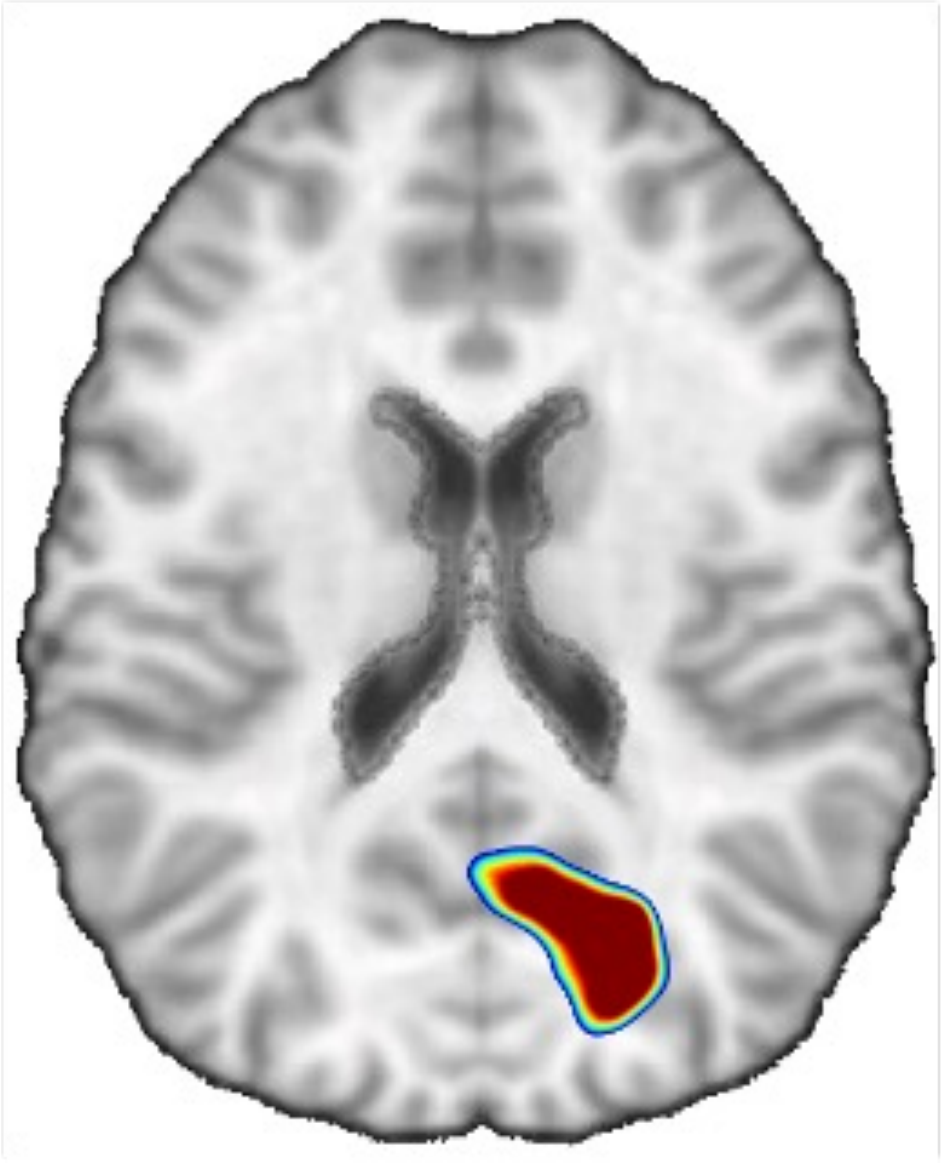} &
        \includegraphics[width=0.11\linewidth]{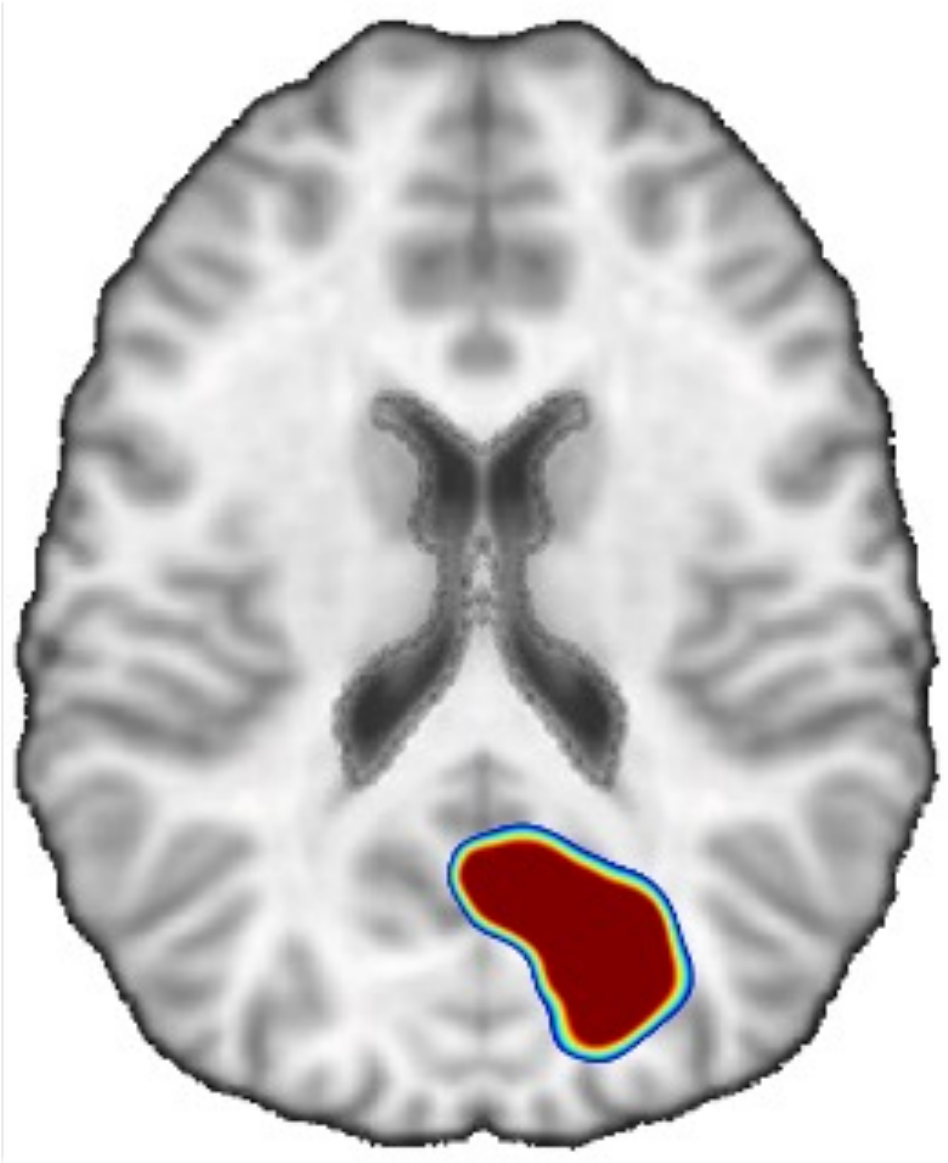} &
        \hspace{-0.05in}\includegraphics[width=0.021\linewidth]{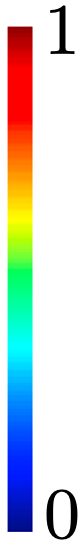} \vspace{-0.075in}\\

        \footnotesize $t=0$ day & 
        \footnotesize $t=50$ day & 
        \footnotesize $t=100$ day & 
        \footnotesize $t=150$ day & 
        \footnotesize $t=200$ day & 
        \footnotesize $t=250$ day & 
        \\
    \end{tabular}

    \vspace{0.125in} 
    \hspace{-6in}\large{\textbf{b}}\vspace{-0.4in}\\
    \footnotesize
    \begin{minipage}[t]{0.4\textwidth}
        \vspace{0pt} 
        \centering
        \footnotesize
        \setlength{\tabcolsep}{1pt}
        \begin{tabular}{c c c c}
         & $m_1$ & $m_2$ & $m_3$ \\
        \raisebox{0.5cm}{{\shortstack{Finite \\ element \\ $u_h$}}} & 
        \includegraphics[width = 0.23\linewidth]{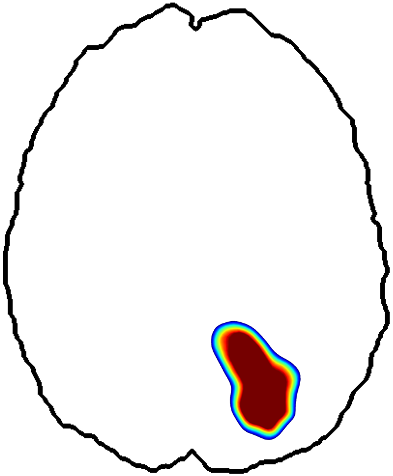} &
        \includegraphics[width = 0.23\linewidth]{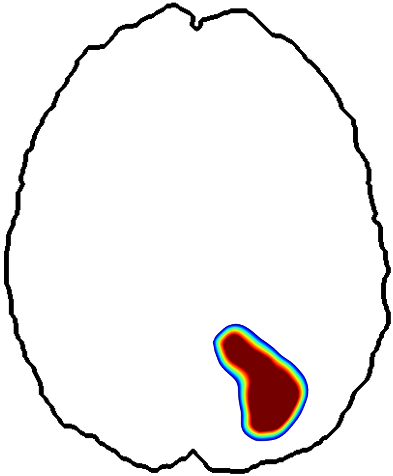} &
        \includegraphics[width = 0.23\linewidth]{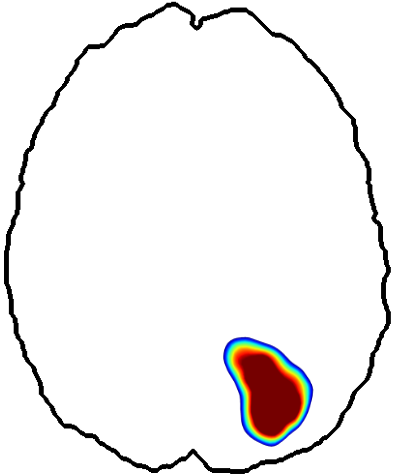} \\
        \raisebox{0.5cm}{{\shortstack{State \\ prediction\\ $\overline{u}_{\mathrm{vBNO}}$}}}
        &
        \includegraphics[width = 0.23\linewidth]{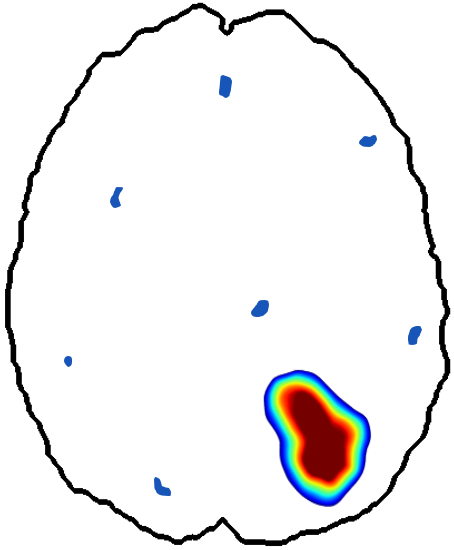} &
        \includegraphics[width = 0.23\linewidth]{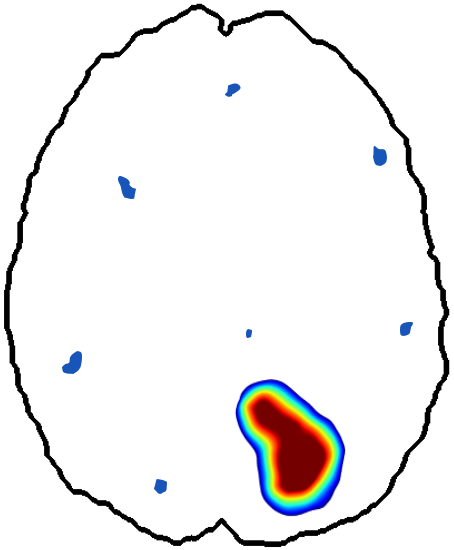} &
        \includegraphics[width = 0.23\linewidth]{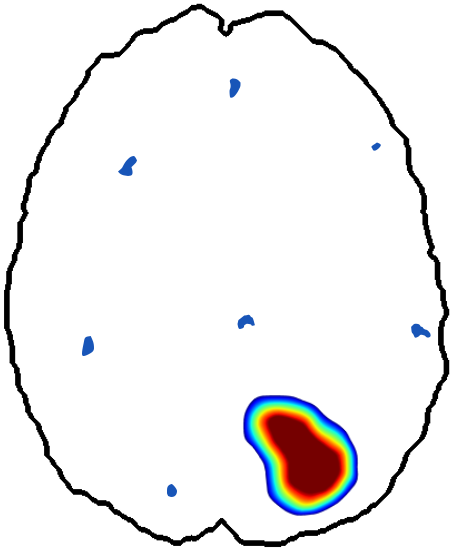} \\
        \raisebox{0.5cm}{{\shortstack{State \\ prediction\\ $\overline{u}_{\mathrm{RI}}$}}} &
        \includegraphics[width = 0.23\linewidth]{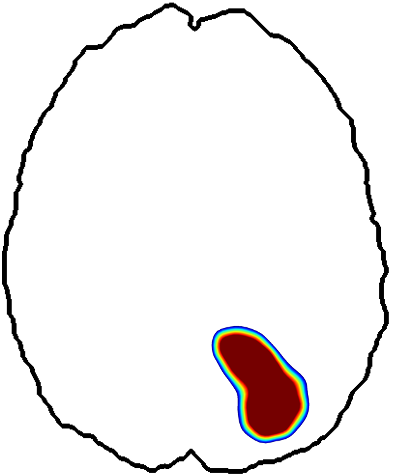} &
        \includegraphics[width = 0.23\linewidth]{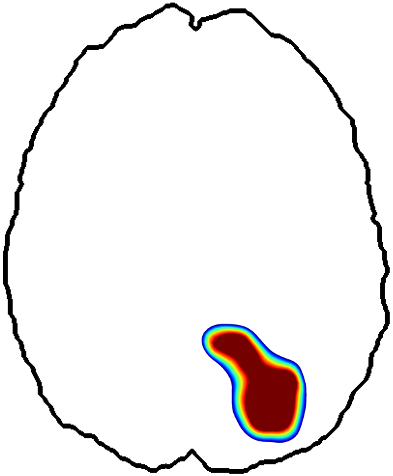} &
        \includegraphics[width = 0.23\linewidth]{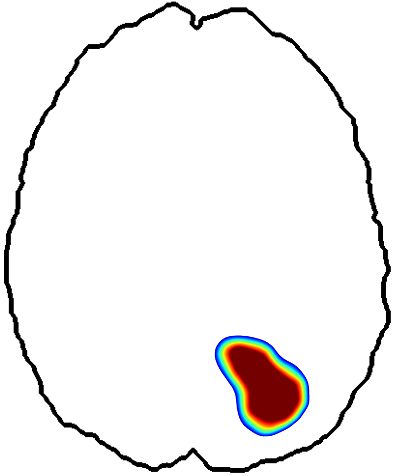} \\
        \end{tabular}
    \end{minipage}
    \hspace{0.035in}
    \begin{minipage}[t]{0.22\textwidth}
        \vspace{0pt} 
        \centering            
        \includegraphics[width=0.995\linewidth]{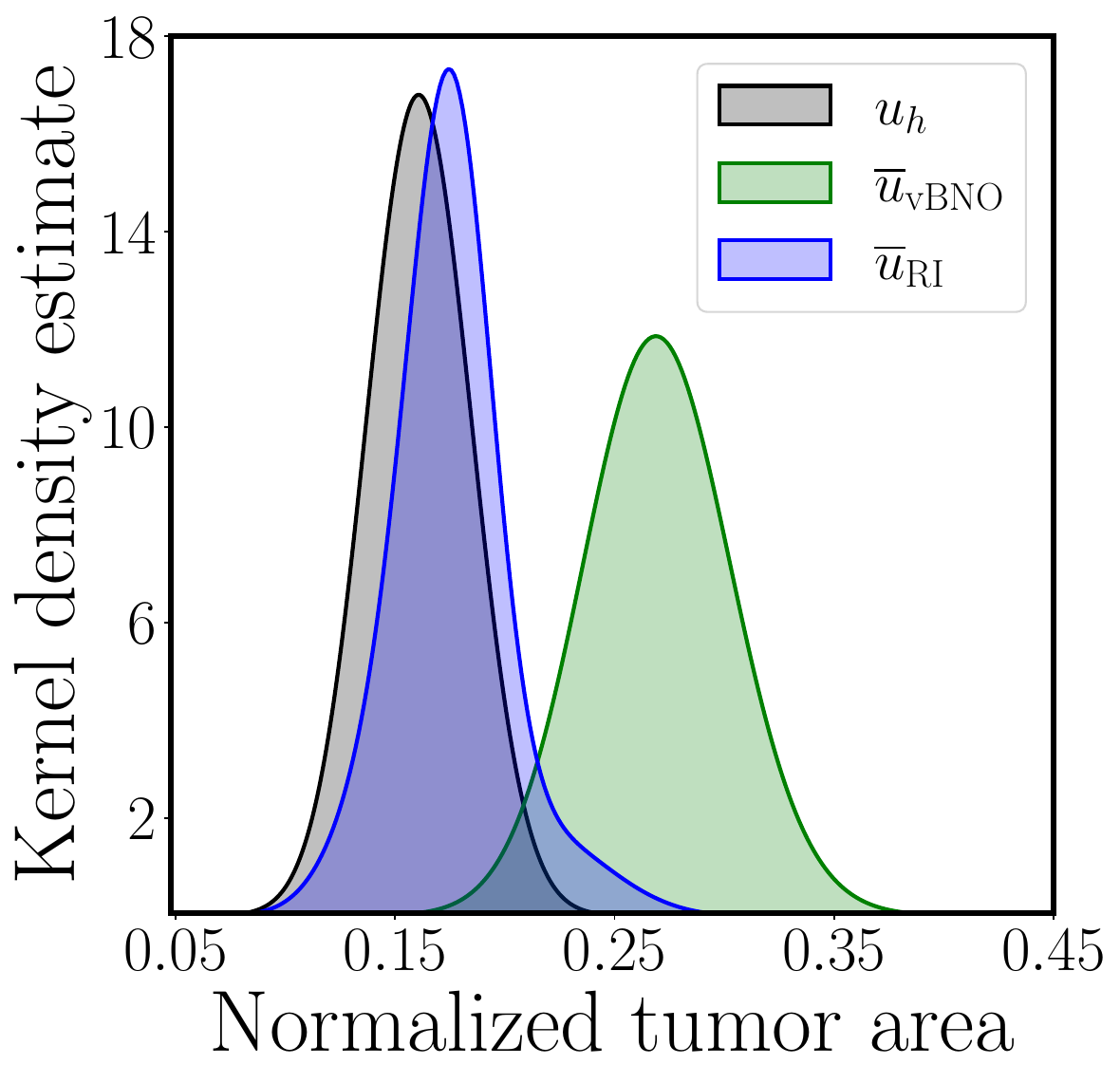}\\
        \hspace{-0.07in}\includegraphics[width=0.99\linewidth]{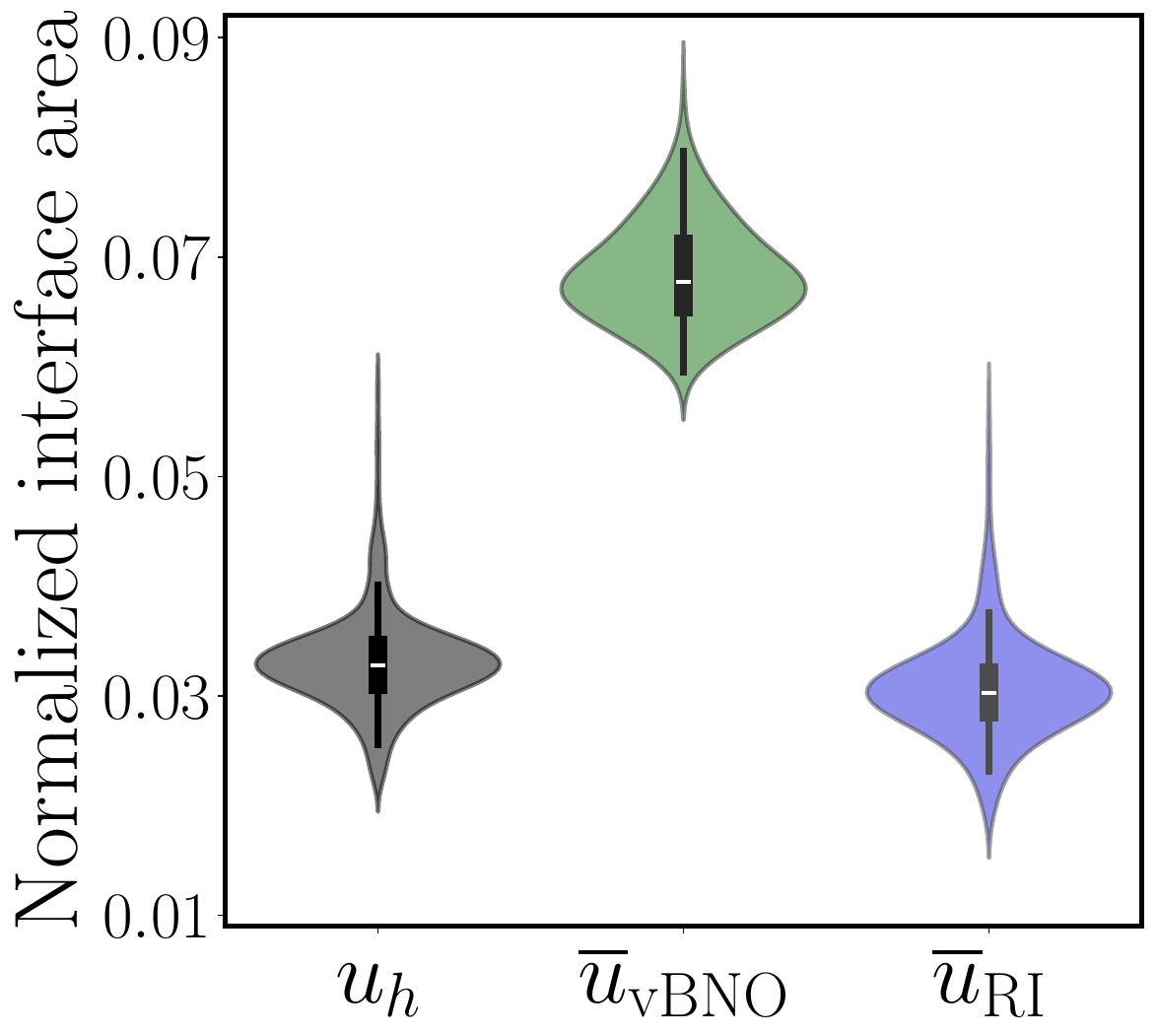}\\
    \end{minipage}
    \hspace{0.15in}
    \begin{minipage}[t]{0.22\textwidth}
        \vspace{0pt} 
        \footnotesize
        \begin{tabular}{c c}
        &\vspace{-0.07in}\\
        \hspace{-1.1in}\large{\textbf{c}} &\vspace{-0.12in}\\
        {\shortstack{Inadequacy\\mean}} & \vspace{0in}\\
        \includegraphics[width=0.48\linewidth]{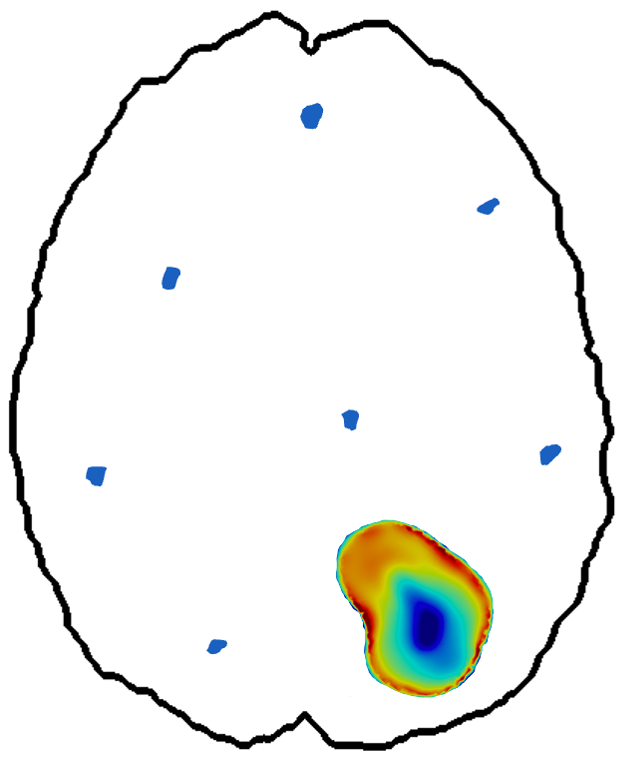} &
        \hspace{-0.2in}\includegraphics[width=0.19\linewidth]{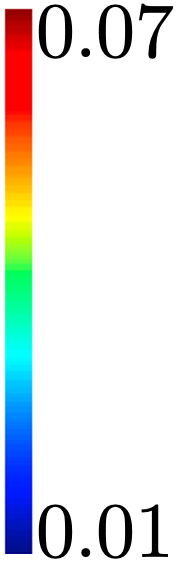}
        \vspace{0.1in}\\
        {\shortstack{Inadequacy\\variance}} & \vspace{0in}\\
        \includegraphics[width=0.48\linewidth]{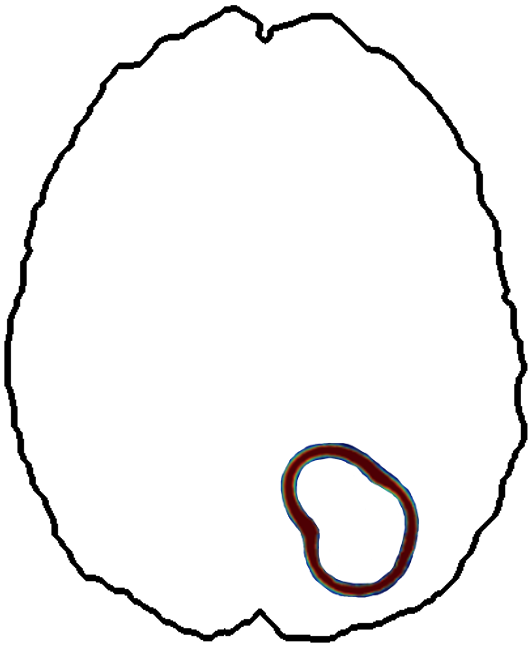} &
        \hspace{-0.15in}\includegraphics[width=0.2\linewidth]{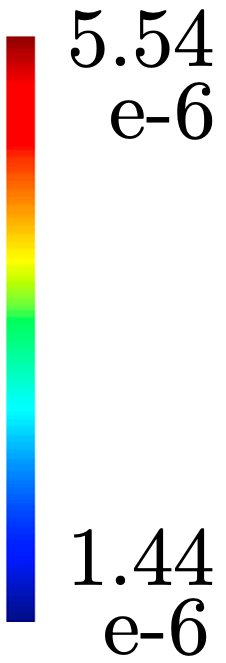}
        \vspace{0.0in}\\
        \end{tabular}
    \end{minipage}
    \vspace{0in} 
    \begin{minipage}[t]{0.7\textwidth}
    \hspace{-0.9in}\large{\textbf{d}}\vspace{-0.4in}\\
        \vspace{0in}\\
        \begin{tabular}{c c}
    \hspace{-1.75in}\footnotesize{vBNO}
    \hspace{1.05in}\footnotesize{RI-BNO} & \vspace{0.01in}\\
    
    \hspace{-0.7in}
    \includegraphics[
        trim={0.0in 0.0in 0.0in 0.0in},
        clip,
        width=0.95\linewidth
    ]{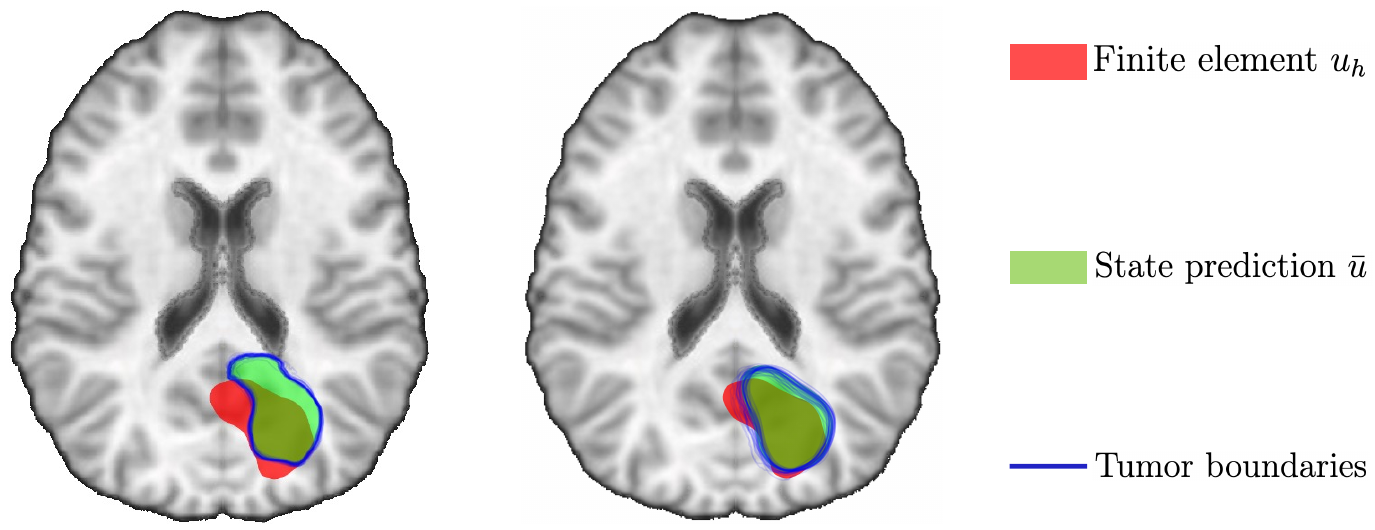}
    &
    
    \hspace{0.0in}
    \raisebox{0.125in}{%
        \hspace{-0.1in}\includegraphics[
            width=0.25\linewidth
        ]{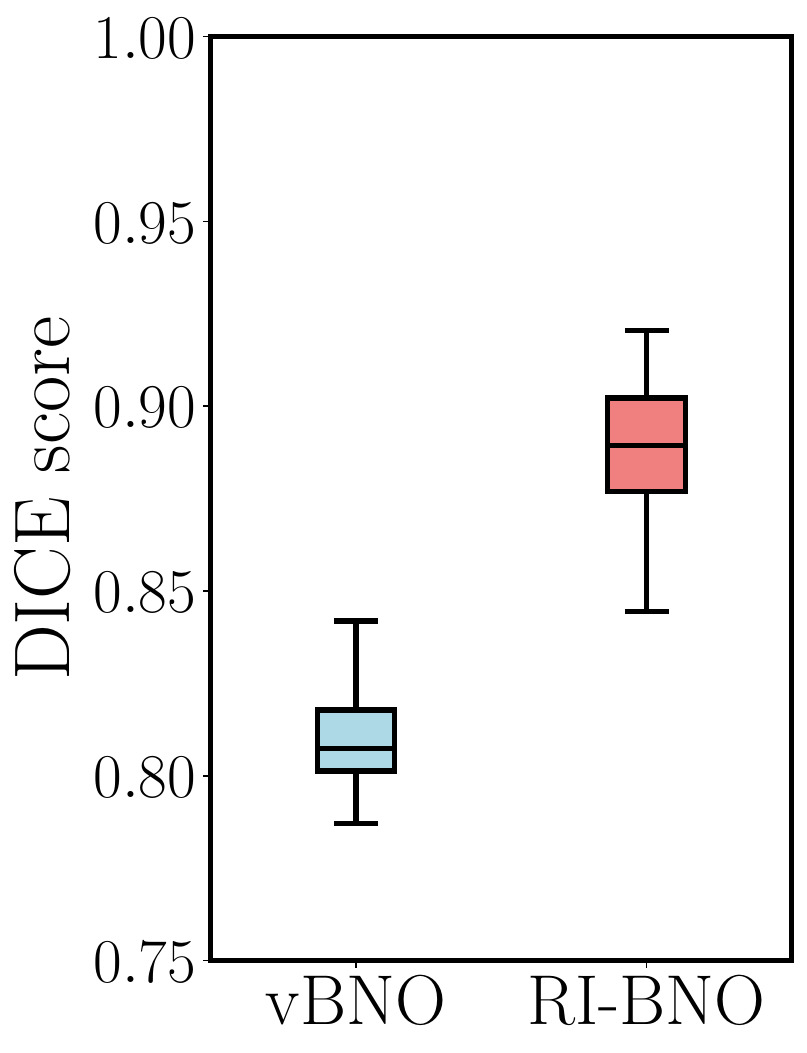}
    }\vspace{0in}\\
\end{tabular}
    \end{minipage}
    \caption{
\textbf{Forward prediction and inadequacy representation for synthetic glioma growth.}
Panel (a) shows the finite element evolution of the tumor volume fraction
$\phi_T(\bs x,t)$ from the prescribed initial condition to the terminal time
$t_F=250$ days under the mean parameter fields
$m=( \log D, \log G)$, overlaid on an axial slice of a healthy brain atlas \cite{Cocosco1997BrainWebOI}.
Panel (b) shows the terminal tumor fields for representative unseen
realizations of $m$ from the finite element
reference $u_h$
with the vBNO and RI-BNO state prediction
$\overline{u}_{\rm vBNO}$ and $\overline{u}_{\rm RI}$. 
The accompanying distributions, computed over 100 unseen realizations of
$m$, show the normalized tumor area
$\mathrm{NTA}=(\int_{\Omega} u(\bs x)\,\mathrm d\bs x)/|\Omega|$,
and the normalized diffuse-interface area
$\mathrm{NIA}=
(\int_{\Omega}\mathbbm{1}_{\{0.1<u(\bs x)<0.9\}}\,\mathrm d\bs x)/|\Omega|$,
where $u=\phi_T(\bs x,t_F)$.
Panel (c) shows the inadequacy mean $\mu_\eta$ and pointwise marginal variance
$\bs{C}_\eta$ of the learned neural-operator inadequacy. 
Panel (d) shows terminal tumor boundaries for one fixed test realization of $m$. 
Red denotes the finite element tumor region, green the state prediction \(\overline u\), and the 50 blue curves posterior tumor boundary realizations.
For vBNO, the boundary realizations are obtained by sampling \(\bs w\sim\pi_{\rm post}(\bs w\mid\mathcal D)\). For RI-BNO, \(\bs w\sim\pi_{\rm post}(\bs w\mid\mathcal D,\mathcal E)\) is combined with a realization of the learned inadequacy field \(\eta\).
The box plot summarizes the Dice similarity coefficients between the predicted and finite element tumor regions over the same 100 test inputs, ranging from 0 for no overlap to 1 for perfect overlap.
}
    \label{fig:tumor_forward}
\end{figure}

We first examine whether RI-BNO preserves the diffuse tumor interface while correcting structured prediction errors by vBNO. In Figure~\ref{fig:tumor_forward},
both vBNO and RI-BNO use the same latent-subspace architecture to learn the map from the spatially varying mobility and proliferation fields, to the terminal tumor volume fraction $u=\phi_T(\bs x,t_F)$. Training states were generated by propagating independent Gaussian random field realizations of \(\log D\) and \(\log G\) through the nonlinear phase field model using finite elements. The vBNO was trained with 1024 parameter–state pairs, whereas the RI-BNO used the same total observation budget, comprising 512 parameter–state pairs in $\mathcal D$ and 512 residual-informed observations in $\mathcal E$. 
Each parameter–state pair in \(\mathcal D\) requires a full transient FE simulation. In contrast, each residual-informed observation in \(\mathcal E\) is obtained from a linearized variational problem solve using the preliminary neural operator predictions of the state $u$, and the auxiliary fields $u^{-}$ and $n$.
The cost per residual-informed observation is therefore negligible compared to that of generating a parameter–state pair in $\mathcal D$, which requires a full phase field trajectory comprising approximately 250 nonlinear FE solves.

Figure~\ref{fig:tumor_forward}(a) illustrates the phase field tumor evolution from the prescribed initial tumor state to the terminal state. 
For unseen in-distribution parameter fields, Figure~\ref{fig:tumor_forward}(b) shows that vBNO produces an overly diffuse interface and disconnected false-positive tumor regions outside the finite element prediction. These false positives could be mistaken for tumor burden quantities used in downstream personalized treatment optimization workflows. RI-BNO suppresses these artifacts and more faithfully preserves the finite element tumor spread and infiltration.

The normalized tumor area and diffuse-interface area distributions confirm that vBNO overestimates both quantities, including an interface area approximately twice the finite element value, whereas RI-BNO closely recovers both reference distributions.

Figure~\ref{fig:tumor_forward}(c) shows that the learned inadequacy mean and variance are concentrated around the diffuse tumor interface, where small operator errors produce the largest changes in glioma extent and morphology. The mean captures the systematic interfacial and false-positive correction, whereas the variance represents uncertainty in that correction.
Figure~\ref{fig:tumor_forward}(d) 
shows vBNO boundary samples tightly concentrated around a state prediction displaced from the FE reference, suggesting overconfidence. RI-BNO produces a broader spread of boundary samples with closer agreement to the FE boundary. Its higher DICE similarity coefficient scores indicate improved spatial overlap with the FE tumor region.



\subsection{Patient-specific MRI inversion using pretrained neural operators}
\begin{figure}[ht!]
    \centering
    
    \begin{minipage}[t]{0.22\textwidth}
        \vspace{0pt} 
        \centering
        \footnotesize
        \shortstack{\hspace*{-0.4in} \large{\textbf{a}} \hspace{0.1in} \footnotesize{T1-CE MRI} \\with tumor core} \vspace{0.2in}\\ 
        \includegraphics[trim=0in 0.01in 0in 0in, clip, width=0.875\linewidth]{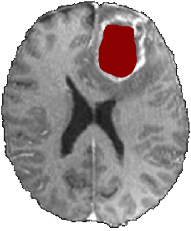}\\
    \end{minipage}
    \hspace{-0.0in}
    \begin{minipage}[t]{0.34\textwidth}
        \vspace{0pt} 
        \raisebox{0ex}{\hspace{-1.3in}\large\textbf{b}}\vspace{-0.1in}\\
        \centering
        \includegraphics[width=0.99\linewidth]{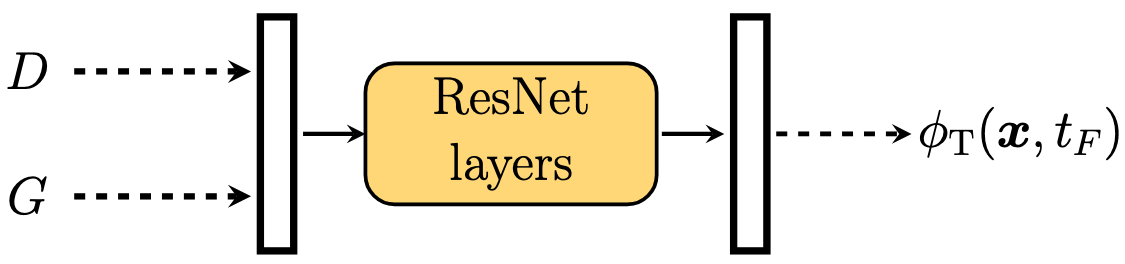}\vspace{0.1in}\\
        \hspace{-0.1in}\includegraphics[width=0.99\linewidth]{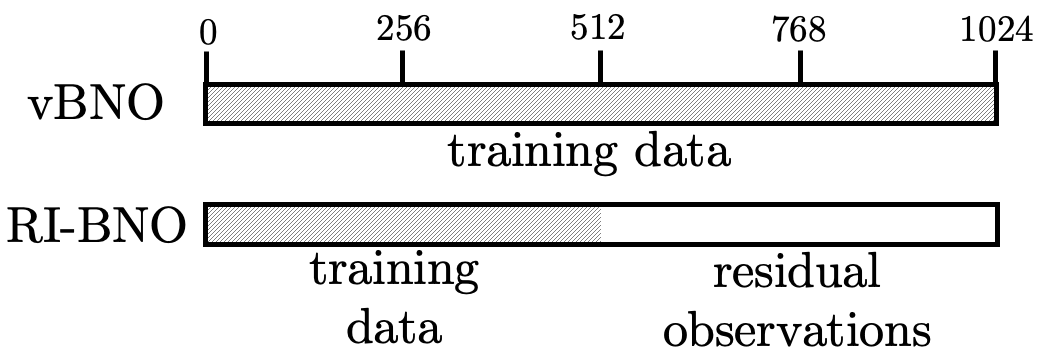}\vspace{0.075in}\\
        \hspace{-0.2in}\includegraphics[width=1.0\linewidth]{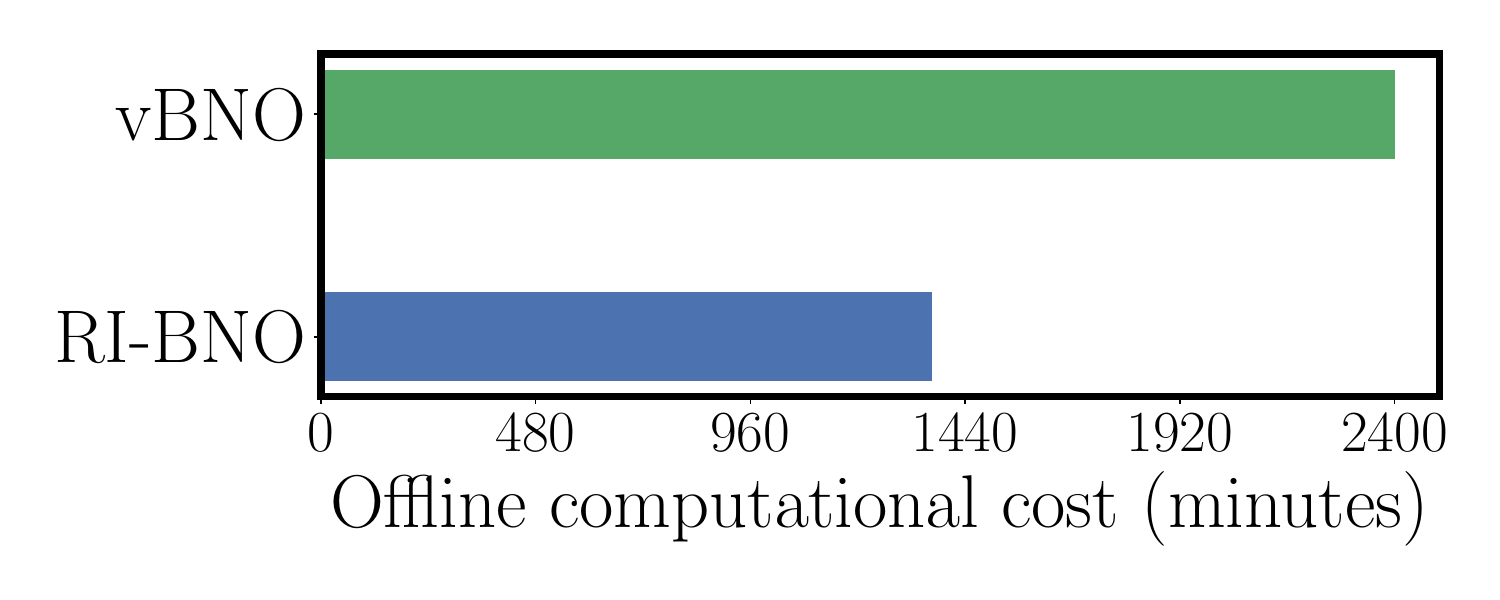}\\
    \end{minipage}
    \begin{minipage}[t]{0.41\textwidth}
        \vspace{0pt} 
        \vspace{-0.1in}
        \centering
        \setlength{\tabcolsep}{1pt}
        \begin{tabular}{c c c c}
         \includegraphics[trim=0.01in 0.01in 0.01in 0.01in, clip,width=0.25\linewidth]{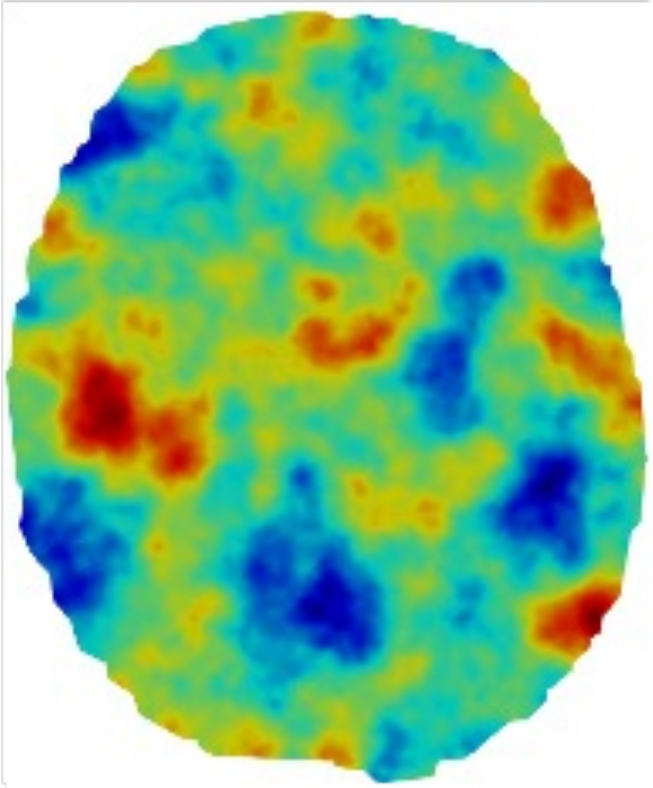} &
        \includegraphics[trim=0.01in 0.01in 0.01in 0.01in, clip,width=0.25\linewidth]{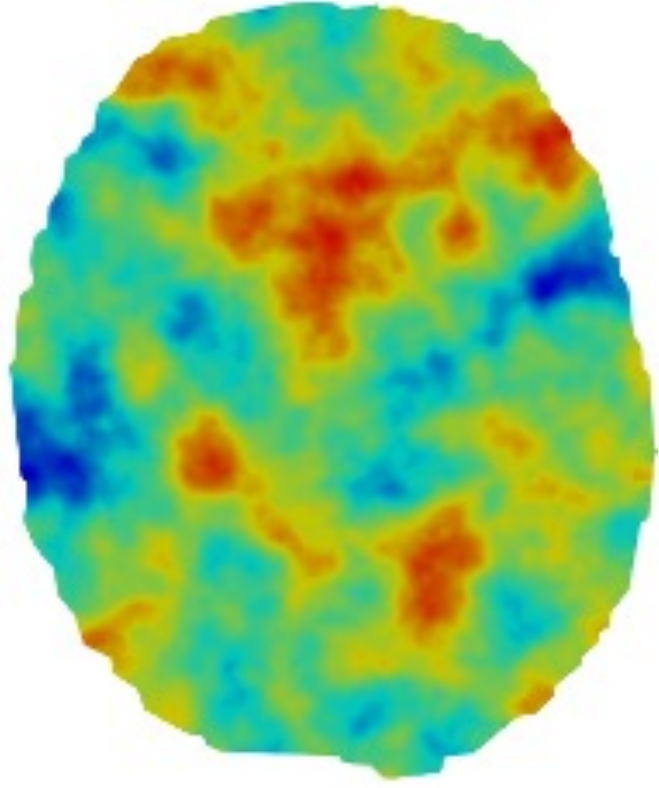} &
        \includegraphics[trim=0.01in 0.01in 0.01in 0.01in, clip,width=0.25\linewidth]{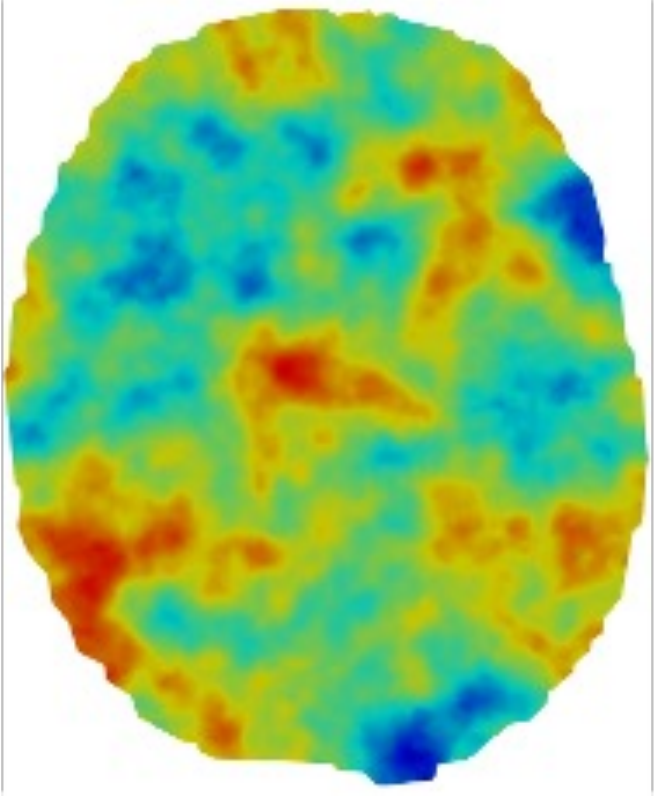} &
        \includegraphics[width=0.1\linewidth]{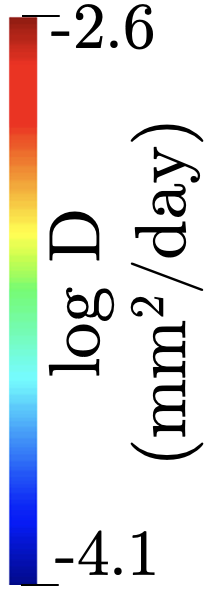}
        \\
        \vspace{-0.2in}\\
         \includegraphics[trim=0.01in 0.01in 0.01in 0.01in, clip,width=0.25\linewidth]{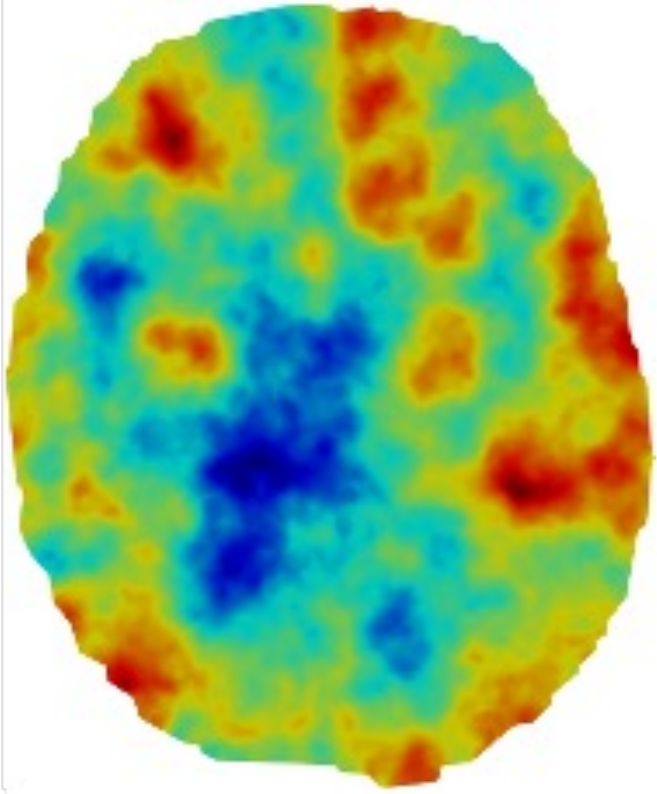} &
        \includegraphics[trim=0.01in 0.01in 0.01in 0.01in, clip,width=0.25\linewidth]{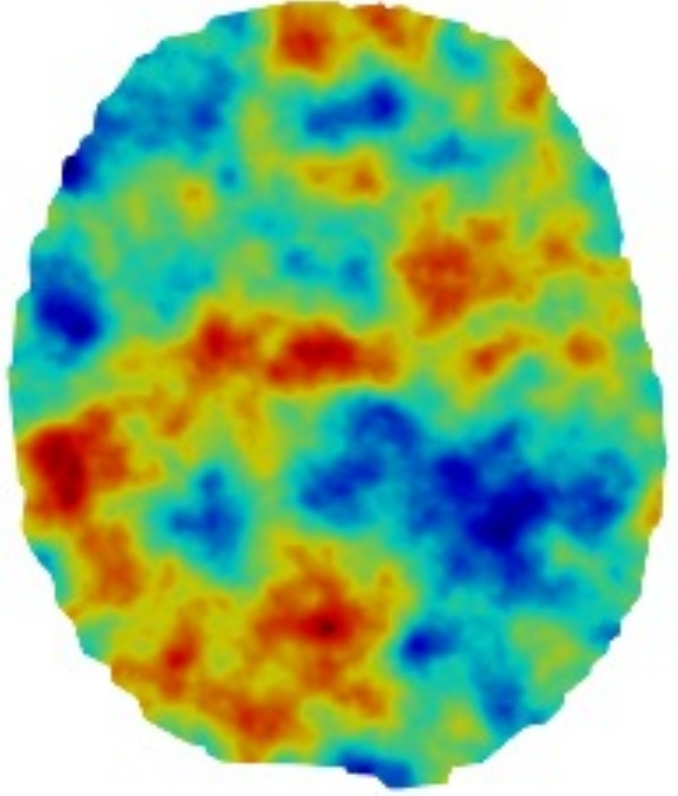} &
        \includegraphics[trim=0.01in 0.01in 0.01in 0.01in, clip,width=0.25\linewidth]{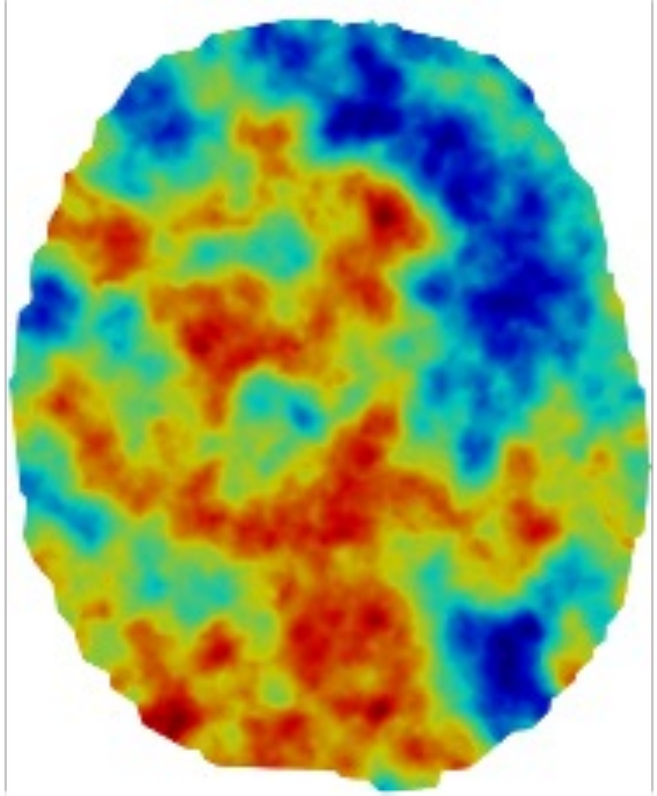} &
        \hspace{0in}\includegraphics[width=0.1\linewidth]{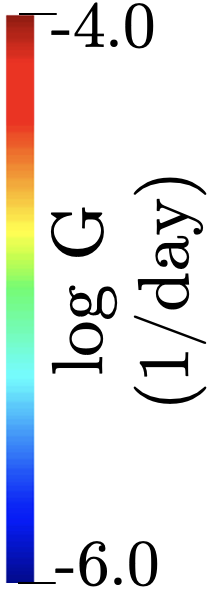}\\\vspace{-0.2in}\\
    \includegraphics[trim=0.01in 0.01in 0.01in 0.0125in, clip,width=0.25\linewidth]{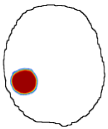} &
        \includegraphics[trim=0.01in 0.01in 0.01in 0.0125in, clip,width=0.25\linewidth]{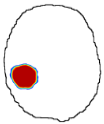} &
        \includegraphics[trim=0.01in 0.01in 0.01in 0.0125in, clip,width=0.25\linewidth]{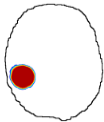}&
        \hspace{-0.08in}\includegraphics[width=0.0625\linewidth]{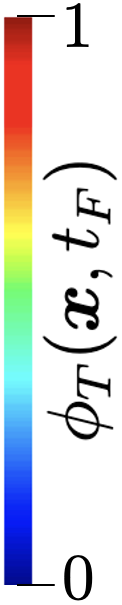}
        \\
        \end{tabular}
    \end{minipage}
    \hspace{-0.2in}
    \vspace{0.05in}\\
    \begin{minipage}[t]{0.5\textwidth}
        \vspace{0pt} 
        \hspace{-0.125in}\raisebox{8.5ex}{\large\textbf{c}}
        \setlength{\tabcolsep}{3.25pt}
        \footnotesize
        \begin{tabular}{c c c}
             FE & vBNO & RI-BNO \\
             \scriptsize{(DICE=0.932)} & \scriptsize{(DICE=0.724)} & \scriptsize{(DICE=0.918)}\vspace{0.025in}\\
             \includegraphics[trim=0.0105in 0.015in 0.0105in 0.015in, clip,width=0.28\linewidth]{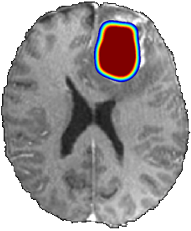}&
             \includegraphics[trim=0.0105in 0.015in 0.0105in 0.015in, clip,width=0.28\linewidth]{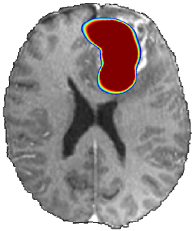}&
             \includegraphics[trim=0.0105in 0.015in 0.0105in 0.015in, clip,width=0.28\linewidth]{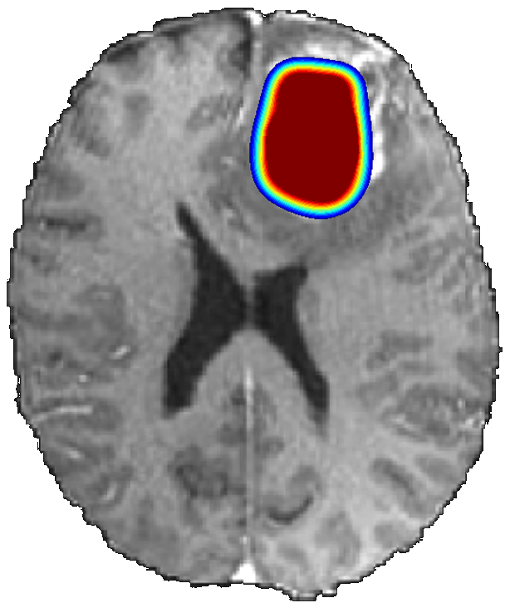}
             \\
        \end{tabular}
    \end{minipage}
    \hspace{0.0in}
    \begin{minipage}[t]{0.44\textwidth}
        \vspace{0pt} 
        \hspace{-1.5in}\raisebox{7.5ex}{\large\textbf{d}} \hspace{-0.05in}\vspace{-0.75in}\\
        \footnotesize \hspace*{0.04in} FE \hspace{0.375in} vBNO \hspace{0.25in}  RI-BNO \\
        \centering
        \includegraphics[trim=0.0085in 0.009in 0.0085in 0.0095in, clip,width=0.23\linewidth]{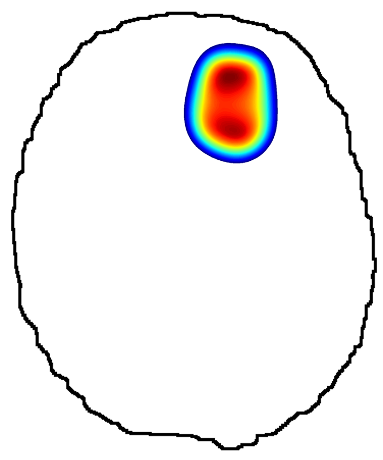} \hspace{0.04in}
        \includegraphics[trim=0.0085in 0.009in 0.0085in 0.0095in, clip,width=0.23\linewidth]{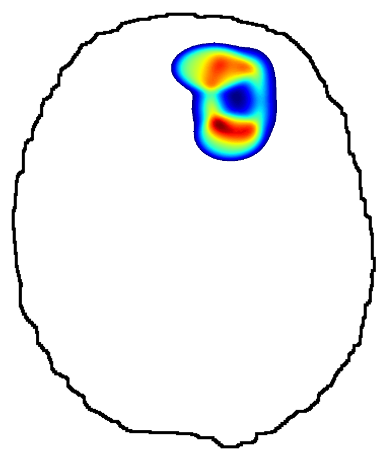} \hspace{0.04in}
        \includegraphics[trim=0.0085in 0.009in 0.0085in 0.0095in, clip,width=0.23\linewidth]{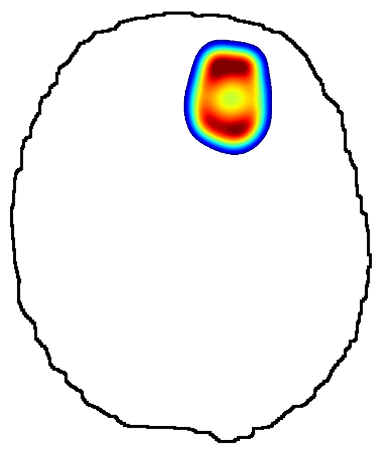} 
        \includegraphics[width=0.1\linewidth]{Figure_3/contour_D.png}
        \\
        \hspace{0.025in}\includegraphics[trim=0.0085in 0.009in 0.0085in 0.0095in, clip,width=0.23\linewidth]{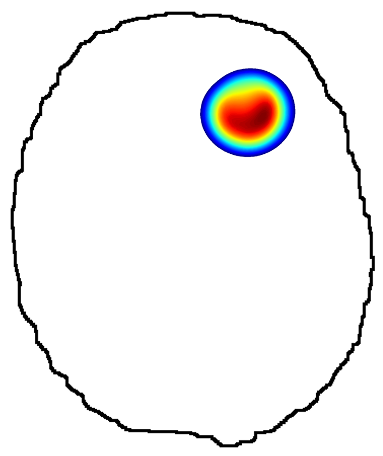} \hspace{0.04in}
        \includegraphics[trim=0.0085in 0.0095in 0.0085in 0.0095in, clip,width=0.23\linewidth]{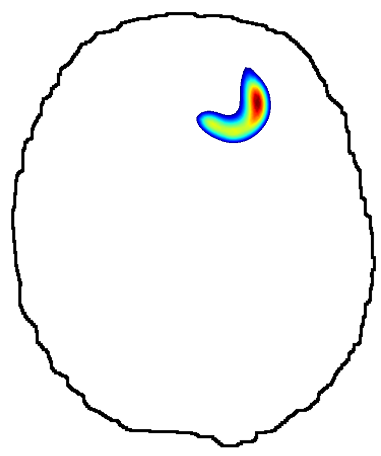} \hspace{0.04in}
        \includegraphics[trim=0.0085in 0.0095in 0.0085in 0.0095in, clip,width=0.23\linewidth]{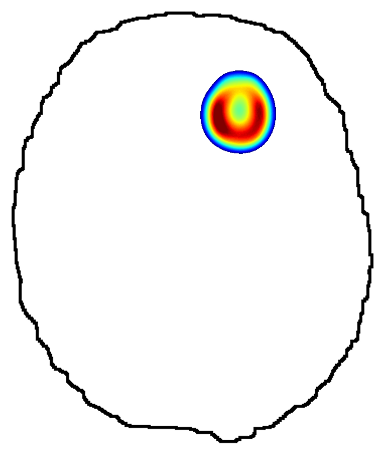}
        \hspace{0.0in}\includegraphics[width=0.1\linewidth]{Figure_3/contour_G.png}\\
    \end{minipage}
    \vspace{0.02in}\\
   \begin{minipage}[t]{0.45\textwidth}
        \vspace{0pt} 
        \footnotesize
        \setlength{\tabcolsep}{0.5pt}
        \begin{tabular}{c c c}
             \hspace{-0.65in}\large\textbf{e} \hspace{0.4in}\footnotesize FE & vBNO & RI-BNO \vspace{0.025in}\\
             \hspace{-0.105in}\includegraphics[width=0.36\linewidth]{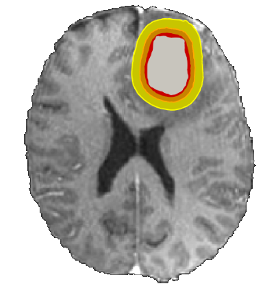}&
             \hspace{-0.045in}\includegraphics[width=0.36\linewidth]{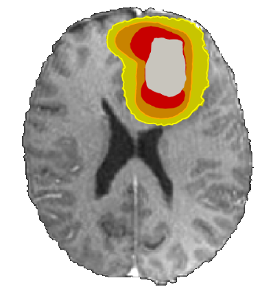}&
             \hspace{-0.065in}\includegraphics[width=0.36\linewidth]{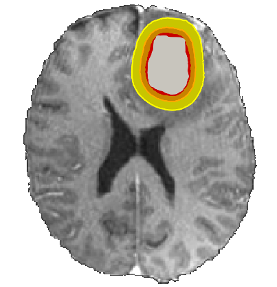}
             \\
            \multicolumn{3}{c}{%
    \hspace{-0.1in}\includegraphics[width=1.0\linewidth]{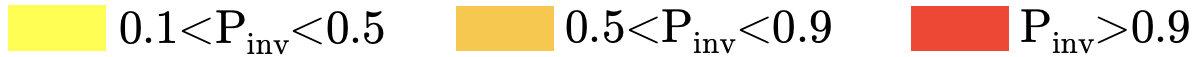}
} \\
        \end{tabular}
    \end{minipage}
    \hspace{0.05in}
    \begin{minipage}[t]{0.45\textwidth}
        \vspace{0pt} 
        \centering
        \hspace{-2.6in}\large\textbf{f} \vspace{-0.1in}\\
        \hspace{0.05in}\raisebox{0.01in}{\includegraphics[width=0.47\linewidth]{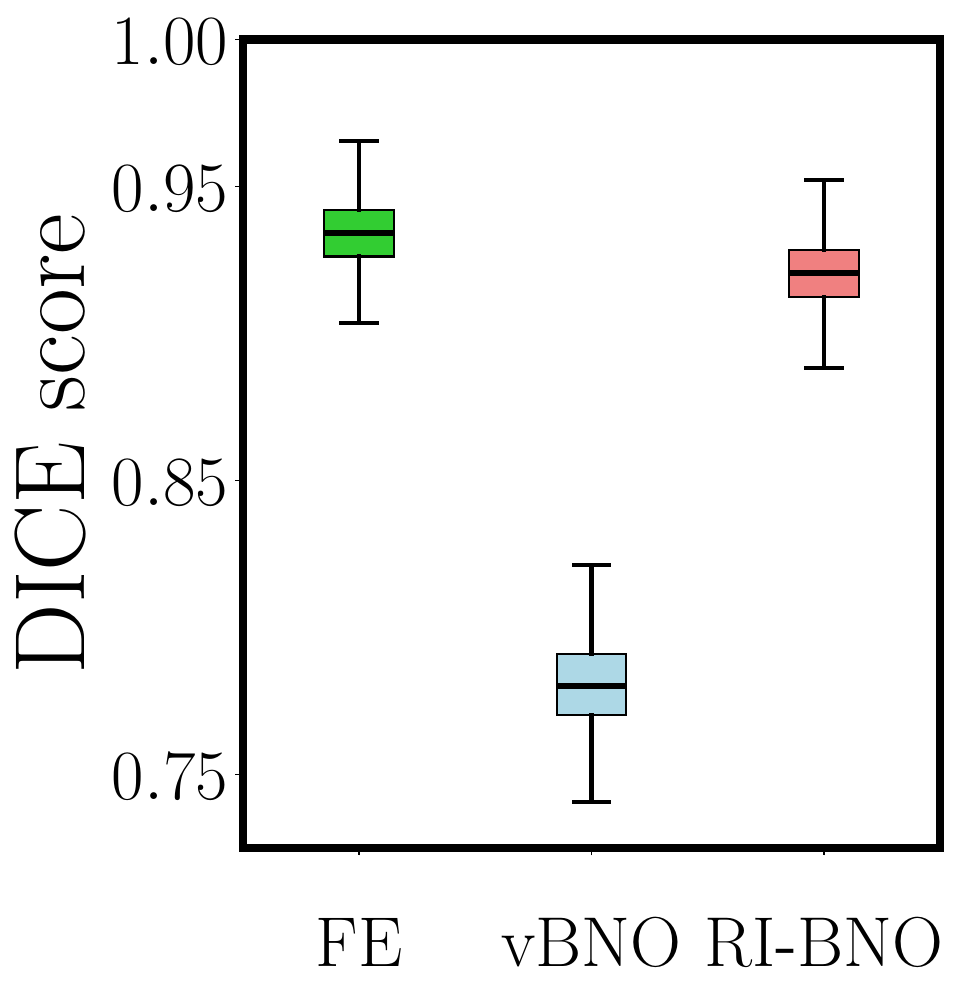}}
        \includegraphics[width=0.48\linewidth]{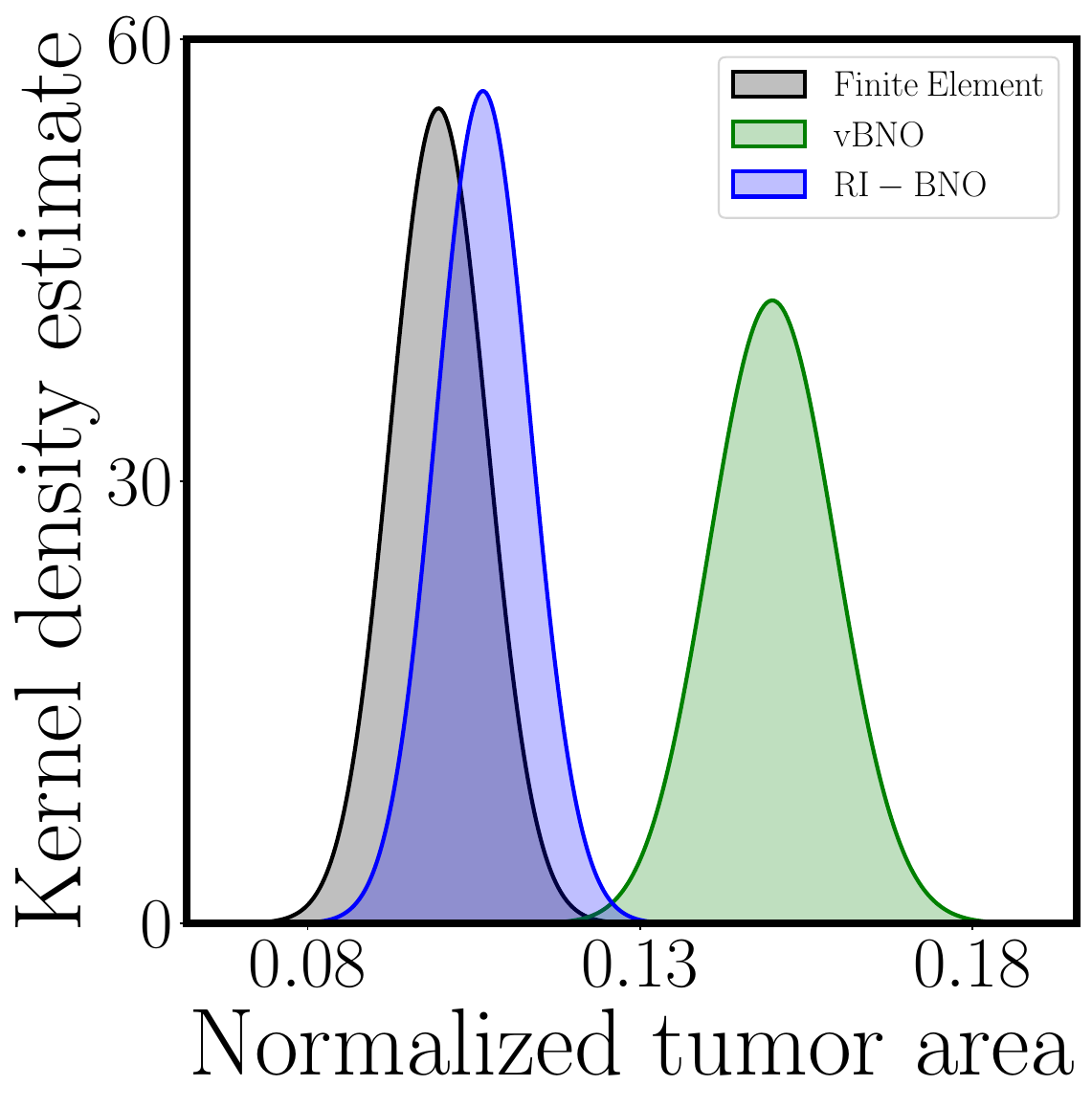} 
        \\
    \end{minipage}
\caption{\textbf{Patient-specific MRI inference using pretrained neural operators.}
Panel (a) shows an axial T1-CE MRI slice from the BraTS 2019 high-grade glioma case BraTS19\_CBICA\_AQP\_1, together with the segmented tumor-core mask used for the two-dimensional inversion.
Panel (b) summarizes offline operator construction, including the latent-subspace architecture, allocation of the matched total observation budget, offline neural operator construction cost, and representative synthetic realizations of training data of the
spatially varying mobility and proliferation fields, \(m=(\log D, \log G)\), and their
terminal tumor states, \(u_h=\phi_T(\bs x,t_F)\).
Panel (c) compares the finite-element, vBNO and RI-BNO tumor reconstructions and their Dice similarity coefficients relative to the MRI-derived tumor core mask.
Panel (d) shows the corresponding MAP estimates of the spatially varying fields \(\log D\) and \(\log G\).
Panel (e) shows bands of posterior infiltration exceedance probability, computed from $250$ retained MCMC samples using
$\phi_{\rm inv}=0.075$. The bands $0.1<P_{\rm inv}<0.5$, $0.5<P_{\rm inv}<0.9$, and
$P_{\rm inv}>0.9$ indicate increasing posterior support for subvisual tumor extension outside the MRI-visible tumor core boundary, defined using the MRI-detection threshold $\phi=0.2$.
Panel (f) shows the posterior distributions of the DICE similarity coefficient and brain-normalized tumor area computed from \(500\) retained MCMC
samples.
}
    \label{fig:tumor_mri}
\end{figure}

The spatially varying tumor mobility and proliferation fields, \(m=(\log D, \log G)\), govern tumor invasion and growth but cannot be observed directly from clinical MRI \cite{subramanian2020multiatlas, liang2023bayesian, noman2026dynamic}. 
We test whether neural operators pretrained on synthetic phase field simulations can infer these patient-specific fields from an unseen clinical MRI (Figure~\ref{fig:tumor_mri}(a)), without patient-specific operator retraining or additional FE training trajectories.
For glioblastoma, diffuse infiltration beyond the MRI-visible tumor core motivates patient-specific estimates of subvisual spread, which can inform tumor forecasting and treatment-planning \cite{swan2018patient,lipkova2019personalized}.
We consider a representative axial T1-weighted contrast-enhanced (T1-CE) MRI slice from a de-identified high-grade glioma case in the BraTS 2019 training dataset \cite{menze2014multimodal,bakas2017advancing,bakas2018identifying}. The observation \(u_{\rm MRI}\) shown in Figure~\ref{fig:tumor_mri}(a) is constructed from the annotated tumor core, defined as the union of the enhancing and necrotic tumor regions.

For offline training, Gaussian random field realizations of \(m=(\log D,\log G)\) are propagated through the phase-field model \eqref{eq:governing_tumor} from a prescribed initial tumor to \(t_F=240\) days, generating synthetic parameter–state pairs (Figure~\ref{fig:tumor_mri}(b) and Supplementary Figure~\ref{fig:MRI_training_predictions}(a)). For FE-based MRI inversion, a different initial tumor is placed within the observed tumor core adopted from \cite{subramanian2019simulation} as shown in Figure~\ref{fig:MRI_training_predictions}(b). The pretrained neural operators are evaluated without patient-specific retraining or an explicit initial condition input, although their learned maps reflect the initialization used during training. We therefore assess agreement with FE-based inversion under different initialization assumptions. Such reuse could support cancer digital twins that update patient-specific tumor models as new imaging becomes available, reducing the computational delay between observation and model-based assessment.
Under the matched total observation budget \(N_{\rm tot}=1024\), vBNO uses \(N_{\rm train}=N_{\rm tot}\), whereas RI-BNO uses \(N_{\rm train}=N_{\rm err}=N_{\rm tot}/2\). Both models employ the same latent-subspace architecture, with 25-dimensional active-subspace and POD representations and a 15-layer ResNet of width 50.


%

The MRI-informed inverse problem is
\begin{equation}\label{eq:bayes_mri}
\pi_{\rm post}(m\mid u_{\rm MRI})
\propto
\exp\!\left[
-\frac{1}{2}
\left\|
\mathcal B(u_{\rm MRI})-u(m)
\right\|_{\bs\Sigma_{\rm MRI}^{-1}}^{2}
\right]
\pi_{\rm pr}(m),
\end{equation}
where $\pi_{\rm pr}(m)$ is the Gaussian random field prior on $m$, 
$u_{\rm MRI}$ is the voxel-wise tumor field observation from MRI and $\mathcal B$
denotes its projection onto the
finite element mesh, and $\bs\Sigma_{\rm MRI}$ is the imaging noise covariance in the projected
space.
We solve the inverse problem \eqref{eq:bayes_mri} separately using three forward maps $u(m)$: the finite element reference \(u_h\), vBNO prediction \(u_{\rm vBNO}\), and RI-BNO prediction \(u_{\rm RI}\) and using a Markov Chain Monte Carlo (MCMC) based on generalized preconditioned Crank-Nicolson \cite{cotter2013mcmc,VillaPetraGhattas18,VillaPetraGhattas21}. 
A Newton--Conjugate Gradient solve first provides the MAP estimate, and the local posterior curvature defines a Laplace covariance for constructing an informed proposal. This function space formulation retains mesh-independent sampling for the high-dimensional fields \(D\) and \(G\), each discretized on a brain mesh with \(44{,}967\) finite element nodes, while improving mixing relative to a prior-based proposal. Because posterior inference requires many forward evaluations, replacing repeated nonlinear transient phase field simulations with accurate neural operator evaluations is computationally consequential.
For the neural operator models, the likelihood in \eqref{eq:bayes_mri} is evaluated by Monte Carlo averaging over posterior weight samples and, for RI-BNO, realizations of the learned inadequacy field.

Figure~\ref{fig:tumor_mri}(c,d) compares the MAP estimates of the parameter fields and corresponding tumor reconstructions. Finite element inversion reconstructs the MRI-visible tumor core with a DICE score of \(0.932\). In contrast, vBNO inadequacy propagates through the inverse problem, producing substantially different \(D\) and \(G\) fields and a distorted tumor reconstruction with a DICE coefficient of \(0.724\). RI-BNO closely recovers the finite element-inferred parameter structure and tumor extent, achieving a DICE score of \(0.918\), within \(0.014\) of the finite-element result. 
In this case, RI-BNO yields inferred parameter fields and tumor reconstructions that agree more closely with the FE-based results than vBNO.
FE forward simulations using the vBNO-/RI-BNO-inferred MAP parameter fields provide a further consistency check (Supplementary Figure~\ref{fig:MRI_training_predictions}(b)).
Using the same patient-specific initialization as FE-based inversion, FE simulations at the RI-BNO- and vBNO-inferred MAP parameters achieve Dice scores of \(0.919\) and \(0.727\) with the MRI-visible tumor core, respectively, providing an FE consistency check at the inferred MAP points.
Additionally, the same pretrained operators are also applied to four additional BraTS patients without patient-specific retraining or additional FE training trajectories (Supplementary Figure~\ref{fig:MRI_multiple_patients}). RI-BNO achieves a mean Dice score of 0.912, compared with 0.732 for vBNO and 0.936 for FE-based inversion, supporting reuse of the offline-trained models across additional MRI observations.
Figure~\ref{fig:tumor_mri}(e,f) extends the comparison beyond the MAP
estimates using samples from the MCMC posterior. 
We estimate the posterior infiltration exceedance probability \(P_{\rm inv}(\bs x)\) as the fraction of retained samples satisfying \(\phi_T^{(s)}(\bs x,t_F)>\phi_{\rm inv}\), with \(\phi_{\rm inv}=0.075\) used to probe subvisual tumor extension.
Figure~\ref{fig:tumor_mri}(e) displays
$P_{\rm inv}$ outside the MRI-visible tumor core boundary using the probability
bands $0.1<P_{\rm inv}<0.5$, $0.5<P_{\rm inv}<0.9$, and
$P_{\rm inv}>0.9$. The finite element and RI-BNO posteriors yield similar
spatial patterns of probable infiltration, whereas vBNO predicts a markedly
displaced and enlarged region at risk. 
These maps show how neural operator inadequacy can alter posterior estimates of subvisual infiltration, with potential implications for personalized treatment planning.
RI-BNO instead closely preserves the finite element posterior spatial risk
pattern.
Finally, Figure~\ref{fig:tumor_mri}(f) compares the posterior distributions of the DICE similarity coefficient relative to the MRI-derived tumor core mask and the brain-normalized tumor area. vBNO places substantial posterior mass on enlarged false-positive tumor regions, inflating the predicted tumor area and reducing spatial agreement with the MRI-derived core. RI-BNO instead closely recovers the finite element posterior distributions of both quantities.

Detailed computational costs are reported in the \textit{Supplementary Information}. RI-BNO reduces the MAP optimization time from \(134.2\) minutes to \(12.9\) seconds and the MCMC sampling time from \(175.1\) hours to \(135.6\) minutes, corresponding to an approximately \(77\times\) reduction in MCMC wall time and a \(70.4\times\) increase in effective sampling speed. Thus, RI-BNO closely reproduces the FE-based posterior for tumor extent and subvisual infiltration while reducing online inversion from more than seven days to approximately \(2.3\) hours, making eventual posterior forecasting and treatment-optimization studies substantially more tractable within cancer digital twin workflows.

\section{Example 3: Risk-averse design of fiber-reinforced aerogel thermal breaks}\label{sec:thermal_break}

We finally embed RI-BNO within a risk-averse topology optimization problem for
superinsulating thermal breaks for net-zero building. The design controls the spatial
fiber distribution under additive manufacturing uncertainty to suppress heat transfer
while maintaining mechanical stability, requiring repeated Monte Carlo
evaluations of coupled thermal and stress fields and their tail-risk measures.

\subsection{Three-phase thermomechanical model of wheat straw aerogel composites}
Silica aerogels are promising superinsulating materials for commercial and
residential building envelopes because their highly porous structure combines
low density with exceptionally low thermal conductivity
\cite{an2021wearable,tan2022predictive}. Their mechanically fragile silica
skeleton, however, limits use in insulation components such as thermal breaks,
which must interrupt heat flow across structural junctions while sustaining
service loads. Additively manufactured wheat-straw fiber reinforcement offers a
lightweight, low-cost, and renewable route to improve mechanical
stability without substantially compromising thermal insulation
\cite{guo2025wheat}. The composite is modeled as an interacting mixture of silica skeleton ($s$), gas ($g$), and reinforcing fibers
($f$), with $\phi_s+\phi_g+\phi_f=1$. The coupled thermal and mechanical
response is governed by a three-phase continuum-mixture model.
The thermodynamic derivation and finite-element formulation are provided in \cite{singh2025continuum}; the governing PDEs relevant here are summarized as follows:
\begin{eqnarray}
C p
+\nabla\cdot\bs u_s
+\nabla\cdot\bs u_f
=0,& \bs x\in\Omega,
\nonumber\\
\nabla\cdot\bs T_s'
-\phi_s\nabla p
+\bs X_{sf}
=\bs 0,
& \bs x\in\Omega,
\nonumber\\
\nabla\cdot\bs T_f'
-\phi_f\nabla p
-\bs X_{sf}
=\bs 0,
& \bs x\in\Omega,
\nonumber\\
\nabla\cdot
\left(
\phi_\alpha K_\alpha\nabla\theta_\alpha
\right)
-
\sum_{\beta\neq\alpha}
H_{\alpha\beta}
\left(\theta_\alpha-\theta_\beta\right)^3
=0,
& \bs x\in\Omega,
\nonumber\\
\bs u_\alpha =\bs u_\alpha^\dagger,
\quad \bs x \in\Gamma_D,
\qquad
\bs T_\alpha\bs n =\bs t_\alpha^\dagger,
& \bs x \in\Gamma_N,
\nonumber\\
-\phi_\alpha K_\alpha\nabla\theta_\alpha\cdot\bs n
=
H_{\rm amb}
\left(\theta_\alpha-\theta_{\rm amb}\right),
& \bs x \in\Gamma_R,
\nonumber\\
-\phi_\alpha K_\alpha\nabla\theta_\alpha\cdot\bs n =0
& \bs x \in\Gamma_A,
\label{eq:mixture_pde}
\end{eqnarray}
Here, \(p\) is the pore gas pressure, \(\bs u_s\) and \(\bs u_f\) are the solid-phase displacements, and \(\theta_\alpha\) is the temperature of phase \(\alpha\in\{s,g,f\}\). The coefficients \(C\), \(K_\alpha\), \(H_{\alpha\beta}\), and \(H_{\rm amb}\) govern gas storage, phase conductivity, interphase heat exchange, and ambient convection, respectively; their values are provided in the  \textit{Supplementary Information}.
The interaction force \(\bs X_{sf}\), which enters the two solid-phase
momentum balances with opposite signs, represents load transfer between the silica skeleton and fibers, including loss of coupling upon debonding. 
Assuming isotropic elasticity for the silica skeleton and
fibers, their effective stresses are
$
\bs T_\alpha'
=
\mu_\alpha
\left(
\nabla\bs u_\alpha+\nabla\bs u_\alpha^\top
\right)
+
\lambda_\alpha
\left(\nabla\cdot\bs u_\alpha\right)\bs I,
\alpha\in\{s,f\},
$
where \(\mu_\alpha\) and \(\lambda_\alpha\) are the corresponding Lamé
moduli. The partial stresses are
\(\bs T_\alpha=-\phi_\alpha p\bs I+\bs T_\alpha'\) for
\(\alpha\in\{s,f\}\) and \(\bs T_g=-\phi_g p\bs I\).
Neglecting relative
phase momentum, the mixture stress field is therefore
$\bs T=\bs T_s'+\bs T_f'-p\bs I$.
The mixture von Mises stress and heat flux are
\[
T_{\rm VM}
=
\sqrt{\frac{3}{2}\operatorname{dev}(\bs T):
\operatorname{dev}(\bs T)},
\qquad
\bs q
=
-\sum_{\alpha}
\phi_\alpha K_\alpha\nabla\theta_\alpha,
\]
where
\(\operatorname{dev}(\bs T)
=\bs T-\tfrac{1}{3}\operatorname{tr}(\bs T)\bs I\).
%
The residual of this PDE can be derived as,
\begin{align}\label{eq:residual_duu}
\left\langle
\mathcal R,v
\right\rangle_{\mathcal V^\ast,\mathcal V}
:={}&
\left\langle
Cp +\nabla\cdot\bs u_s +\nabla\cdot\bs u_f, \, v_p \right\rangle_\Omega
\nonumber\\
&
-
\left\langle
\bs T_s', \, \nabla\bs v_s
\right\rangle_\Omega
+
\left\langle
p, \nabla\cdot(\phi_s \bs{v}_s )
\right\rangle_\Omega
+
\left\langle
\bs X_{sf}, \,\bs v_s
\right\rangle_\Omega
+
\left\langle
\bs t_s^\dagger, \, \bs v_s
\right\rangle_{\Gamma_N}
\nonumber\\
&
-
\left\langle
\bs T_f', \, \nabla\bs v_f
\right\rangle_\Omega
+
\left\langle
p, \nabla\cdot(\phi_f \bs{v}_f )
\right\rangle_\Omega
-
\left\langle
\bs X_{sf}, \,\bs v_f
\right\rangle_\Omega
+
\left\langle
\bs t_f^\dagger, \, \bs v_f
\right\rangle_{\Gamma_N}
\nonumber\\
&
+
\sum_{\alpha}
\Bigg[
+
\langle
\phi_\alpha K_\alpha\nabla\theta_\alpha \, ,
\nabla{z}_\alpha
\rangle_\Omega
+
\langle
\sum_{\beta\neq\alpha}
H_{\alpha\beta}
(\theta_\alpha-\theta_\beta)^3, \, z_\alpha
\rangle_\Omega
\nonumber\\
&
+
\left\langle
H_{\rm amb}
(\theta_\alpha-\theta_{\rm amb}), \, z_\alpha
\right\rangle_{\Gamma_R}
\Bigg], \qquad \forall v\in\mathcal V.
\end{align}
where $u
=\left(\theta_s,\,\theta_g,\,\theta_f, p,\,\bs u_s,\,\bs u_f\right)$ is the state vector and $v
=\left(z_s,\,z_g,\,z_f, \,v_p,\,\bs v_s,\,\bs v_f \right)$ is the corresponding test vector.

\subsection{Risk-averse design under spatial manufacturing
variability}
\begin{figure}

\begin{minipage}[t]{0.52\textwidth}
    \vspace{0pt} 
    \vspace{-0.1in}%
    \noindent\hspace{0.05in}\large{\textbf{a}}\par\vspace{0.1in}
    \centering
    \includegraphics[trim=0in 0in 0.0in 0in, clip,width=0.975\linewidth]{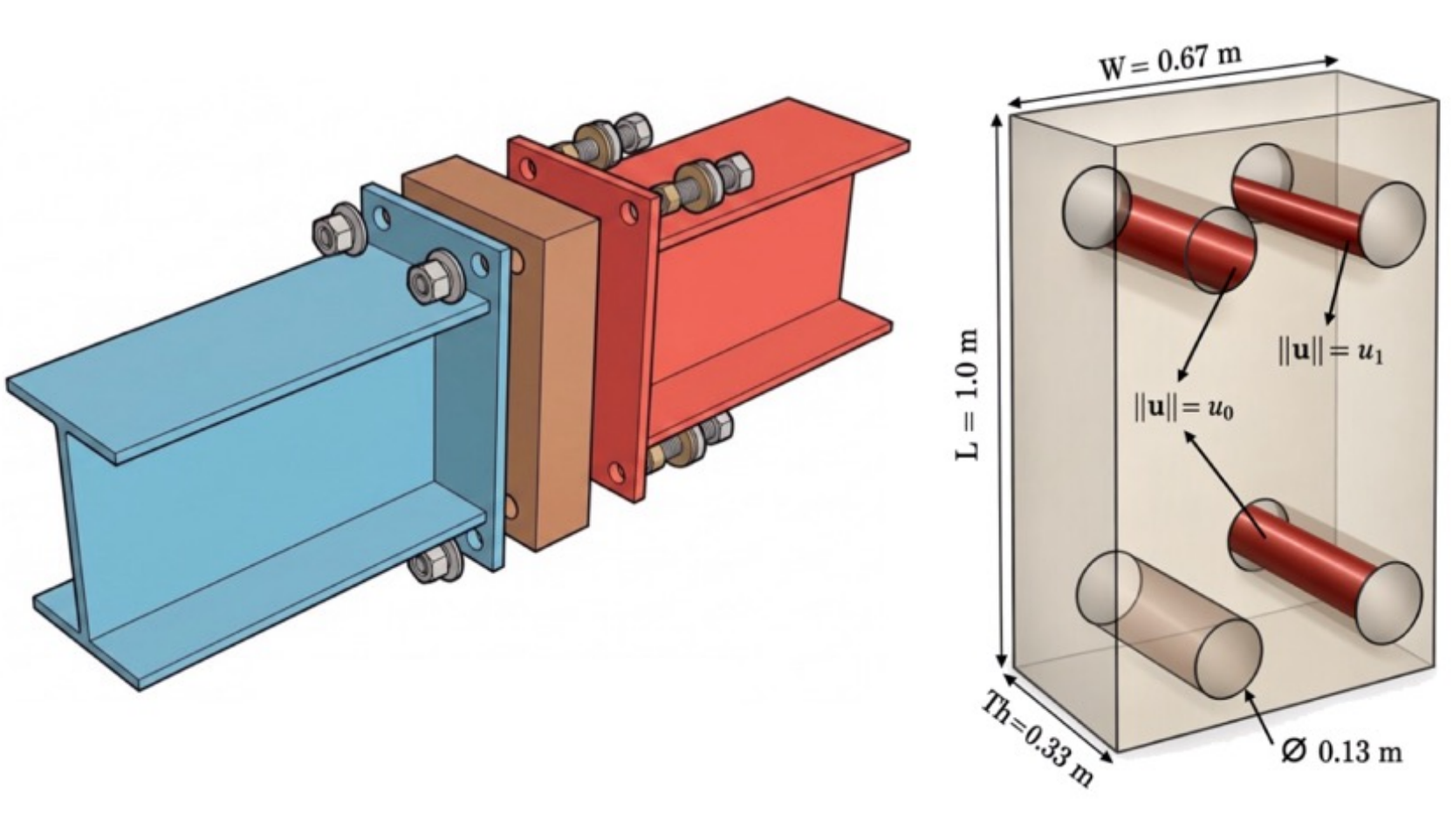}\vspace{0.1in}
\end{minipage}
    \begin{minipage}[t]{0.45\textwidth}
        \hspace{-0.0in}\large{\textbf{b}} \vspace{-0.1in}\\
\hspace*{0.15in}\includegraphics[width=0.975\linewidth]{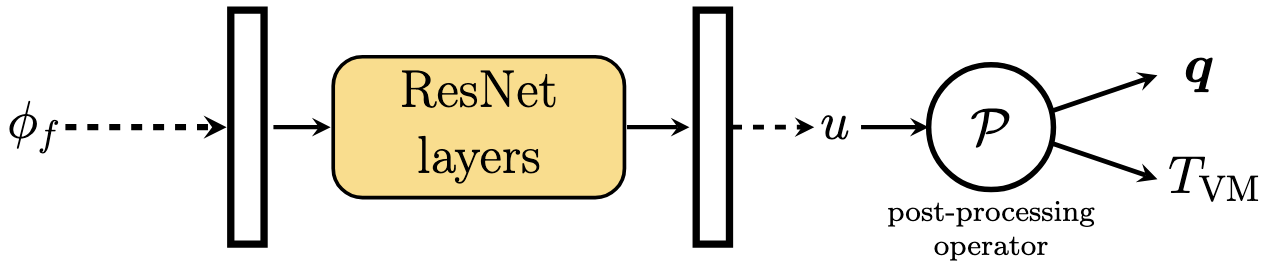}\vspace{0.025in}\\
\setlength{\tabcolsep}{2.0pt}
\begin{tabular}{c c c c c c} 
 \includegraphics[trim=0.01in 0.012in 0.01in 0.012in, clip,width=0.2\linewidth]{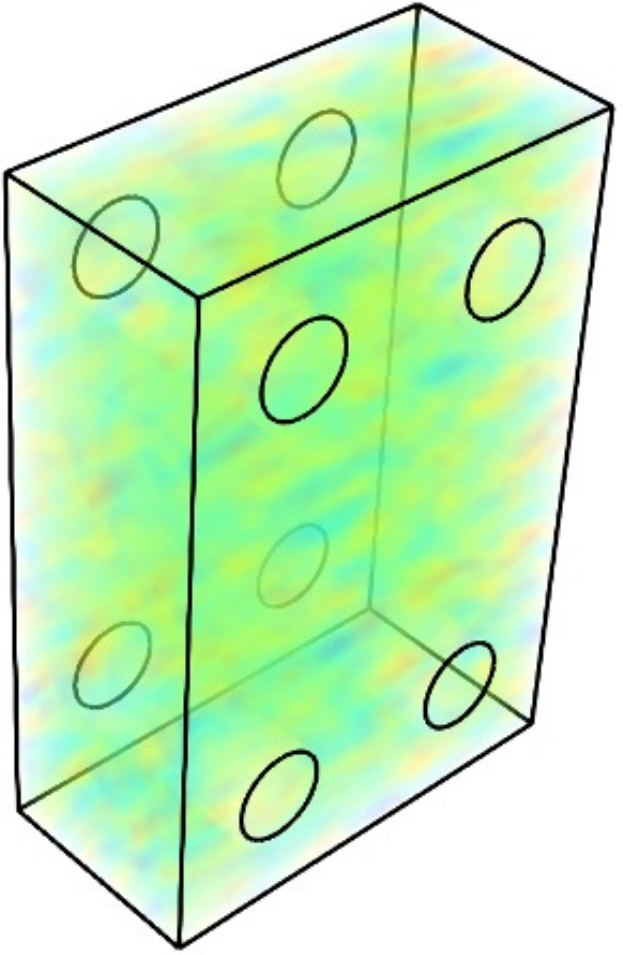}&
 \includegraphics[width=0.0525\linewidth]{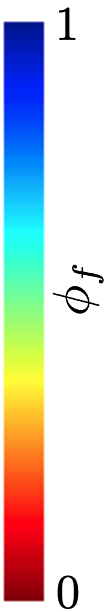} &
 \includegraphics[trim=0.012in 0.012in 0.01in 0.012in, clip,width=0.2\linewidth]{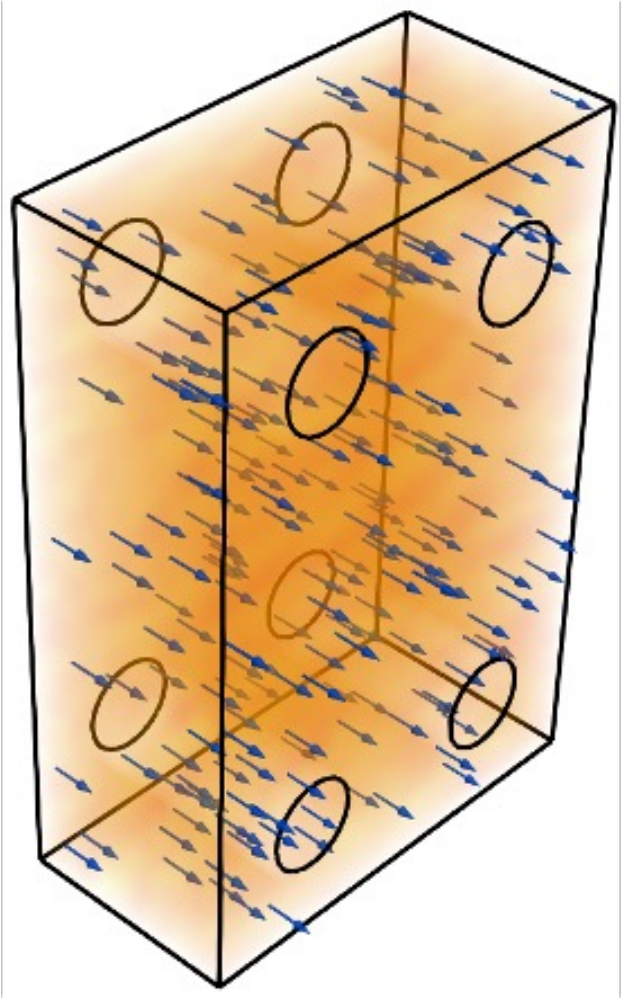}&
 \includegraphics[width=0.0685\linewidth]{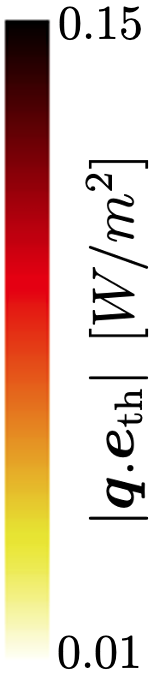}&
 \includegraphics[trim=0.01in 0.012in 0.01in 0.012in, clip,width=0.2\linewidth]{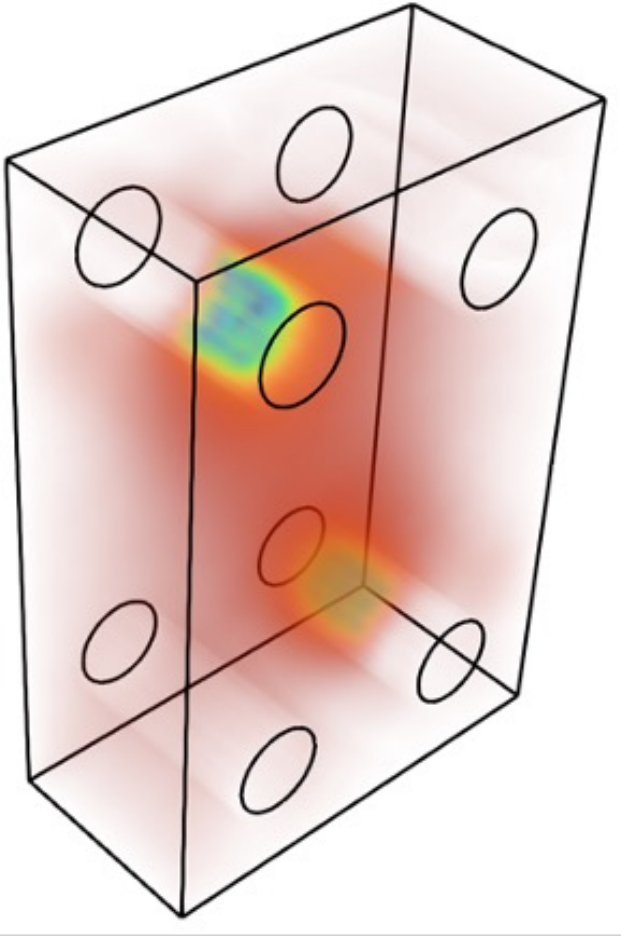}&
 \includegraphics[width=0.0685\linewidth]{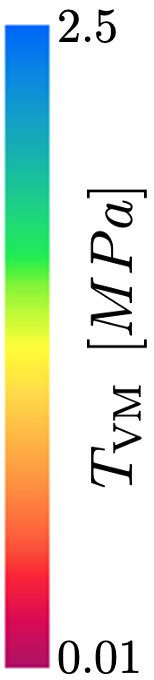}
 \vspace{-0.05in}\\
 \includegraphics[trim=0.01in 0.012in 0.01in 0.012in, clip,width=0.2\linewidth]{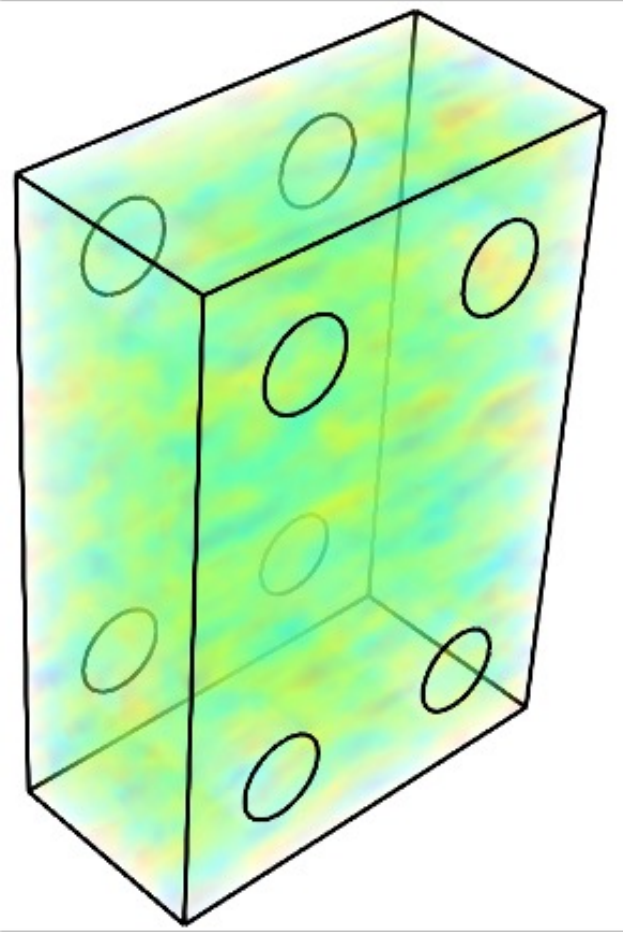}&
 \includegraphics[width=0.0525\linewidth]{Figure_4/contour_samples.png} &
 \includegraphics[trim=0.012in 0.012in 0.01in 0.012in, clip,width=0.2\linewidth]{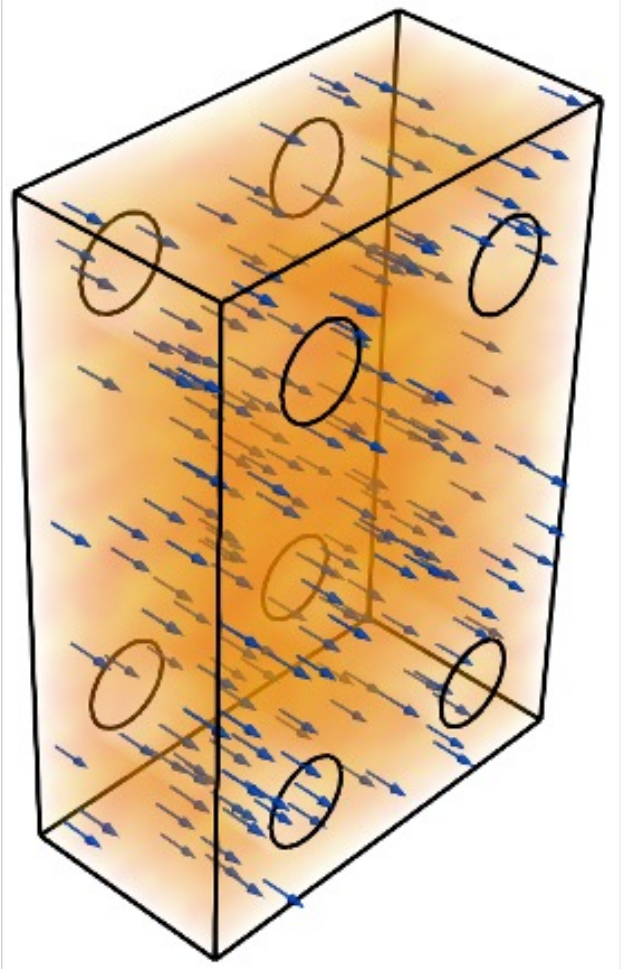}&
 \includegraphics[width=0.0685\linewidth]{Figure_4/contour_heat_flux_training.png}&
 \includegraphics[trim=0.01in 0.012in 0.01in 0.012in, clip,width=0.2\linewidth]{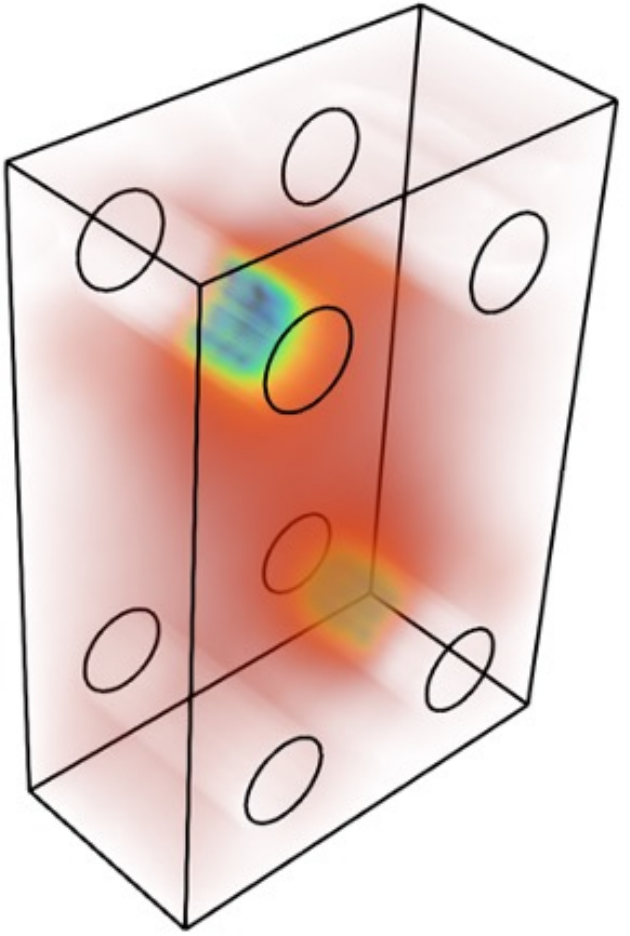}&
 \includegraphics[width=0.0685\linewidth]{Figure_4/contour_von_mises_training.png}
 \\
\end{tabular}
    \end{minipage}
    \begin{minipage}{0.54\textwidth}
        \vspace{-0.5in} 
        \centering
\normalsize
\setlength{\tabcolsep}{2.5pt}
\begin{tabular}{c c c c} 
 \hspace*{-0.525in}\large{\textbf{c}} \hspace{0.25in} \normalsize FE & vBNO & RI-BNO & \vspace{-0.0in}\\
 \includegraphics[trim=0.01in 0.012in 0.01in 0.012in, clip,width=0.23\linewidth]{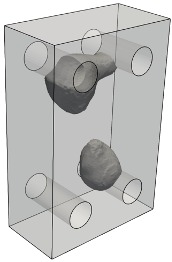}&
 \includegraphics[trim=0.01in 0.012in 0.01in 0.012in, clip,width=0.23\linewidth]{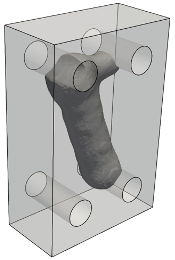}&
 \includegraphics[trim=0.01in 0.012in 0.013in 0.012in, clip,width=0.23\linewidth]{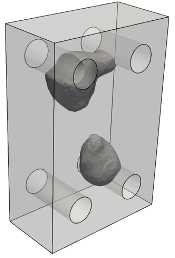}&
 \includegraphics[width=0.07\linewidth]{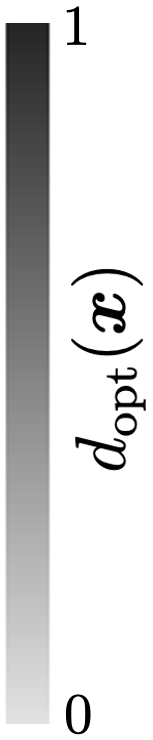}
 \\
 \includegraphics[trim=0.011in 0.012in 0.0105in 0.012in, clip,width=0.23\linewidth]{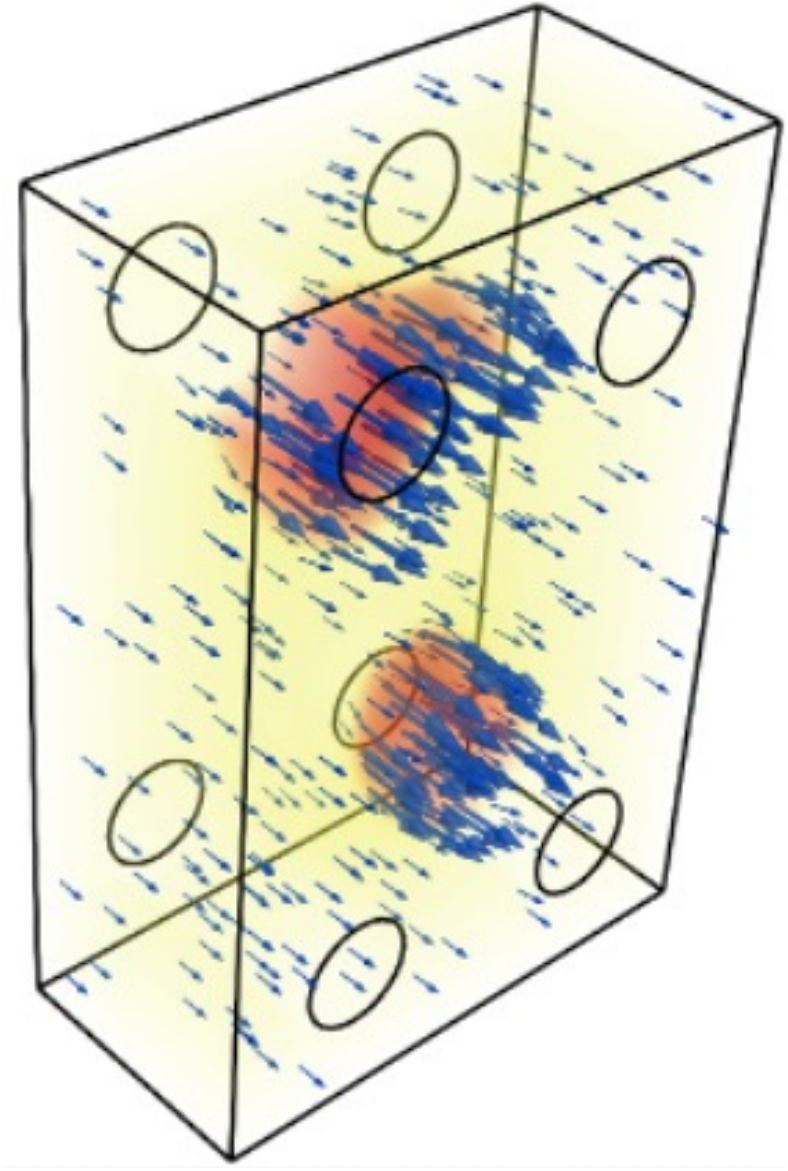}&
 \includegraphics[trim=0.011in 0.012in 0.01in 0.012in, clip,width=0.23\linewidth]{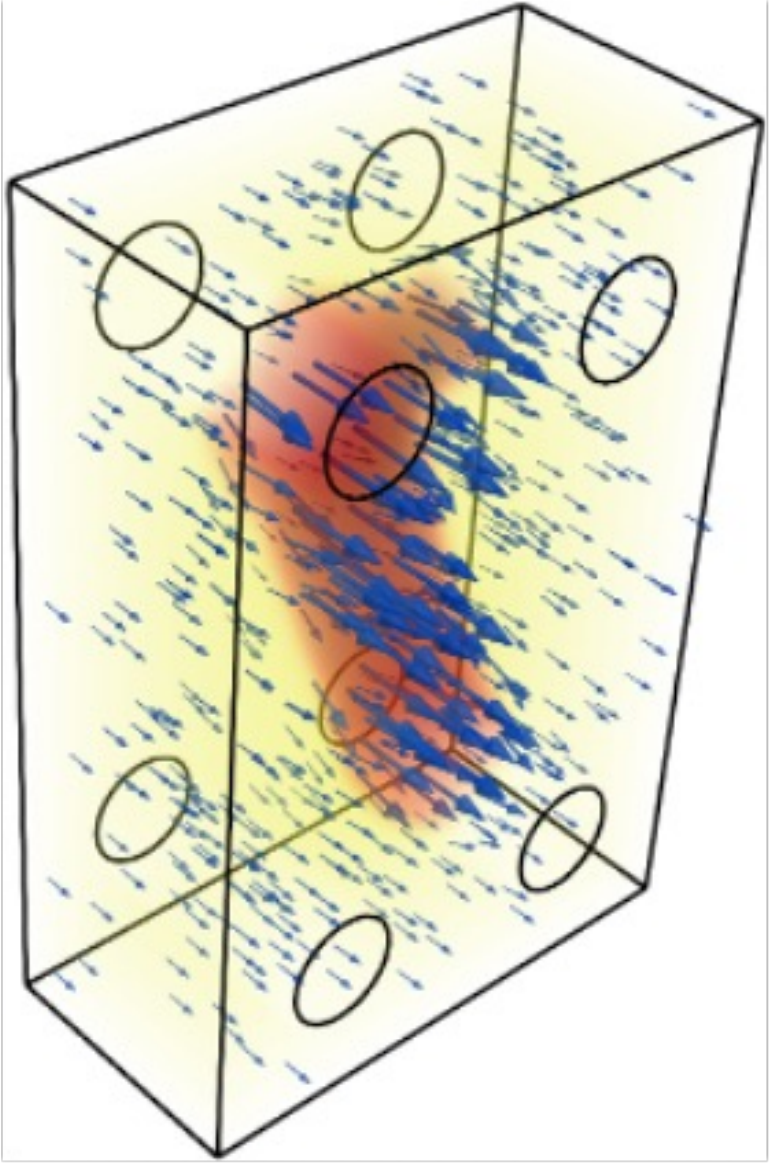}&
 \includegraphics[trim=0.011in 0.012in 0.01in 0.012in, clip,width=0.23\linewidth]{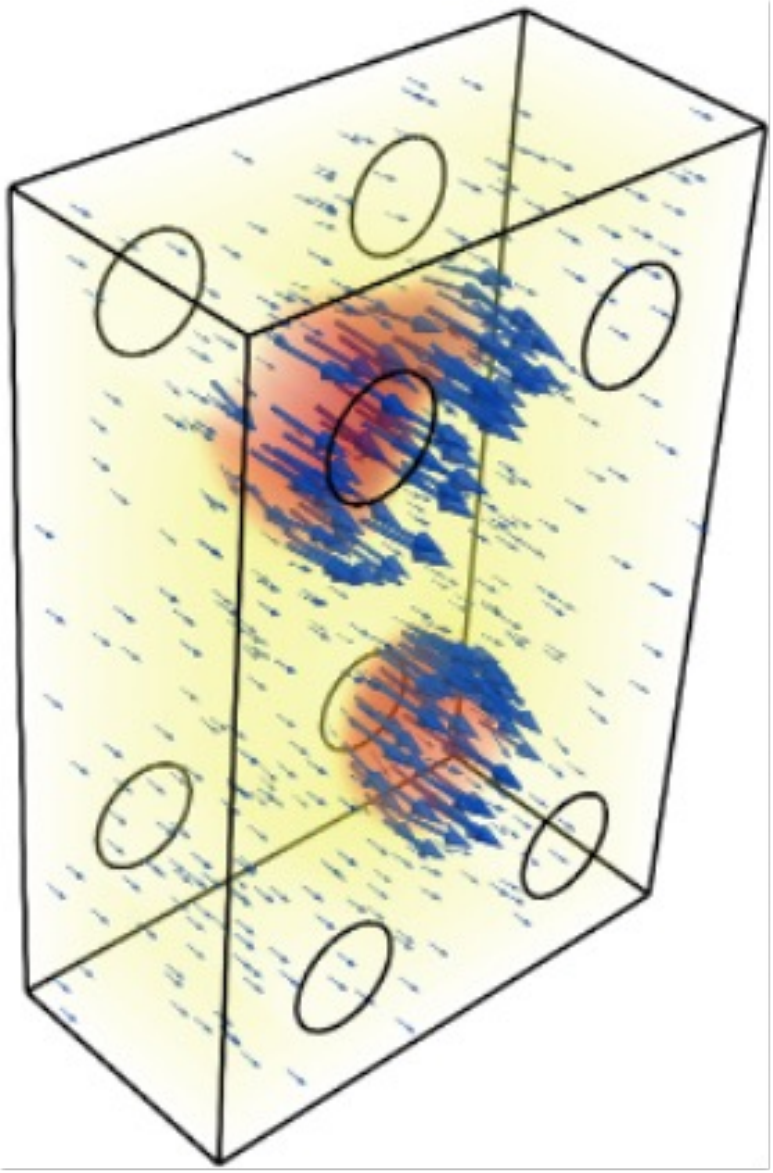}&
 \includegraphics[width=0.08\linewidth]{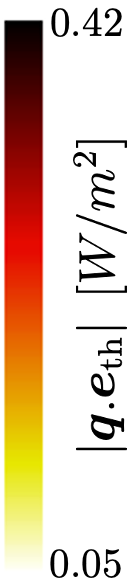}
 \\
 \includegraphics[trim=0.011in 0.012in 0.01in 0.012in, clip,width=0.23\linewidth]{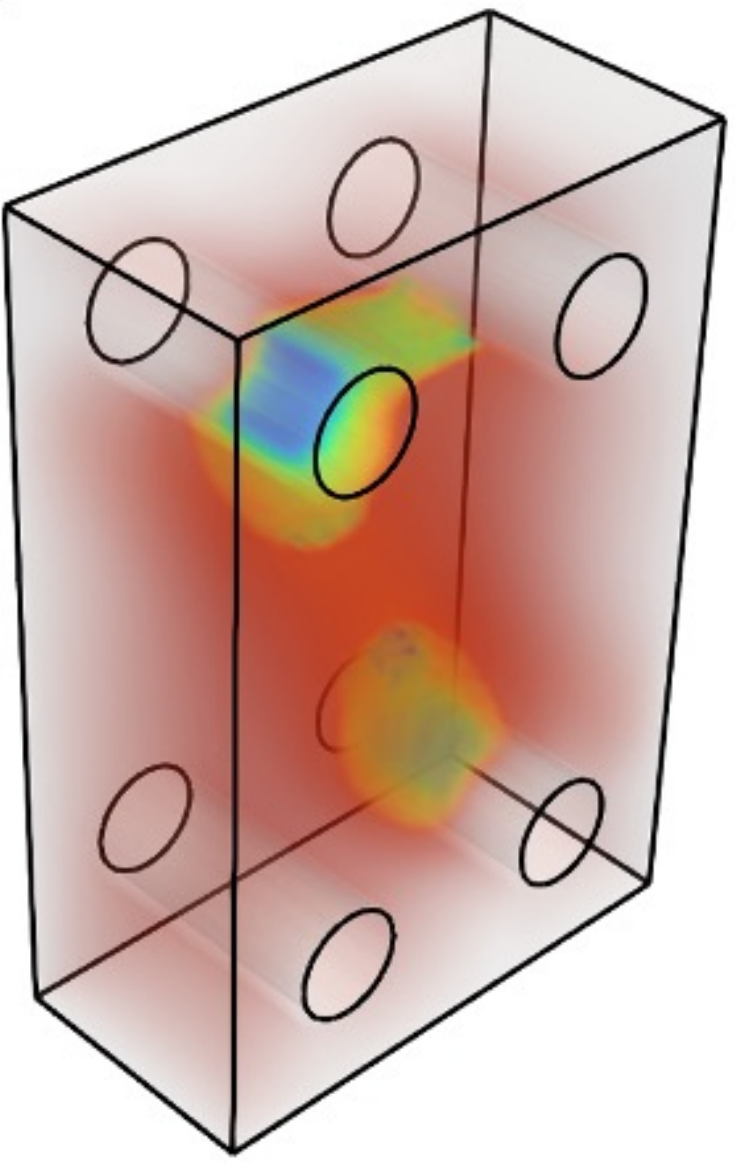}&
 \includegraphics[trim=0.011in 0.012in 0.01in 0.012in, clip,width=0.23\linewidth]{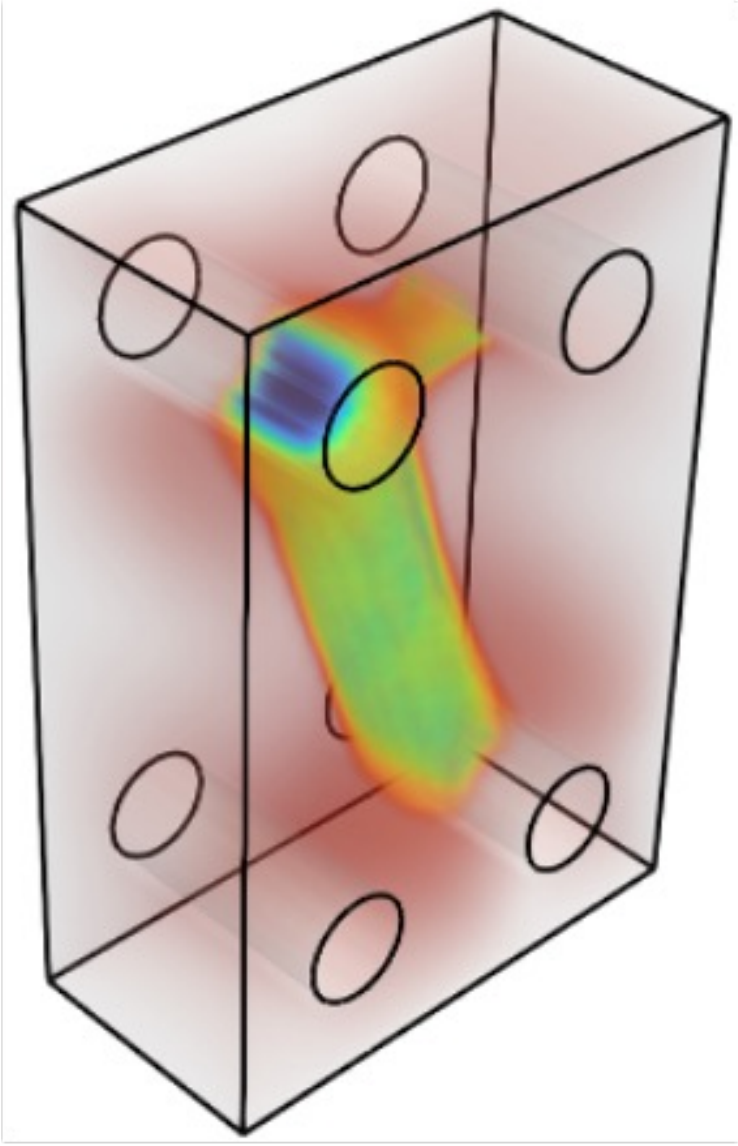}&
 \includegraphics[trim=0.011in 0.012in 0.01in 0.012in, clip,width=0.23\linewidth]{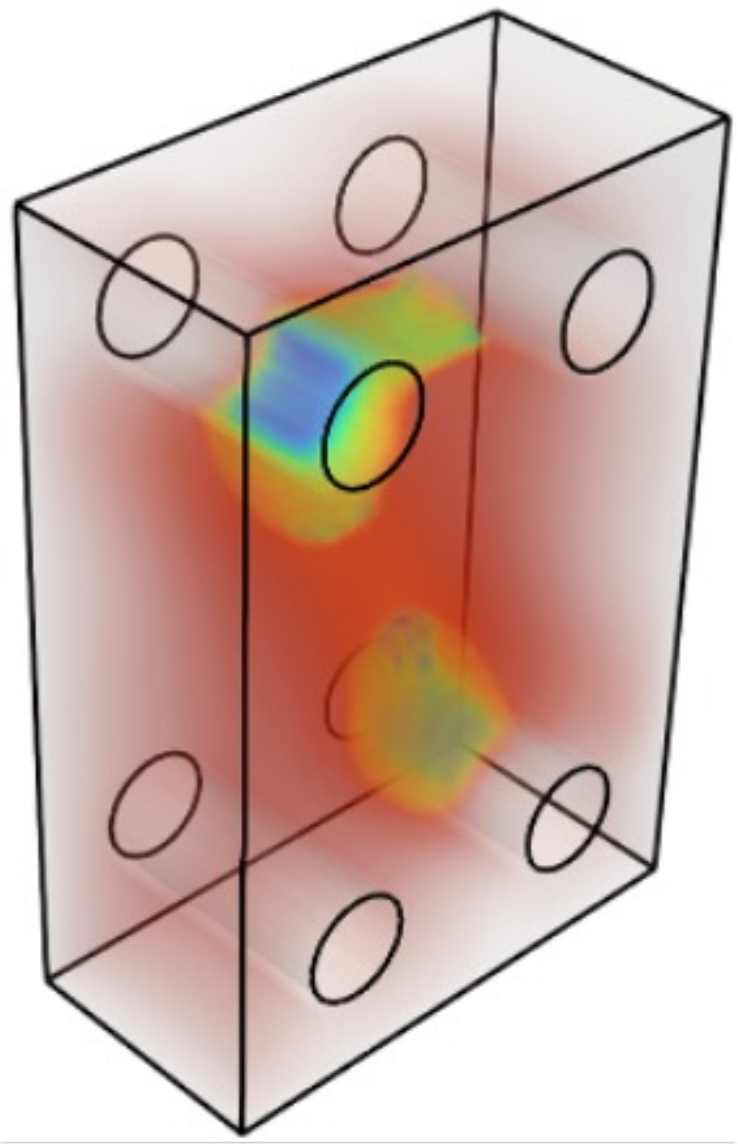}&
 \includegraphics[width=0.08\linewidth]{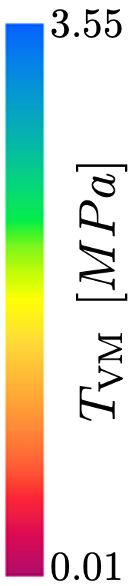}
 \\
\end{tabular}
    \end{minipage}
    \begin{minipage}{0.52\textwidth}
        \vspace{0.0in} 
        \hspace{-1.5in}\large{\textbf{d}}\vspace{-0.4in}\\
        \centering
\\
\hspace{-0.3in}\includegraphics[width=0.8\linewidth]{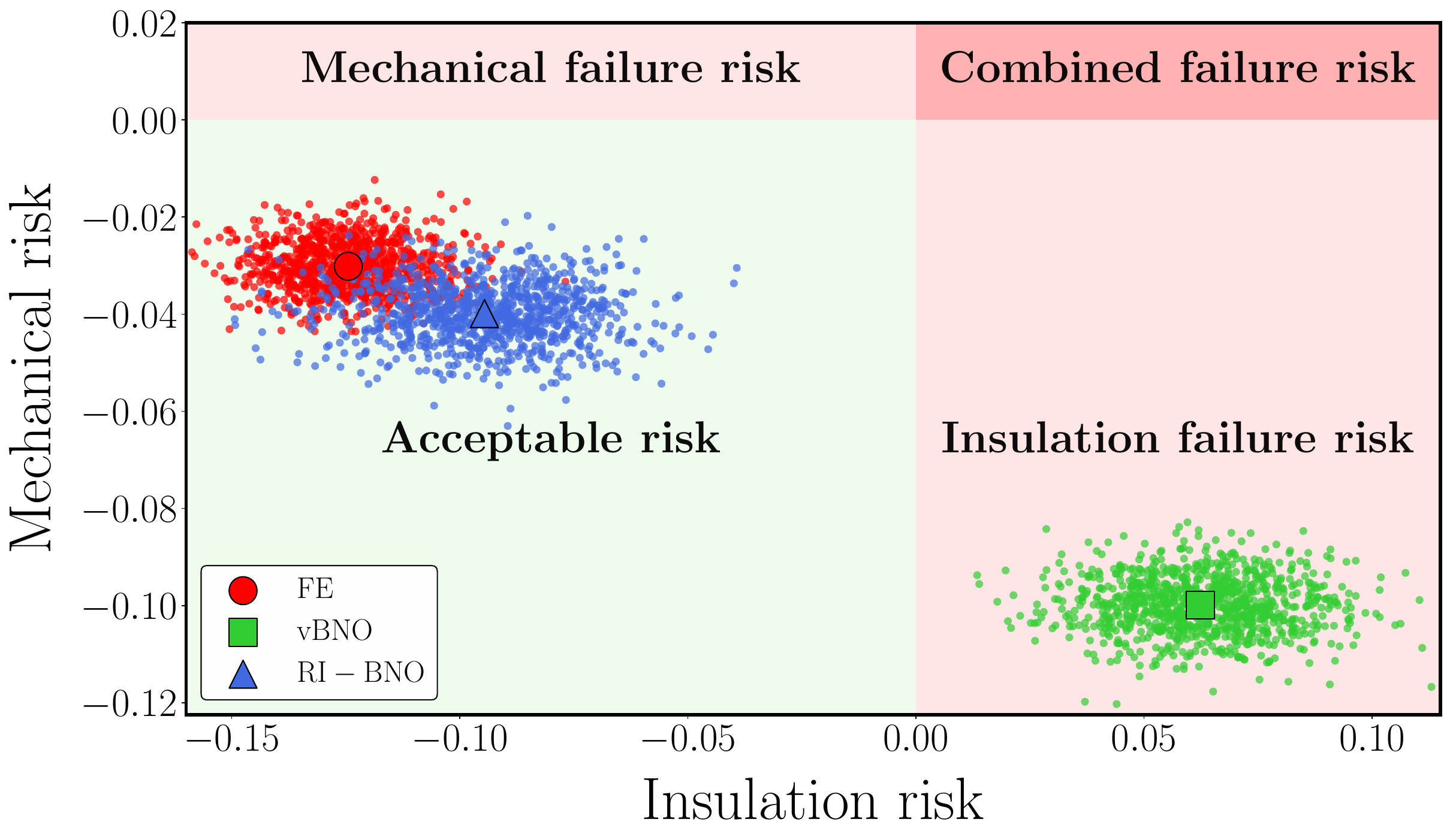}\vspace{-0.1in}\\
\hspace{-3.0in}\large{\textbf{e}}\vspace{-0.0in}\\
\hspace*{-0.3in}\includegraphics[width=0.43\linewidth]{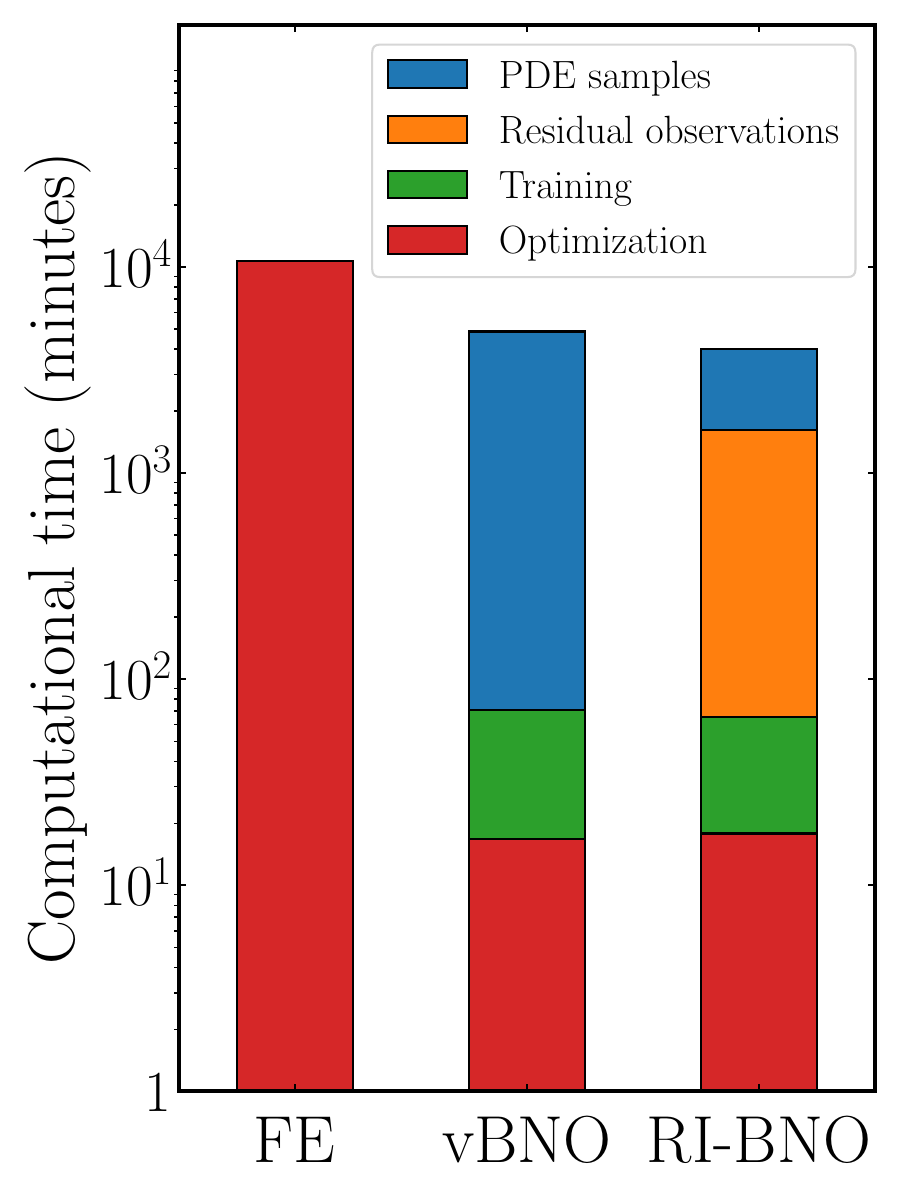} \hspace{-0.1in}
\includegraphics[width=0.42\linewidth]{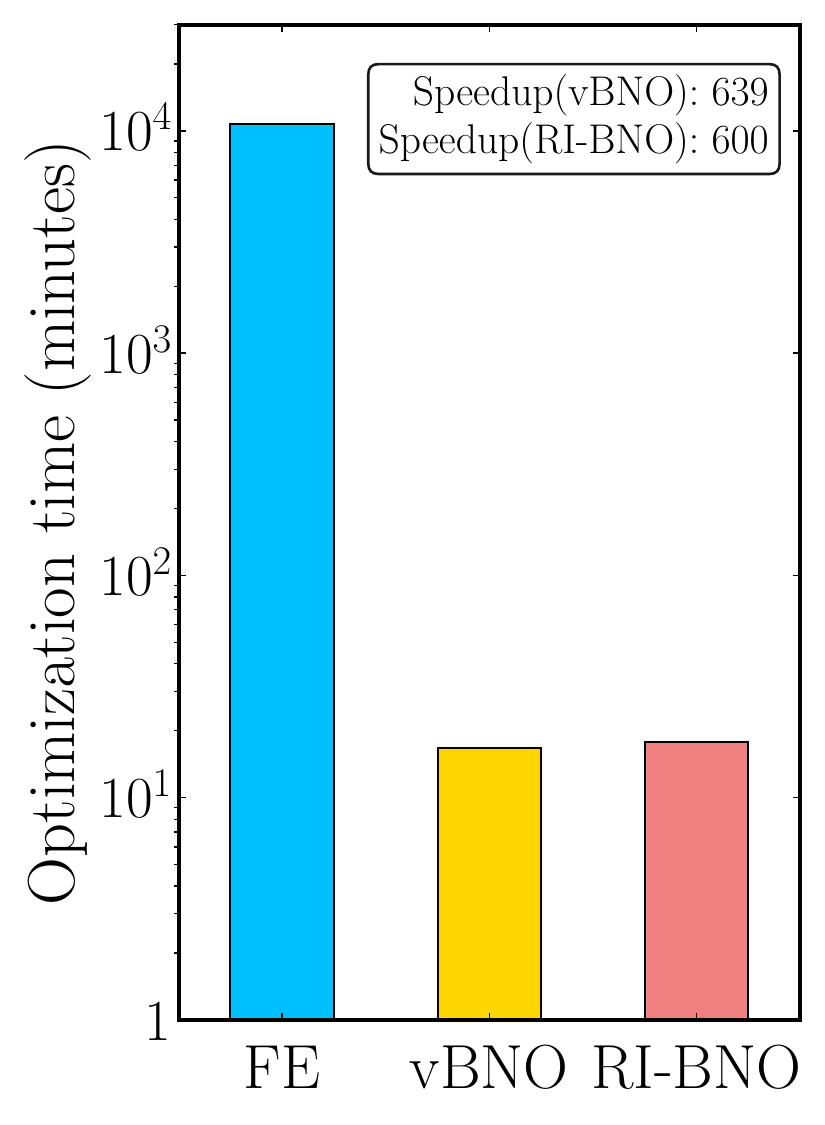} 
\\
    \end{minipage}
    \vspace{-0.3in}\\
    \caption{\textbf{Risk-averse design of a fiber-reinforced aerogel thermal break under spatial manufacturing uncertainty.} 
Panel (a) shows the thermal break inserted between hot (red) and cold (blue) steel beams, along with geometry and bolt-hole displacements.
Panel (b) illustrates the latent-subspace NO architecture. The NO learns the mapping from fiber volume fraction $\phi_f$ to the coupled PDE state $u$, while the heat flux $\bs q$ and von Mises stress $T_{\rm VM}$ are recovered through post-processing operators applied to $u$. The two rows present training realizations of $\phi_f$ and the corresponding through thickness heat flux magnitude $\lvert\bs q\cdot\bs e_{\rm th}\rvert$ and von Mises stress fields.
Panel (c) compares the optimal designs and physical fields obtained using FE-constrained, vBNO-accelerated and RI-BNO-accelerated optimization. The rows show, from top to bottom, the optimal design $d_{\rm opt}(\bs x)$, the through thickness heat flux magnitude with arrows showing direction and relative magnitude, and the von Mises stress. 
Panel (d) compares risk indicators \((r_Q,r_T)\) under  realizations \(m\sim\nu_{\rm man}\), with each optimized design evaluated using its corresponding forward model: FE, vBNO or RI-BNO. Small markers denote individual realizations, while large markers denote the marginal \(\mathrm{CVaR}_{0.95}\) values.
Panel (e) compares the FE optimization cost with the total construction and optimization costs of vBNO and RI-BNO, including nonlinear PDE data generation, residual-informed observation generation, operator training and online risk-averse optimization. The right plot reports the online optimization times, with speedups of \(639\times\) and \(600\times\) for vBNO and RI-BNO relative to FE, respectively.
     }
     \label{fig:thermal_break}
    \end{figure}

Steel elements penetrating the building envelope create localized thermal bridges that increase heat loss and condensation risk at interior surfaces. Motivated by practical
steel beam and canopy connections, Figure~\ref{fig:thermal_break}(a) considers
a thermal break pad inserted between the end
plates of a bolted steel splice
\cite{seacrmsca2023thermal,schoeck2018design}. The opposing pad faces
are exposed to warm and cold conditions across the building envelope, while
prescribed displacements along portions of the bolt-hole boundaries represent
localized load transfer through the fasteners.
We consider a thermal break made from wheat straw fiber reinforced silica aerogel \cite{guo2025wheat}. Although fiber reinforcement improves mechanical stability, its higher thermal conductivity can create conductive pathways. The design must therefore place fibers only where mechanically necessary while preserving the superinsulating aerogel matrix elsewhere.
For graded multimaterial additive manufacturing using a fiber-rich feedstock, we represent spatial deposition variability through the realized fiber volume fraction, represented through
$
    \phi_f(\bs x;d,m)
    =
    \operatorname{sigmoid}\!\left(d(\bs x)+m(\bs x)\right),
$
where \(d \in [0,1]\) is the nominal design field and \(m \sim \nu_{\rm man}\) is a centered Gaussian random
field describing spatially correlated layer-by-layer additive manufacturing variability \cite{singh2025chance, tan2024scalable}.
For each realization \(m\), we define the thermal objective \(Q\) and stress violation functional \(g\) as
\begin{equation}
    Q(d,m)
    = \int_\Omega
    \left[
    \bs q\cdot\bs e_{\rm th}
    \right]^2
    \,{\rm d}\bs x,
    \qquad
    g(d,m)=
    \left[ \frac{1}{|\Omega|}
    \int_\Omega
    T^p_{\rm VM}
    \,{\rm d}\bs x
    \right]^{1/p}
    -T_{\rm cr},
\end{equation}
where \(\bs e_{\rm th}\) is the through-thickness direction, so minimizing $Q$ suppresses thermal bridging.
Here \(|\Omega|\) is the domain volume, and \(p\) controls the emphasis on regions of high stress, and the threshold \(T_{\rm cr}\) defines the admissible stress level. The condition \(g(d,m)\leq0\) enforces the prescribed stress constraint, indicating mechanical feasibility.

To limit sensitivity to adverse realizations of the spatial manufacturing
variability, we formulate the design problem in terms of conditional value
at risk (CVaR), which penalizes the upper tails of both heat transfer and
mechanical stress:
\begin{equation}
\begin{aligned}
(d_{\rm opt},\zeta_{Q,\rm opt},\zeta_{g,\rm opt})
=
\underset{
d\in H^1(\Omega),\zeta_Q,\zeta_g\in\mathbb R
}{\operatorname{argmin}}
\quad&
\zeta_Q
+\frac{1}{1-\beta}
\mathbb E_{\nu_{\rm man}}
\left[
\bigl(Q(d,m)-\zeta_Q\bigr)_+
\right]
+\lambda_{\rm Reg}
    \int_{\Omega}
    \left\lvert\nabla d\right\rvert^2
    \,{\rm d}\bs x
\\
\text{subject to}\quad&
\zeta_g
+\frac{1}{1-\alpha}
\mathbb E_{\nu_{\rm man}}
\left[
\bigl(g(d,m)-\zeta_g\bigr)_+
\right]
\leq 0,
\\
&
\left\langle
\mathcal R\bigl(u(d,m);d,m\bigr),v
\right\rangle_{\mathcal V^\ast,\mathcal V}
=0,
\qquad \forall v\in\mathcal V.
\end{aligned}
\label{eq:risk_averse_design}
\end{equation}
where \((a)_+=\max(a,0)\), 
\(\zeta_Q\) and \(\zeta_g\) are the thermal and mechanical CVaR thresholds, and \(\lambda_{\rm Reg}\) controls spatial regularization of the design. We set \(\alpha=\beta=0.95\), so the objective penalizes the highest heat transfer realizations while the constraint limits the upper tail of the stress violation and mechanical failure.

Figure ~\ref{fig:thermal_break}(b) summarizes the neural operator construction. Both vBNO and RI-BNO learn the map from the uncertain fiber volume fraction field $\phi_f$ to the coupled PDE state $u$. The quantities of interest, namely the heat flux and von Mises stress fields$(\bs q,T_{\rm VM})$, are subsequently evaluated using parameter-free post-processing operator $\mathcal P$ acting on the predicted state.
Training data are generated around a spatially uniform nominal design, \(d(\bs x)=0.5\), using realizations of the manufacturing field \(m(\bs x)\), with the corresponding outputs obtained from finite element (FE) solutions of \eqref{eq:mixture_pde}. The resulting data capture manufacturing variability around a deliberately simple nominal design but do not include the strongly heterogeneous layouts encountered during optimization. Both models use the same latent-subspace architecture, with 25 active-subspace modes, 25 POD modes, and a 20-layer ResNet of width 100. Under the matched total observation budget \(N_{\rm tot}=2048\), vBNO uses \(N_{\rm train}=2048\) nonlinear PDE solution pairs, whereas RI-BNO uses \(N_{\rm train}=N_{\rm err}=1024\), replacing half of these pairs with lower cost residual-informed observations. During optimization, the trained operators provide Monte Carlo estimates of the tail risks, while automatic differentiation supplies the sensitivities required by Newton CG optimizer.
Figure~\ref{fig:thermal_break}(c) compares the optimal designs and physical fields obtained using FE-constrained, vBNO-accelerated, and RI-BNO-accelerated optimization. Each iteration uses \(5{,}000\) realizations of \(m\); at the adopted \(95\%\) risk levels, each empirical upper tail therefore contains 250 adverse realizations. The FE solution concentrates fibers around stress concentrations induced by bolt misalignment, forming mechanically stronger pathways between the displaced holes. vBNO instead produces a substantially more extensive fiber network, reducing stress beyond what is required while strengthening the thermal bridge through the more conductive fibers. This behavior is consistent with the restricted training coverage when optimization moves from a uniform nominal design to strongly heterogeneous layouts. RI-BNO closely recovers the FE-optimal fiber distribution, heat-flux field, and stress localization despite using only half as many nonlinear PDE training solutions.

To quantify the competing risks under manufacturing variability, we define the insulation-risk indicator \(r_Q=(Q-Q_{\rm cr})/Q_{\rm cr}\) and mechanical-risk indicator \(r_T=g/T_{\rm cr}\). Positive values indicate violation of the corresponding insulation or mechanical criterion. Figure~\ref{fig:thermal_break}(d) shows realizations of \((r_Q,r_T)\), together with the marginal tail-risk summaries \((\operatorname{CVaR}_{0.95}[r_Q],\operatorname{CVaR}_{0.95}[r_T])\). The FE and RI-BNO designs remain within the acceptable-risk regime and yield closely aligned tail-risk summaries. By contrast, the vBNO design shifts the response into the insulation-failure regime, such that its excessive fiber reinforcement drives the mechanical risk far below its admissible limit but creates a conductive pathway that compromises superinsulation.
Evaluating all three optimized designs using FE on a common independent set of \(5{,}000\) manufacturing realizations confirms that the RI-BNO design retains thermal and mechanical tail-risk performance close to the FE-optimized design, whereas the vBNO design exceeds the insulation-risk threshold (see Figure~\ref{fig:realization_design}).
Figure~\ref{fig:thermal_break}(e) compares the computational cost of FE-constrained optimization with the end-to-end vBNO and RI-BNO workflows, including offline data generation, residual-informed observation construction, operator training and online optimization solution \eqref{eq:risk_averse_design}. Once trained, vBNO and RI-BNO reduce the optimization time from \(178.13\) hours to \(16.7\) and \(17.8\) minutes, corresponding to online speedups of \(639\times\) and \(600\times\), respectively. Although vBNO is marginally faster online, RI-BNO requires fewer full nonlinear PDE solutions and reduces the total construction-and-optimization cost from \(81.1\) to \(67.0\) hours, corresponding to a saving of \(14.1\) hours, or \(17.4\%\). Thus, RI-BNO recovers FE-level thermal--mechanical risk control while substantially reducing the required nonlinear training data and retaining approximately \(600\times\) online acceleration.

%

\section{Discussion}\label{sec:conclusions}

The central finding of this study is that a Bayesian treatment of the neural operator weights alone did not account for structured inadequacy in approximating the governing PDE solution, allowing this discrepancy to propagate into downstream quantities. RI-BNO improved agreement with the FE reference across
a forward benchmark and two downstream applications. In the controlled reaction–diffusion benchmark, vBNO accuracy plateaued over the observation budgets examined, whereas RI-BNO continued to improve for both the latent subspace operator and DeepONet. In the MRI-driven glioma inversion, forward inadequacy altered the inferred mobility and proliferation fields and displaced posterior infiltration maps, while RI-BNO more closely reproduced the FE reference results. In the thermal break design problem, inadequacy changed the optimized fiber distribution and shifted its response into the insulation-failure risk regime, whereas RI-BNO recovered a design and tail risk profile consistent with FE-constrained optimization. These examples show how modest forward errors can materially alter posterior inference, estimated risks and optimized designs.

RI-BNO builds on residual-based error estimation, which uses the governing residual evaluated at a neural operator prediction to obtain approximate observations of unresolved state error \cite{CAO_jcp2023,JHA_cmame2024}. These residual-informed observations train a probabilistic inadequacy operator whose mean captures the systematic correction and whose covariance describes uncertainty in that correction. 
Both enter the likelihood used to infer the final neural operator
weight posterior from the original prior. The residual-informed
inadequacy model therefore informs both Bayesian training and
posterior prediction.
This formulation complements physics-informed operator learning,
which incorporates governing equations into the training objective
\cite{li2024physics}, by using residual-derived observations to learn
an explicit probabilistic model of operator inadequacy.
The preliminary operator and inadequacy learning concentrate PDE-based
error estimation in the offline construction stage, allowing online
predictions to use the trained neural operator and inadequacy model
without additional residual solves. The computational benefit depends
on whether the savings in reference simulations outweigh the costs of
preliminary training, residual-observation generation, inadequacy
learning and final Bayesian inference.
Under the observation budgets examined, RI-BNO used half as many
full FE training simulations, supplementing them with residual-informed
error observations.
Time-dependent problems offer a particular opportunity when the
temporal and coupled state information required for residual evaluation
can be obtained from operator predictions and limited auxiliary
computations. In the glioma example, this construction avoids generating
additional full transient trajectories of highly nonlinear phase-field model.
In the thermal break example, RI-BNO reduced the online optimization time from 178.13 hours to 17.8 minutes, corresponding to approximately (600$\times$) acceleration, while its total workflow cost was 17.4\% lower than that of vBNO. The computational benefit is greatest when the operator is reused across repeated inversions, uncertainty analyses or design studies and when linearized residual solves are substantially cheaper than additional nonlinear FE solutions.

The theoretical result is local and applies to the residual-informed estimator before inadequacy learning and full Bayesian inference. Under the stated stability and smoothness conditions, the observation error is quadratic in the local neural operator error, so the corresponding ideal corrected state has locally higher-order accuracy. This result should not be interpreted as a global guarantee or as a calibration result for the final RI-BNO posterior. Practical use also requires access to the governing residual and its state derivative. 
The sparse Gaussian process inadequacy model and modular inference
provide a scalable implementation, using Gaussian correction
uncertainty and an approximate treatment of dependence between the
neural operator weights and inadequacy field.
The results across two operator architectures and two downstream applications are encouraging, while broader studies across clinical cases, geometries, manufacturing regimes and calibration diagnostics would further establish the scope of the approach.
%
Future work could extend RI-BNO in several directions.
Residual-informed observations could be selected adaptively using
inadequacy uncertainty or the sensitivity of downstream quantities,
concentrating computation where it most affects an inverse problem,
risk measure or design objective \cite{prudhomme1999goal}.
Joint inference of the neural operator weights and inadequacy field
could better represent their posterior dependence.
Extending RI-BNO to operators learned from longitudinal observations
could support digital twins that combine information from governing
biophysical and physiological PDEs with incoming measurements to update predictions over time.
This extension would require accounting for inadequacy in the
underlying physical model alongside neural operator inadequacy.
Further studies across geometries, boundary conditions, patient cases
and manufacturing settings could establish the transfer capabilities
and limits of the approach. Prospective clinical studies and controlled
manufacturing experiments would assess predictive validity beyond
agreement with the FE reference.

RI-BNO provides a practical connection between residual-based error
estimation and Bayesian operator learning, allowing information about
systematic inadequacy to inform training and prediction.
Across the examples studied, this approach improved agreement with
FE-constrained inference and design while requiring fewer full reference
simulations. These findings support assessing neural operators
through the downstream conclusions and decisions they enable,
alongside their forward accuracy and computational speed.

\section*{Acknowledgments}
\noindent
Primary support for P.K.S. and D.F. was provided by the U.S. National Science Foundation (NSF) through CAREER Award CMMI-2143662. D.F. and E.B.P. also acknowledge partial support from the NSF FDT-BioTech program under Award DMS-2533946. The authors further acknowledge the computational resources provided by the Center for Computational Research at the University at Buffalo. Any opinions, findings and conclusions or recommendations expressed in this work are those of the authors and do not necessarily reflect the views of the NSF.


\clearpage \appendix 
\begin{center} 
{\Large Supplementary Information
\par} 
\end{center}
\vspace{0.5cm} 
\setcounter{section}{0} 
\renewcommand{\thesection}{\arabic{section}} \renewcommand{\thesubsection}{\thesection.\arabic{subsection}} 

\setcounter{figure}{0} \setcounter{table}{0} \setcounter{equation}{0} 
\renewcommand{\thefigure}{S\arabic{figure}} \renewcommand{\thetable}{S\arabic{table}} \renewcommand{\theequation}{S\arabic{equation}} 

\counterwithout{equation}{section}





\setcounter{figure}{0} \setcounter{table}{0} \setcounter{equation}{0} \renewcommand{\thefigure}{S\arabic{figure}} \renewcommand{\thetable}{S\arabic{table}} \renewcommand{\theequation}{S.\arabic{equation}}

\section{Neural operator architectures and Bayesian framework implementation}
\label{sec:la-lrsfn}
In this section, we discuss our Bayesian solution through the Laplace approximation and the description of neural operator architectures. This work evaluates RI-BNO using two neural operator architectures. The main results use a latent-subspace neural operator, and the Deep Operator Network (DeepONet) is used as an additional architecture to demonstrate that RI-BNO is not tied to a specific neural operator parameterization.
\subsection{Laplace approximation to the posterior}
Given training data \(\mathcal D\), Bayesian neural operator inference defines the posterior over network weights as
\begin{equation}\label{eq:bayes_si}
 \pi_{\mathrm{post}}(\bs w|\mathcal D)
 =
 \frac{
 \pi_{\mathrm{like}}(\mathcal D|\bs w)\,
 \pi_{\rm pr}(\bs w)}
 {\pi_{\rm evid}(\mathcal D)} .
\end{equation}
Direct inference in \(\pi_{\mathrm{post}}\) is intractable for modern neural operators because the number of weights is large and the posterior covariance cannot be formed explicitly. We therefore use a Laplace approximation, which represents the posterior by a Gaussian centered at the maximum a posteriori estimate, $\pi_{\mathrm{post}}(\bs w|\mathcal D) \approx \mathcal N(\bs w_{\rm MAP},\bs\Gamma_{\mathrm{post}})$.
The MAP point is obtained by minimizing the negative log posterior,
\begin{equation}\label{eq:map_training_si}
    \bs w_{\rm MAP}
    = \underset{\bs{w}}{\rm{argmin}} \; \{- \log \pi_{\mathrm{post}} (\bs{w} \mid \mathcal D) \} 
    =
    \underset{\bs{w}}{\rm{argmin}}
    \left\{
    -\log \pi_{\mathrm{like}}(\mathcal D|\bs w)
    -
    \log \pi_{\rm pr}(\bs w)
    \right\}.
\end{equation}
For a Gaussian prior with covariance \(\bs\Gamma_{\rm pr}\), the Laplace covariance is
\begin{equation}\label{eq:post_cov_si}
\bs \Gamma_{\mathrm{post}}
=
 \mathbf{H}_{\mathrm{post}}^{-1}
=
\left(
\mathbf{H}_{\mathrm{like}}
+
\bs \Gamma_{\rm pr}^{-1}
\right)^{-1},
\end{equation}
where 
$\mathbf{H}_{\rm{post}} = -\nabla^2_{\bs{w}} \left.\log \pi_{\mathrm{post}}(\bs w \mid \mathcal D) \right|_{\bs{w} = \bs{w}_{\rm{MAP}}}$
and
$\mathbf{H}_{\rm{like}} = -\nabla^2_{\bs{w}} \left.\log \pi_{\mathrm{like}}(\mathcal D \mid \bs{w}) \right|_{\bs{w} = \bs{w}_{\rm{MAP}}}$ denote the Hessians of the log posterior and log likelihood evaluated at the MAP point, respectively.
The main computational challenge is therefore to obtain useful curvature information near $\bs w_{\rm MAP}$ without forming or inverting the full Hessian. The LRSFN strategy used for this purpose is described next.

\subsection{MAP and posterior covariance estimation with Low Rank Saddle Free Newton}
The Laplace approximation requires both a MAP estimate and local curvature information at that point. For neural operators, this is challenging because the loss landscape is non-convex and the number of trainable weights is large. Forming or factorizing the full Hessian is therefore infeasible. We address both issues using the Low Rank Saddle Free Newton (LRSFN) method.
A key feature of our use of LRSFN is that the low-rank curvature information computed during MAP training is reused to construct the Laplace posterior covariance, rather than using LRSFN only as an optimizer.
Let
$\mathcal J(\bs w) = -\log \pi_{\mathrm{like}}(\mathcal D|\bs w) -\log \pi_{\rm pr}(\bs w)$
be the negative log posterior.
During MAP optimization, LRSFN approximates the dominant rank $r$ eigenspace of the likelihood Hessian using randomized singular value decomposition (SVD) with Hessian-vector products of the negative log-likelihood
\begin{equation}
    \mathbf{H}_{\mathrm{like}}
    \approx
    \mathbf{V}_r \bs \Lambda_r \mathbf{V}_r^T,
    \qquad
    \mathbf{V}_r\in\mathbb R^{N_w\times r},
    \quad
    \bs \Lambda_r\in\mathbb R^{r\times r}.
\end{equation}
Here, $\mathbf{V}_r$ contains the $r\ll N_w$ retained dominant eigenvectors of the likelihood curvature, and $\bs \Lambda_r$ is a diagonal matrix containing the corresponding $r$ eigenvalues. 
To obtain a saddle free search direction, the eigenvalues are replaced by their absolute values,
$|\mathbf{H}_{\mathrm{like}}|_r =\mathbf{V}_r |\bs \Lambda_r| \mathbf{V}_r^T$ \cite{oleary2019inexact}.
The resulting positive curvature approximation is combined with the prior precision,
$\bs B=|\mathbf{H}_{\mathrm{like}}|_r + \bs \Gamma_{\rm pr}^{-1}$,
and the LRSFN update direction is
$\Delta \bs w = - \bs B^{-1}\nabla \mathcal J(\bs w)$.
Using the Sherman-Morrison-Woodbury identity, this inverse action is evaluated as
\[
    \Delta \bs w
    =
    -
    \left[
    \bs \Gamma_{\rm pr}
    -
    \bs \Gamma_{\rm pr}
    \mathbf{V}_r
    \left(
    |\bs \Lambda_r|^{-1}
    +
    \mathbf{V}_r^T
    \bs \Gamma_{\rm pr}
    \mathbf{V}_r
    \right)^{-1}
    \mathbf{V}_r^T
    \bs \Gamma_{\rm pr}
    \right]
    \nabla \mathcal J(\bs w).
\]
For an isotropic Gaussian prior, \(\bs\Gamma_{\rm pr}=\sigma_{\rm pr}^2\bs I\), the inner matrix is only \(r\times r\), so the update avoids dense Hessian storage and the \(\mathcal O(N_w^3)\) cost of full Newton methods.

After MAP convergence, the retained low-rank basis vectors $\mathbf{V}_r$ are reused to construct the Laplace covariance, avoiding a new randomized SVD. Importantly, the covariance is constructed from the true likelihood curvature at the MAP point, rather than the modified absolute Hessian used to stabilize optimization. 
In practice, we keep the basis $\mathbf{V}_r$ fixed and recompute the signed curvature eigenvalues $\bs\Lambda_r$ on this retained subspace at \(\bs w_{\rm MAP}\), yielding the updated approximation
$\mathbf{H}_{\mathrm{like}}(\boldsymbol{w}_{\rm MAP})\approx\mathbf{V}_r \bs \Lambda_r \mathbf{V}_r^T$.
This requires only $r$ Hessian-vector products and avoids forming or factorizing the full $N_w\times N_w$ Hessian.
Using the Sherman-Morrison-Woodbury formula, the posterior covariance is then approximated by
\begin{equation}
\bs \Gamma_{\mathrm{post}}
=
\left(
\bs \Gamma_{\rm pr}^{-1}
+
\mathbf{H}_{\mathrm{like}}
\right)^{-1}
\approx
\bs \Gamma_{\rm pr}
-
\bs \Gamma_{\rm pr}
\mathbf{V}_r
\left(
\bs \Lambda_r^{-1}
+
\mathbf{V}_r^T
\bs \Gamma_{\rm pr}
\mathbf{V}_r
\right)^{-1}
\mathbf{V}_r^T
\bs \Gamma_{\rm pr}.
\end{equation}
Consequently, curvature refresh scales with the cost of $r$ Hessian-vector products, while all subsequent covariance construction and inversion operations are restricted to the $r$-dimensional subspace. Forming the low-rank covariance requires only matrix products involving $\mathbf V_r$ together with the inversion of an $r\times r$ matrix, yielding an overall cost of $\mathcal O(N_w r^2 + r^3)$ and storage requirement of $\mathcal O(N_w r)$. Since $r\ll N_w$, this represents a substantial reduction from the $\mathcal O(N_w^3)$ computational cost and $\mathcal O(N_w^2)$ storage requirements associated with dense Hessian methods.
In the reported experiments, the retained signed eigenvalues of the likelihood Hessian at the MAP estimate were nonnegative. The Laplace covariance is constructed from the posterior precision \(\mathbf H_{\rm post}=\mathbf H_{\rm like}+\boldsymbol\Gamma_{\rm pr}^{-1}\), which must be positive definite. Negative eigenvalues of the likelihood Hessian alone do not invalidate this construction, provided the prior contribution makes the posterior precision positive definite. Low-rank posterior approximations of this form are also consistent with optimal low-rank constructions for Bayesian inverse problems \cite{spantini2015optimal}.
\subsection{Latent-subspace neural operator architecture}
The main neural operator architecture used in this work is a latent-subspace neural operator following \cite{CAO_jcp2023,ohippyflow,OLeary2022derivative,OLeary2024derivative,cao2025derivative}. The input parameter field is projected onto a low dimensional active subspace \cite{zahm2020gradient,constantine2014active, constantine2016accelerating}, the output state field is represented in a Proper Orthogonal Decomposition (POD) basis \cite{manzoni2016dimensionality}, and a stack of residual network (ResNet) layers \cite{he2016deep} learns the map between the resulting latent coordinates. This construction avoids learning directly between high dimensional nodal fields and
keeps the trainable map independent of the dimensions of the full discretized input and output fields.

Let \(m\in\mathcal M\) denote the parameter field and
\(u_h=\mathcal F_h(m)\in\mathcal U\) the corresponding state. The input
encoder is constructed from the leading active subspace directions
\(\bs\Psi_m=[\bs\psi_1,\ldots,\bs\psi_{r_m}]\), obtained from the empirical
Jacobian information matrix of the parameter-to-state map. 
$\bs H_{\rm AS} = \mathbb E_{\varphi} \left[ \bs J_m(m)^T \bs J_m(m) \right],$
where \(\bs J_m(m)=\partial \mathcal F_h/\partial m\) is the Jacobian of the discrete solution map with respect to the input parameter and $m\sim\varphi=\mathcal{N}(\bar m, \bs C_m)$.
The active subspace directions are the eigenvectors associated with the largest eigenvalues of the generalized eigenvalue problem
\[
    \bs H_{\rm AS} \bs\psi_i = \lambda_i^{\rm AS} \bs C_m^{-1} \bs\psi_i,
\]
where the basis vectors are normalized such that the $\boldsymbol{C}_m^{-1}$-orthogonality condition holds $\bs\Psi_m^T \bs C_m^{-1} \bs\Psi_m = \mathbf{I}_{r_m}$.
Then the reduced input
coordinates are
\[
    \boldsymbol{\xi}_m
    =
    \bs\Psi_m^T \bs C_m^{-1}(m-\bar m),
\]
where \(\bs C_m\) is the input covariance. 
The number of active subspace
directions \(r_m\) is chosen as the smallest integer satisfying
\[
    \frac{\sum_{i=1}^{r_m}\lambda_i^{\rm AS}}
         {\sum_{i=1}^{r_{\rm AS}}\lambda_i^{\rm AS}}
    \geq \tau_{\rm AS},
\]
where \(\lambda_i^{\rm AS}\) denotes the \(i\)th active subspace eigenvalue, ordered as
\(\lambda_1^{\rm AS}\geq \lambda_2^{\rm AS}\geq \cdots\), and quantifies the average sensitivity of the parameter-to-state map along the corresponding active direction. 
Throughout this work, we use \(\tau_{\rm AS}=0.99\) as in \cite{CAO_jcp2023,OLeary2024derivative,cao2025derivative}, unless otherwise stated.
The output decoder is constructed from a POD basis
\(\bs\Phi_u=[\bs\phi_1,\ldots,\bs\phi_{r_u}]\), computed from centered solution snapshots using the state-space inner product. For a given discretization, let \(\bs M_u\) denote the matrix associated with this inner product. The POD basis satisfies the orthonormality condition
$\bs\Phi_u^T\bs M_u\bs\Phi_u=\mathbf I_{r_u}.$
The state is represented as
\[
    u_h(m)
    \approx
    \bar u_h+\bs\Phi_u\boldsymbol{\alpha}(m),
\]
where \(\boldsymbol{\alpha}(m)\in\mathbb R^{r_u}\) are the POD coefficients.
The number of POD modes \(r_u\) is chosen as the smallest integer satisfying
\[
    \frac{\sum_{i=1}^{r_u}\lambda_i^{\rm POD}}
         {\sum_{i=1}^{r_{\rm POD}}\lambda_i^{\rm POD}}
    \geq \tau_{\rm POD},
\]
where \(\lambda_i^{\rm POD}\) denotes the \(i\)th POD eigenvalue, ordered as
\(\lambda_1^{\rm POD}\geq \lambda_2^{\rm POD}\geq \cdots\), and represents the energy captured by the corresponding POD mode.
In the numerical studies, we used \(\tau_{\rm POD}=0.99\), unless otherwise stated.

The trainable part of the latent-subspace neural operator is
\(\mathcal S_{\bs w}:\mathbb R^{r_m}\rightarrow\mathbb R^{r_u}\), composed of ResNet layers, which maps
active subspace coordinates to POD coefficients $\boldsymbol{\alpha}_{\bs w}=\mathcal S_{\bs w}(\boldsymbol{\xi}_m)$.
The latent-subspace neural operator is then
\[
    \mathcal F_{\bs w}(m)
    =
    \bar u_h+\bs\Phi_u\boldsymbol{\alpha}_{\bs w}(m).
\]
Thus, the trainable network learns only the nonlinear map between the latent coordinates, while the active subspace and POD operators provide the projection and lifting between the physical fields and the latent representations.
The retained dimensions $r_m$ and $r_u$, together with the number, width and depth of the ResNet layers, define the neural operator architecture used for each numerical example and are reported with the corresponding results.

\subsection{DeepONet architecture}
As an additional neural operator architecture, we use a Deep Operator Network (DeepONet) \cite{lu2021learning}, a well-established neural operator framework motivated by the universal approximation theorem for nonlinear operators \cite{chen1995universal}. DeepONet has demonstrated strong performance across a broad spectrum of operator learning problems in science and engineering \cite{lin2023operator,yin2022simulating,cai2021deepm,lin2021operator,lu2022comprehensive}.
DeepONet represents
the operator by separating the dependence on the input function from the
dependence on the output coordinates. The branch network $\mathcal B_{\bs w_b}:
    \mathbb R^{n_s}\rightarrow \mathbb R^{q}$ maps discrete sensor
evaluations of the input parameter field to latent coefficients,
$
    \bs b
    =
    \mathcal B_{\bs w_b}
    \left(
    m(\bs x_1),\ldots,m(\bs x_{n_s})
    \right),
$
where \(\{\bs x_j\}_{j=1}^{n_s}\) are sensor locations. 
The trunk network $\mathcal T_{\bs w_t}: \Omega\rightarrow \mathbb R^{q}$ maps
the evaluation coordinate to latent basis functions,
$\bs t = \mathcal T_{\bs w_t}(\bs x)$.
The DeepONet prediction is then
\[
    \mathcal F_{\bs w}(m)
    =
    \sum_{\ell=1}^{q}
    b_{\ell}(m)\,
    t_{\ell}(\bs x)
    + b_0,
\]
where \(\bs w=(\bs w_b,\bs w_t)\) denotes the trainable weights and \(b_0\) is a
trainable bias. Thus, the branch network encodes the input field, while the
trunk network provides coordinate dependent basis functions for evaluating the
predicted state at arbitrary spatial locations.
The Bayesian DeepONet baseline in this work is implemented using the variational inference formulation proposed by \cite{garg2023vb}.
\clearpage
\subsection{Gaussian random field prior with Mat\'ern type covariance}
Spatially varying input parameters is represented using a Gaussian random field with a Matérn type covariance structure. Specifically, the parameter field $m$ is assumed to follow the Gaussian measure
$m \sim \mathcal{N}(\bar{m},\mathcal{C})$,
where $\bar{m}$ denotes the prior mean and $\mathcal{C}$ is the covariance operator.

Rather than constructing the covariance matrix explicitly, realizations of the random field are generated through the solution of the stochastic elliptic problem
\begin{equation}
    \begin{aligned}
       -\gamma_{\rm M} \nabla \cdot \bigl(\boldsymbol{\Theta}\nabla (m - \bar{m})\bigr) + \delta (m - \bar{m}) &=  \: \dot{W} \quad \text{in} \quad \Omega \\
        \frac{\sqrt{\gamma_{\rm M} \delta}}{1.42} \: (m - \bar{m})  +  \gamma_{\rm M} (\boldsymbol{\Theta}\nabla (m - \bar{m})) \cdot \: \mathbf{n} &=  \: 0 \quad \text{on} \quad \partial \Omega
    \end{aligned}
    \label{eq:matern_pde}
\end{equation}
where $\dot W$ denotes spatial Gaussian white noise and $\mathbf{n}$ is the outward unit normal on the boundary.
The positive constants $\gamma_{\rm M}$ and $\delta$ determine the statistical properties of the field which are the correlation length $L_{\mathrm{CR}}$ and the marginal variance $\sigma^2$. 
The correlation length $L_{\mathrm{CR}}$ is defined in $n_d$ spatial dimensions as
\begin{equation}
    L_{\mathrm{CR}} = \sqrt{\frac{(16 - 4 n_d)\gamma_{\rm M}}{\delta}},
    \label{eq:correlation_length}
\end{equation}
while the marginal variance $\sigma^2$ is given by
\begin{equation}
    \sigma^2 = \frac{\mathrm{\Gamma_E}\left(2 - \frac{n_{d}}{2}\right)}{(4\pi)^{n_{d}/2} \, \gamma_{\rm M}^{{n_d}/2} \, \delta^{2 - {n_d}/2}},
    \label{eq:marginal_variance_d}
\end{equation}
where $\mathrm{\Gamma_E}(\cdot)$ denotes the Gamma function.
The Robin boundary condition is scaled by the factor $\sqrt{\delta\gamma_{\rm M}}/1.42$, following \cite{roininen2014}, to reduce artificial boundary effects in finite domain realizations. The tensor $\boldsymbol{\Theta}$ specifies the anisotropy of the covariance and allows for different correlation lengths along different spatial directions. Specially, for the 3D thermal break design problem, we have $\boldsymbol{\Theta} = \rm{diag}$ $(a_x, a_y, a_z)$, where $a_x \gg a_y, a_z$ to mimic the layer-by-layer manufacturing process, resulting in a larger correlation length along the deposited layers than across them.

\clearpage

\section{Residual-informed error estimation}
\subsection{A posteriori estimation of neural operator inadequacy}

Assume that the residual functional \(\mathcal R(m,u)\) is twice
G\^{a}teaux differentiable with respect to the state \(u\). For
directions \(v,q\in\mathcal U\), the first and second variations are
\begin{align}
    \delta_u \mathcal{R}(m,u)[v]
    &:=
    \lim_{\epsilon \to 0}
    \frac{
    \mathcal{R}(m,u+\epsilon v)-\mathcal{R}(m,u)}
    {\epsilon}, \\
    \delta_u^2 \mathcal{R}(m,u)[v,q]
    &:=
    \lim_{\epsilon \to 0}
    \frac{
    \delta_u\mathcal{R}(m,u+\epsilon q)[v]
    -
    \delta_u\mathcal{R}(m,u)[v]}
    {\epsilon}.
\end{align}
Let \(u=\mathcal F(m)\) be the PDE solution satisfying
\(\mathcal R(m,u)=0\), and let
\(u_{\bs w}=\mathcal F_{\bs w}(m)\) be the neural operator prediction.
The neural operator error is $e := u-u_{\bs w}$.
Expanding the residual about \(u_{\bs w}\) gives
\begin{equation}\label{eq:taylor_expansion}
\begin{aligned}
0
&=
\mathcal{R}(m,u)
=
\mathcal{R}(m,u_{\bs w}+e) \\
&=
\mathcal{R}(m,u_{\bs w})
+
\delta_u \mathcal{R}(m,u_{\bs w})[e]
+
\int_0^1 (1-s)\,
\delta_u^2 \mathcal{R}(m,u_{\bs w}+se)[e,e]\,ds .
\end{aligned}
\end{equation}
Neglecting the second-order remainder yields the first order residual
error equation \cite{CAO_jcp2023, JHA_cmame2024}
\begin{equation}\label{eq:error_state_variational}
    \delta_u \mathcal{R}(m,u_{\bs w})[e_{\rm est}]
    =
    -
    \mathcal{R}(m,u_{\bs w}),
\end{equation}
where $\mathcal{R}(m, u_{\bs w})$ indicates the degree to which $u_{\bs w}$ fails to satisfy the PDE.
This equation defines the a posteriori neural operator inadequacy
estimate following earlier error estimation techniques in \cite{prudhomme1999goal, oden2002estimation, ODEN2001}.
When the linearized residual operator is
invertible, the estimate can be written as
\begin{equation}\label{eq:error_state_inverse}
    e_{\rm est}
    =
    -
    \left[
    \delta_u \mathcal{R}(m,u_{\bs w})
    \right]^{-1}
    \mathcal{R}(m,u_{\bs w}).
\end{equation}
Equation \eqref{eq:error_state_variational} is equivalent to one Newton
step for solving \(\mathcal R(m,u)=0\), initialized at \(u_{\bs w}\). Thus,
under the standard local assumptions of Newton--Kantorovich theory and
when \(u_{\bs w}\) is sufficiently close to \(u\), the corrected state
\(u_{\bs w}+e_{\rm est}\) is a locally higher order approximation to the
PDE solution \cite{CAO_jcp2023, jha2022goal}.

\subsection{Residual-informed approximation of neural operator error}

Let $\mathcal M$, $\mathcal U$, and $\mathcal V$ be Banach spaces, and let
$\nu_{\mathcal{M}}$ be the Borel probability measure describing the distribution of inputs
in $\mathcal M$. Consider a residual
$\mathcal R:\mathcal M\times\mathcal U\to\mathcal V^\ast$ and strongly
$\nu_{\mathcal{M}}$-measurable operators
$\mathcal F,\mathcal F_{\bs w}:\mathcal M\to\mathcal U$, where
\begin{equation}
    \mathcal R\left(m,\mathcal F(m)\right)=0
    \quad\text{for }\nu_{\mathcal{M}}\text{-almost every }m.
    \label{eq:residual_consistency}
\end{equation}
Fix the neural operator weights $\bs w$ and define
\begin{equation}
    \eta(m):=\mathcal F(m)-\mathcal F_{\bs w}(m),
    \qquad
    A_m:=\delta_u\mathcal R
    \left(m,\mathcal F_{\bs w}(m)\right).
    \label{eq:inadequacy_and_linearization}
\end{equation}
We assume that $\mathcal F(m)$ and $\mathcal F_{\bs w}(m)$ satisfy the same
strongly imposed boundary and initial conditions. Otherwise, the following
analysis applies to the corresponding lifted homogeneous state variables.
Suppose there exist constants $\gamma>0$ and $L>0$ that uniformly bound the inverse stability and derivative Lipschitz constants over the admissible parameter regime, and a measurable radius
$r(m)>0$, determined by the residual problem and
independent of neural operator training. For $\nu_{\mathcal{M}}$-almost every $m$, assume
\begin{equation}
    \|\eta(m)\|_{\mathcal U}\leq r(m),
    \label{eq:newton_neighborhood_condition}
\end{equation}
and let
$B_m:=\overline B_{\mathcal U}(\mathcal F(m),r(m))$.
Assume that $\mathcal R(m,\cdot)$ is continuously Fr\'{e}chet differentiable
on an open neighborhood of $B_m$ and, for $v,z\in B_m$,
\begin{align}
    \left\|{\left[\delta_u\mathcal R(m,v)\right]}^{-1}\right\|_{
    \mathcal L(\mathcal V^\ast,\mathcal U)}
    &\leq\gamma,
    \nonumber\\
    \|\delta_u\mathcal R(m,v)-\delta_u\mathcal R(m,z)\|_{
    \mathcal L(\mathcal U,\mathcal V^\ast)}
    &\leq L\|v-z\|_{\mathcal U}.
    \label{eq:newton_stability_assumptions}
\end{align}
Define the residual-informed error estimate by
\begin{equation}
    A_m e_{\rm est}(m)
    =-\mathcal R\left(m,\mathcal F_{\bs w}(m)\right).
    \label{eq:residual_error_estimate_si}
\end{equation}
This is one Newton correction initialized at the neural operator
prediction~\cite{Ortega1968,CAO_jcp2023,JHA_cmame2024}.
We define a global constant $\omega_{\mathcal R}$ that characterizes the nonlinear conditioning and curvature of the underlying PDE residual
\begin{equation}
    \omega_{\mathcal R}
    :=\frac{1}{2}\gamma L.
    \label{eq:residual_conditioning_weight}
\end{equation}
The constants $\gamma$, $L$, and the radius function $r(m)$ are fixed a priori from analysis of the
residual problem; they are not enlarged after observing a trained neural
operator. Condition~\eqref{eq:newton_neighborhood_condition} only determines
whether a given neural operator prediction lies in the regime covered by the theorem.

\begin{ribnotheorem}[Residual-informed operator error bound]\label{thm:residual_operator_error}%
Under equations~\eqref{eq:newton_neighborhood_condition}
and~\eqref{eq:newton_stability_assumptions}, for $\nu_{\mathcal{M}}$-almost every $m$,
\begin{equation}
    \|e_{\rm est}(m)-\eta(m)\|_{\mathcal U}
    \leq
    \omega_{\mathcal R}\|\eta(m)\|_{\mathcal U}^2.
    \label{eq:pointwise_residual_error_bound}
\end{equation}
If $e_{\rm est}$ is strongly measurable and
$\|\eta(\cdot)\|_{\mathcal U}^2
\in L^1_{\nu_{\mathcal{M}}}(\mathcal M)$, then
\begin{equation}
    \|e_{\rm est}-\eta\|_{L^1_{\nu_{\mathcal{M}}}(\mathcal M;\mathcal U)}
    \leq
    \omega_{\mathcal R}\mathbb E_{m\sim\nu_{\mathcal{M}}}
    \left[
    \|\eta(m)\|_{\mathcal U}^2
    \right].
    \label{eq:residual_operator_bound}
\end{equation}
\end{ribnotheorem}

\subsection{Proof of Theorem~\ref{thm:residual_operator_error}}

\begin{proof}
For $\nu_{\mathcal{M}}$-almost every $m$, the fundamental theorem of calculus gives
\begin{align}
    0
    &=\mathcal R\left(m,\mathcal F(m)\right)
    \nonumber\\
    &=\mathcal R\left(m,\mathcal F_{\bs w}(m)\right)
    +A_m\eta(m)+ \Delta_{\mathcal{R}}(m),
    \label{eq:residual_expansion}\\
    \Delta_{\mathcal{R}}(m)
    &:={\int}_0^1
    \left[
    \delta_u\mathcal R
    \left(m,\mathcal F_{\bs w}(m)+t\eta(m)\right)-A_m
    \right]\eta(m)\,\mathrm dt.
    \nonumber
\end{align}
Combining equations~\eqref{eq:residual_error_estimate_si}
and~\eqref{eq:residual_expansion} yields
\begin{equation}
    A_m\left(e_{\rm est}(m)-\eta(m)\right)=\Delta_{\mathcal{R}}(m).
\end{equation}
Because $\mathcal F_{\bs w}(m)+t\eta(m)\in B_m$ for $t\in[0,1]$,
equation~\eqref{eq:newton_stability_assumptions} implies
\begin{equation}
    \|\Delta_{\mathcal{R}}(m)\|_{\mathcal V^\ast}
    \leq
    \int_0^1tL\|\eta(m)\|_{\mathcal U}^2\,\mathrm dt
    =\frac{L}{2}\|\eta(m)\|_{\mathcal U}^2.
\end{equation}
Applying the inverse-stability bound in
equation~\eqref{eq:newton_stability_assumptions} proves
equation~\eqref{eq:pointwise_residual_error_bound}.
Integrating equation~\eqref{eq:pointwise_residual_error_bound} proves
equation~\eqref{eq:residual_operator_bound}.
\end{proof}

The pointwise estimate is the standard one-step Newton correction
bound~\cite{CAO_jcp2023,JHA_cmame2024}. Integrating it with respect to the input
measure gives the operator level statement used by the RI-BNO methodology.
Thus, the theorem is a consistency result for the residual-informed
observation operator rather than a new Newton convergence theorem.

For any strongly measurable learned correction
$\widetilde\eta:\mathcal M\to\mathcal U$ with
$e_{\rm est}-\widetilde\eta\in L^1_{\nu_{\mathcal{M}}}(\mathcal M;\mathcal U)$, the triangle
inequality gives
\begin{equation}
\begin{aligned}
    \left\|\mathcal F-
    \left(\mathcal F_{\bs w}+\widetilde\eta\right)
    \right\|_{L^1_{\nu_{\mathcal{M}}}(\mathcal M;\mathcal U)}
    \leq{}&
    \omega_{\mathcal R}\mathbb E_{m\sim\nu_{\mathcal{M}}}
    \left[
    \|\eta(m)\|_{\mathcal U}^2
    \right]\\
    &+\|e_{\rm est}-\widetilde\eta\|_{L^1_{\nu_{\mathcal{M}}}(\mathcal M;\mathcal U)}.
\end{aligned}
    \label{eq:learned_residual_operator_bound}
\end{equation}

\paragraph{Learning the correction operator}
For fixed neural operator weights $w$, suppose that the learned correction $\widetilde{\eta}_N$, trained using $N$ residual observations, converges to $e_{\mathrm{est}}$ in $L^1_{\nu_{\mathcal M}}(\mathcal M;\mathcal U)$. This assumption is motivated by approximation results for vector-valued kernels and kernel-based operator learning~\cite{carmeli2010vector,batlle2024kernel}, together with Gaussian-process posterior contraction theory under suitable assumptions~\cite{van2008rates}. The corrected operator $F_w+\widetilde{\eta}_N$ then converges to the one-step Newton-corrected predictor $F_w+e_{\mathrm{est}}$, whose error is bounded by $\omega_R\,\mathbb{E}_{m\sim\nu_{\mathcal M}} [\|\eta(m)\|_{\mathcal U}^{2}]$ under the assumptions of Theorem~\ref{thm:residual_operator_error}. Therefore, \eqref{eq:learned_residual_operator_bound} provides a theoretical foundation for transferring the local quadratic accuracy of residual-based correction to a learned operator. The bound separates the residual approximation error from the correction-learning error, identifying two complementary routes to improved prediction: refining the preliminary operator and learning its residual correction more accurately.

\paragraph{Bayesian scope}
The theorem is conditional on fixed $\bs w$ and controls the consistency of
the residual-informed error observation. By itself, it does not imply
posterior contraction or calibration of the GP covariance, the Laplace
weight posterior, or the RI-BNO posterior predictive distribution.

\clearpage
\section{Gaussian process representation of operator inadequacy}
This section details the sparse variational Gaussian process (GP) that models the inadequacy map $\mathcal{G}_{\mathbf{g}}:\mathcal{M}\rightarrow\mathcal{U}$. The model is trained on the expanded residual observation dataset
\begin{equation}
\mathcal{E} = \left\{ \left((m^{(i)},\mathbf{x}_j), e_{\mathrm{est}}^{(i)}(\mathbf{x}_j)\right) \right\}_{i=1,j=1}^{N_{\mathrm{err}},N_{\mathrm{pt}}},
\end{equation}
where $N_{\mathrm{T}}=N_{\mathrm{err}}N_{\mathrm{pt}}$, pairing continuous parameter fields $m^{(i)} \in \mathcal{M}$ with spatial evaluation coordinates $\mathbf{x}_j \in \Omega$.
Functional inputs can be incorporated into GP covariance kernels through norm-based metrics defined on finite-dimensional basis representations of the underlying fields \cite{muehlenstaedt2017computer, betancourt2020gaussian, tan2019gaussian}. 
To obtain a common finite-dimensional representation of parameter fields across observation discretizations, we represent $m(\mathbf{x})\in H^1(\Omega)$ using a fixed spatial basis, $m(\mathbf{x}) = \sum_{i=1}^{N_{\mathrm{basis}}} m_i\phi_i(\mathbf{x}),$ where $\mathbf m = [m_1,\ldots,m_{N_{\mathrm{basis}}}]^T$ denotes the coefficient vector. Motivated by Sobolev-space approximation theory and the connections between reproducing kernel Hilbert spaces, Gaussian processes, and PDE approximation \cite{fasshauer2013reproducing,batlle2025error}, field similarity is measured using a normalized $H^1$ type distance
\begin{equation}
d_{\mathcal{M}}^2(m,m') = \frac{(\mathbf{m}-\mathbf{m}')^T \mathbf{W}_{L^2} (\mathbf{m}-\mathbf{m}')}{|\Omega|s_m^2} + \frac{(\mathbf{m}-\mathbf{m}')^T \mathbf{W}_{\nabla} (\mathbf{m}-\mathbf{m}')}{|\Omega|s_{\nabla m}^2},
\end{equation}
where the Gram matrices $\mathbf{W}_{L^2}, \mathbf{W}_{\nabla} \in \mathbb{R}^{N_{\mathrm{basis}} \times N_{\mathrm{basis}}}$ encode the continuous $L^2$ and $H^1$ semi-inner products:
\begin{equation}
[\mathbf{W}_{L^2}]_{ij} = \int_\Omega \phi_i(\mathbf{x}) \phi_j(\mathbf{x}) \, \mathrm{d}\mathbf{x}, \qquad 
[\mathbf{W}_{\nabla}]_{ij} = \int_\Omega \nabla \phi_i(\mathbf{x}) \cdot \nabla \phi_j(\mathbf{x}) \, \mathrm{d}\mathbf{x}.
\end{equation}
Here, $s_m$ and $s_{\nabla m}$ are empirical variance scales. The functional kernel uses a squared exponential formulation over $H^1(\Omega)$,
\begin{equation}
k_{\mathcal{M}}(m,m') = \exp\left(-\frac{d_{\mathcal{M}}^2(m,m')}{2\ell_m^2}\right),
\end{equation}
parameterized by length scale $\ell_m$. Assuming separability between functional and spatial correlations, the joint covariance is modeled using the product kernel. Combining $k_{\mathcal{M}}$ with a spatial Matérn $5/2$ kernel $k_\Omega(\mathbf{x},\mathbf{x}')$ yields the joint operator product kernel
\begin{equation}
k\big((m,\mathbf{x}),(m',\mathbf{x}')\big) = k_{\mathcal{M}}(m,m') \, k_\Omega(\mathbf{x},\mathbf{x}').
\end{equation}
This kernel induces a Gaussian process prior over the inadequacy field and defines the covariance matrix associated with all observations in $\mathcal E$. As $N_{\mathrm{T}}$ can be large, standard exact GP inference over the residual observation dataset $\mathcal{E}$ requires an intractable $\mathcal{O}(N_{\mathrm{T}}^3)$ inversion of the full covariance matrix $\mathbf{K}$. We employ a sparse variational Gaussian process (SVGP) \cite{titsias09a,hensman2013gaussian} inference to achieve scalability. Specifically, by approximating the full covariance through a low rank factorization governed by $M$ inducing points ($M \ll N_{\mathrm{T}}$), SVGP reduces the computational complexity from $\mathcal{O}(N_{\mathrm{T}}^3)$ to $\mathcal{O}(N_{\mathrm{T}} M^2)$. To ensure that the sparse approximation remains robust across varying resolutions, the inducing set $\mathcal Z=\{(m_a,\mathbf z_a)\}_{a=1}^{M} $ is defined independently of any discretization. Each inducing field $m_a(\mathbf x)\in H^1(\Omega)$ is represented in the same fixed basis space through coefficient vector $\mathbf m_a\in\mathbb R^{N_{\mathrm{basis}}}$, and kernel evaluations employ the corresponding Gram matrices $\mathbf W_{L^2}$ and $\mathbf W_\nabla$. The inducing variables are therefore continuous functional quantities, $ \mathbf f_{\mathcal Z}=f(\mathcal Z)\sim \mathcal N(\mathbf 0,\mathbf K_{MM})$, where $\mathbf{K}_{{M}{M}} \in \mathbb{R}^{M \times M}$ is the covariance matrix among inducing inputs. The inducing variable construction leads to the Nyström type approximation of the prior covariance matrix $\mathbf{K} \approx \mathbf{Q} = \mathbf{K}_{N_{\mathrm{T}}{M}} \mathbf{K}_{MM}^{-1} \mathbf{K}_{{M}N_{\mathrm{T}}}$, where $\mathbf{K}_{N_{\mathrm{T}}{M}} \in \mathbb{R}^{N_{\mathrm{T}} \times M}$ represents the cross covariance between residual observation inputs and inducing fields. Under the variational family $q(\mathbf{f}_{\mathcal{Z}}) = \mathcal{N}(\mathbf{a}, \mathbf{S})$, marginalizing over $\mathbf{f}_{\mathcal{Z}}$ yields the approximate posterior mean $\boldsymbol{\mu}$ and covariance $\mathbf{K}_q$
\begin{align}
\boldsymbol{\mu} &= \mathbf{K}_{N_{\mathrm{T}}{M}} \mathbf{K}_{{M}{M}}^{-1} \mathbf{a}, \\
\mathbf{K}_q &= \mathbf{K}_{N_{\mathrm{T}}N_{\mathrm{T}}} - \mathbf{K}_{N_{\mathrm{T}}{M}} \mathbf{K}_{{M}{M}}^{-1} \mathbf{K}_{{M} N_{\mathrm{T}}} + \mathbf{K}_{N_{\mathrm{T}}{M}} \mathbf{K}_{{M}{M}}^{-1} \mathbf{S} \mathbf{K}_{{M}{M}}^{-1} \mathbf{K}_{{M} N_{\mathrm{T}}}.
\end{align}
where $\mathbf{a}$ and $\mathbf{S}$ are the mean and covariance of the variational distribution $q(\mathbf{f}_{\mathcal{Z}})$. Substituting this low rank structure into the marginal likelihood reduces matrix inversion operations to the $M \times M$ matrix $\mathbf{K}_{{M}{M}}$. 
We optimize the GP hyperparameter set $\mathbf{g}$ and residual noise parameters $\mathbf{\Sigma}_{\mathrm{err}}$ alongside inducing inputs $\mathcal Z$, and variational parameters $(\mathbf{a}, \mathbf{S})$ by maximizing the evidence lower bound (ELBO)
\begin{equation}
\mathcal{L} = \sum_{n=1}^{N_{\mathrm{T}}} \mathbb{E}_{q(f_n)}\!\left[\log \pi_{\mathrm{like}}(e_{\mathrm{est}, n} \mid f_n, \mathbf{\Sigma}_{\mathrm{err}})\right] - \operatorname{KL}\!\left[q(\mathbf{f}_{\mathcal{Z}})\,\|\,p(\mathbf{f}_{\mathcal{Z}})\right].
\end{equation}
where $n$ indexes the training dataset. For a new query input  $z^*=(m^*,\mathbf{x}^*)$, comprising an unseen continuous parameter field $m^*\in\mathcal{M}$ and a spatial evaluation coordinate $\mathbf{x}^*\in\Omega$, the predictive distribution is obtained by conditioning on the optimized variational posterior $q(\mathbf{f}_{\mathcal Z})$. Denoting the cross-covariance vector between the test input and the $M$ inducing fields by $\mathbf{k}_{*M}\in\mathbb{R}^{1\times M}$, the predictive mean and variance are given by 
\begin{align} 
\mu^* &= \mathbf{k}_{*M}\mathbf{K}_{MM}^{-1}\mathbf{a}, \\ \sigma^{*2} &= k(z^*,z^*) - \mathbf{k}_{*M}\mathbf{K}_{MM}^{-1}\mathbf{k}_{M*} + \mathbf{k}_{*M}\mathbf{K}_{MM}^{-1} \mathbf{S} \mathbf{K}_{MM}^{-1} \mathbf{k}_{M*}, 
\end{align} 
where $k(z^*,z^*)$ is the prior variance evaluated using the product kernel and $ \mathbf{k}_{M*}=\mathbf{k}_{*M}^{T} \in \mathbb{R}^{M\times 1}$. Predictions for an entire inadequacy field are obtained by evaluating the predictive distribution at a collection of spatial coordinates $\{\mathbf{x}_j\}_{j=1}^{N_q}\subset\Omega,$ allowing the posterior mean and covariance field to be assembled on an arbitrary target discretization. This is enabled by both the discretization independent kernel construction and the SVGP inducing representation. The inducing inputs $\mathcal Z=\{(m_a,\mathbf{z}_a)\}_{a=1}^{M}$ are defined in the same fixed functional basis space as the parameter fields, making the variational parameters $(\mathbf a,\mathbf S)$ independent of any particular discretization. Consequently, the learned sparse posterior can be evaluated consistently across discretizations. 
\clearpage
\section{Comparison of model inadequacy learning through residual error estimates and direct FE prediction errors}
To assess the quality of residual-informed error estimates as training observations, we compare RI-BNO against an alternative framework that learns model inadequacy directly from finite element (FE) prediction errors $e_{\rm FE}$. The FE prediction error can be computed for any given input parameter field $m$ as
\begin{equation}\label{eq:fe_error_bno}
    e_{\rm FE} = \mathcal{F}_h (m) - \mathcal{F}_{\bs{w}^0_{\rm MAP}}(m)
\end{equation}
where $\mathcal{F}_h (m)$ is the FE solution for the corresponding input parameter $m$. A Gaussian process is then trained on these error observations to construct the inadequacy model.
\setlength{\tabcolsep}{2.5pt}
\begin{figure}[h]
\centering
\footnotesize
\raisebox{-0.1in}{
    \begin{tabular}{c}
        \hspace{-0.3in}$m$ \\
        \includegraphics[width=0.16\linewidth]{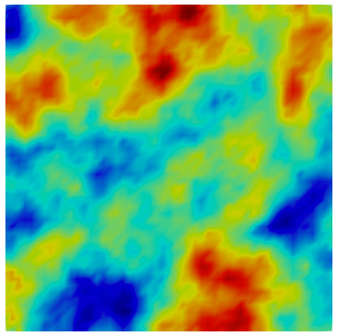}
        \hspace{0.02in}%
        \includegraphics[width=0.0575\linewidth]{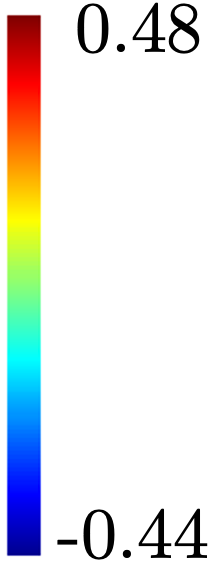} \\[0.3in] 
        
        \hspace{-0.3in}FE $u_h$\\
        \includegraphics[width=0.16\linewidth]{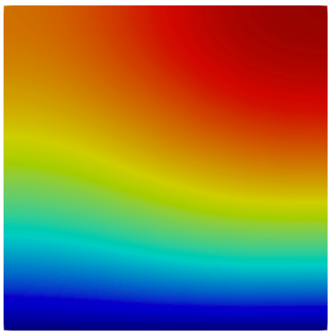} \hspace{0.02in}%
        \includegraphics[width=0.0575\linewidth]{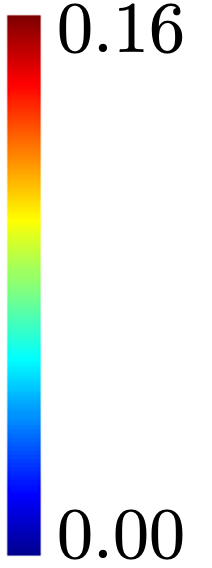}
    \end{tabular}%
}%
\hspace{0.15in}
\begin{tabular}{c c c c c}
    & 
    vBNO & 
    RI-BNO & 
    \shortstack{FE-error\\BNO} & 
    \\
    
    \raisebox{0.13in}{\shortstack{State\\prediction\\$\overline{u}$}} & 
    \includegraphics[width=0.16\linewidth]{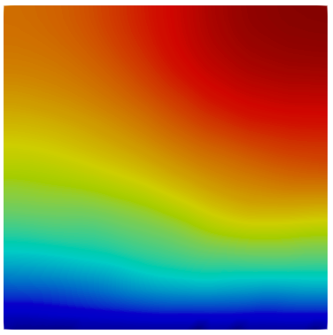} & 
    \includegraphics[width=0.16\linewidth]{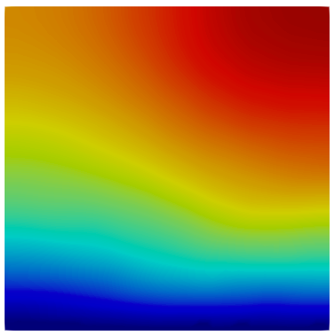} & 
    \includegraphics[width=0.16\linewidth]{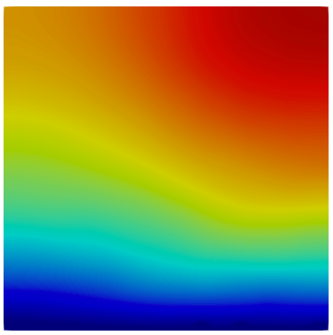} & 
    \includegraphics[width=0.0575\linewidth]{Figure_SI_1/u_contour.png} \\
    
    \raisebox{0.13in}{\shortstack{Relative\\Error\\$\varepsilon$}} & 
    \includegraphics[width=0.16\linewidth]{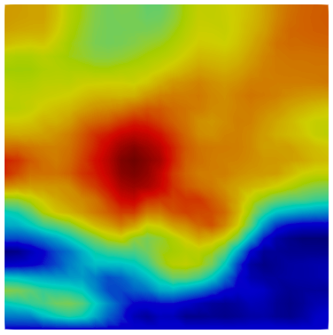} & 
    \includegraphics[width=0.16\linewidth]{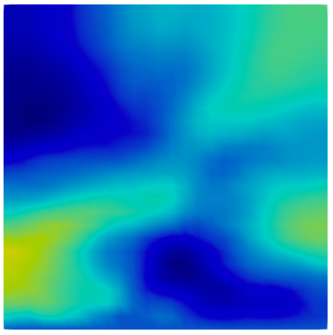} & 
    \includegraphics[width=0.16\linewidth]{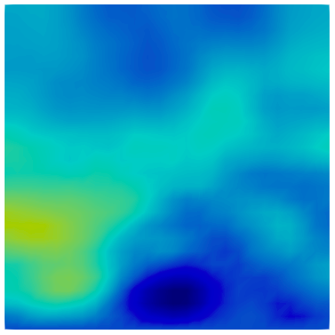} & 
    \includegraphics[width=0.0575\linewidth]{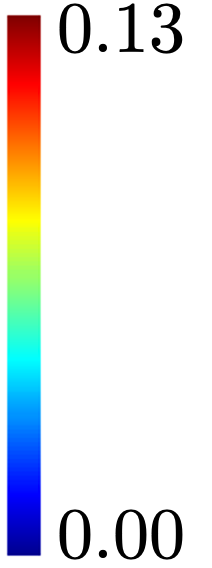} \\
    
    \raisebox{0.13in}{\shortstack{State\\variance\\$\sigma^2$}} & 
    \includegraphics[width=0.16\linewidth]{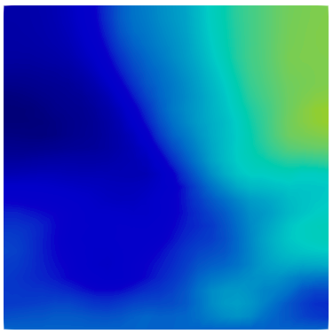} & 
    \includegraphics[width=0.16\linewidth]{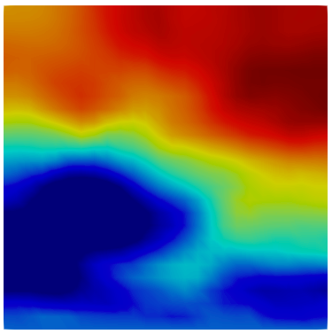} & 
    \includegraphics[width=0.16\linewidth]{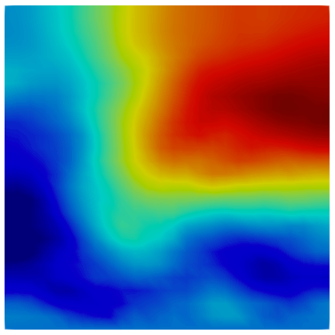} & 
    \includegraphics[width=0.0575\linewidth]{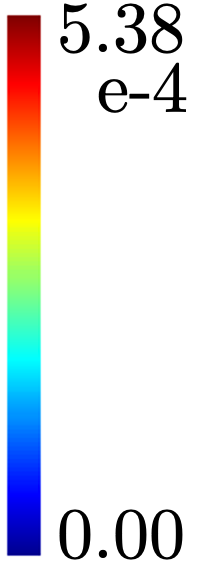}
\end{tabular}
\caption{Comparison of state predictions across frameworks (vBNO, RI-BNO, and FE-error BNO) against FE solution for an unseen in distribution parameter field $m$ using the latent subspace architecture.}
\end{figure}
\\
\begin{table}[h]
\centering
\small
\begin{tabular}{|c|c|c|}
    \hline
         Method &  \shortstack{\\Test Error  (\%) \\(using GP mean)} & \shortstack{\\Test Error (\%)\\   (Final)} 
         \\
     \hline Latent Subspace RI-BNO & 2.65 & 1.02 \\
     \hline Latent Subspace FE-error BNO & 3.53 & 1.83\\
     \hline
\end{tabular}
\caption{Comparison of test errors for the latent subspace architecture evaluated using the preliminary operator with GP mean correction versus the full inadequacy-aware Bayesian inference.}
\label{tab:LS_comparison}
\end{table}
\begin{table}[h]
\centering
\small
\begin{tabular}{|c|c|c|}
    \hline
         Method &  \shortstack{\\Test Error  (\%) \\(using GP mean)} & \shortstack{\\Test Error (\%)\\   (Final)}\\
     \hline DeepONet RI-BNO & 2.34 & 0.75\\
     \hline DeepONet FE-error BNO & 3.17 & 1.52\\
     \hline
\end{tabular}
\caption{Comparison of test errors for the DeepONet architecture evaluated using the preliminary operator with GP mean correction versus the full inadequacy-aware Bayesian inference.}
\label{tab:DON_comparison}
\end{table}

This approach to learn the model inadequacy through direct FE prediction error is called FE-error BNO. For the comparison of FE-error BNO with RI-BNO , we utilize the same preliminary operator, GP fitting, and final Bayesian fitting procedure. We also compare the performance of both RI-BNO and FE-error BNO using the GP posterior mean for correction and, separately, the final Bayesian inference using the inadequacy aware likelihood. For a given input parameter field $m$, the prediction using only the GP mean is $\mathcal F_{\bs w_{\rm MAP}^0}(m)+\mu_{\eta,\rm est}$ for RI-BNO and $\mathcal F_{\bs w_{\rm MAP}^0}(m)+\mu_{\eta,\rm FE}$ for FE-error BNO, where $\mu_{\eta,\rm est}$ and $\mu_{\eta,\rm FE}$ denote the GP mean learned for RI-BNO and FE-error BNO respectively. Test error can be defined using relative $L^2$ norm as
\begin{equation}
    \text{Test error}
=\mathbb E_{m\sim\nu_{\mathcal M}}
\left[
\frac{\|u_h(m)-\bar u(m)\|_{L^2(\Omega)}}
     {\|u_h(m)\|_{L^2(\Omega)}}
\right]
\end{equation}
For our numerical experiment, our data $\mathcal{D}$ comprises of 512 samples of $(m,u)$ pairs. We split the data $\mathcal{D}$ into $\mathcal{D}_{\rm pre}$ and $\mathcal{D}_{\rm fit}$, both having 256 samples each. For RI-BNO, the MAP estimate $\bs{w}^0_{\rm MAP}$ learned using $\mathcal{D}_{\rm pre}$ is used to create the inadequacy set $\mathcal{E}$ for 512 inputs in $\nu_{\mathcal{M}}$ which are different from $\mathcal{D}$ , through linear variational equation for posteriori error estimation using $\bs{w}^0_{\rm MAP}$. For FE-error BNO, the MAP estimate $\bs{w}^0_{\rm MAP}$ learned using $\mathcal{D}_{\rm pre}$ is used to compute error samples using \eqref{eq:fe_error_bno}. 
Mean Predictive Interval Width (MPIW) for the 95 \% intervals can be defined as
\begin{equation}
\text{MPIW}
=\mathbb E_{m\sim\nu_{\mathcal M}}
\left[
\frac{1}{|\Omega|} \int_{\Omega} \left( \bar{u}_{97.5\%}(m, x) - \bar{u}_{2.5\%}(m, x) \right) dx
\right]
\label{eq:mpiw}
\end{equation}
Prediction Interval Coverage Probability (PICP) metric can be used to evaluate the quality of prediction intervals which is computed through
\begin{equation}
\text{PICP}
 =\mathbb E_{m\sim\nu_{\mathcal M}}
\left[
\frac{1}{|\Omega|} \int_{\Omega} \mathbf{1} \left( \bar{u}_{2.5\%}(m, x) \le u_h(m, x) \le \bar{u}_{97.5\%}(m, x) \right) dx
\right]
\label{eq:picp}
\end{equation}
The indicator function $\mathbf{1}(\cdot)$ takes the value one when the high-fidelity solution lies within the predictive interval and zero otherwise. Thus, the
quantity inside the expectation represents the fraction of the spatial domain covered by the 95\% predictive interval for a given input $m$, while the
expectation averages this spatial coverage over the input distribution $\nu_{\mathcal M}$.
Tables \ref{tab:LS_comparison} and \ref{tab:DON_comparison} report the prediction test error using the preliminary operator and the GP mean, compared to the prediction with the final Bayesian fitting procedure. It can be observed that for both architectures the test error with GP mean is comparable for RI-BNO and FE-error BNO. Moreover, the final Bayesian inference reduces the test error further to below 2 \%.
Tables \ref{tab:LS_validation_metric} and \ref{tab:DON_validation_metric} show the validation metrics with Test Error, MPIW and PICP for vBNO, RI-BNO and FE-error BNO. The joint analysis of MPIW and PICP clearly demonstrates that vBNO is severely overconfident as its narrow prediction intervals yield coverage much below the nominal $95\%$ region, underestimating predictive uncertainty. In contrast, both RI-BNO and FE-error BNO achieve average pointwise coverage closer to the nominal 95\% level ($\mathrm{PICP} \approx 95\%$).  Furthermore, vBNO exhibits higher test errors across both architectures compared to RI-BNO and FE-error BNO, confirming that inadequacy learning leads to better predictive accuracy.
\setlength{\tabcolsep}{2.5pt}
\begin{figure}[htbp]
\centering
\footnotesize
\raisebox{-0.1in}{
    \begin{tabular}{c}
        \hspace{-0.3in}$m$ \\
        \includegraphics[width=0.16\linewidth]{Figure_SI_1/m_2.png}
        \hspace{0.02in}%
        \includegraphics[width=0.0575\linewidth]{Figure_SI_1/m_contour.png} \\[0.3in] 
        \hspace{-0.3in}FE $u_h$\\
        \includegraphics[width=0.16\linewidth]{Figure_SI_1/u_FE_2.png} \hspace{0.02in}%
        \includegraphics[width=0.0575\linewidth]{Figure_SI_1/u_contour.png}
    \end{tabular}
}
\hspace{0.15in}
\begin{tabular}{c c c c c}
    & 
    vBNO & 
    RI-BNO & 
    \shortstack{FE-error\\BNO} & 
    \\
    
    \raisebox{0.13in}{\shortstack{State\\prediction\\$\overline{u}$}} & 
    \includegraphics[width=0.16\linewidth]{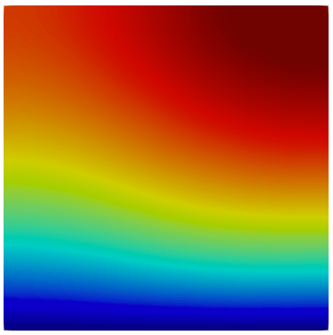} & 
    \includegraphics[width=0.16\linewidth]{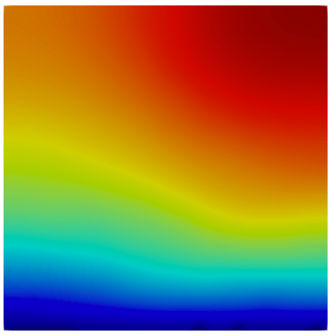} & 
    \includegraphics[width=0.16\linewidth]{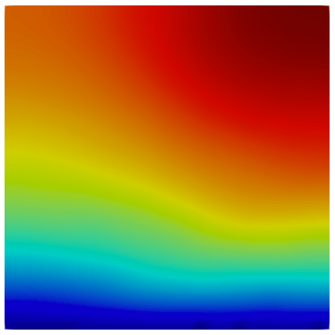} & 
    \includegraphics[width=0.0575\linewidth]{Figure_SI_1/u_contour.png} \\
    
    \raisebox{0.13in}{\shortstack{Relative\\Error\\$\varepsilon$}} & 
    \includegraphics[width=0.16\linewidth]{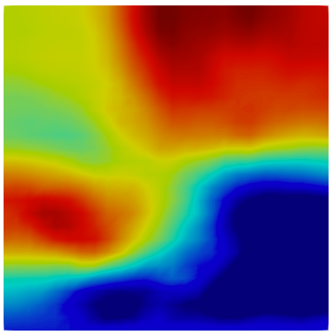} & 
    \includegraphics[width=0.16\linewidth]{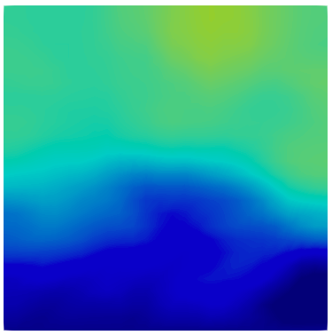} & 
    \includegraphics[width=0.16\linewidth]{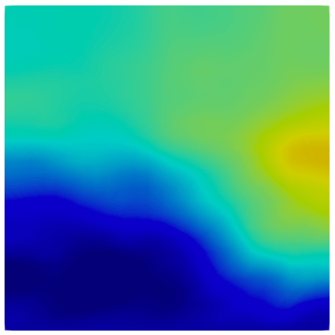} & 
    \includegraphics[width=0.0575\linewidth]{Figure_SI_1/re_contour.png} \\
    
    \raisebox{0.13in}{\shortstack{State\\variance\\$\sigma^2$}} & 
    \includegraphics[width=0.16\linewidth]{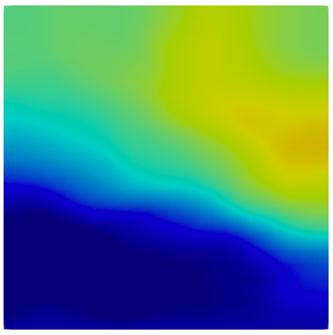} & 
    \includegraphics[width=0.16\linewidth]{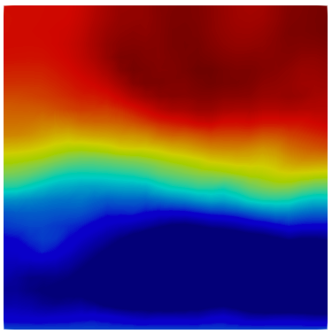} & 
    \includegraphics[width=0.16\linewidth]{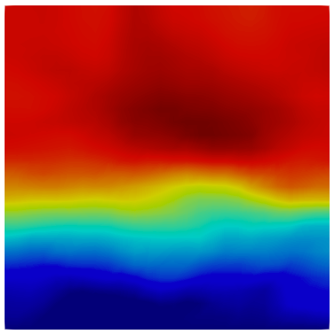} & 
    \includegraphics[width=0.0575\linewidth]{Figure_SI_1/var_contour.png}
\end{tabular}
\caption{Comparison of state predictions across frameworks (vBNO, RI-BNO, and FE-error BNO) against FE solution for an unseen in-distribution parameter field $m$ using the DeepONet architecture.}
\end{figure}

\begin{figure}[h]
        \centering
        \includegraphics[width=0.3\linewidth]{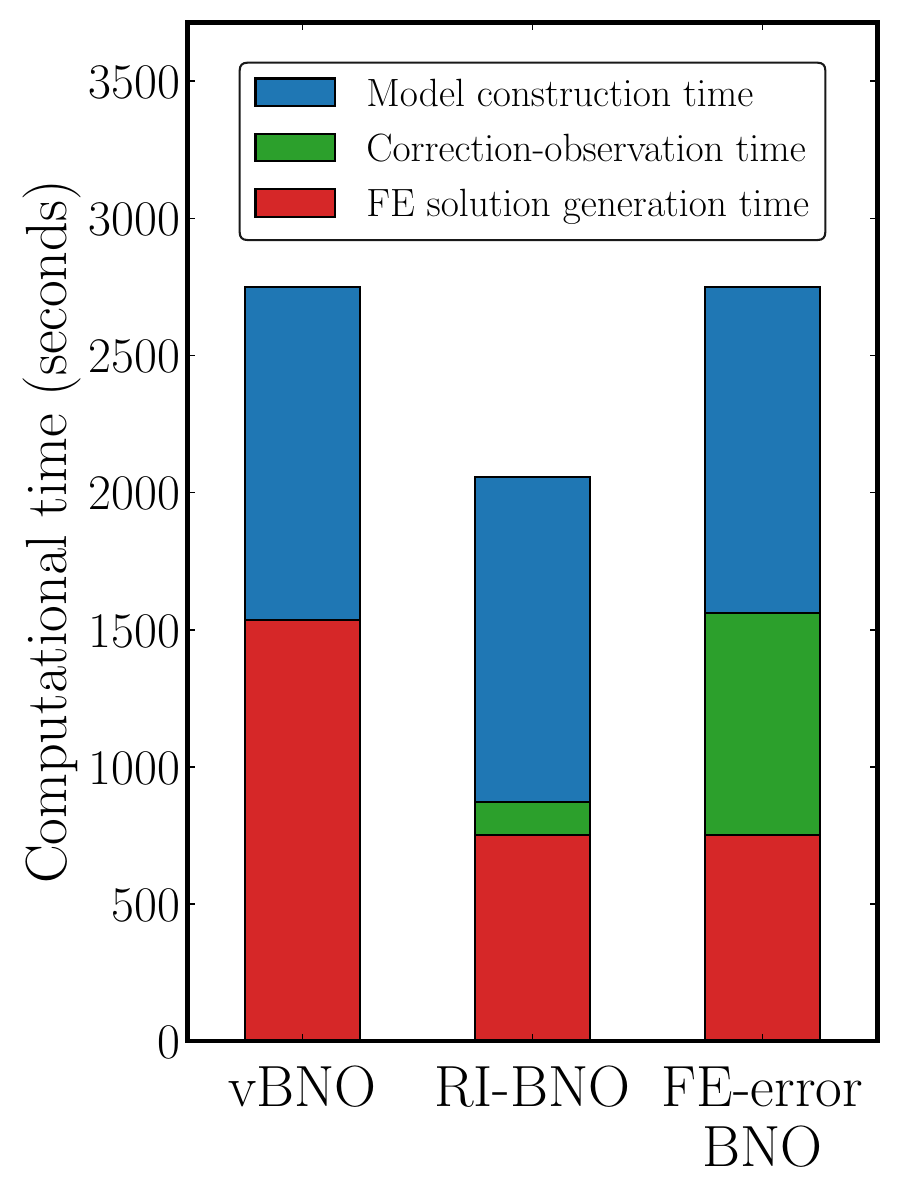} \vspace{-0.15in}\\
        \caption{Comparison of computational costs for latent subspace architecture for vBNO, RI-BNO and FE-error BNO. The total wall clock time is partitioned into finite element (FE) solution generation, correction-observation generation, and model construction time across the vBNO, RI-BNO, and FE-error BNO frameworks.}
        \label{fig:cost}
\end{figure}

 Table \ref{tab:time_LS} shows the computational cost associated with the FE solution generation, correction-observation generation, preliminary training, GP fitting, final Bayesian inference using inadequacy-aware likelihood and the total time for vBNO, RI-BNO and FE-error BNO. It is important to note that RI-BNO reduces the data generation time by around 45 \% and the total construction time by 25 \% in comparison to both vBNO and FE-error BNO. Figure \ref{fig:cost} shows the computational cost for different phases for the three different approaches of vBNO, RI-BNO and FE-error BNO.

\begin{table}[h]
\centering
\small
\begin{tabular}{|c|c|c|c|}
    \hline
         Method &\shortstack{\\Test Error (\%)\\   (Final)} 
         & MPIW & PICP (\%)
         \\
     \hline Latent Subspace vBNO  & 10.5 & 1.05 e-2 & 84.2\\
     \hline Latent Subspace RI-BNO & 1.02 &1.61 e-2 & 94.7\\
     \hline Latent Subspace FE-error BNO & 1.83 & 1.57 e-2 & 94.4\\
     \hline
\end{tabular}
\vspace{-0.05in}
\caption{Test accuracy and UQ metrics for the latent subspace architecture, comparing vBNO, RI-BNO, and FE-error BNO. Test Error denotes the relative $L^2$ error at the final time, while MPIW and PICP denote the mean prediction interval width and prediction interval coverage probability of the 95\% predictive intervals, respectively.}
\label{tab:LS_validation_metric}
\end{table}

\begin{table}[h]
\centering
\small
\begin{tabular}{|c|c|c|c|}
    \hline
         Method &  \shortstack{\\Test Error (\%)\\   (Final)} & MPIW & PICP (\%) \\
     \hline DeepONet vBNO  &7.73 & 0.93e-2 & 86.3\\
     \hline DeepONet RI-BNO  & 0.75& 1.52e-2 & 94.2\\
     \hline DeepONet FE-error BNO & 1.52& 1.46e-2 & 93.8\\
     \hline
\end{tabular}
\vspace{-0.05in}
\caption{Test accuracy and UQ metrics for the DeepONet architecture, comparing vBNO, RI-BNO, and FE-error BNO. Test Error denotes the relative $L^2$ error at the final time, while MPIW and PICP denote the mean prediction interval width and prediction interval coverage probability of the 95\% predictive intervals, respectively.}
\label{tab:DON_validation_metric}
\end{table}
\begin{table}[h]
\centering
\small
\begin{tabular}{|c|c|c|c|c|c|c|}
     \hline Method & \shortstack{FE solution\\ generation} & \shortstack{Correction\\ Observation }& \shortstack{\\Preliminary \\training}  & \shortstack{GP\\fitting} & \shortstack{Final\\training} & \shortstack{Total\\time}\\
     \hline vBNO & 
     1535.7& ---
     &---
     & ---
     & 
     1214.6& 
     2750.3 \\
     \hline RI-BNO & 
     751.3& 
     119.1&
     303.6& 
     512.9& 
     371.2& 2058.1 \\
     \hline FE-error BNO & 
     752.4& 
     807.9&
     308.5& 
     504.4& 
     378.3&
     2751.6\\
     \hline
\end{tabular}
\vspace{-0.05in}
\caption{Computational cost (seconds) per phase for  latent subspace architecture: (a) FE data generation, (b) correction observation generation, (c) preliminary BNO training, (d) GP fitting, (e) final training using inadequacy-aware likelihood, and (f) total time.}
\label{tab:time_LS}
\end{table}
\vspace{0.0in}
\clearpage

\section{Glioma growth and MRI inversion}
This section presents additional results for the glioma growth and MRI inversion study. We describe the Laplace approximation and gpCN sampling strategy employed for Bayesian inference, validate the forward predictions of vBNO and RI-BNO together with the preliminary neural operators, and illustrate the synthetic training data used for neural operator learning. We also provide a detailed comparison of the computational costs associated with MAP estimation and MCMC sampling using FE, vBNO, and RI-BNO.
\subsection{Glioma growth: validation of forward prediction}
\begin{figure}[!b]
    \centering
    \begin{tabular}{c c c c c c}
    \raisebox{0.3in}{\shortstack{Finite\\Element\\$u_h$}} &
    \includegraphics[trim={3.4in 0.75in 3.4in 0.75in},clip,width = 0.11\linewidth]{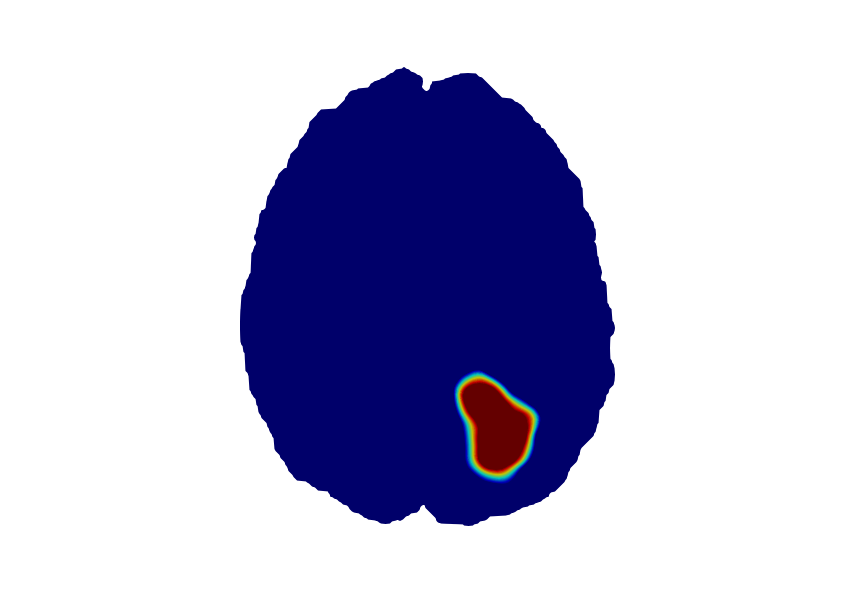}&
    \includegraphics[trim={3.4in 0.75in 3.4in 0.75in},clip,width = 0.11\linewidth]{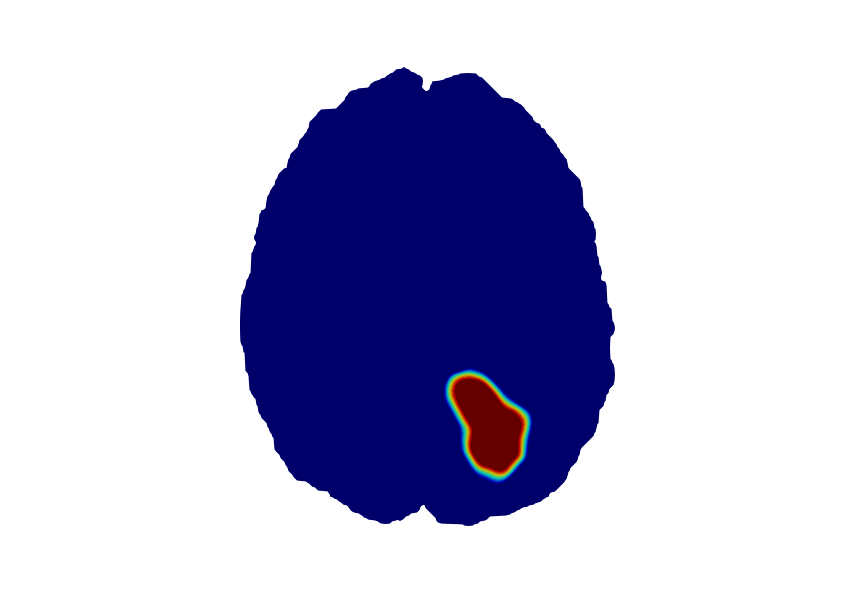}&
    \includegraphics[trim={3.4in 0.75in 3.4in 0.75in},clip,width = 0.11\linewidth]{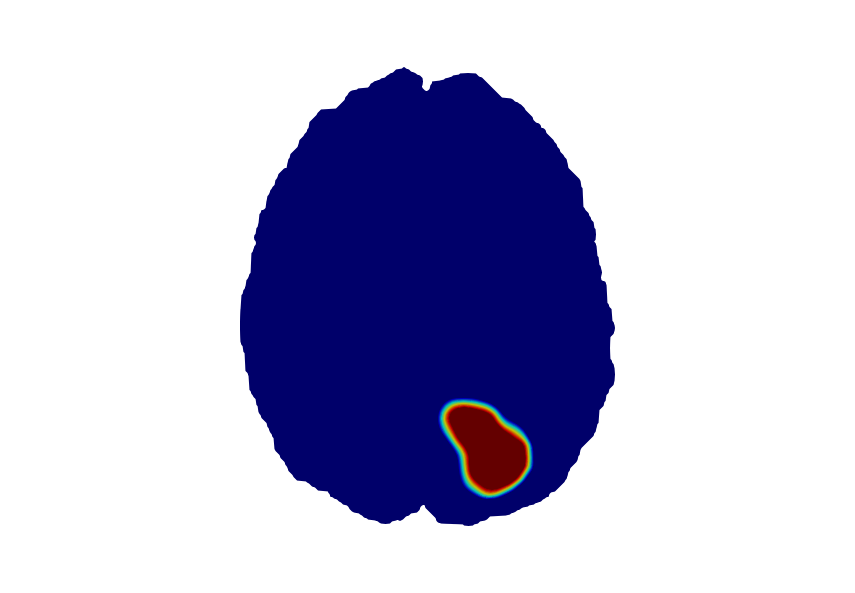}&
    \includegraphics[trim={3.4in 0.75in 3.4in 0.75in},clip,width = 0.11\linewidth]{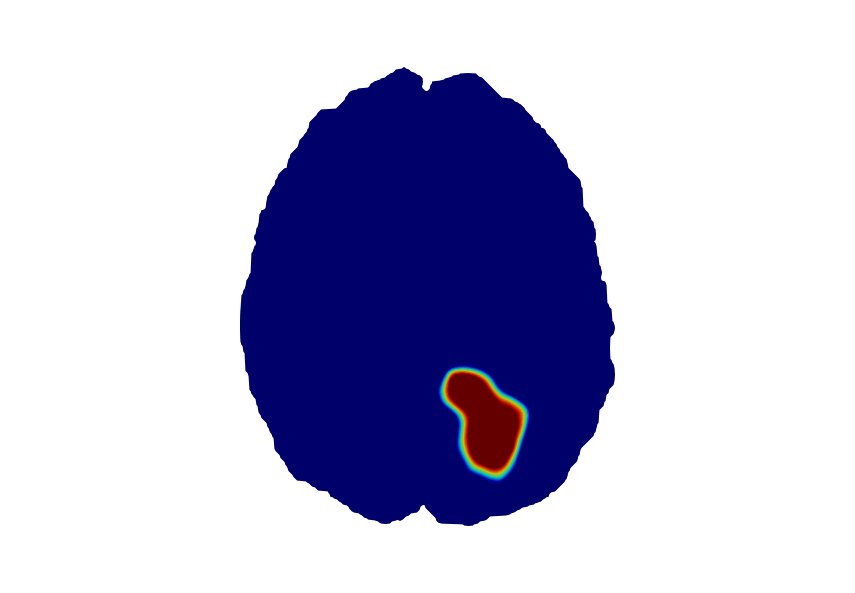}&
    \includegraphics[width = 0.023\linewidth]{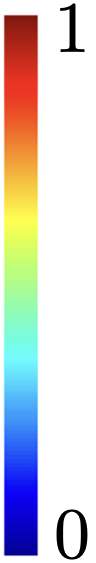}
    \\
    \raisebox{0.3in}{\shortstack{State\\Prediction\\$\overline{u}_{\text{vBNO}}$}} &
    \includegraphics[trim={3.4in 0.75in 3.4in 0.75in},clip,width = 0.11\linewidth]{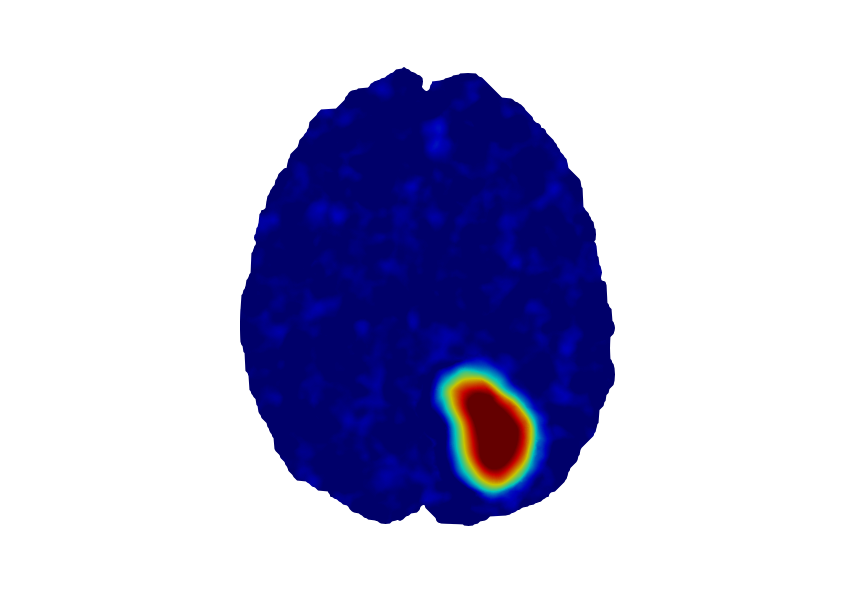}&
    \includegraphics[trim={3.4in 0.75in 3.4in 0.75in},clip,width = 0.11\linewidth]{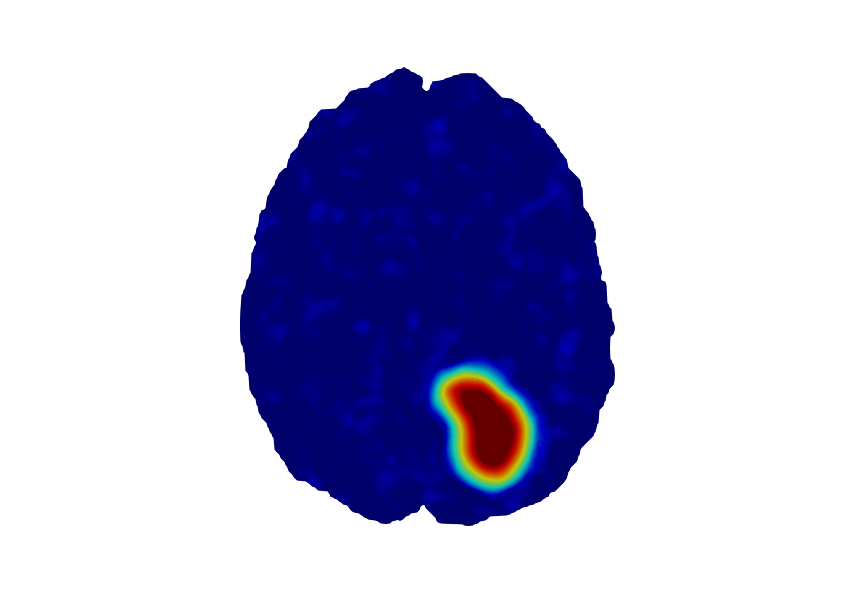}&
    \includegraphics[trim={3.4in 0.75in 3.4in 0.75in},clip,width = 0.11\linewidth]{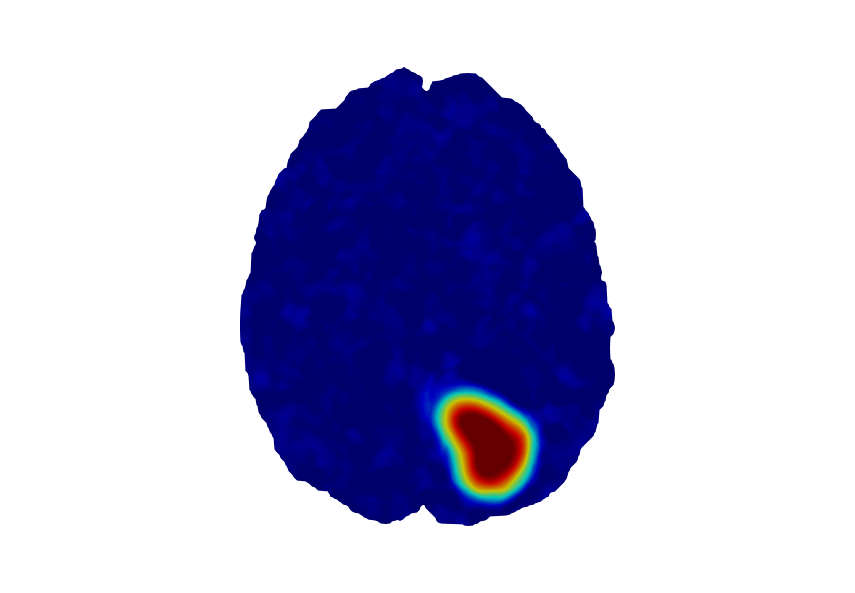}&
    \includegraphics[trim={3.4in 0.75in 3.4in 0.75in},clip,width = 0.11\linewidth]{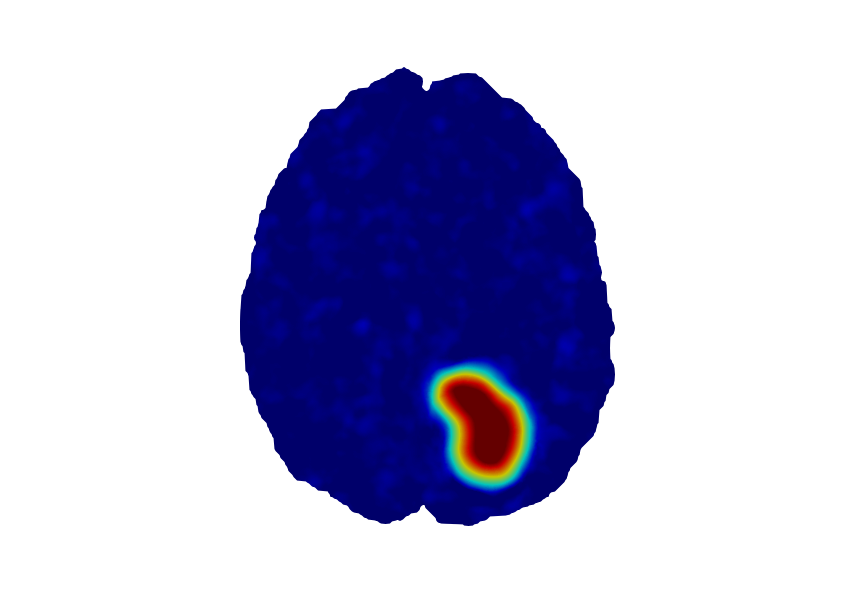}&
    \includegraphics[width = 0.023\linewidth]{Figures_PDE/contour.png}\\
    \raisebox{0.3in}{\shortstack{Pointwise\\ Relative\\Error\\$\varepsilon_{\text{vBNO}}$}} &
    \hspace{0.1in}\includegraphics[trim={4.9in 0.75in 4.9in 0.75in},clip,width = 0.115\linewidth]{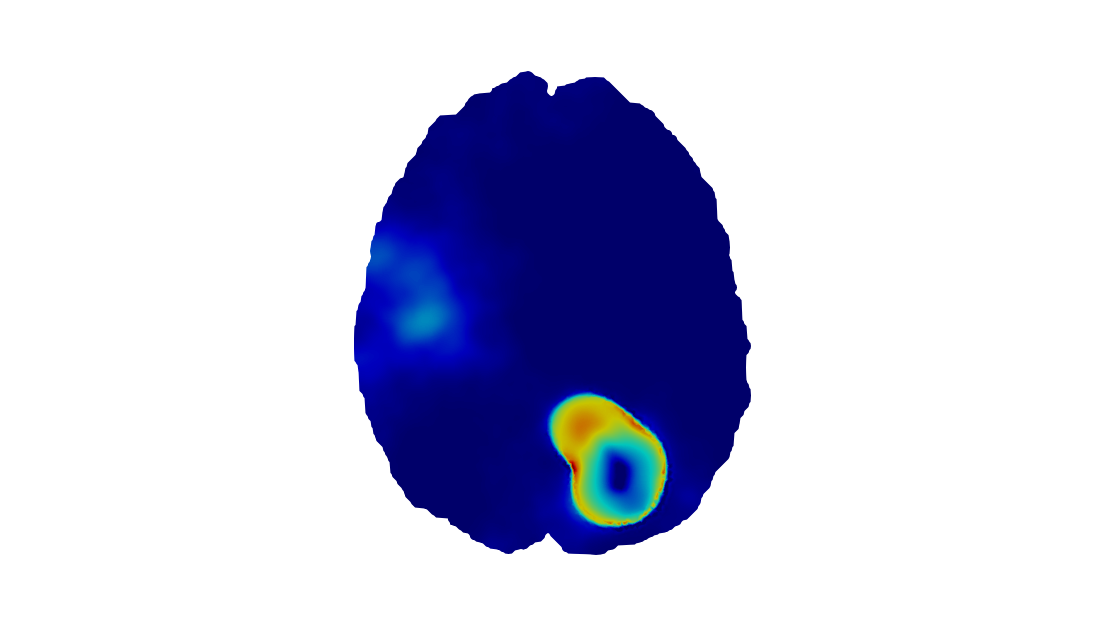}&
    \includegraphics[trim={4.9in 0.75in 4.9in 0.75in},clip,width = 0.115\linewidth]{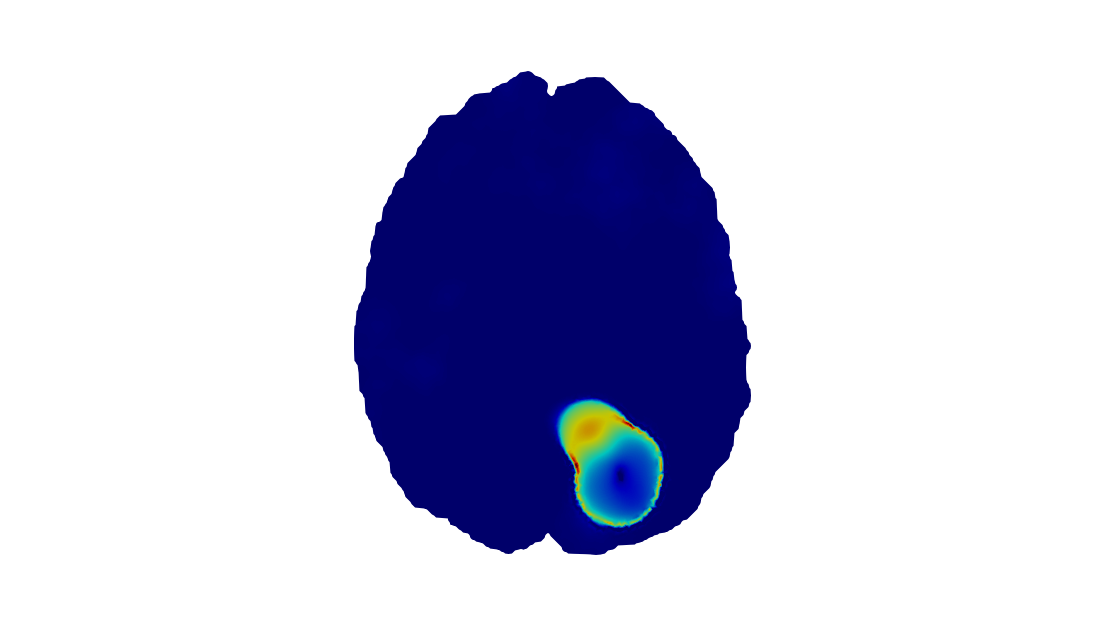}&
    \includegraphics[trim={4.9in 0.75in 4.9in 0.75in},clip,width = 0.115\linewidth]{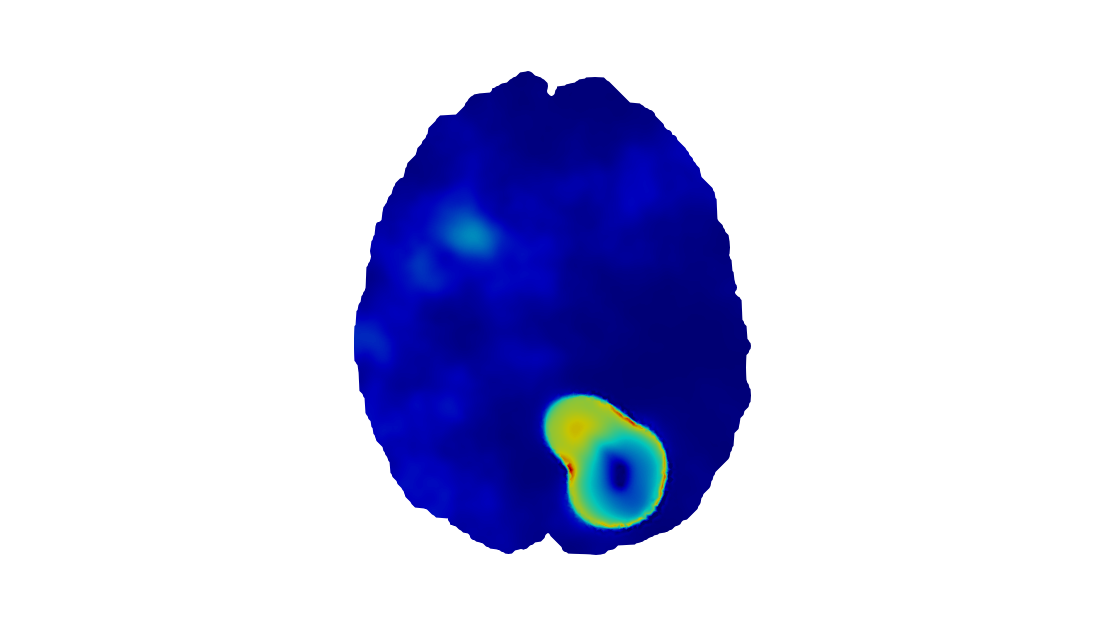}&
    \includegraphics[trim={4.9in 0.75in 4.9in 0.75in},clip,width = 0.115\linewidth]{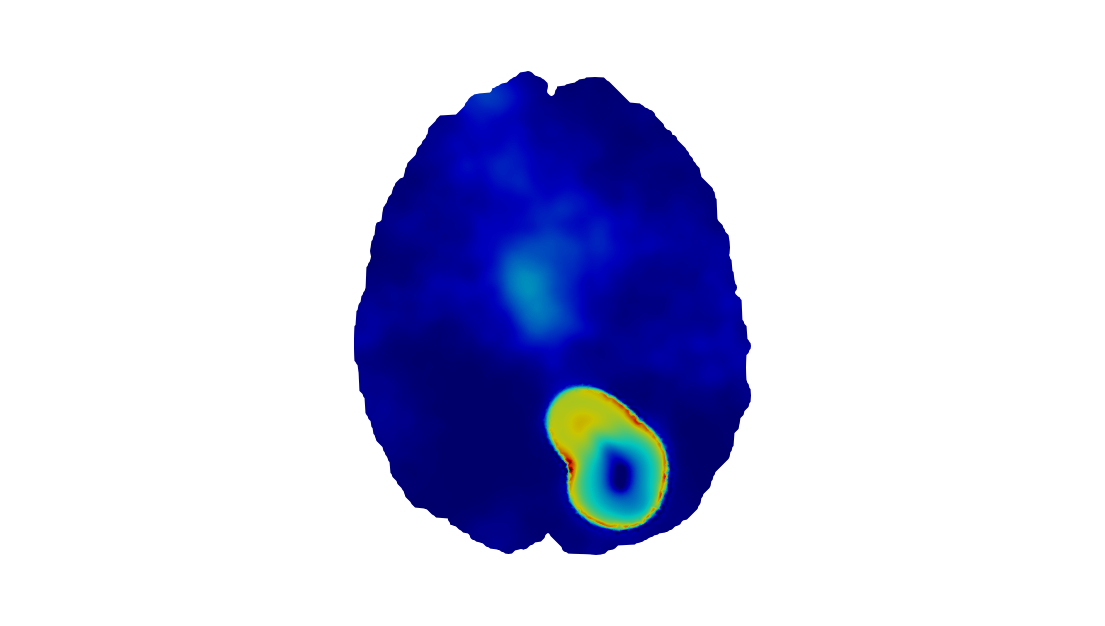}&
    \hspace{0.15in}\includegraphics[width = 0.0475\linewidth]{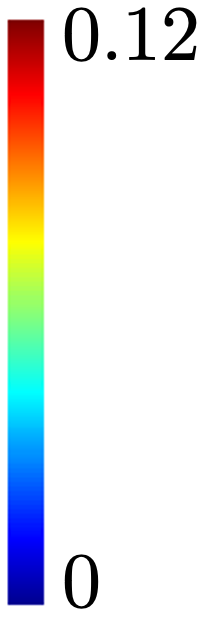}
    \\
    \raisebox{0.3in}{\shortstack{State\\Prediction\\$\overline{u}_{\text{RI}}$}}&
    \includegraphics[trim={3.4in 0.75in 3.4in 0.75in},clip,width = 0.11\linewidth]{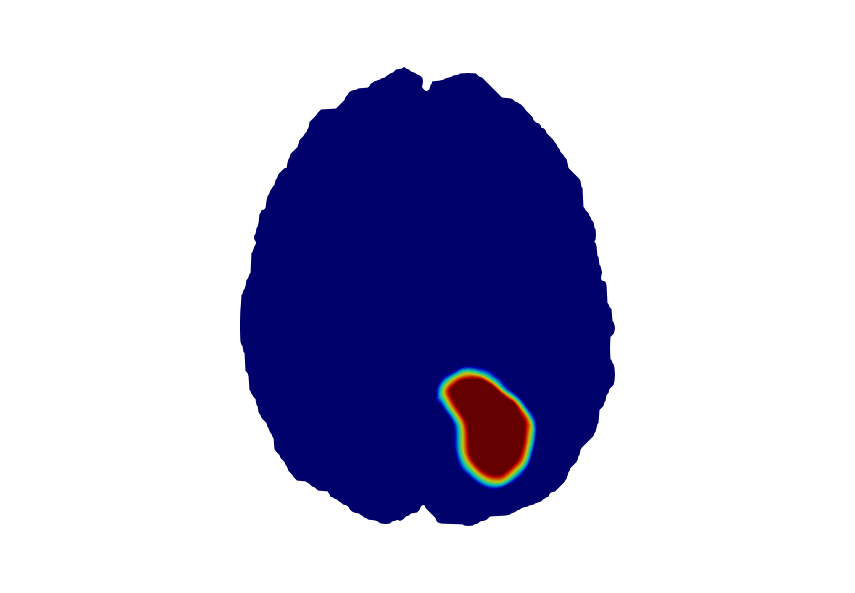}&
    \includegraphics[trim={3.4in 0.75in 3.4in 0.75in},clip,width = 0.11\linewidth]{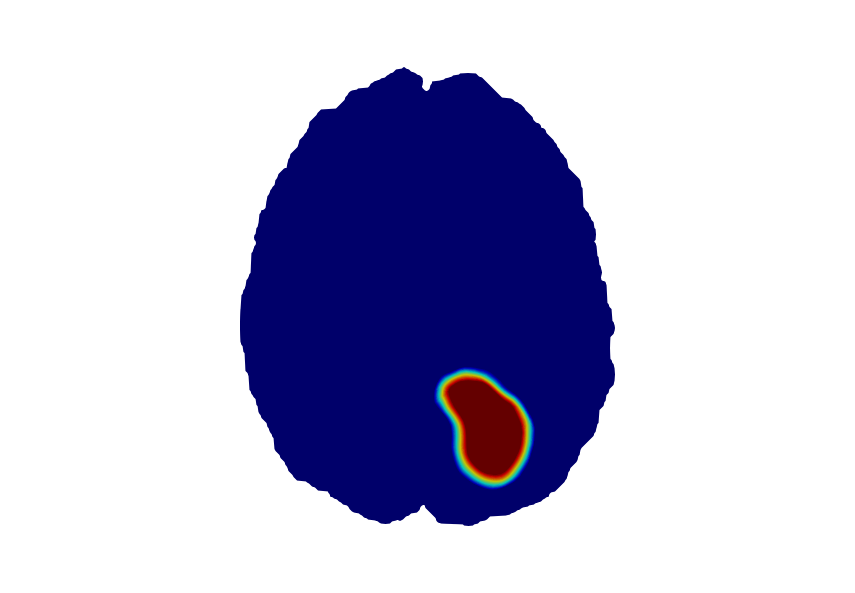}&
    \includegraphics[trim={3.4in 0.75in 3.4in 0.75in},clip,width = 0.11\linewidth]{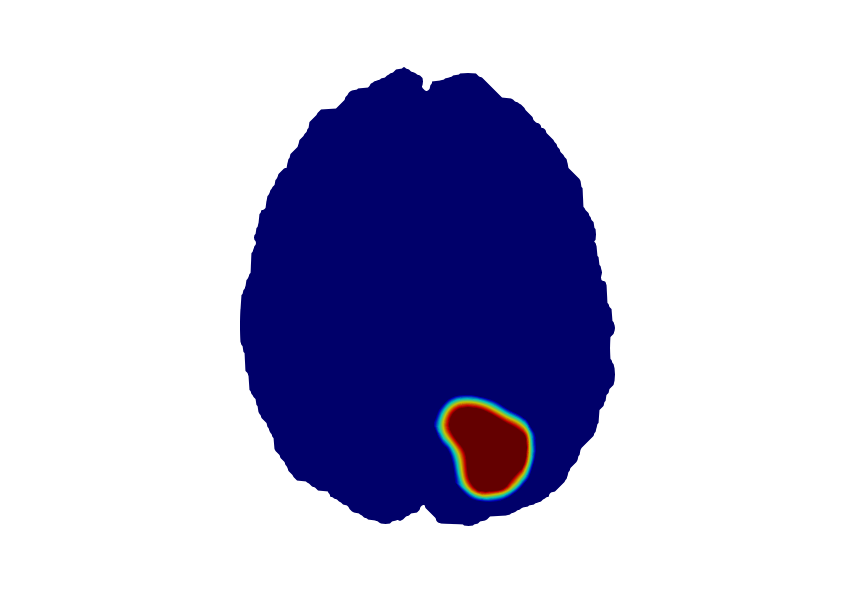}&
    \includegraphics[trim={3.4in 0.75in 3.4in 0.75in},clip,width = 0.11\linewidth]{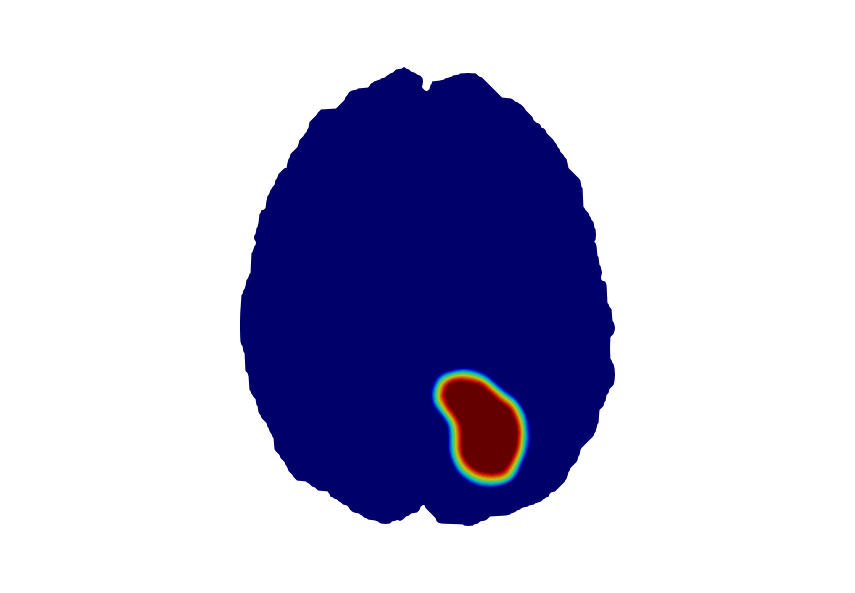}&
    \includegraphics[width = 0.023\linewidth]{Figures_PDE/contour.png}\\
    \raisebox{0.3in}{\shortstack{Pointwise\\ Relative\\Error\\$\varepsilon_{\text{RI}}$}}  &
    \includegraphics[trim={4.9in 0.75in 4.9in 0.75in},clip,width = 0.115\linewidth]{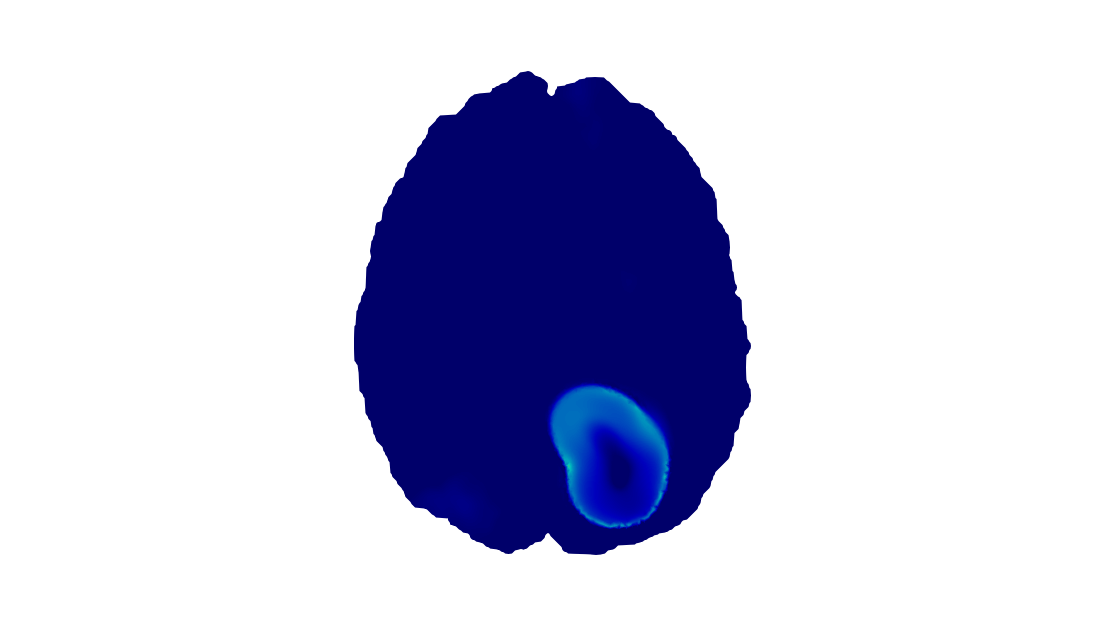}&
    \includegraphics[trim={4.9in 0.75in 4.9in 0.75in},clip,width = 0.115\linewidth]{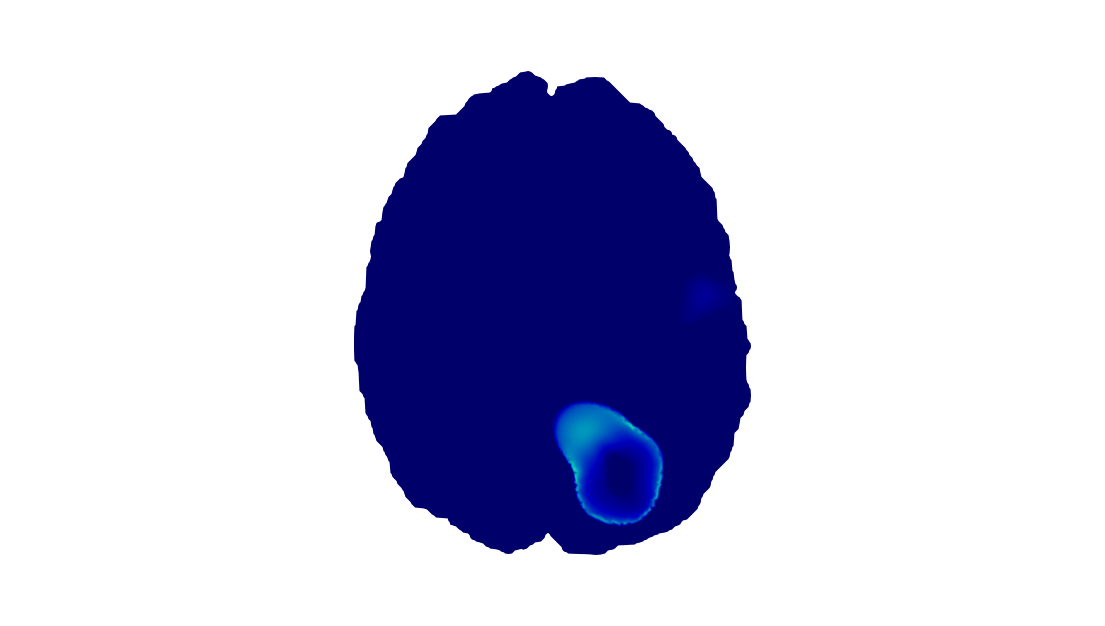}&
    \includegraphics[trim={4.9in 0.75in 4.9in 0.75in},clip,width = 0.115\linewidth]{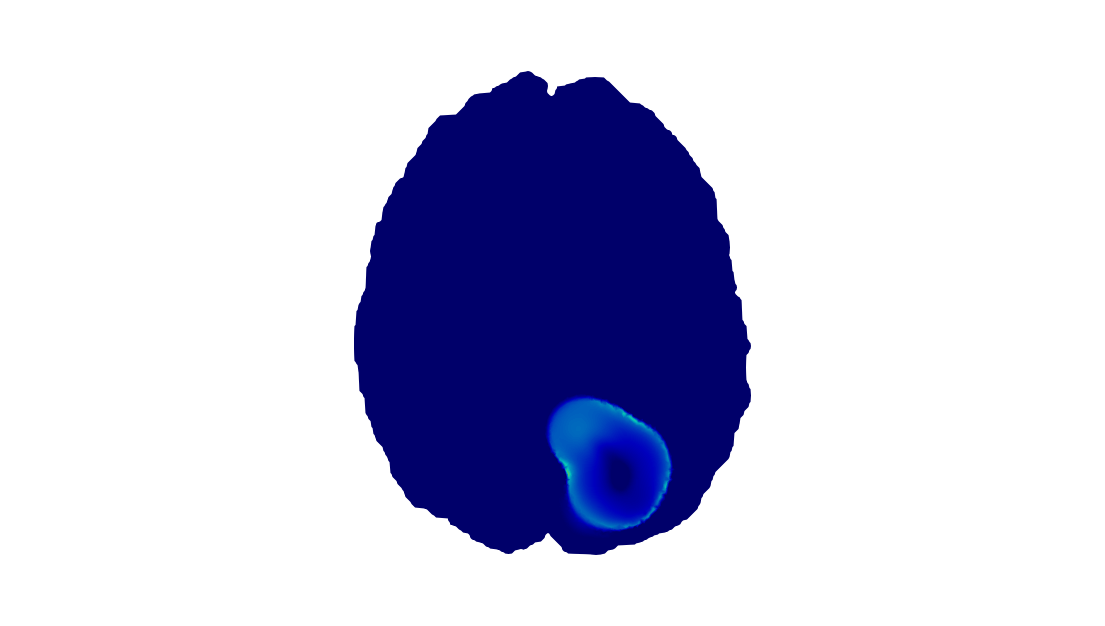}&
    \includegraphics[trim={4.9in 0.75in 4.9in 0.75in},clip,width = 0.115\linewidth]{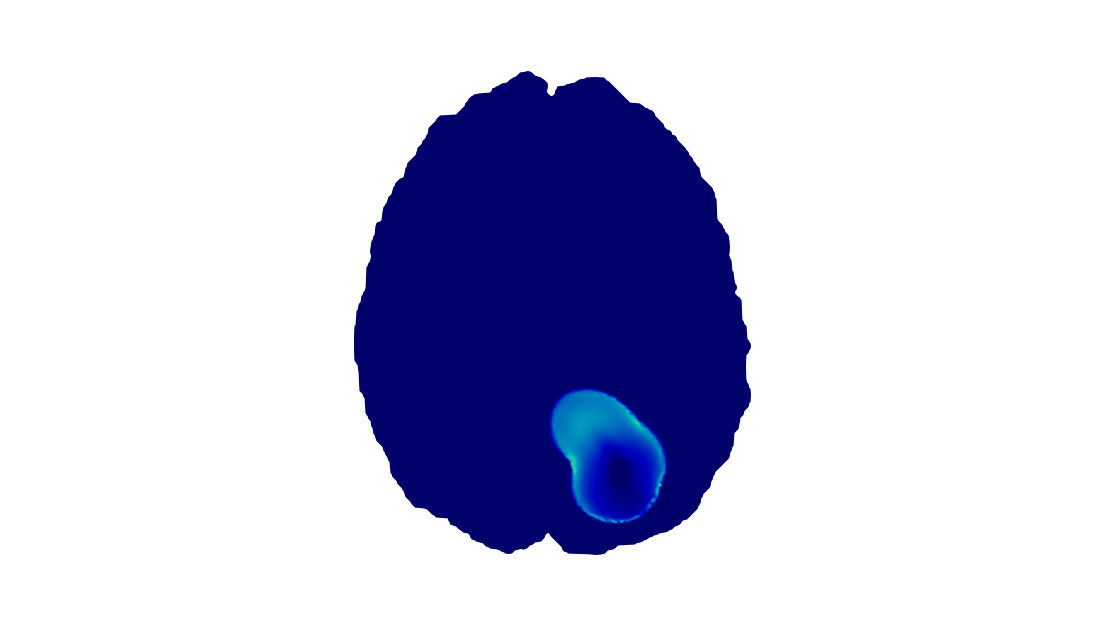}&
    \hspace{0.15in}\includegraphics[width = 0.0475\linewidth]{Figure_er/contour_er.png}
    \\
    \end{tabular}
    \caption{Comparison of predicted tumor volume fraction spatial profiles between FE solution $u_h$, and state predictions through vBNO $\overline{u}_{\text{vBNO}}$, and RI-BNO $\overline{u}_{\text{RI}}$ along with their pointwise relative error fields with respect to FE across four unseen test samples of the input parameter field $m=(\log D, \log G)$.}
    \label{fig:validation_bno}
\end{figure}

\begin{figure}[h]
    \centering
    \footnotesize
    \setlength{\tabcolsep}{1pt}

    \begin{tabular}{c c c c c c}
        & & \hspace{-0.15in}$u$ &\hspace{-0.15in}$u^-$ & \hspace{-0.15in}$n$ & \hspace{-0.1in}$e$ \\
        \multirow{3}{*}{%
            \begin{tabular}{c}
                \vspace{-0.6in}\\
                \hspace*{-0.2in}D \\
                \includegraphics[width=0.1\linewidth]{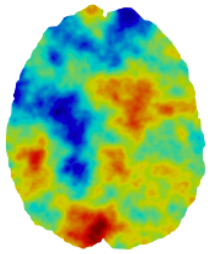}
                \includegraphics[width=0.0375\linewidth]{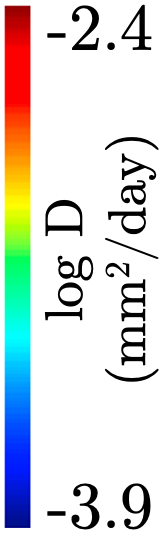}\\[0.1in]
                \hspace*{-0.2in}G\\
                \includegraphics[width=0.1\linewidth]{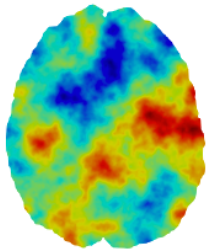}
                \includegraphics[width=0.0375\linewidth]{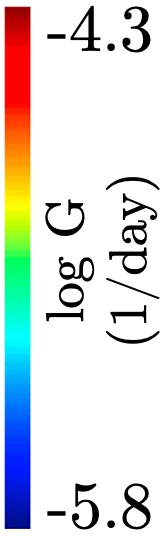}
            \end{tabular}%
        } \hspace{0.1in}
        &
        \raisebox{0.15in}{\shortstack{FE\\ Solution}} &
        \includegraphics[width=0.1\linewidth]{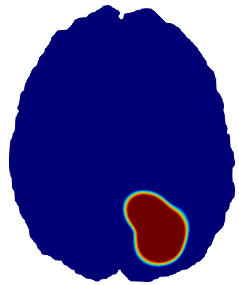}
        \includegraphics[width=0.022\linewidth]{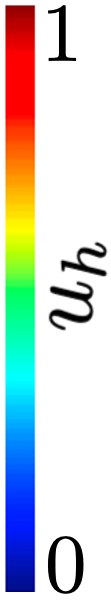}
        &
        \includegraphics[width=0.1\linewidth]{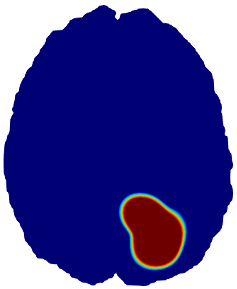}
        \includegraphics[width=0.027\linewidth]{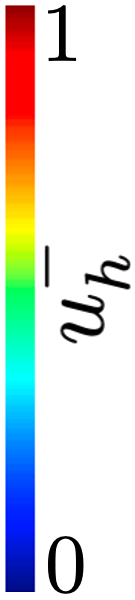}
        &
        \includegraphics[width=0.1\linewidth]{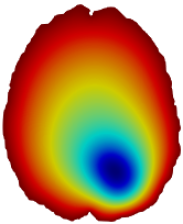}
        \includegraphics[width=0.03\linewidth]{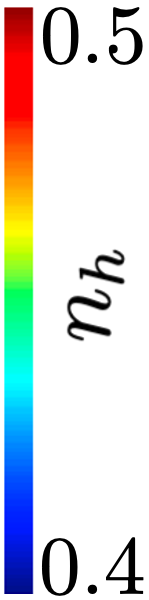}
        &
        \hspace{0.05in}\includegraphics[width=0.1\linewidth]{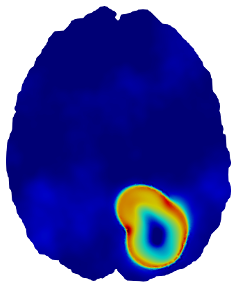}
        \includegraphics[width=0.0365\linewidth]{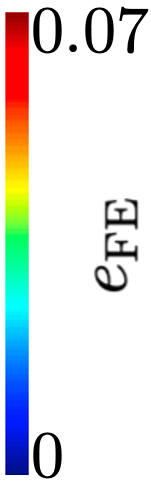}
        \\

        &
        \raisebox{0.15in}{\shortstack{Preliminary\\NO\\ Prediction}} &
        \includegraphics[width=0.1\linewidth]{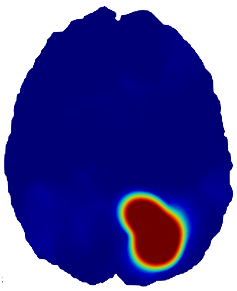}
        \includegraphics[width=0.024\linewidth]{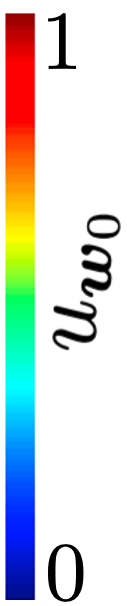}
        &
        \includegraphics[width=0.1\linewidth]{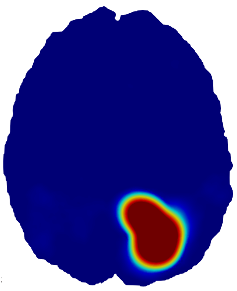}
        \includegraphics[width=0.0275\linewidth]{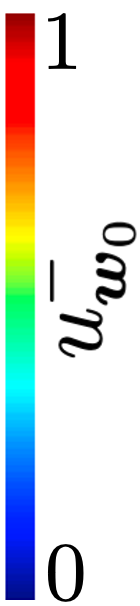}
        &
        \includegraphics[width=0.1\linewidth]{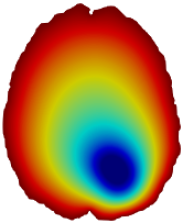}
        \includegraphics[width=0.03\linewidth]{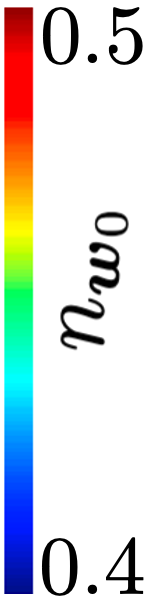}
        &
        \hspace{0.05in}\includegraphics[width=0.1\linewidth]{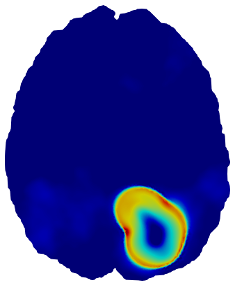}
        \includegraphics[width=0.0365\linewidth]{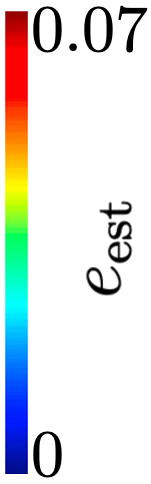}
        \\

        &
        \raisebox{0.175in}{\shortstack{Pointwise\\ Relative\\ Error}} &
        \hspace{0.07in}\includegraphics[width=0.1\linewidth]{Figure_SI_3/u_err_1.png}
        \includegraphics[width=0.036\linewidth]{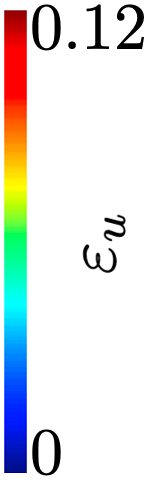}
        &
        \hspace{0.07in}\includegraphics[width=0.1\linewidth]{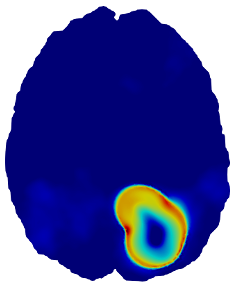}
        \includegraphics[width=0.036\linewidth]{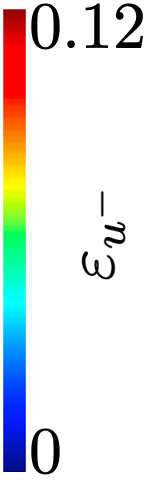}
        &
        \includegraphics[width=0.1\linewidth]{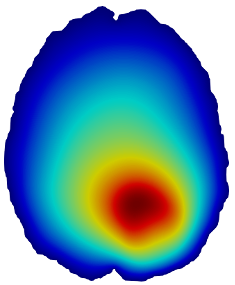}
        \includegraphics[width=0.036\linewidth]{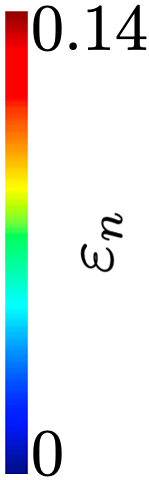}
        &
        \hspace{0.05in}\includegraphics[width=0.1\linewidth]{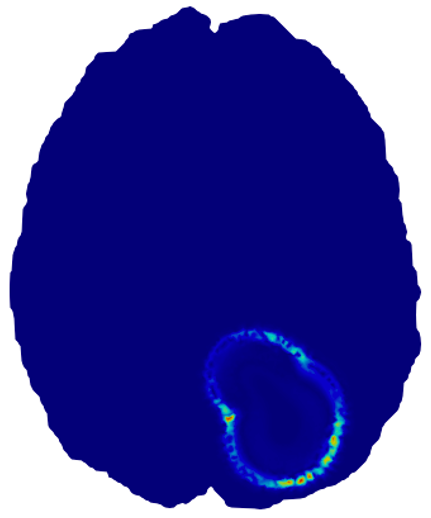}
        \includegraphics[width=0.0365\linewidth]{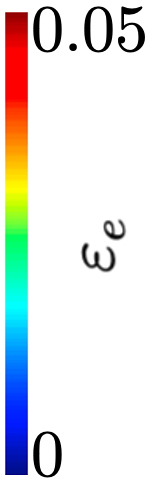}
        \\

        \multicolumn{6}{c}{\footnotesize (a)}\\[0.02in]

        \multirow{3}{*}{%
            \begin{tabular}{c}
                \vspace{-0.6in}\\
                \hspace*{-0.2in}D \\
                \includegraphics[width=0.1\linewidth]{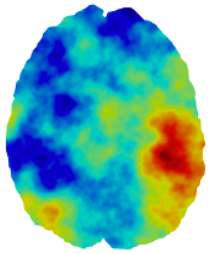}
                \includegraphics[width=0.0375\linewidth]{Figure_SI_3/D_contour.png}\\[0.1in]
                \hspace*{-0.2in}G\\
                \includegraphics[width=0.1\linewidth]{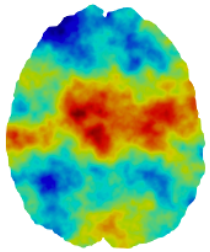}
                \includegraphics[width=0.0375\linewidth]{Figure_SI_3/G_contour.png}
            \end{tabular}%
        } \hspace{0.1in}
        &
        \raisebox{0.15in}{\shortstack{FE\\ Solution}} &
        \includegraphics[width=0.1\linewidth]{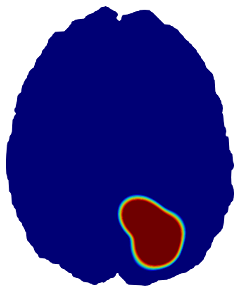}
        \includegraphics[width=0.022\linewidth]{Figure_SI_3/u_FE_contour.png}
        &
        \includegraphics[width=0.1\linewidth]{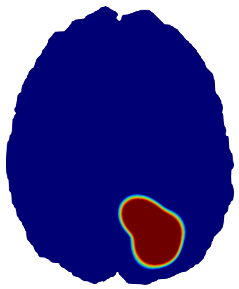}
        \includegraphics[width=0.027\linewidth]{Figure_SI_3/u_p_FE_contour.png}
        &
        \includegraphics[width=0.1\linewidth]{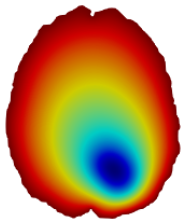}
        \includegraphics[width=0.03\linewidth]{Figure_SI_3/n_FE_contour.png}
        &
        \hspace{0.05in}\includegraphics[width=0.1\linewidth]{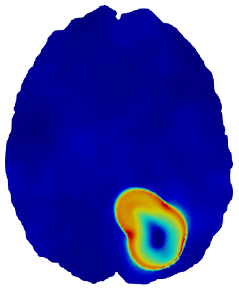}
        \includegraphics[width=0.0365\linewidth]{Figure_SI_3/e_FE_contour.png}
        \\

        &
        \raisebox{0.15in}{\shortstack{Preliminary\\NO \\  Prediction}} &
        \includegraphics[width=0.1\linewidth]{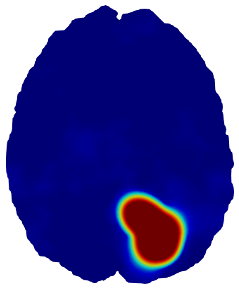}
        \includegraphics[width=0.024\linewidth]{Figure_SI_3/u_NO_contour.png}
        &
        \includegraphics[width=0.1\linewidth]{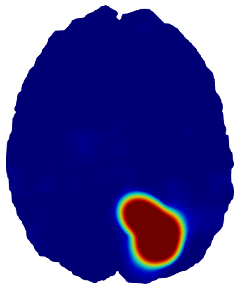}
        \includegraphics[width=0.0275\linewidth]{Figure_SI_3/u_p_NO_contour.png}
        &
        \includegraphics[width=0.1\linewidth]{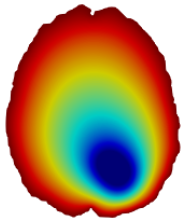}
        \includegraphics[width=0.03\linewidth]{Figure_SI_3/n_NO_contour.png}
        &
        \hspace{0.05in}\includegraphics[width=0.1\linewidth]{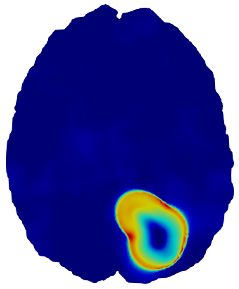}
        \includegraphics[width=0.0365\linewidth]{Figure_SI_3/e_est_contour.png}
        \\

        &
        \raisebox{0.175in}{\shortstack{Pointwise\\ Relative\\ Error}} &
        \hspace{0.07in}\includegraphics[width=0.1\linewidth]{Figure_SI_3/u_err_2.png}
        \includegraphics[width=0.036\linewidth]{Figure_SI_3/u_er_contour.png}
        &
        \hspace{0.07in}\includegraphics[width=0.1\linewidth]{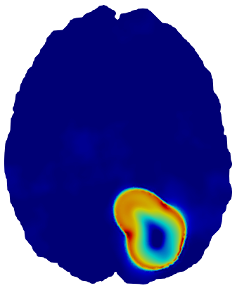}
        \includegraphics[width=0.036\linewidth]{Figure_SI_3/u_p_er_contour.png}
        &
        \includegraphics[width=0.1\linewidth]{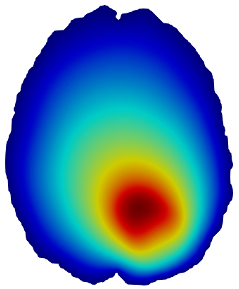}
        \includegraphics[width=0.036\linewidth]{Figure_SI_3/n_er_contour.png}
        &
        \hspace{0.05in}\includegraphics[width=0.1\linewidth]{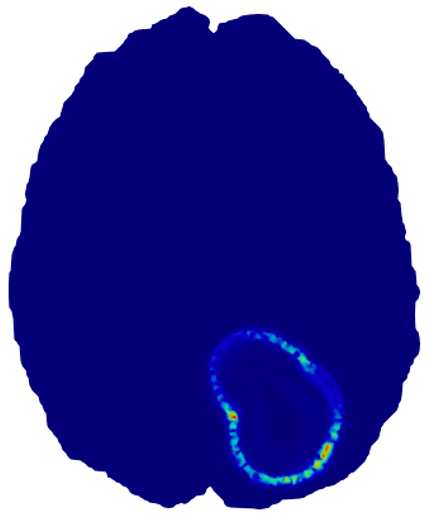}
        \includegraphics[width=0.0365\linewidth]{Figure_SI_3/re_err_contour.png}
        \\

        \multicolumn{6}{c}{\footnotesize (b)}\\[0.02in]

        \multirow{3}{*}{%
            \begin{tabular}{c}
                \vspace{-0.6in}\\
                \hspace*{-0.2in}D \\
                \includegraphics[width=0.1\linewidth]{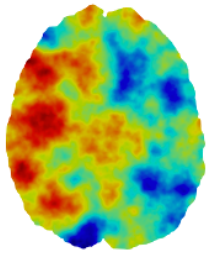}
                \includegraphics[width=0.0375\linewidth]{Figure_SI_3/D_contour.png}\\[0.1in]
                \hspace*{-0.2in}G\\
                \includegraphics[width=0.1\linewidth]{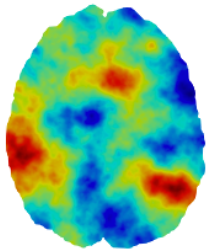}
                \includegraphics[width=0.0375\linewidth]{Figure_SI_3/G_contour.png}
            \end{tabular}%
        } \hspace{0.1in}
        &
        \raisebox{0.15in}{\shortstack{FE\\ solution}} &
        \includegraphics[width=0.1\linewidth]{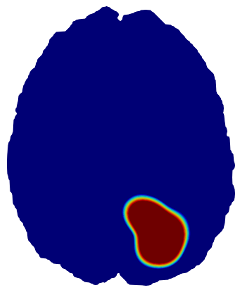}
        \includegraphics[width=0.022\linewidth]{Figure_SI_3/u_FE_contour.png}
        &
        \includegraphics[width=0.1\linewidth]{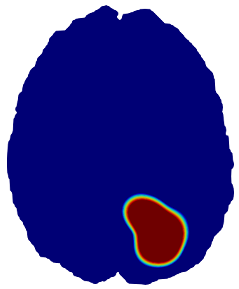}
        \includegraphics[width=0.027\linewidth]{Figure_SI_3/u_p_FE_contour.png}
        &
        \includegraphics[width=0.1\linewidth]{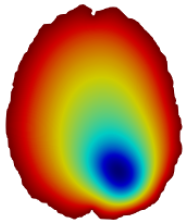}
        \includegraphics[width=0.03\linewidth]{Figure_SI_3/n_FE_contour.png}
        &
        \hspace{0.05in}\includegraphics[width=0.1\linewidth]{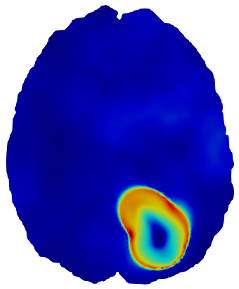}
        \includegraphics[width=0.0365\linewidth]{Figure_SI_3/e_FE_contour.png}
        \\

        &
        \raisebox{0.15in}{\shortstack{Preliminary\\NO\\ Prediction}} &
        \includegraphics[width=0.1\linewidth]{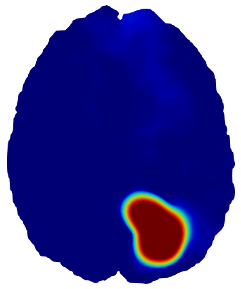}
        \includegraphics[width=0.024\linewidth]{Figure_SI_3/u_NO_contour.png}
        &
        \includegraphics[width=0.1\linewidth]{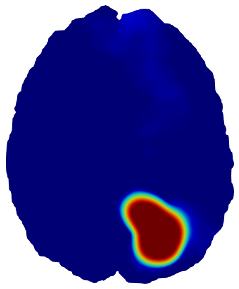}
        \includegraphics[width=0.0275\linewidth]{Figure_SI_3/u_p_NO_contour.png}
        &
        \includegraphics[width=0.1\linewidth]{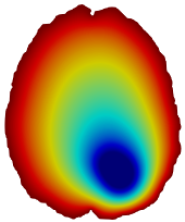}
        \includegraphics[width=0.03\linewidth]{Figure_SI_3/n_NO_contour.png}
        &
        \hspace{0.05in}\includegraphics[width=0.1\linewidth]{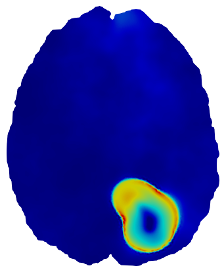}
        \includegraphics[width=0.0365\linewidth]{Figure_SI_3/e_est_contour.png}
        \\

        &
        \raisebox{0.175in}{\shortstack{Pointwise \\ Relative\\ Error}} &
        \hspace{0.07in}\includegraphics[width=0.1\linewidth]{Figure_SI_3/u_err_3.png}
        \includegraphics[width=0.036\linewidth]{Figure_SI_3/u_er_contour.png}
        &
        \hspace{0.07in}\includegraphics[width=0.1\linewidth]{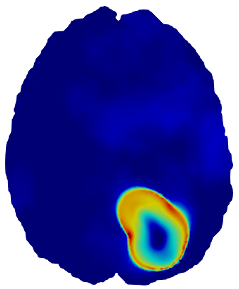}
        \includegraphics[width=0.036\linewidth]{Figure_SI_3/u_p_er_contour.png}
        &
        \includegraphics[width=0.1\linewidth]{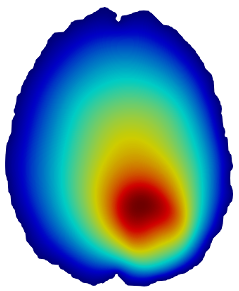}
        \includegraphics[width=0.036\linewidth]{Figure_SI_3/n_er_contour.png}
        &
        \hspace{0.05in}\includegraphics[width=0.1\linewidth]{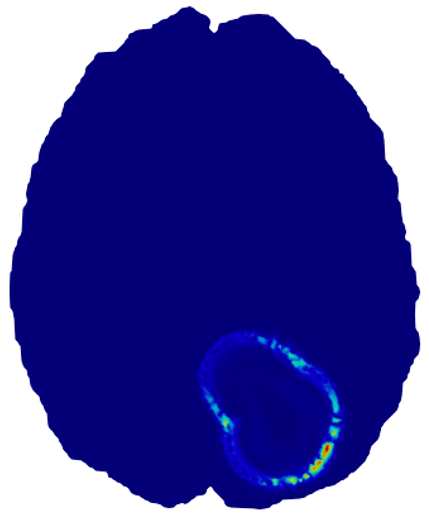}
        \includegraphics[width=0.0365\linewidth]{Figure_SI_3/re_err_contour.png}
        \\

        \multicolumn{6}{c}{\footnotesize (c)}
    \end{tabular}

    \vspace{-0.15in}

    \caption{Comparison of FE solutions (top row) and the preliminary neural operator predictions (middle row) for the state $u$, and auxiliary fields $u^-$ and $n$, alongside pointwise relative error fields (bottom row) for three representative test samples of $m = (\log D, \log G)$ corresponding to panels (a)--(c). The rightmost column displays the error between FE and preliminary NO prediction $e_{\rm FE}$ and the residual estimated error $e_{\rm est}$ using the preliminary NO predictions. The preliminary neural operators were trained on $\mathcal{D}_{\rm pre}$.}
    \label{fig:no_residual}
\end{figure}

We assess the accuracy of vBNO and RI-BNO and evaluate the performance of the preliminary neural operators. Figure \ref{fig:validation_bno} presents a qualitative comparison of the tumor volume fraction fields predicted by vBNO and RI-BNO against the corresponding FE solutions for four unseen samples of the parameter field $m=(\log D,\log G)$. Figure \ref{fig:no_residual} compares the preliminary neural operator predictions of the state and auxiliary fields $u$, $u^{-}$, and $n$ with their FE solutions for three test samples. The figure also shows the true prediction error relative to the FE solution, $e_{\rm FE}$, and the residual estimated error, $e_{\rm est}$,  together with their corresponding pointwise relative error fields. The pointwise relative error field is calculated at each spatial point as the size of the local error divided by the magnitude of the relevant FE solution throughout the computational domain. 
The auxiliary training labels $u^{-}$ and $n$ are obtained from previously saved states corresponding to the samples on the same preliminary training dataset $\mathcal D_{\rm pre}$ used to train the neural operators. For the evaluation of the residual-informed error estimate $e_{\rm est}$, the predicted auxiliary fields $u^{-}$ and $n$ are held fixed, while linearization is done with respect to state field $u$. For both the preliminary neural operators of $u^{-}$ and $n$, we used latent-subspace architecture with 25 dimensional active-subspace and POD representation. For $u^{-}$, we use a 12-layer ResNet of width 40, whereas for $n$, we have a 14-layer ResNet of width 60.

\subsection{MRI inversion: scalable Laplace approximation for physical parameters and gpCN sampling}
\textbf{Maximum A Posteriori (MAP) Estimation}\\
The Laplace approximation models the posterior as a Gaussian distribution centered at the maximum a posteriori (MAP) estimate $m_{\rm MAP}$ with posterior covariance $\bs{\Sigma}_{\rm LA}$, such that $\pi_{\mathrm{post}} \approx \mathcal{N}(m_{\rm MAP},\bs{\Sigma}_{\rm LA})$. Assuming a Gaussian prior distribution $\pi_{\rm pr} = \mathcal{N}(m_{\rm pr}, \bs{\Sigma}_{\rm pr})$, the MAP estimate is defined as the parameter field that minimizes the negative log posterior,
\begin{equation}
m_{\rm MAP} = \underset{m}{\arg\min} \; \frac{1}{2}\left\| \mathcal B(u_{\rm MRI}) - u(m) \right\|_{\bs{\Sigma}_{\rm MRI}^{-1}}^{2} + \frac{1}{2}\left\| m - m_{\rm pr} \right\|_{\bs{\Sigma}_{\rm pr}^{-1}}^{2}.
\end{equation}
To solve this high dimensional optimization problem, we employ an inexact Newton--CG optimization method.\\
\textbf{Laplace Approximation and Posterior Covariance}\\
Once $m_{\rm MAP}$ is found, under the Laplace approximation the posterior covariance $\bs{\Sigma}_{\rm LA}$ is the inverse of the Hessian of the negative log posterior, evaluated at the MAP point,
\begin{equation}
    \bs{\Sigma}_{\rm LA} \approx \bs{\mathcal{H}}(m_{\rm MAP})^{-1} = \left( \bs{\mathcal{H}}_{\rm misfit}(m_{\rm MAP}) + \bs{\Sigma}_{\rm pr}^{-1} \right)^{-1},
\end{equation}
where $\bs{\mathcal{H}}(m_{\rm MAP}) := - \nabla^{2}_{m} \log \pi_{\mathrm{post}} (m \mid u_{\rm MRI})$ is the full Hessian, and $\bs{\mathcal{H}}_{\rm misfit}(m_{\rm MAP}) := - \nabla^2_{m} \log \pi_{\mathrm{like}} (u_{\rm MRI} \mid m)$ denotes the data misfit Hessian.

To quantify parameter uncertainty and explore the full posterior distribution, we employ the generalized preconditioned Crank-Nicolson (gpCN) Markov chain Monte Carlo sampler. By using the Laplace approximation, $\mathcal{N}(m_{\rm MAP},\bs{\Sigma}_{\rm LA})$, as a proposal distribution, gpCN incorporates local posterior curvature information. Posterior samples are then generated according to Algorithm \ref{alg:gpcn}.
\begin{algorithm}[h]
\caption{Generalized Preconditioned Crank-Nicolson (gpCN) Sampler}
\label{alg:gpcn}
\begin{algorithmic}[1]
\State \textbf{Input:} MAP estimate $m_{\mathrm{MAP}}$, posterior covariance from Laplace Approximation $\boldsymbol{\Sigma}_{\mathrm{LA}}$, step size $\Delta s\in(0,1)$, chain length $l_{\rm ch}$, initial state $m^{0}$, observed data $u_{\rm MRI}$
\State \textbf{Define:} $\pi_{\mathrm{LA}}(m \mid u_{\rm MRI}) = \mathcal{N}(m_{\mathrm{MAP}}, \boldsymbol{\Sigma}_{\mathrm{LA}})$

\For{$\ell = 0$ \textbf{to} $l_{\rm ch} - 1$}
    \State Generate a random perturbation: $\boldsymbol{\xi}^{\ell} \sim \mathcal{N}(\mathbf{0}, \boldsymbol{\Sigma}_{\mathrm{LA}})$
    
    \State Propose a candidate state: 
    \Statex \hspace{2em} $m^{\star} \gets m_{\mathrm{MAP}} + \sqrt{1-(\Delta s)^2}\,(m^{\ell}-m_{\mathrm{MAP}}) + \Delta s\,\boldsymbol{\xi}^{\ell}$
    
    \State Solve the forward phase field model using $m^{\star}$ to evaluate $\pi_{\mathrm{post}}(m^{\star}\mid u_{\rm MRI})$
    
    \State Calculate the acceptance probability:
    \Statex \hspace{2em}
    $\varsigma \gets 
    \min\!\left(
    1, 
    \dfrac{\pi_{\mathrm{post}}(m^{\star}\mid u_{\rm MRI})
           \: \pi_{\mathrm{LA}}(m^{\ell}\mid u_{\rm MRI})}
          {\pi_{\mathrm{post}}(m^{\ell}\mid u_{\rm MRI})
           \: \pi_{\mathrm{LA}}(m^{\star} \mid u_{\rm MRI})}
    \right)$
    
    \State Draw a uniform threshold $\chi \sim \mathcal{U}(0,1)$
    
    \If{$\chi \le \varsigma$}
        \State $m^{\ell+1} \gets m^{\star}$ \Comment{Accept proposal}
    \Else
        \State $m^{\ell+1} \gets m^{\ell}$ \Comment{Reject proposal}
    \EndIf
\EndFor

\State \textbf{Output:} Posterior samples $\{m^{\ell}\}_{\ell=1}^{l_{\rm ch}}$
\end{algorithmic}
\end{algorithm}
\begin{table}[h]
    \centering
    \small
    \begin{tabular}{|c|c|c|c|}
    \hline
         \textbf{Parameter} & \textbf{Description} & \textbf{Value}  & \textbf{Reference}\\
         \hline 
         $K$ & phase separation energy of tumor and host tissues &0.025  & \cite{cristini2009nonlinear}\\
         $H$ & interfacial energy among tumor cells and other species &0.01  & \cite{cristini2009nonlinear}\\
         $R$ & coefficient of nutrient consumption&10  & \cite{hinow2009spatial}\\
         $M$ & nutrient mobility & 1 & \cite{hawkins2012numerical,lima2014hybrid}\\
         \hline
    \end{tabular}
    \caption{Summary of physical parameters used in the multiphase glioma growth model.}
    \label{tab:tumor_parameters}
\end{table}
\clearpage
\subsection{MRI inversion: Synthetic training data and FE verification of inferred parameters}
\begin{figure}[h]
    \centering
    \footnotesize
    \setlength{\tabcolsep}{2pt}
    \begin{tabular}{c c c c c c c}
    &
    \includegraphics[width=0.11\linewidth]{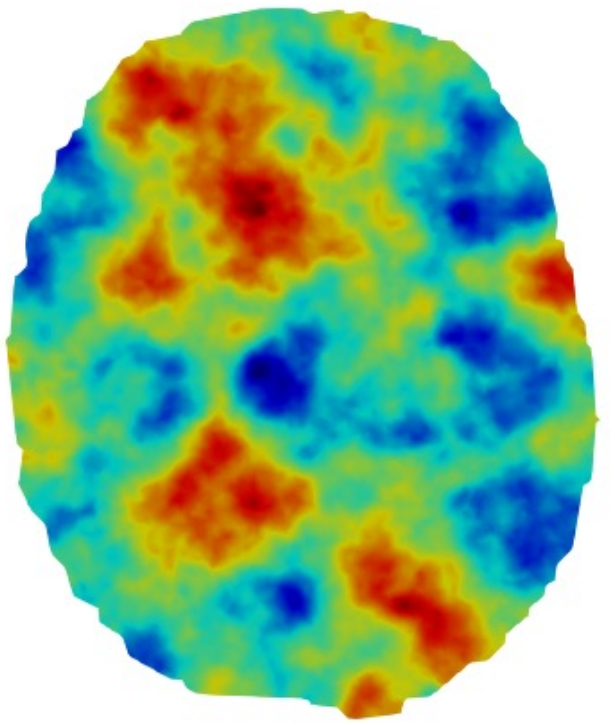} 
    \includegraphics[width=0.045\linewidth]{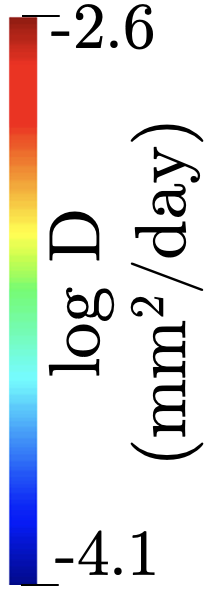}& 
    \includegraphics[width=0.11\linewidth]{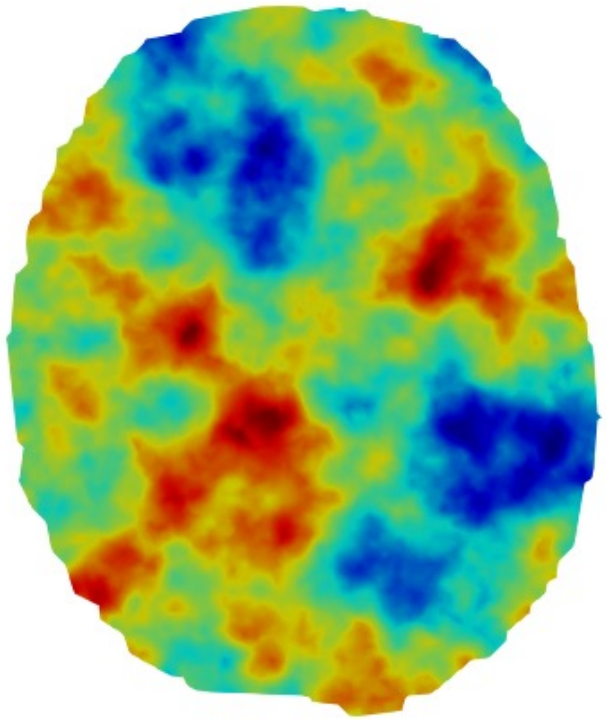} 
    \includegraphics[width=0.045\linewidth]{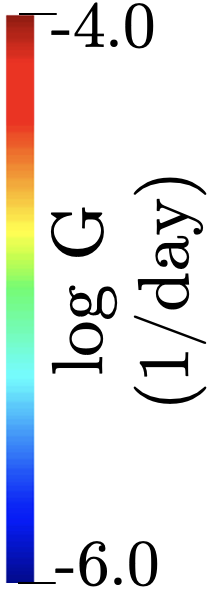} &
    \includegraphics[width=0.11\linewidth]{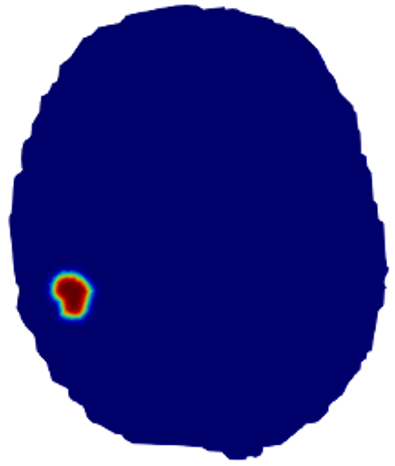} \hspace{0.01in} 
    &
    \includegraphics[width=0.11\linewidth]{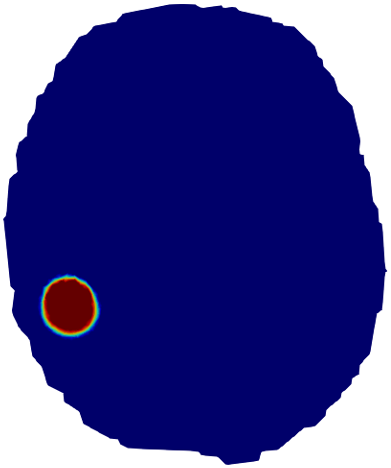}
    \hspace{0.01in} 
    & 
    \includegraphics[width=0.11\linewidth]{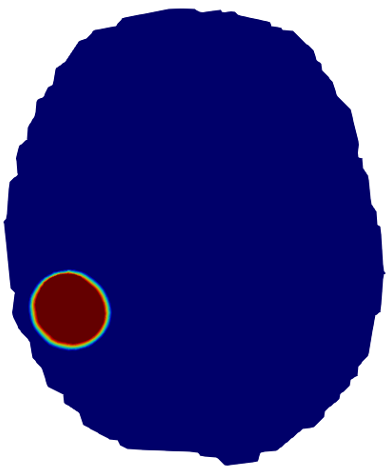}
    \hspace{0.01in} 
    &
    \includegraphics[width=0.11\linewidth]{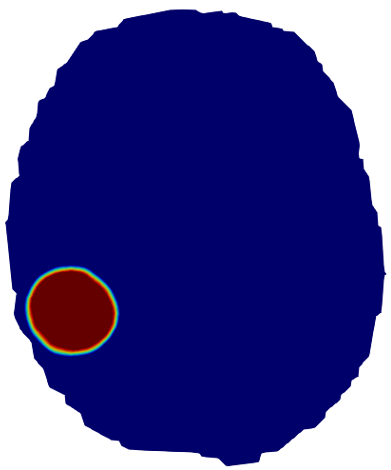}
    \hspace{0.01in} 
    \includegraphics[width=0.027\linewidth]{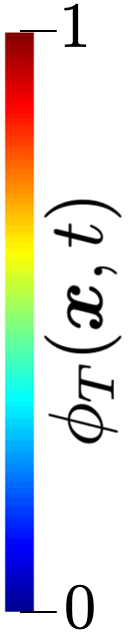}
    \vspace{0.05in}\\
    \multicolumn{7}{c}{(a)}
    \vspace{0.05in}\\
    \raisebox{0.3in}{vBNO}&
    \includegraphics[width=0.11\linewidth]{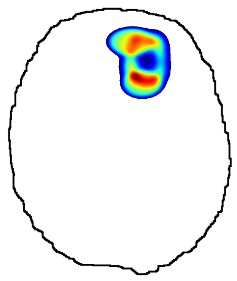} 
    \includegraphics[width=0.045\linewidth]{NO_tumor_training/contour_D.png}& 
    \includegraphics[width=0.11\linewidth]{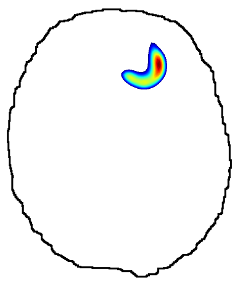} 
    \includegraphics[width=0.045\linewidth]{NO_tumor_training/contour_G.png} &
    \includegraphics[width=0.11\linewidth]{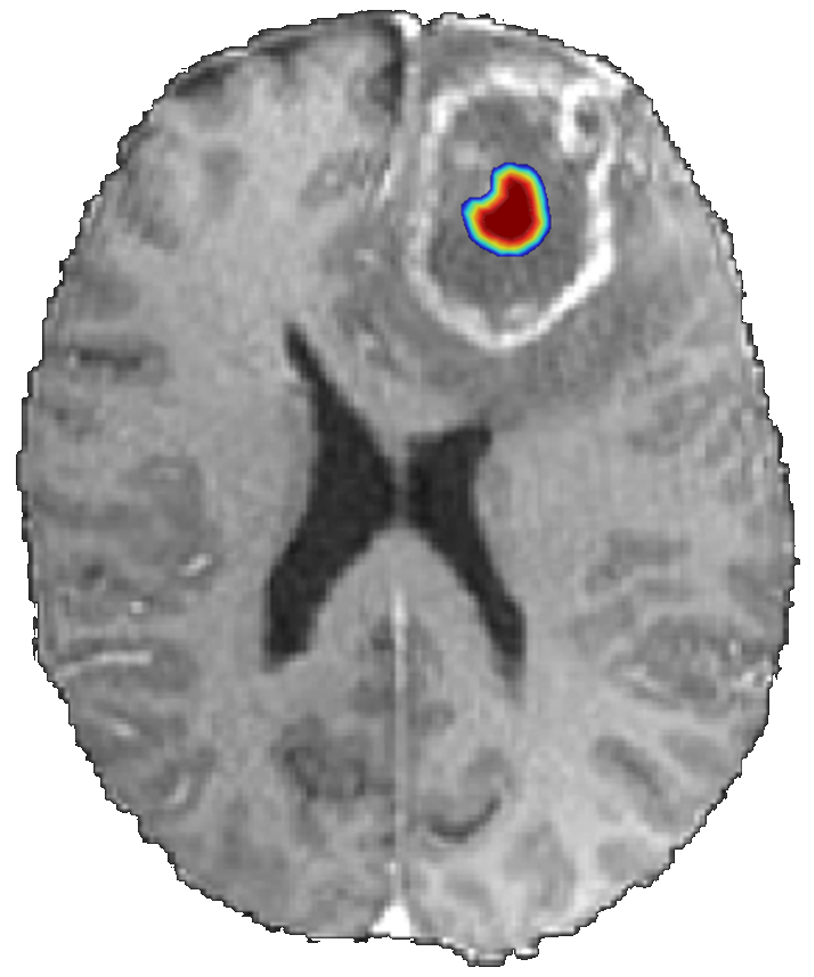} \hspace{0.01in} 
    &
    \includegraphics[width=0.11\linewidth]{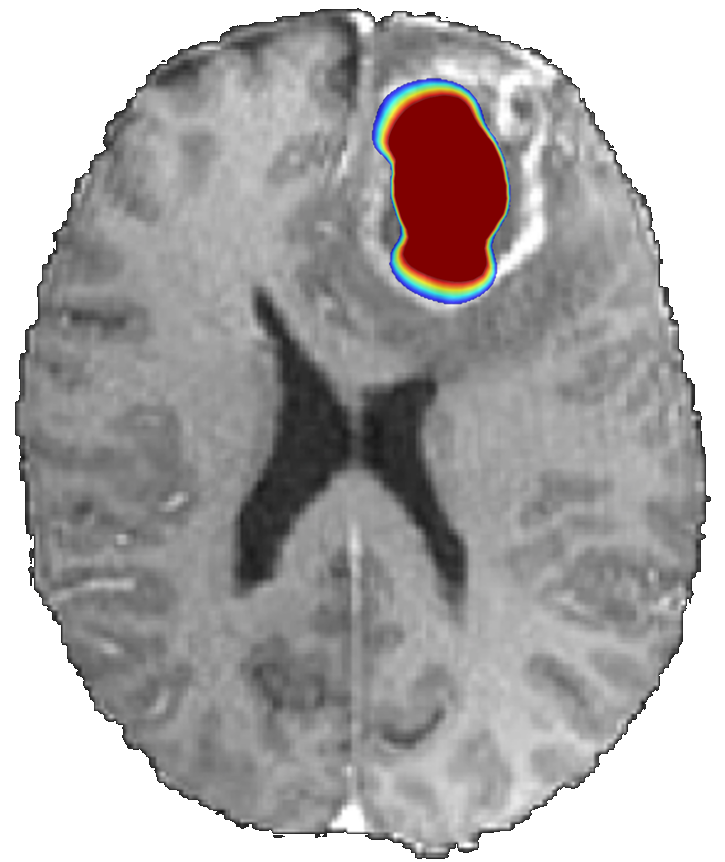}
    \hspace{0.01in} 
    & 
    \includegraphics[width=0.11\linewidth]{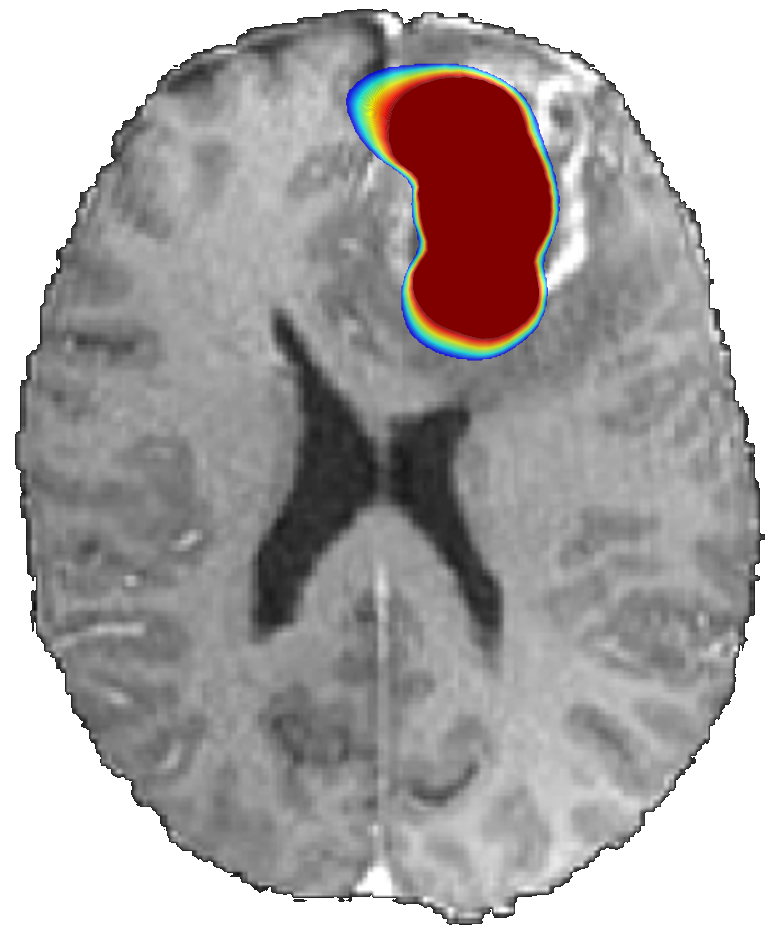}
    \hspace{0.01in} 
    &
    \includegraphics[width=0.11\linewidth]{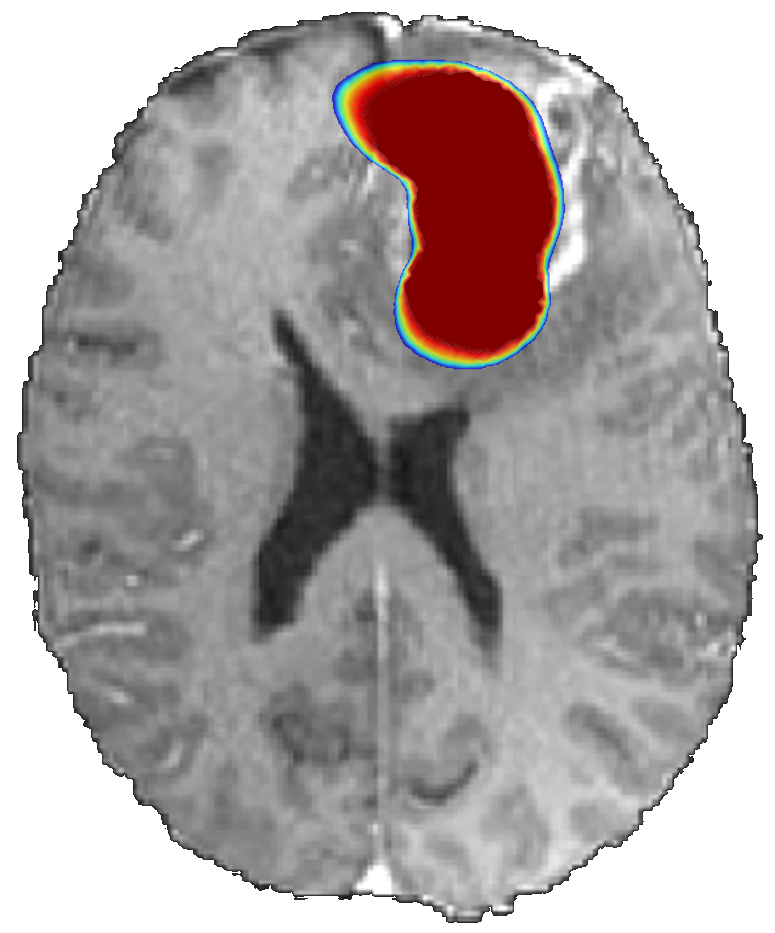}
    \hspace{0.01in} 
    \includegraphics[width=0.027\linewidth]{NO_tumor_training/phi_contour.png}
    \\
    \raisebox{0.3in}{RI-BNO}&
    \includegraphics[width=0.11\linewidth]{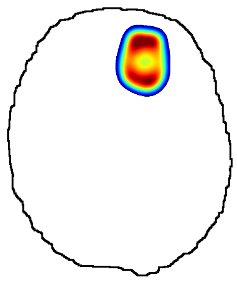} 
    \includegraphics[width=0.045\linewidth]{NO_tumor_training/contour_D.png}& 
    \includegraphics[width=0.11\linewidth]{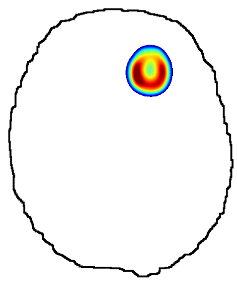} 
    \includegraphics[width=0.045\linewidth]{NO_tumor_training/contour_G.png} &
    \includegraphics[width=0.11\linewidth]{NO_tumor_training/ic.png} \hspace{0.01in} 
    &
    \includegraphics[width=0.11\linewidth]{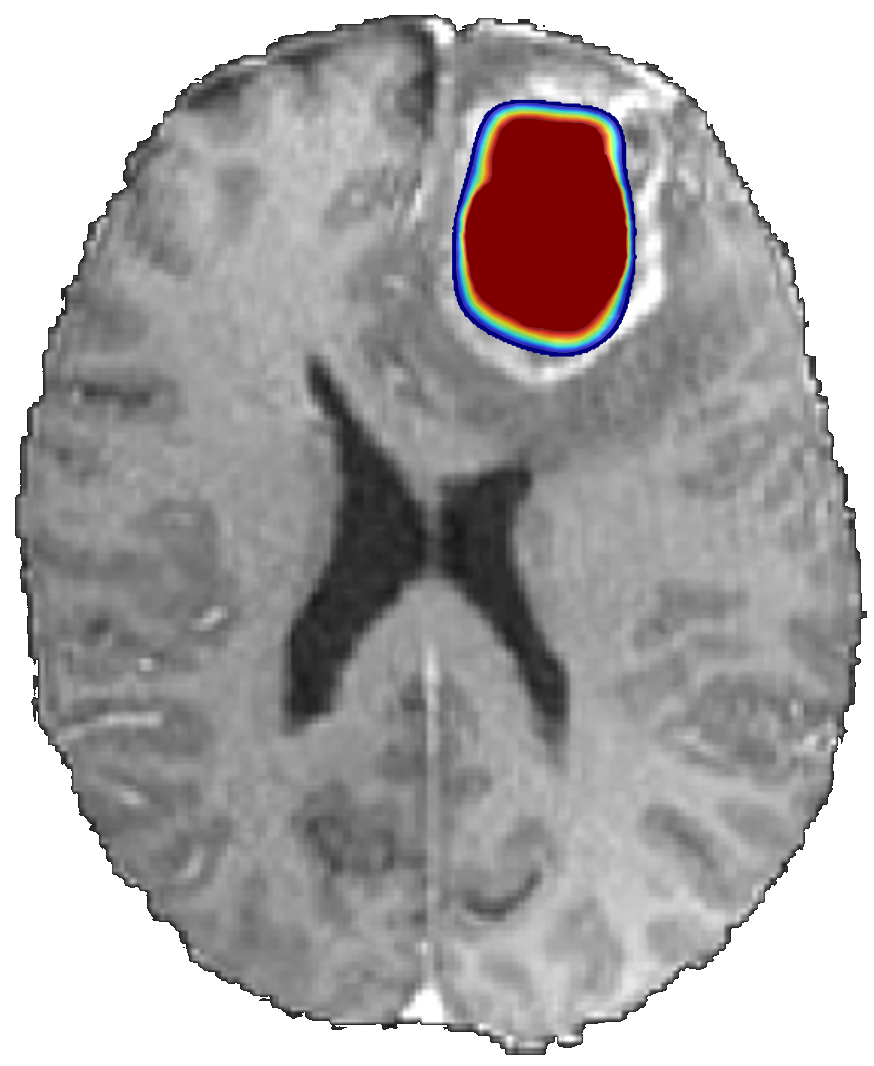}
    \hspace{0.01in} 
    & 
    \includegraphics[width=0.11\linewidth]{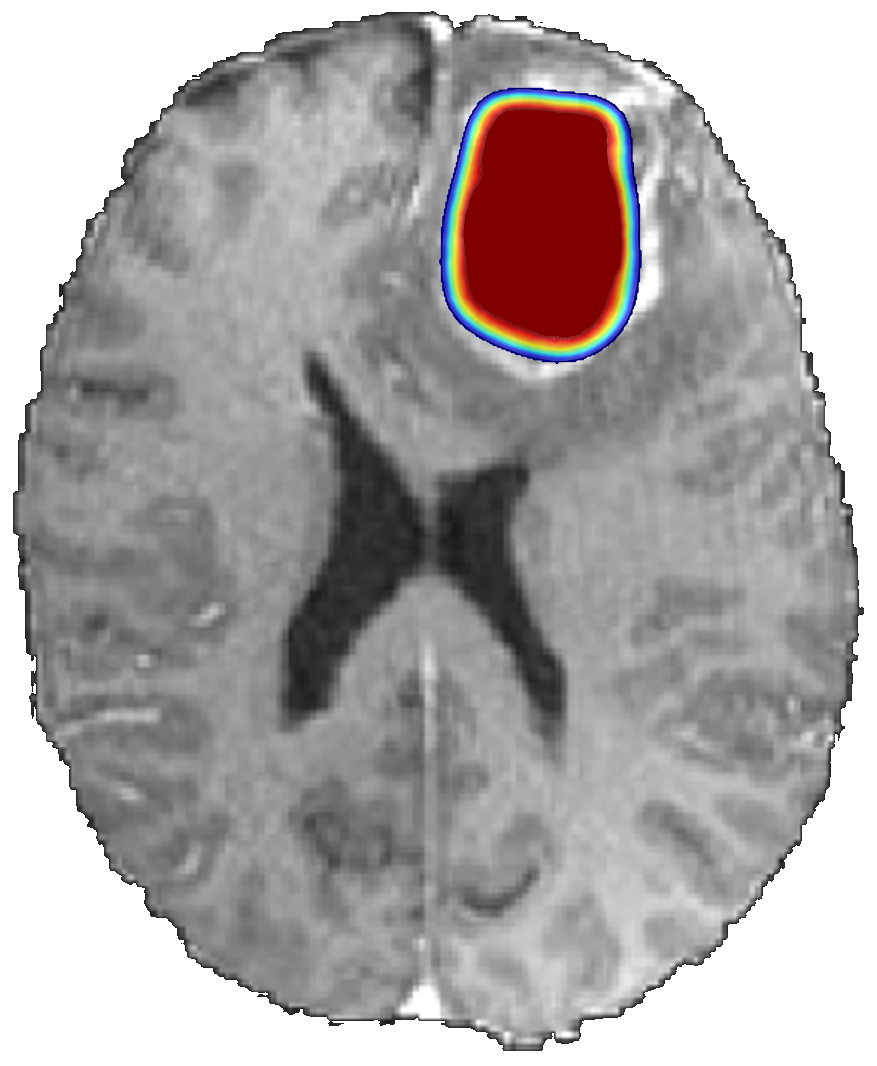}
    \hspace{0.01in} 
    &
    \includegraphics[width=0.11\linewidth]{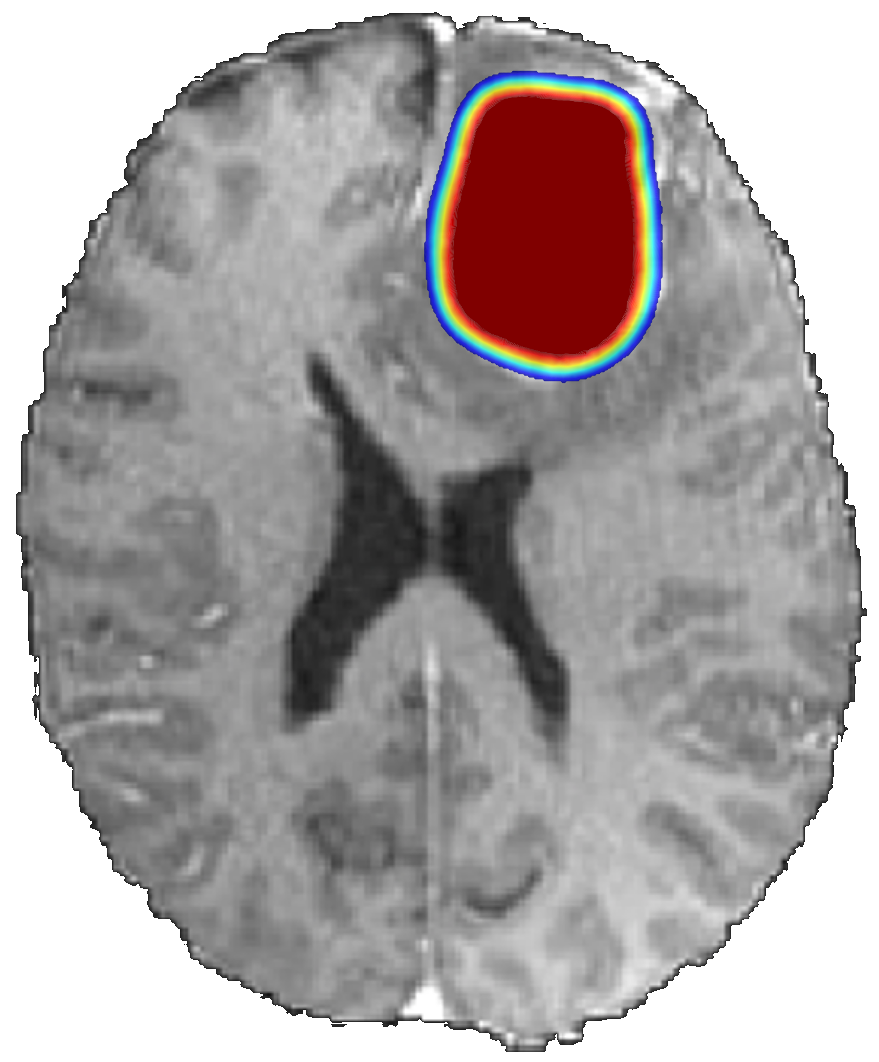}
    \hspace{0.01in} 
    \includegraphics[width=0.027\linewidth]{NO_tumor_training/phi_contour.png}
    \vspace{0.05in}\\
    \multicolumn{7}{c}{(b)}\\
    
\end{tabular}
\vspace{0.05in} \\
\caption{FE tumor trajectories for offline training and verification of MRI-inferred parameter fields.
(a) Representative synthetic training realization, showing the sampled mobility and proliferation fields, \(m=(\log D(\bs x),\log G(\bs x))\), and the corresponding FE tumor evolution from the prescribed training initial condition. The terminal field at \(t_F=240\) days is used for operator learning.
(b) FE tumor trajectories generated using the parameter MAP estimates obtained from vBNO inversion (upper row) and RI-BNO inversion (lower row). Both trajectories use the same patient-specific initial condition adopted in FE-based MRI inversion \cite{subramanian2019simulation}, with the inferred parameter fields held fixed throughout each forward simulation.}
    \label{fig:MRI_training_predictions}
\end{figure}

For offline neural operator training, realizations of the spatially varying parameter fields \(m=(\log D,\log G)\) are propagated through the FE tumor-growth model to \(t_F=240\) days. The simulations use a prescribed initial tumor at a location distinct from the MRI-observed tumor. The sampled parameter fields $m$ and corresponding terminal tumor fields form the training pairs for learning the map $\phi_T(\bs x,t_F)$. Figure~\ref{fig:MRI_training_predictions}(a) shows a representative parameter realization and its FE trajectory. The resulting pretrained operators are subsequently reused for MRI-informed inversion without patient-specific retraining.
Figure~\ref{fig:MRI_training_predictions}(b) shows FE trajectories computed using the parameter MAP estimates obtained from vBNO inversion (upper row) and RI-BNO inversion (lower row). Both simulations use the same patient-specific initial condition and final time adopted in FE-based MRI inversion, without further adjustment of the inferred parameter fields. The terminal FE states are compared with the MRI-derived tumor core and the corresponding neural operator reconstructions to assess whether the reconstruction agreement is retained when the inferred parameters are evaluated through the patient-initialized FE model. The DICE scores for the terminal FE states using MAP estimates from vBNO and RI-BNO are 0.727 and 0.919.
\clearpage

\subsection{MRI inversion: computational cost analysis}
To analyze the performance of the MCMC, we use the following metrics based on integrated autocorrelation time ($\tau_{\rm int}$) and effective sampling speed ($v_{\rm ess}$). First, we would like to define the Effective Sample Size (ESS) percentage $\hat{p}_{\rm{ess}}$ \cite{gelman2013bayesian} as inverse of integrated autocorrelation time ($\tau_{\rm int}$)
\begin{equation} \label{eq:ess_percentage}
    \hat{p}_{\text{ess}} = \frac{100}{\tau_{\text{int}}}
\end{equation}
We also use effective sampling speedup $S_{\rm eff}$ as evaluating the rate of effective sample generation per second of computational effort, ignoring offline training costs. $S_{\rm eff}$ can be defined as
\begin{equation}
    S_{\rm eff} = \frac{\hat{p}_{\rm{ess}}}{\text{Cost of 100 MCMC samples}}
\end{equation}
Tables \ref{tab:ESS}--\ref{tab:mcmc} compare the computational performance of FE, vBNO, and RI-BNO for patient specific MRI inversion. The reported metrics include MCMC sampling statistics, effective sampling speed, and wall clock times for MAP optimization and posterior sampling.
\begin{table}[h!]
\centering
\small
\begin{tabular}{|c|c|c|c|c|}
\hline
\textbf{Method} & \textbf{Forward Eval} & \textbf{Acceptance rate} &  $\tau_{\text{int}}$ & \textbf{ESS \% ($\hat{p}_{\text{ess}}$)} \\ \hline
FE & 2.1 min & 58.3 \% & 25.3 & 3.953 \\ \hline
vBNO & 1.52 s & 43.4 \% & 48.0 & 2.083 \\ \hline
RI-BNO & 1.67 s & 57.2 \%& 27.1 & 3.691 \\ \hline
\end{tabular}
\caption{Comparison of MCMC sampling performance metrics across FE, vBNO, and RI-BNO forward solvers, including wall clock time per iteration for forward evaluation, acceptance rate, integrated autocorrelation time ($\tau_{\text{int}}$), and effective sample size percentage ($\text{ESS}\,\%$).}
\label{tab:ESS}
\end{table}
\begin{table}[h]
\centering
\small
\label{tab:sampling_speed}
\begin{tabular}{|c|c|c|}
\hline
\textbf{Method} & \textbf{Effective Sampling Speed ($S_{\rm{eff}}$)} & \textbf{Speedup vs. FE} \\ \hline
FE  & $3.14 \times 10^{-4}$ ESS/s & 1$\times$ \\ \hline
vBNO & $0.01370$ ESS/s & 43.6$\times$ \\ \hline
RI-BNO & $0.02211$ ESS/s & 70.4$\times$ \\ \hline
\end{tabular}
\caption{Comparison of effective sampling speed $S_{\text{eff}}$ and relative online speedup for MCMC inference using FE, vBNO, and  RI-BNO forward models.}
\end{table}
\begin{table}[h]
\centering
\normalsize
\begin{tabular}{|c|c|c|c|}
\hline
\textbf{Metric} & \textbf{FE} & \textbf{vBNO} & \textbf{RI-BNO} \\ 
\hline
{Wall time(MAP optimization)} & 134.2 min & 21 s & 12.9 s \\
\hline
{Wall time(MCMC)} & 175.1 hrs & 145.2 min & 135.6 min \\
\hline
\end{tabular}
\caption{Wall clock time comparison for MAP optimization and MCMC inference across FE, vBNO, and RI-BNO.}
\label{tab:mcmc}
\end{table}

\clearpage

\clearpage
\subsection{MRI inversion: Pretrained neural operators across patients}
\begin{figure}[h]
    \centering
    \footnotesize
    \setlength{\tabcolsep}{6pt}
    \begin{tabular}{c c c c c}
        &\footnotesize{T1-CE MRI} & \footnotesize{FE} & \footnotesize{vBNO} & \footnotesize{RI-BNO} \vspace{0.05in}\\
        \raisebox{0.4in}{\normalsize{Patient 1}} &
        \includegraphics[width=0.12\linewidth]{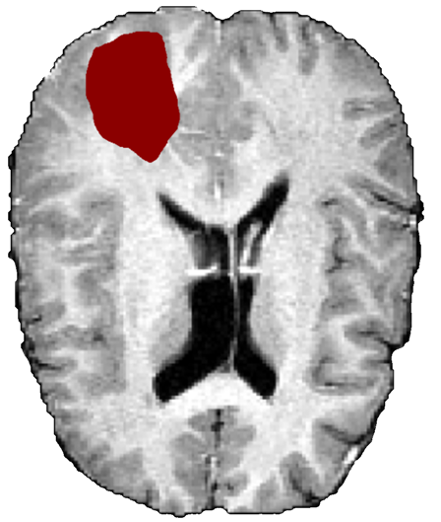}&
        \includegraphics[width=0.12\linewidth]{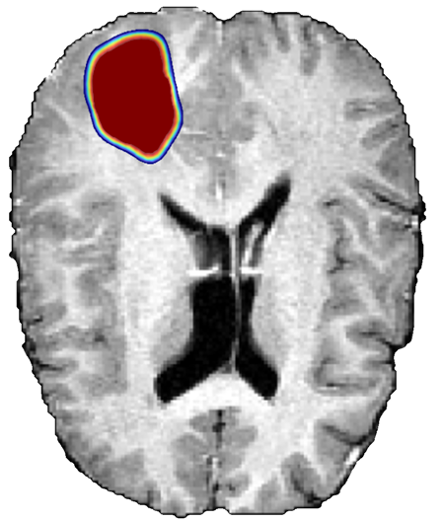}&
        \includegraphics[width=0.12\linewidth]{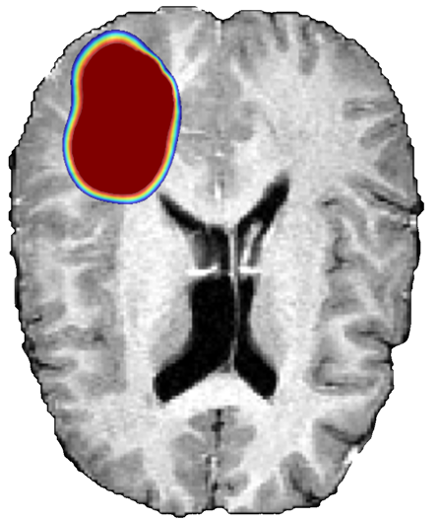}&
        \includegraphics[width=0.12\linewidth]{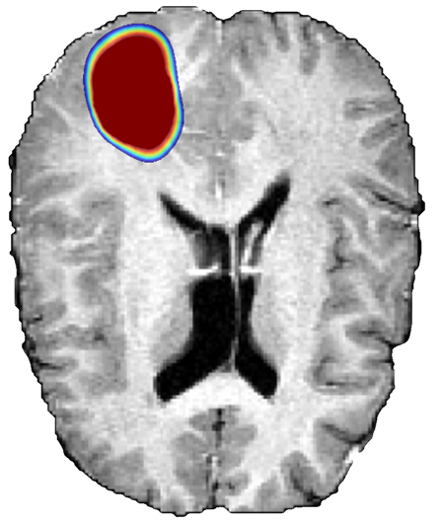}\\
        && DICE=0.937 & DICE=0.715 & DICE=0.912 \\
        \multicolumn{5}{c}{(a)}\\
        \raisebox{0.4in}{\normalsize{Patient 2}} &
        \includegraphics[width=0.12\linewidth]{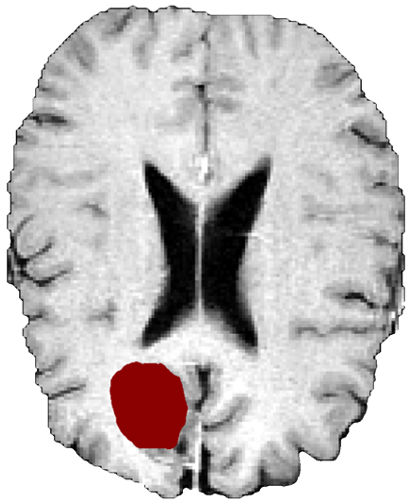}&
        \includegraphics[width=0.12\linewidth]{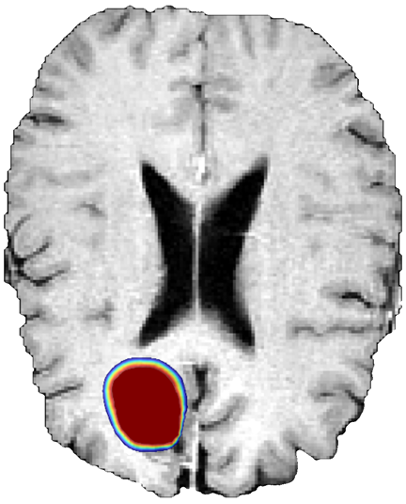}&
        \includegraphics[width=0.12\linewidth]{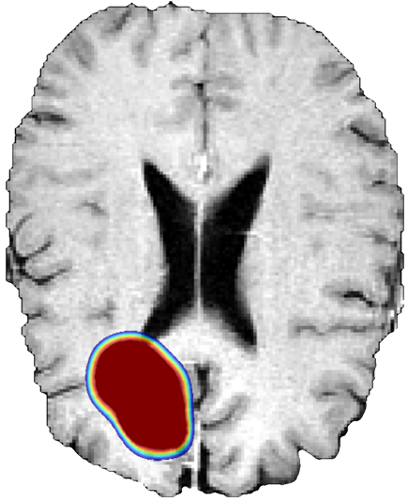}&
        \includegraphics[width=0.12\linewidth]{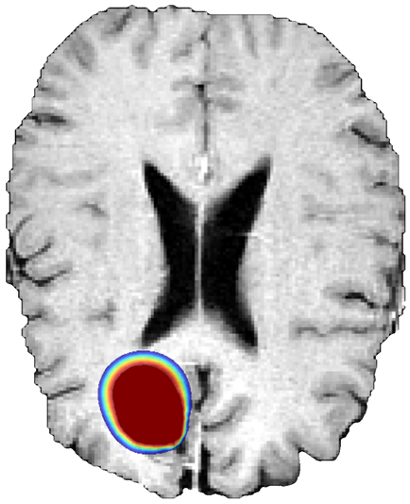}\\
        && DICE=0.939 & DICE=0.734 & DICE=0.914 \\
        \multicolumn{5}{c}{(b)}\\
        \raisebox{0.4in}{\normalsize{Patient 3}} &
        \includegraphics[width=0.12\linewidth]{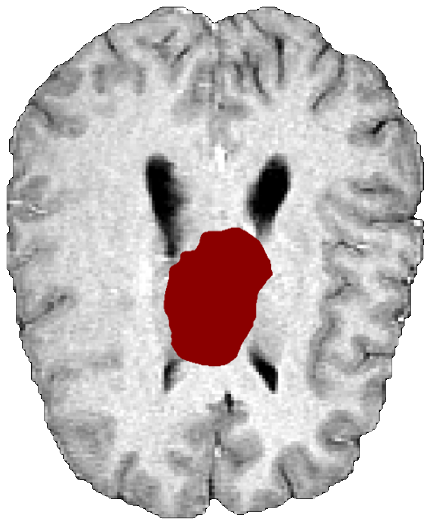}&
        \includegraphics[width=0.12\linewidth]{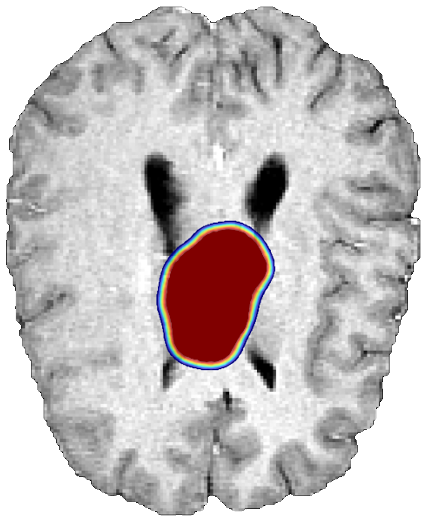}&
        \includegraphics[width=0.12\linewidth]{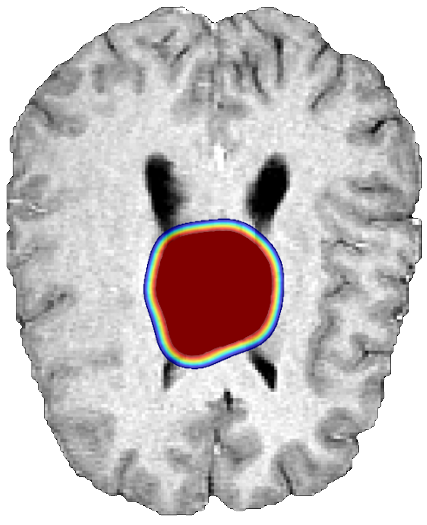}&
        \includegraphics[width=0.12\linewidth]{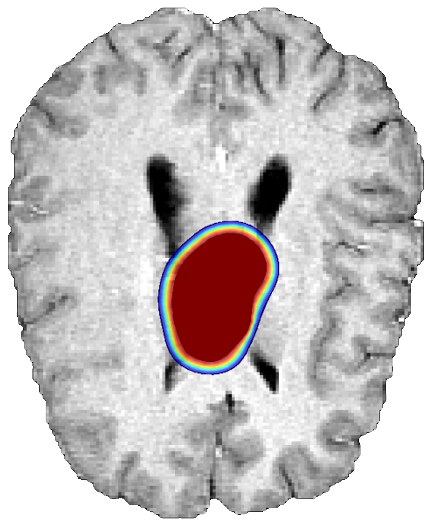}\\
        && DICE=0.928 & DICE=0.742 & DICE=0.908 \\
        \multicolumn{5}{c}{(c)}\\
        \raisebox{0.4in}{\normalsize{Patient 4}} &
        \includegraphics[width=0.12\linewidth]{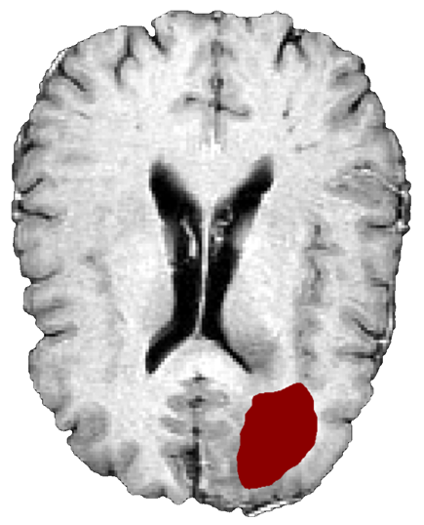}&
        \includegraphics[width=0.12\linewidth]{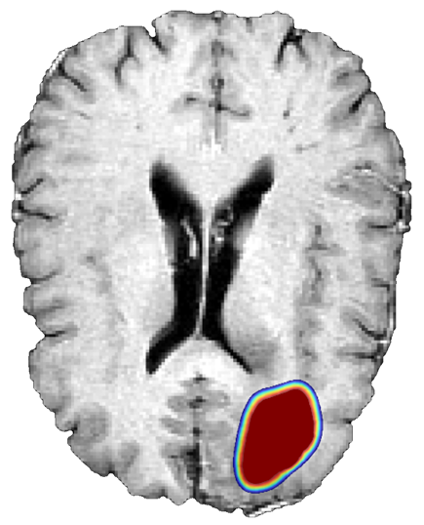}&
        \includegraphics[width=0.12\linewidth]{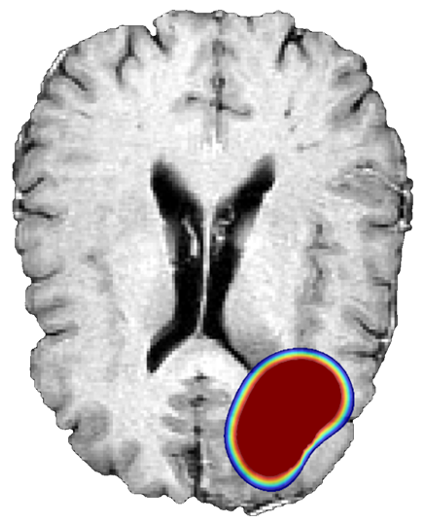}&
        \includegraphics[width=0.12\linewidth]{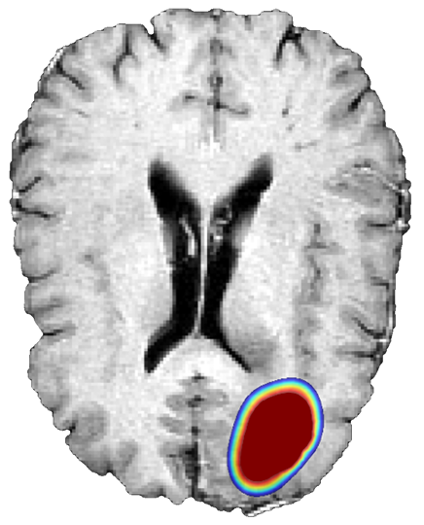}\\
        && DICE=0.941 & DICE=0.736 & DICE=0.914 \\
        \multicolumn{5}{c}{(d)}\\
    \end{tabular}
    \caption{MRI-informed tumor reconstruction using pretrained neural operators across four additional patients.
Rows correspond to BraTS 2019 cases (a) BraTS19\_TCIA10\_413\_1, (b) BraTS19\_TCIA10\_266\_1, (c) BraTS19\_TCIA10\_351\_1, and (d) BraTS19\_TCIA10\_276\_1.
Columns show the axial T1-CE MRI with its annotated tumor core and the tumor inversion obtained using FE, vBNO and RI-BNO, evaluated at the parameter MAP estimates from the corresponding inverse problems.
The same pretrained neural operators are reused for all patients.}
    \label{fig:MRI_multiple_patients}
\end{figure}

To assess the reuse of pretrained neural operators beyond the primary example in the manuscript, we perform MRI-informed inversion for four additional patients from the BraTS 2019 dataset \cite{menze2014multimodal,bakas2017advancing,bakas2018identifying}. All cases use the same pretrained vBNO and RI-BNO constructed from the synthetic training data shown in Figure~\ref{fig:MRI_training_predictions}(a), without patient-specific operator retraining or additional FE training trajectories. These training simulations originate from a prescribed initial tumor at a location distinct from the MRI-observed tumors.
All cases use the fixed modeling horizon \(t_F=240\) days, interpreted as an assumed growth duration rather than an estimate of patient-specific tumor age. For FE-based inversion, the initial tumor is placed within the MRI-visible tumor core using a common placement rule, with its size and interface width held fixed across cases.
Figure~\ref{fig:MRI_multiple_patients} compares tumor reconstructions evaluated at the parameter MAP estimates obtained using FE, vBNO and RI-BNO inversion. Across the four additional patients, the mean Dice scores relative to the MRI-derived tumor core masks are 0.936, 0.732, and 0.912, respectively. RI-BNO improves reconstruction accuracy over vBNO in every case, with Dice scores differing from the FE results by 0.020--0.027. These results support the reuse of the pretrained RI-BNO for parameter inference from additional MRI observations, while maintaining reconstruction accuracy close to that of FE-based inversion.

\clearpage

\section{Thermal break design and finite element validation}
This section provides additional results supporting the risk averse design optimization study. We primarily present FE based risk analysis of the optimal designs obtained using FE, vBNO, and RI-BNO, along with a brief discussion on the associated computational costs for the neural operator learning and risk averse optimization.
\subsection{Validation of optimal design through Finite Element risk evaluation}
\begin{figure}[h]
    \centering
    \setlength{\tabcolsep}{10pt}
    \begin{tabular}{c c c c}
    \shortstack{FE\\optimized\\design}& 
    \shortstack{vBNO\\optimized\\design}& 
    \shortstack{RI-BNO\\optimized\\design}& \vspace{0.1in}\\
         \includegraphics[width=0.125\linewidth]{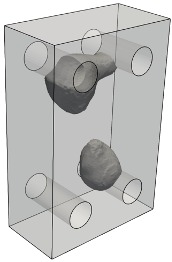}& 
         \includegraphics[width=0.125\linewidth]{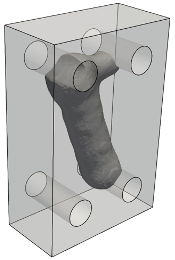}& 
         \includegraphics[width=0.125\linewidth]{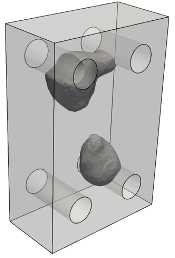}&
         \includegraphics[width=0.038\linewidth]{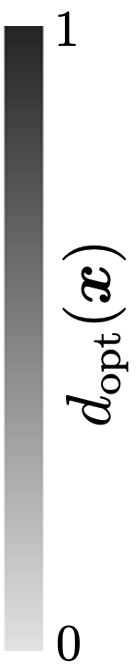}\\
        \multicolumn{4}{c}{\footnotesize{(a)}} \\
        \noalign{\smallskip}
         \includegraphics[width=0.125\linewidth]{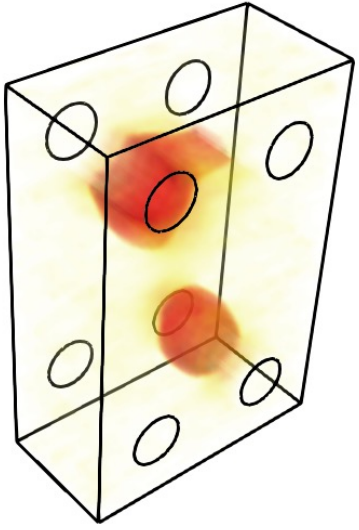}& 
         \includegraphics[width=0.125\linewidth]{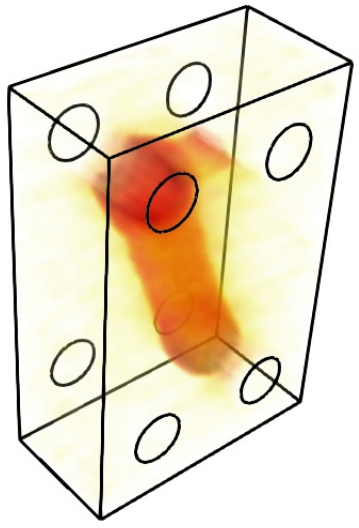}& 
         \includegraphics[width=0.125\linewidth]{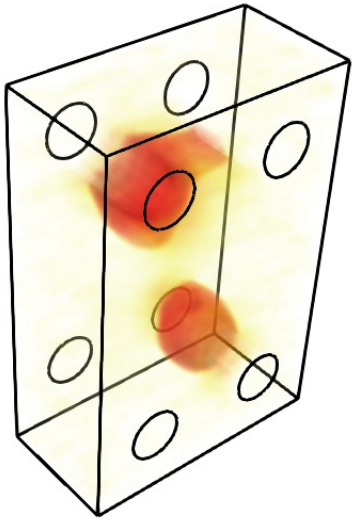}&  
         \includegraphics[width=0.042\linewidth]{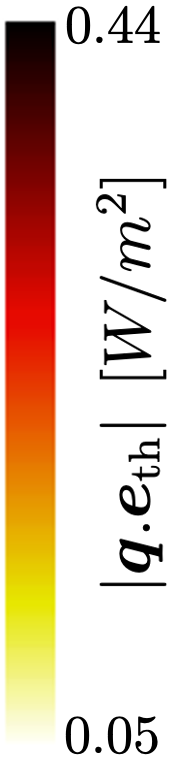}\\
         \includegraphics[width=0.125\linewidth]{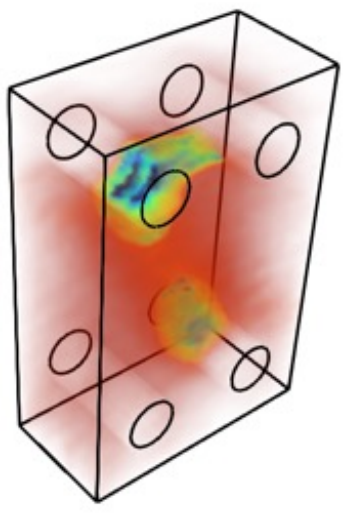}& 
         \includegraphics[width=0.125\linewidth]{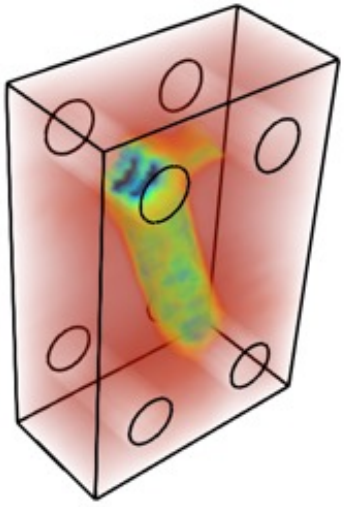}& 
         \includegraphics[width=0.125\linewidth]{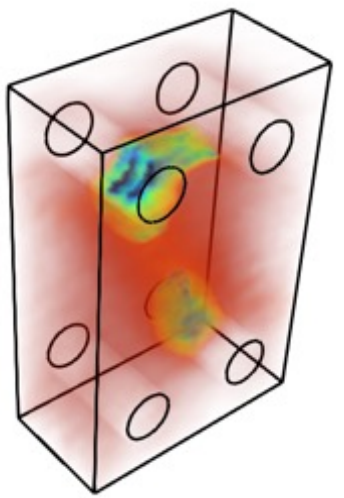}&   
         \includegraphics[width=0.042\linewidth]{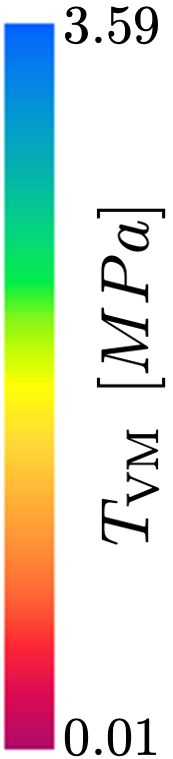}\\
        \multicolumn{4}{c}{\footnotesize{(b)}} \\
        \noalign{\smallskip}
         \includegraphics[width=0.125\linewidth]{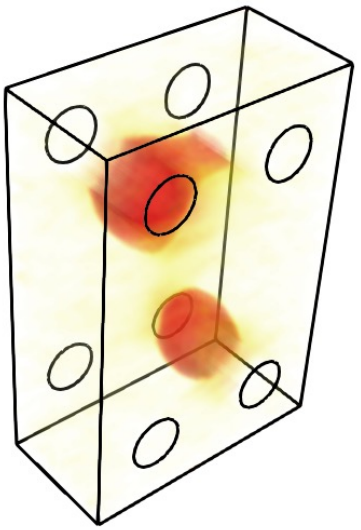}& 
         \includegraphics[width=0.125\linewidth]{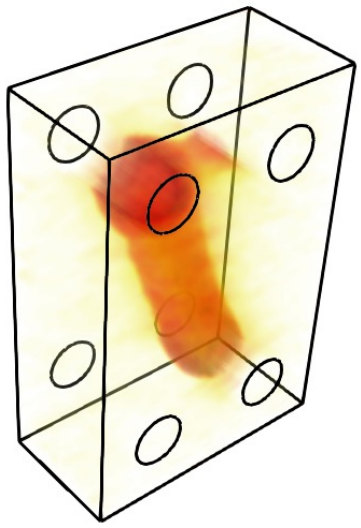}& 
         \includegraphics[width=0.125\linewidth]{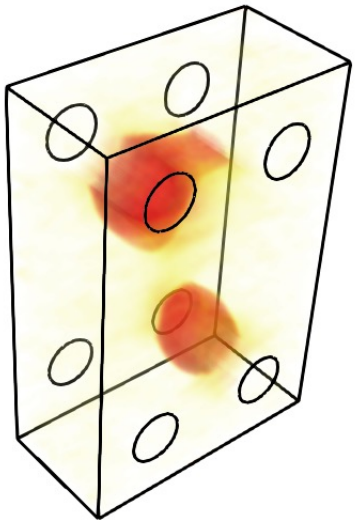}&   
         \includegraphics[width=0.042\linewidth]{Figure_SI_2/contour_hf.png}\\
         \includegraphics[width=0.125\linewidth]{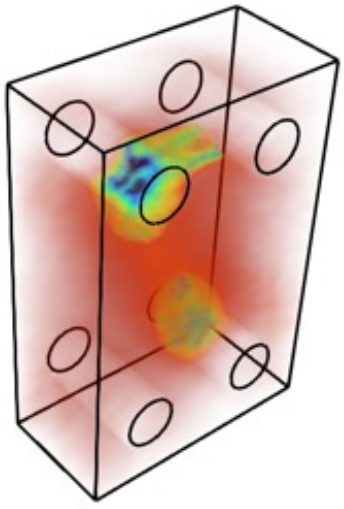}& 
         \includegraphics[width=0.125\linewidth]{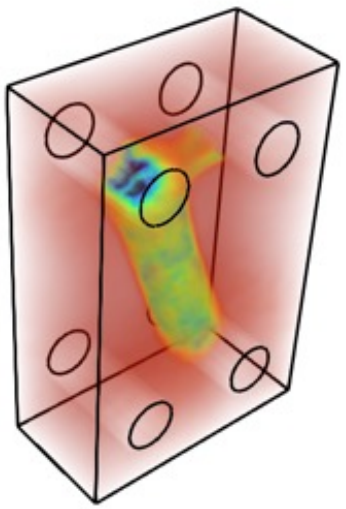}& 
         \includegraphics[width=0.125\linewidth]{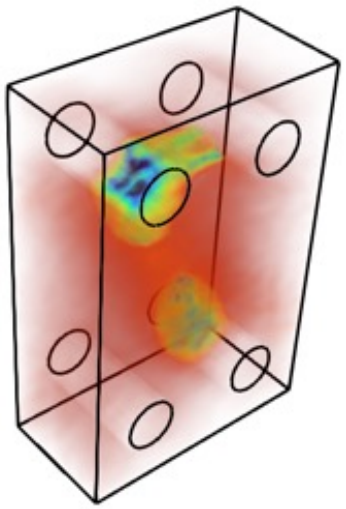}&   
         \includegraphics[width=0.042\linewidth]{Figure_SI_2/contour_vm.png}\\
        \multicolumn{4}{c}{\footnotesize{(c)}} \\
        \noalign{\smallskip}
    \end{tabular}
    \caption{Comparison of optimal designs and FE-evaluated fields : (a) Optimal design $d_{\rm{opt}}(\boldsymbol{x})$ obtained via FE, vBNO, and RI-BNO ; (b) through thickness heat flux magnitude $|\bs{q}\cdot \mathbf{e}_{\rm th}|$ and von Mises stress $T_{\rm VM}$ fields for manufacturing realization $m_A$; and (c) corresponding field predictions for manufacturing realization $m_B$. Both fields shown in (b) and (c) are evaluated using FE.}
    \label{fig:realization_design}
\end{figure}

We evaluate the three optimized designs using FE simulations for the optimal designs obtained through FE, vBNO and RI-BNO. For this, we select a common set of independent manufacturing realizations of $m \sim \nu_{\rm man}$, which are used to evaluate the states using the FE model. We then compute the corresponding thermal objective $Q$ and the stress violation functional $g$. We also evaluate the risk indicators, which are insulation-risk indicator $r_Q = (Q-Q_{\rm cr})/Q_{\rm cr}$ and mechanical risk indicator $r_T = g/T_{\rm cr}$. The mean, Value at Risk (VaR) and Conditional Value at Risk (CVaR) are calculated for the risk analysis of our designs. Figure \ref{fig:realization_design} shows the magnitude of heat flux through thickness and the von Mises stress evaluated for two representive manufacturing realizations. Qualitatively, the heat flux through thickness and von Mises stress distributions for RI-BNO are closely aligned with FE, whereas the vBNO design clearly exhibits excessive fiber network across both realizations. 

Tables \ref{tab:heat_leakage} and \ref{tab:stress_violation} summarize the FE-evaluated risk metrics for \(Q\), \(g\), and their associated risk indicators over \(5{,}000\) manufacturing realizations. The RI-BNO design closely reproduces the FE benchmark in both mean and tail-risk statistics. In particular, the thermal risk indicators remain negative, indicating desired operation within the superinsulating regime. In contrast, the vBNO design experiences significant thermal degradation, producing positive thermal risk indicators and entering the insulation failure regime. Although all three designs satisfy the mechanical stress constraint, the lower mechanical risk observed for vBNO arises from excessive fiber reinforcement rather than a balanced design. The VaR and CVaR statistics further demonstrate that RI-BNO accurately captures not only the mean behavior but also the adverse tail of the manufacturing distribution, whereas vBNO exhibits substantially larger thermal tail risk.
\begin{table}[h]
    \centering
    \small
    \begin{tabular}{|c|c|c|c|c|c|c|}
    \hline
    \rule{0pt}{2.5ex}
         Method & $\overline{Q}$ & \(\operatorname{VaR}_{0.95}[Q]\) & \(\operatorname{CVaR}_{0.95}[Q]\) & $\overline{r_Q}$& \(\operatorname{VaR}_{0.95}[r_Q]\) & \(\operatorname{CVaR}_{0.95}[r_Q]\) \\
         \hline
         FE      & 1.971  & 2.015  & 2.031 & -0.150  & -0.131 & -0.125\\
         \hline
         vBNO    & 2.379  & 2.437  & 2.459 & 0.025 & 0.051  & 0.061\\
         \hline
         RI-BNO  & 2.024  & 2.068  & 2.088 & -0.127 & -0.108 & -0.101\\
         \hline
    \end{tabular}
    \caption{Comparison of statistical risk measures (expected value, 95\% VaR, and 95\% CVaR) for heat leakage $Q \: (\mathrm{W}^2/\mathrm{m})$ and insulation risk $r_Q$ evaluated across 5,000 Monte Carlo realizations. The optimal designs were generated using the FE, vBNO, and RI-BNO frameworks, while all performance metrics were evaluated using the FE model.}
    \label{tab:heat_leakage}
\end{table}
\begin{table}[h]
    \centering
    \small
    \begin{tabular}{|c|c|c|c|c|c|c|}
    \hline
    \rule{0pt}{2.5ex}
         Method & $\overline{g}$ & \(\operatorname{VaR}_{0.95}[g]\) & \(\operatorname{CVaR}_{0.95}[g]\) & $\overline{r_T}$& \(\operatorname{VaR}_{0.95}[r_T]\) & \(\operatorname{CVaR}_{0.95}[r_T]\) \\
         \hline
         FE      & -0.161  & -0.128  & -0.121 & -0.041  & -0.033 & -0.031\\
         \hline
         vBNO    & -0.433  & -0.398  & -0.391 & -0.111 & -0.102  & -0.100\\
         \hline
         RI-BNO  & -0.190  & -0.162  & -0.156 & -0.049 & -0.042 & -0.040\\
         \hline
    \end{tabular}
    \caption{Comparison of statistical risk measures (expected value, 95\% VaR, and 95\% CVaR) for stress violation $g \: (\rm{MPa)}$ and mechanical risk $r_T$ evaluated across 5,000 Monte Carlo realizations for full FE, vBNO, and RI-BNO models. The optimal designs were generated using the FE, vBNO, and RI-BNO frameworks, while all performance metrics were evaluated using the FE model.}
    \label{tab:stress_violation}
\end{table}
\begin{table}[h]
    \centering
    \small
    \begin{tabular}{|c|c|c|c|}
    \hline
         \textbf{Parameter} & \textbf{Description} & \textbf{Value}  & \textbf{Reference}\\
         \hline 
         $K_f$ & thermal conductivity of fiber phase &0.066 $\mathrm{W m^{-1} K^{-1}}$  & \cite{guo2025wheat}\\
         $K_s$ & thermal conductivity of solid phase &0.5 $\mathrm{W m^{-1} K^{-1}}$  & \cite{guo2021cost}\\
         $K_g$ & thermal conductivity of gaseous phase &0.08 $\mathrm{W m^{-1} K^{-1}}$  & \cite{guo2021cost}\\
         $\lambda_f$ & lam\'{e} first parameter for fiber phase &5.77 MPa  & \cite{guo2025wheat}\\
         $\mu_f$ & lam\'{e} second parameter for fiber phase &3.84 MPa  & \cite{guo2025wheat}\\
         $\lambda_s$ & lam\'{e} first parameter for solid phase &0.7 MPa  & \cite{guo2021cost}\\
         $\mu_s$ & lam\'{e} second parameter for solid phase &0.27 MPa  & \cite{guo2021cost}\\
         $H_{\rm amb}$ & convective heat transfer coefficient & 25 $\mathrm{W m^{-2} K^{-1}}$ & ---\\
         $C$ & pore compressibility coefficient & 8.5 $\times 10^{-9}$ $\mathrm{Pa^{-1}}$ & \cite{tan2022predictive}\\
         \hline
    \end{tabular}
    \caption{Summary of physical parameters used in the three phase thermomechanical model of wheat straw aerogel composites.}
    \label{tab:duu_parameters}
\end{table}
\clearpage
\subsection{Input and output fields for neural operator training dataset}
\begin{figure}[h]
    \centering
    \setlength{\tabcolsep}{2pt}
    \begin{tabular}{c c c c}
    \multirow{3}{*}[2.5em]{
    \includegraphics[width= 0.13\linewidth]{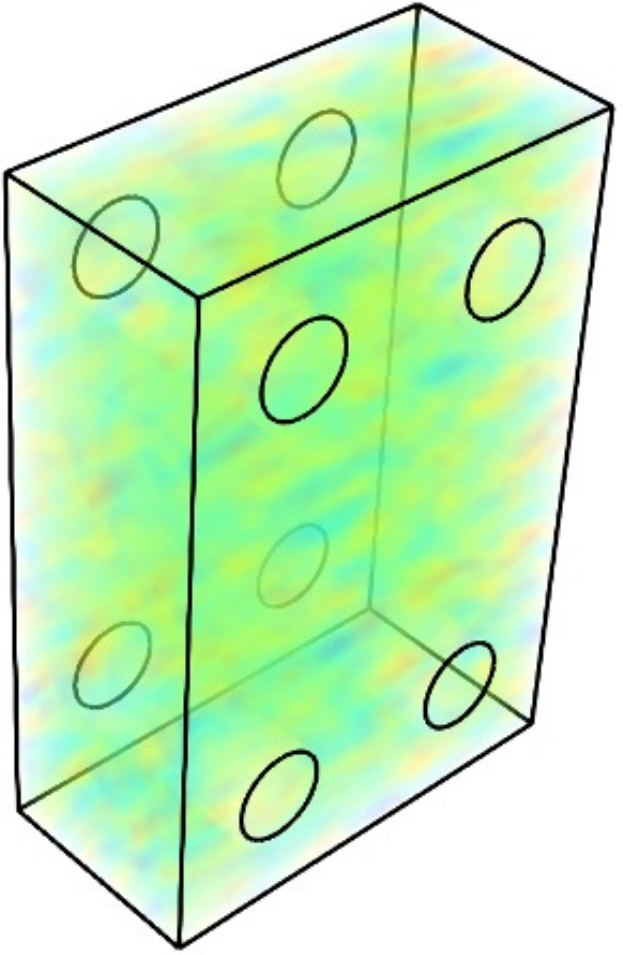} \includegraphics[width=0.04\linewidth]{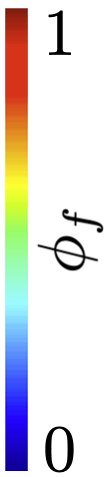}} \hspace{0.2in}
     & 
    \hspace{0.15in}\includegraphics[width=0.13\linewidth]{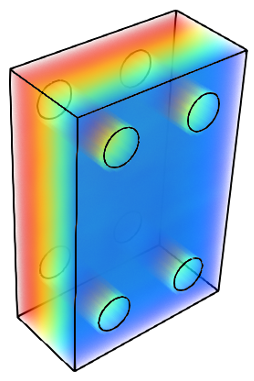}%
    \includegraphics[width=0.057\linewidth]{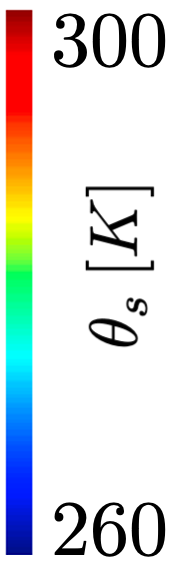} &
    \hspace{0.15in}\includegraphics[width=0.13\linewidth]{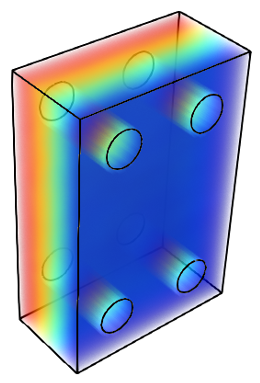}%
    \includegraphics[width=0.057\linewidth]{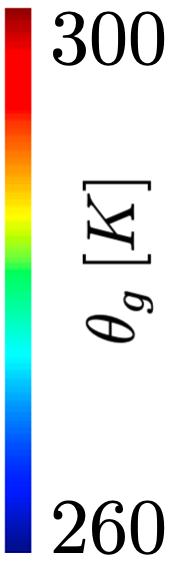} &
    \hspace{0.15in}\includegraphics[width=0.13\linewidth]{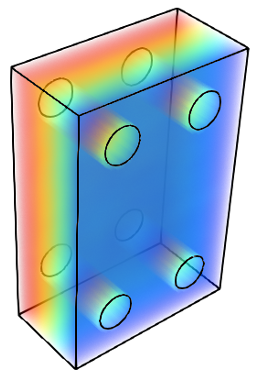}%
    \includegraphics[width=0.057\linewidth]{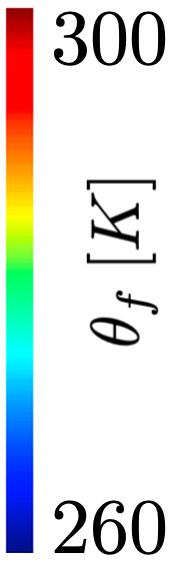} \\
    & 
    \hspace{0.175in}\includegraphics[width=0.13\linewidth]{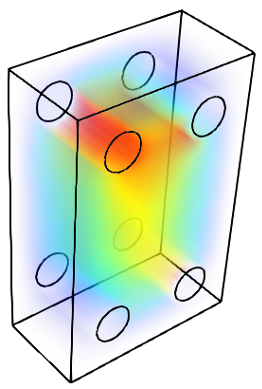}%
    \includegraphics[width=0.06\linewidth]{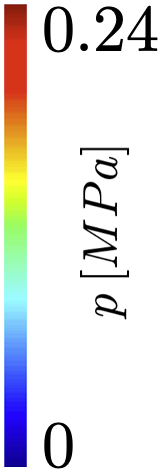} &
    \hspace{0.175in}\includegraphics[width=0.13\linewidth]{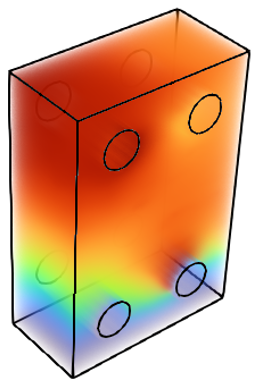}%
    \includegraphics[width=0.06\linewidth]{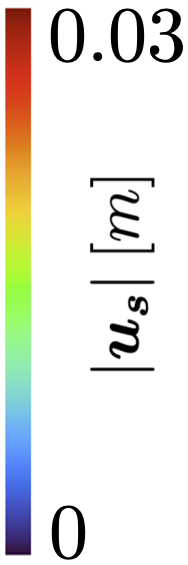} &
    \hspace{0.175in}\includegraphics[width=0.13\linewidth]{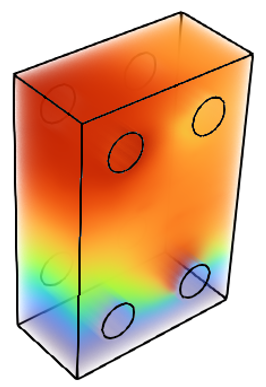}%
    \includegraphics[width=0.06\linewidth]{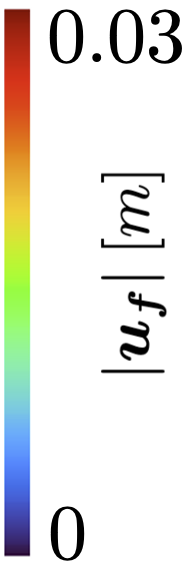} \\
    \multicolumn{3}{c}{\hspace{1.5in}\small{(a)}}\vspace{0.025in}\\
    \multirow{3}{*}[2.5em]{
    \includegraphics[width= 0.13\linewidth]{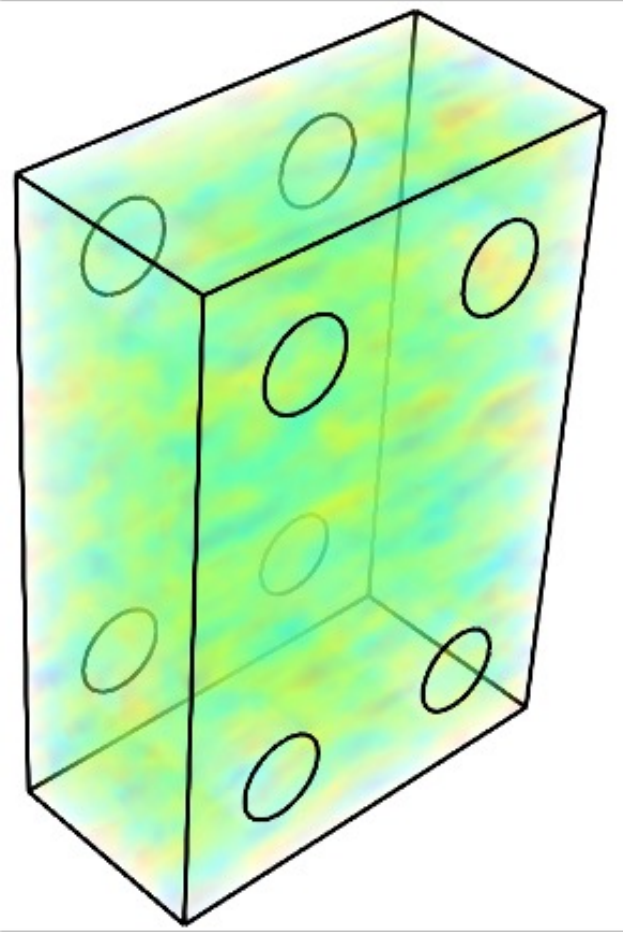} \includegraphics[width=0.04\linewidth]{Figure_SI_4/phif_contour.png}} \hspace{0.2in} & 
    \hspace{0.15in}\includegraphics[width=0.13\linewidth]{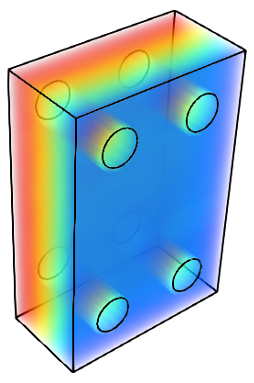}%
    \includegraphics[width=0.057\linewidth]{Figure_SI_4/T_s_contour.png} &
    \hspace{0.15in}\includegraphics[width=0.13\linewidth]{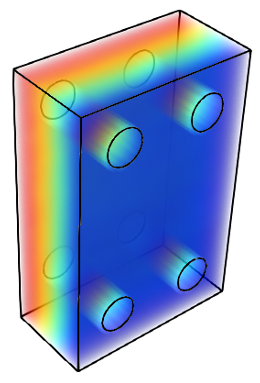}%
    \includegraphics[width=0.057\linewidth]{Figure_SI_4/T_g_contour.png} &
    \hspace{0.15in}\includegraphics[width=0.13\linewidth]{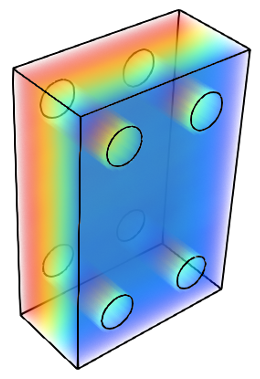}%
    \includegraphics[width=0.057\linewidth]{Figure_SI_4/T_f_contour.png} \\
    & 
    \hspace{0.175in}\includegraphics[width=0.13\linewidth]{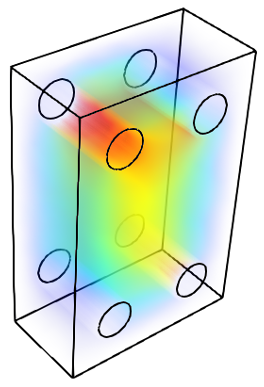}%
    \includegraphics[width=0.06\linewidth]{Figure_SI_4/p_contour.png} &
    \hspace{0.175in}\includegraphics[width=0.13\linewidth]{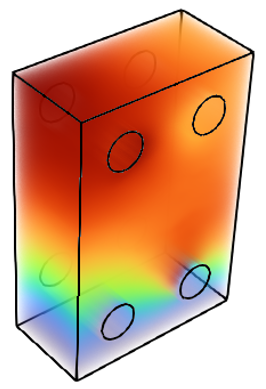}%
    \includegraphics[width=0.06\linewidth]{Figure_SI_4/D_s_contour.png} &
    \hspace{0.175in}\includegraphics[width=0.13\linewidth]{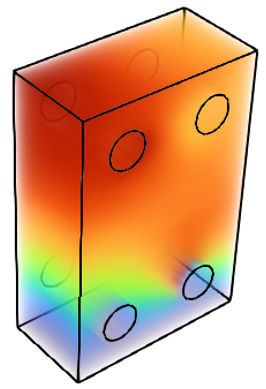}%
    \includegraphics[width=0.06\linewidth]{Figure_SI_4/D_f_contour.png} \\
    \multicolumn{3}{c}{\hspace{1.5in}\small{(b)}} \vspace{0.025in}\\
    \end{tabular}
    \caption{Realizations of the training data for input $\phi_f$ and the six output states which are solid temperature $\theta_s$, gas temperature $\theta_g$, fiber temperature $\theta_f$, pore gas pressure $p$, solid displacement  $\bs{u_s}$ and fiber displacement $\bs{u_f}$ shown for two different samples in (a) and (b). The displacement states are visualized using their magnitudes, $|\bs{u_s}|$ and $|\bs{u_f}|$. 
    }
    \label{fig:training_design}
\end{figure}

The training data consist of the spatially varying input fiber volume fraction field $\phi_f$ and the corresponding output states $u=(\theta_s,\theta_g,\theta_f,p,\bs{u_s},\bs{u_f})$. Figure \ref{fig:training_design} illustrates these six output fields, namely the solid temperature $\theta_s$, gas temperature $\theta_g$, fiber temperature $\theta_f$, pore gas pressure $p$, solid displacement $\bs{u_s}$, fiber displacement $\bs{u_f}$, for two representative realizations of the input spatial fiber volume fraction field $\phi_f$.
\clearpage
\subsection{Computational cost analysis for risk-averse design}
To assess the efficiency of the proposed framework, we examine both the offline and online computational costs associated with risk-averse optimization. The offline stage includes data generation, residual evaluation, and model training, whereas the online stage consists of repeated model evaluations and gradient computations during optimization.
Table~\ref{tab:cost_duu} summarizes the overall computational cost of the risk-averse optimization workflow for all the stages of workflow.  Table~\ref{tab:iteration_cost_duu} compares the average online cost per optimization iteration for CVaR gradient computation and observable evaluation.
\vspace{0.1in}\\
\begin{table}[htbp]
    \centering
    \small
    \begin{tabular}{|c|c|c|c|c|c|}
    \hline
    \textbf{Method} & \textbf{Data} & \textbf{Residual} & \textbf{Training} & \textbf{Optimization} & \textbf{End-to-end} \\
    \hline
    vBNO & 4795.7 min & --- & 54.4 min & 16.7 min & 4866.9 min \\
    \hline
    RI-BNO & 2397.9 min & 1558.2 min & 47.6 min & 17.8 min & 4021.5 min \\
    \hline
    FE & ---  & --- & --- & 10688.2 min & 10688.2 min \\
    \hline
\end{tabular}
    \caption{Breakdown of wall clock computational time (in minutes) across offline data generation, residual evaluation, model training, and online risk-averse optimization stages for vBNO, RI-BNO, and FE baselines.}
    \label{tab:cost_duu}
\end{table} \vspace{0.1in}\\
\begin{table}[htbp]
    \centering
    \small
    \begin{tabular}{|c|c|c|c|}
        \hline
         \textbf{Method}& \textbf{Observable evaluation}& \textbf{Gradient} & \textbf{Total} \\
        \hline
        vBNO & 0.066 s & 0.098 s & 0.164 s \\
        \hline
        RI-BNO & 0.071 s & 0.105 s & 0.176 s \\
        \hline
    \end{tabular}
    \caption{Average wall-clock time per optimization iteration (in seconds) for risk-averse optimization with vBNO and RI-BNO models. Each iteration consists of a CVaR objective evaluation and its corresponding gradient computation using 5000 samples of $m \sim \nu_{\rm man}$. The reported times are averaged over all optimization iterations.}
    \label{tab:iteration_cost_duu}
\end{table}

\clearpage



\bibliographystyle{elsarticle-num} 
 \bibliography{references}
\end{document}